\pdfoutput=1
\documentclass[12pt]{colt2026}
\usepackage[T1]{fontenc}
\usepackage{times}
\usepackage{mathtools}
\usepackage{enumitem}
\usepackage{booktabs}
\usepackage{tikz}
\usetikzlibrary{arrows.meta,calc,decorations.pathreplacing,patterns}
\newenvironment{restated}[1]{\par\medskip\noindent\textbf{#1 (restated).}\ \itshape\ignorespaces}{\par\medskip}
\newcommand{\NB}{\mathsf{NB}}
\newcommand{\LRW}{\mathsf{LRW}}
\newcommand{\StmtChenLi}{%
Let $P$ be a reversible nearest-neighbour curriculum kernel on a connected, locally finite graph $G$, with reversible measure $\pi$,
$P(x,y)>0$ on every edge and $\sup_x\pi(x)<\infty$, and let $D_0$ be a cold start with $\|D_0\|_{\ell^2(1/\pi)}<\infty$. For every $\epsilon\in(0,1)$ and every
admissible prover sequence for $P$ started from $D_0$ (without replay),
\[
|K_t|\ \ge\ \frac{(1-\epsilon)^2}{\sup_x\pi(x)\,\|D_0\|^2_{\ell^2(1/\pi)}}\Big(\frac{1-\epsilon}{1-\lambda_0}\Big)^{2t}\qquad\text{for every }t\ge0 .
\]
Consequently $\epsilon^*_{\mathrm{div}}(P;G,D_0)\ge\lambda_0\ge\Phi^2/2$. On $T_3^{(L)}$ the simple random walk satisfies
$\epsilon^*_{\mathrm{div}}(P;T_3^{(L)},o)\ge1/(2(4L-1)^2)$.}

\newcommand{\StmtRandomWalk}{%
\begin{enumerate}
\item[(a)] Let $P$ be a reversible nearest-neighbour curriculum kernel on a connected, locally finite graph $G$ with reversible measure $\pi$ and
$P(x,y)>0$ on every edge. Then
$\lambda_0\le\epsilon_D(P)\le\lambda_0/e_{\min}(P)$. For every $\epsilon\in(\epsilon_D(P),1)$ there are a finite window $U$ and a
quasi-stationary distribution $\nu$ of $P_U$ such that, from the cold start $D_0=\nu$ and with any replay weight, the prover sequence
that holds $U$ is admissible at level $\epsilon$ and $K_t=\mathrm{int}\,U$ for every $t\ge0$. If $\sup_x\pi(x)<\infty$, then
\[
\lambda_0\ \le\ \inf_{D_0}\epsilon^*_{\mathrm{div}}(P;G,D_0)\ \le\ \inf_{D_0}\epsilon^*_{\mathrm{bdd}}(P;G,D_0)\ \le\ \epsilon_D(P)\ \le\ \lambda_0/e_{\min}(P),
\]
the infima being over finitely supported cold starts.
\item[(b)] On $T_3^{(L)}$, $L\ge1$, the lazy random walk $\mathsf{LRW}$ started from $\delta_o$ satisfies
\[
\frac1{4(4L-1)^2}\ \le\ \epsilon^*_{\mathrm{div}}(\mathsf{LRW};T_3^{(L)},o)\ \le\ \epsilon^*_{\mathrm{bdd}}(\mathsf{LRW};T_3^{(L)},o)\ \le\ \tfrac32\big(1-2^{-1/(2L)}\big)^2,
\]
where $\frac32(1-2^{-1/(2L)})^2<3(\ln2)^2/(8L^2)$; the upper bound holds for every replay weight.
\end{enumerate}}

\newcommand{\StmtUniform}{%
Let $k\ge1$ and $C_k:=6\cdot10^{31}k^5$. For every $L\ge10^5k$, every $p\in(0,1)$, every $\mathcal C\in\mathcal P_0(k)$ on $T_3^{(L)}$ and every
$\epsilon\in[C_k/L^2,1)$, there is an admissible prover sequence for $\mathcal C$ started from $D_0=\delta_o$ such that every round is
defined and $\sup_t|K_t|<\infty$. Consequently
\[
\epsilon^*_{\mathrm{div}}(\mathcal C;T_3^{(L)},o)\ \le\ \epsilon^*_{\mathrm{bdd}}(\mathcal C;T_3^{(L)},o)\ \le\ \frac{C_k}{L^2}.
\]}

\newcommand{\StmtNB}{%
For every $L\ge1$,
\[
\epsilon^*_{\mathrm{div}}(\mathsf{NB};T_3^{(L)},o)=\epsilon^*_{\mathrm{bdd}}(\mathsf{NB};T_3^{(L)},o)=1-2^{-1/L},\qquad
\frac{\ln2}{L}-\frac{(\ln2)^2}{2L^2}\ \le\ 1-2^{-1/L}\ \le\ \frac{\ln2}{L}.
\]
Moreover, if $\epsilon<1-2^{-1/L}$, then every admissible prover sequence for $\mathsf{NB}$ from $\delta_o$ satisfies
$|K_t|\ge\frac32(1-\epsilon)\big(2^{1/L}(1-\epsilon)\big)^t$ for every $t\ge1$, and $2^{1/L}(1-\epsilon)>1$.}

\newcommand{\StmtSeparation}{%
Let $k\ge1$, write $\epsilon^*(\mathsf{NB})$ for the common value in Theorem~\ref{thm:nb}, and let all tolerances be on $T_3^{(L)}$ from $\delta_o$
without replay.
\begin{enumerate}
\item[(a)] For every fixed $k$, as $L\to\infty$,
\[
\frac{\epsilon^*(\mathsf{NB})}{\sup_{\mathcal C\in\mathcal P_0(k)}\epsilon^*_{\mathrm{bdd}}(\mathcal C)}=\Theta_k(L)
\quad\text{and}\quad
\frac{\epsilon^*(\mathsf{NB})}{\sup_{\mathcal C\in\mathcal P_0^{=}(k)}\epsilon^*_{\mathrm{bdd}}(\mathcal C)}=\Theta_k(L);
\]
both ratios are at most $64\ln2\cdot L$ for every $L\ge1$.
\item[(b)] There is $L_k$ such that for every $L\ge L_k$ the interval $[C_k/L^2,\,1-2^{-1/L})$ is nonempty, and for every $\epsilon$ in it every
$\mathcal C\in\mathcal P_0(k)$ has an admissible prover sequence from $\delta_o$ with every round defined and $\sup_t|K_t|<\infty$, while every
admissible prover sequence for $\mathsf{NB}$ from $\delta_o$ makes $|K_t|$ grow exponentially.
\item[(c)] For every $L\ge1$, with $a:=2^{-1/(2L)}$,
\[
\frac{\epsilon^*(\mathsf{NB})}{\epsilon^*_{\mathrm{bdd}}(\mathsf{LRW})}\ \ge\ \frac{2(1+a)}{3(1-a)}\ >\ 3L .
\]
\end{enumerate}}

\newcommand{\StmtBoundaryA}{%
(Batch selection with bounded shell weight.) Let $\varpi$ be a selection weight with $\varpi(p)>0$ and $\varpi(s^\ast)\le\varpi(p)$ for some
$s^\ast\in[p/2,p)$. For every $\mathcal C\in\mathcal P_0(k)$ with corridor variance $\sigma^2>0$, the selection dynamics of
$(\mathcal C,\varpi)$ on $T_3^{(L)}$ satisfies $\epsilon^*_{\mathrm{bdd}}=O(k^2/L^2)$ from a quasi-stationary cold start of its cycle
operator, and for every lazy $\mathcal C\in\mathcal P_0(1)$ it satisfies $\epsilon^*_{\mathrm{bdd}}=O(1/L^2)$ from $\delta_o$; here $L\to\infty$ and the
constants are absolute.}
\newcommand{\StmtBoundaryB}{%
(Novelty certificate with replay.) On every connected, locally finite graph without dead ends, the outward curriculum with replay weight
$w\in(0,1)$ satisfies $|K_t|\ge(1-\epsilon/w)t$ for every $t\ge1$, every $\epsilon<\min\{w,1-w\}$ and every admissible prover sequence. On
$T_3^{(L)}$ from $\delta_o$ without replay, with the outward kernel uniform on the children of the current vertex, its knowledge satisfies
the bound of Theorem~\ref{thm:nb} for every $\epsilon<1-2^{-1/L}$.}
\newcommand{\StmtBoundaryC}{%
(Proof library.) On $T_3^{(L)}$ from $\{o\}$, the curriculum that stores the set of proved problems and proposes its whole outer
boundary makes the knowledge grow exponentially against every admissible prover sequence when $\epsilon<1/(3(2L-1))$, and some admissible
prover sequence keeps it bounded when $\epsilon>1/(2L)$.}
\newcommand{\StmtBoundary}{%
Let $p\in(0,1)$.
\begin{enumerate}
\item[(a)] \StmtBoundaryA
\item[(b)] \StmtBoundaryB
\item[(c)] \StmtBoundaryC
\end{enumerate}}

\makeatletter
\def\ps@jmlrtps{%
  \let\@mkboth\@gobbletwo
  \def\@oddhead{}%
  \let\@evenhead\@oddhead
  \def\@oddfoot{\hfill\small\rmfamily\thepage\hfill}%
  \let\@evenfoot\@oddfoot
}
\makeatother

\hypersetup{pdftitle={Can Self-Play Keep Improving Itself? One Step of Memory Raises Its Error Tolerance},pdfauthor={Yilin Jiang}}

\title[Can Self-Play Keep Improving Itself?]{Can Self-Play Keep Improving Itself? One Step of Memory Raises Its Error Tolerance}
\coltauthor{%
 \Name{Yilin Jiang} \Email{lien.jiang@ieee.org}\\
 \addr The Hong Kong University of Science and Technology (Guangzhou)%
}

\begin{document}
\maketitle

\begin{abstract}
In self-play theorem proving without external data, a curriculum proposes problems from the prover's pass rates under a fixed
verifier, and we ask whether this loop can keep growing when the prover may forget every problem it is not currently trained on.
\citet{chenli2026} proved that a random-walk curriculum grows the knowledge exponentially below training error $\Phi^2/2$, where
$\Phi$ is the conductance. We show that on the $3$-regular tree with edges subdivided into paths of length $L$, for each fixed $k$, every memoryless,
time-homogeneous curriculum whose proposals lie within distance $k$ and depend only on the anonymous local graph and its pass
rates can be stalled at training error $O(k^5/L^2)$, the order of $\Phi^2$, by a prover that meets this
error bound in every round. This class contains abstractions of
the local scoring rules of STP, R-Zero and Absolute Zero. One step of memory changes the order. The curriculum that stores the
parent of each sample and proposes away from it tolerates exactly $1-2^{-1/L}$, about $(\ln2)/L$, at the per-proposal oracle cost
of the lazy random walk. The stored parent supplies the outward direction that an anonymous local view cannot detect, and on these
trees it keeps the loop growing at training errors larger by a factor of order $L$ for each fixed $k$.
\end{abstract}

\begin{keywords}
self-play, curriculum learning, forgetting, quasi-stationary distributions
\end{keywords}

\section{Introduction}\label{sec:intro}

Recent systems for automated reasoning train a prover on problems that the system generates itself. In STP
\citep{dong2025stp}, R-Zero \citep{huang2026rzero} and Absolute Zero \citep{zhao2025absolute}, a proposer generates new problems,
often conditioned on earlier problems of the system, the problems are selected or rewarded by the prover's empirical pass rate,
and the prover is trained on the selected problems. The loop uses no external data; its only signals are the verdicts of a
fixed verifier or consistency check and the pass rates computed from them. The learning-theoretic question is whether the local
scoring of proposals by pass rates, which turns these signals into a curriculum, lets the loop keep improving, and what else the
component that proposes problems needs.

In STP the expert-iteration baseline plateaus \citep[Sections~1 and~4.4]{dong2025stp}, R-Zero reports degradation after
several iterations \citep[Section~4.2]{huang2026rzero}, and in the ablation of SGS without its guide a conjecturer rewarded only
through the solver's pass rate collapses to artificially complex problems \citep[Section~4.4]{bailey2026sgs}. In each case the loop
stops expanding. Training controls the error only on the current training distribution, so the prover may forget problems on which
it is no longer trained, and growth then depends on how much training error per round the curriculum can tolerate.

\citet{chenli2026} model the loop on a theorem graph whose vertices are problems and whose edges join problems on which every
prover of the class has pass rates within $p/2$ (their Assumption~3). With a random-walk conjecturer, a filter that keeps
proposals with pass rate at least $p/2$ and a prover trained to accuracy $1-\epsilon$ on the kept proposals, they prove that the
knowledge, the set of problems solved with pass rate at least $p$, grows exponentially whenever $\epsilon<\Phi^2/2$, where $\Phi$
is the conductance of the walk \citep[Theorem~4.2]{chenli2026}. Their Section~6 lists the lack of adaptation to the current prover as
a limitation. Related analyses of self-improving generators take the expansion of coverage as an
assumption \citep[Proposition~4.1]{yang2026ancora} or as a separate mechanism \citep[Section~2]{avestimehr2026nova}. It has
remained open whether the condition of Chen and Li is necessary, whether a curriculum that reads the prover's pass rates can
tolerate more error, and what else a curriculum could use.
\begin{center}
\begin{minipage}{0.88\linewidth}
\centering\itshape
When the prover may forget whatever it is not currently trained on, how much training error can a self-play curriculum
tolerate while its knowledge keeps growing, and how much memory does the curriculum need?
\end{minipage}
\end{center}
We answer this question on the $3$-regular tree $T_3^{(L)}$ whose edges are subdivided into paths of length $L$, on which $\Phi$ is
of order $1/L$. For each fixed $k$, a memoryless, time-homogeneous curriculum
that proposes within distance $k$ from the anonymous local graph and its pass rates tolerates training error at most of order
$L^{-2}$, whatever use it makes of the pass rates, while a curriculum that remembers one parent per sample tolerates
exactly $1-2^{-1/L}$, of order $L^{-1}$, at the per-proposal oracle cost of the lazy random walk.

\begin{figure}[t]
\centering
\includegraphics[width=0.56\textwidth]{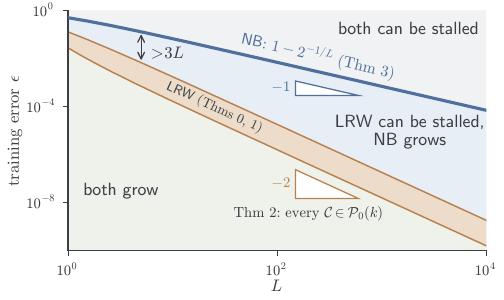}
\caption{\textbf{The memoryless class tolerates order $L^{-2}$ and $\NB$ order $L^{-1}$.} From $o$ without replay; the
triangles mark slopes, only the order of Theorem~\ref{thm:uniform} is drawn, and $>3L$ is the ratio of Corollary~\ref{cor:sep}(c).}
\label{fig:regions}
\end{figure}

We measure a curriculum by two forgetting tolerances in the model of Chen and Li. The divergence tolerance is the supremum of the
training errors at which the knowledge tends to infinity against every admissible prover sequence, and the stall threshold is the
infimum of those at which some admissible prover sequence keeps it bounded (Definitions~\ref{def:main-admissible}
and~\ref{def:main-tolerance}). Table~\ref{tab:main} and Figure~\ref{fig:regions} collect the bounds. For reversible random-walk
curricula the spectral gap determines the smallest stall threshold up to the inverse of the smallest edge probability, which gives a
converse to the growth condition of Chen and Li, and on $T_3^{(L)}$ the lazy random walk $\LRW$ can be stalled at training error of
order $L^{-2}$ already from the fixed start $o$ (Theorem~\ref{thm:rw}). The class $\mathcal P_0(k)$ consists of the memoryless,
local, time-homogeneous and view-equivariant curricula. Their proposals depend only on the anonymous view of radius $k$, which shows
the local graph and its pass rates but no problem identities or coordinates. For each fixed $k$, every curriculum in
$\mathcal P_0(k)$ can be stalled from $o$ at training error $O(k^5/L^2)$, and $\LRW\in\mathcal P_0(1)$ attains this order, so the
condition of Chen and Li is tight for the whole class up to a factor that depends only on $k$ (Theorem~\ref{thm:uniform}).

One stored parent changes the order. The non-backtracking curriculum $\NB$ stores the parent of each sample, proposes a uniform
neighbour other than the parent and tolerates exactly $1-2^{-1/L}\approx(\ln2)/L$ (Theorem~\ref{thm:nb}). For each fixed $k$ this
exceeds the largest tolerance in $\mathcal P_0(k)$ by a factor of order $L$ as $L\to\infty$, at no greater per-proposal oracle cost
(Corollary~\ref{cor:sep}). With replay weight $1/2$, $\NB$ has divergence tolerance at least $1/2$, while $\LRW$ with any replay
weight can still be stalled at training error $O(L^{-2})$ (Corollary~\ref{cor:nb-tree-replay}).

\begin{table}[t]
\caption{\textbf{Forgetting tolerances on $T_3^{(L)}$.} Without replay, from $o$ unless stated otherwise. Theorem~0 is due to
\citet{chenli2026}, $C_k$ is from Theorem~\ref{thm:uniform}, and batch selection has shell weight at most $1$ and a base rule of
positive corridor variance $\sigma^2$.}
\label{tab:main}
\centering
\small\setlength{\tabcolsep}{4pt}
\begin{tabular}{@{}llll@{}}
\toprule
Curriculum & Extra information & Tolerance & Source\\
\midrule
$\LRW$ (in $\mathcal P_0(1)$) & none &
$\frac1{4(4L-1)^2}\le\epsilon^*_{\mathrm{div}}\le\epsilon^*_{\mathrm{bdd}}\le\frac32\big(1-2^{-1/(2L)}\big)^2$ & Thms~\ref{thm:chenli}, \ref{thm:rw}\\
Every $\mathcal C\in\mathcal P_0(k)$ & none & $\epsilon^*_{\mathrm{bdd}}\le C_k/L^2$ for $L\ge10^5k$ & Thm~\ref{thm:uniform}\\
Batch selection & none & $\epsilon^*_{\mathrm{bdd}}=O(k^2/L^2)$ from a QSD start & Prop.~\ref{prop:boundary}(a)\\
\midrule
$\NB$ & parent & $\epsilon^*_{\mathrm{div}}=\epsilon^*_{\mathrm{bdd}}=1-2^{-1/L}$ & Thm~\ref{thm:nb}\\
Outward & novelty certificate & $\epsilon^*_{\mathrm{div}}=\epsilon^*_{\mathrm{bdd}}=1-2^{-1/L}$ & Rem.~\ref{rem:nb-outward}\\
Proof library & proved problems & $\frac1{3(2L-1)}\le\epsilon^*_{\mathrm{div}}\le\epsilon^*_{\mathrm{bdd}}\le\frac1{2L}$ & Prop.~\ref{prop:boundary}(c)\\
\bottomrule
\end{tabular}
\end{table}

The mechanism behind the separation is the difference between diffusion and direction. At a vertex of an arm whose view of
radius $k$ is a path on which every pass rate equals $p$, view equivariance and the reflection of the path force every curriculum in
$\mathcal P_0(k)$ to take a symmetric step of range at most $k$. A prover that holds the mass at such views for several rounds lets
it spread along the arm like a symmetric random walk and then forgets the edge of the knowledge for one round. For each fixed $k$
this keeps the training error at $O(L^{-2})$ and the knowledge bounded, and a rule whose step at these views has zero variance is
stalled by a prover that holds a fixed set (Section~\ref{sec:overview}).
The stored parent marks the inward side, so $\NB$ crosses an arm in $L$ rounds, its spheres double every
$L$ levels, and the retained knowledge grows exponentially when $2(1-\epsilon)^L>1$.

Our contributions are as follows. First, we define the class $\mathcal P_0(k)$ by four conditions, show that it contains
abstractions of the local scoring rules of STP, R-Zero and Absolute Zero and the local form of Chen and Li's diffusion
conjecturer (Appendix~\ref{app:B_model}), and give one stall construction for every rule of the class from the fixed start, with no
assumption on a rule at views with non-constant pass rates. Second, we determine the exact forgetting tolerance of $\NB$ and show that a stored
parent, a certificate that a proposal is farther from the start and a library of proved problems used for training each supply the
outward direction that lets a curriculum reach order $L^{-1}$, while batch selection that weights the frontier no more than the
knowledge has stall threshold $O(k^2/L^2)$ from a quasi-stationary start for every rule of $\mathcal P_0(k)$ that moves along an arm
at constant pass rates (Proposition~\ref{prop:boundary}). Third, for random-walk curricula we bound the stall threshold from both sides by the inner-boundary
mass of quasi-stationary distributions of killed walks, which gives the converse of the growth theorem of Chen and Li on every graph with $\sup_x\pi(x)<\infty$
and its fixed-start form on subdivided trees through the Keilson--Fill representation of birth--death passage times.

\subsection{Related work}\label{sec:related}

\paragraph{Growth theory of self-play and self-improvement.}
\citet{chenli2026} introduced the theorem-graph model and proved exponential growth for a reversible random-walk conjecturer
below training error $\Phi^2/2$ (their Theorem~4.2, with the Cheeger inequality of \citealp[Theorem~10.3]{woess2000}). We prove
the converse direction, bound the tolerance of every curriculum in $\mathcal P_0(k)$, which may adapt to the current pass rates,
and show that one stored parent changes the order. The growth proposition of \citet[Proposition~4.1]{yang2026ancora} assumes that each step admits a new problem with
probability bounded away from zero, and we give the error level at which such an admission rate fails against a prover that may
forget. \citet{avestimehr2026nova} show that anchoring preserves exposure within the support of a base distribution (Theorem~3
and Corollary~4) and note that expanding the support needs a separate mechanism; the local proposal analysed here is one such
mechanism.

\paragraph{Curriculum learning theory.}
\citet{rajaraman2026curriculum} prove that selecting prompts from a fixed distribution by the model's correctness lowers the cost
of supervised and reinforcement-learning fine-tuning, and list self-play and coverage expansion as open questions.
\citet{rajaraman2026curriculumii} expand coverage by composition, and \citet{diazbone2025discover} analyse target-directed goal
selection assuming that goals near previously achieved ones become achievable (their Assumption~C.5). In our model the curriculum
generates problems without a pool or a target, and the prover may forget.

\paragraph{Undirected exploration and local information.}
In the reinforcement-learning exploration models of \citet{whitehead1991} and
\citet{thrun1992}, undirected random-walk exploration of a static environment can take exponential time, and counter-based
exploration \citep{thrun1992} or a suitable reward representation \citep{koenig1996} makes it polynomial. Related results concern
memoryless policies in POMDPs \citep{singh1994}, graph exploration by finite automata \citep{budach1978,fraigniaud2005} and faster
mixing of non-backtracking walks \citep{alon2007nonbacktracking}. In our model the orders $L^{-2}$ and $L^{-1}$ of $\LRW$ and $\NB$
match the inverse crossing times of a diffusive and a directed walk on an arm, the prover sets the pass rates in every round, and the
bound $O(k^5/L^2)$ holds for every rule of $\mathcal P_0(k)$, which may adapt to them. \citet{kim2025metastable} prove
statistical-query hardness for learners restricted to local neighbourhoods of a reasoning graph (their Theorem~5.3 and
Corollary~5.6), and view equivariance is an analogous restriction of the curriculum.

\paragraph{Empirical self-play systems.}
In concurrent work, \citet{pu2026compassplay} observe that a reward based only on success rate cannot distinguish tasks with
equal success rates that contribute differently to target performance; in our model every such rule of $\mathcal P_0(k)$ tolerates
at most order $k^5/L^2$ (Theorem~\ref{thm:uniform}). Appendix~\ref{app:A_related} discusses further related work.

\section{Model and preliminaries}\label{sec:model}

We use the theorem-graph model of \citet{chenli2026}. The problems are the vertices of a connected, locally finite graph
$G=(V,E)$ with a start vertex $o$, and an edge joins two similar problems. The curriculum plays the role of Chen and Li's conjecturer
and of the proposers above, and the filter that keeps proposals with pass rate at least $p/2$ is their
Algorithm~2. In round $t$ the prover has a pass-rate profile $s_t:V\to[0,1]$ and, for a threshold $p\in(0,1)$, knowledge
$K_t=\{x:s_t(x)\ge p\}$. As in their Assumption~2, the prover is required to be accurate only on its current training distribution.
We impose their Assumptions~2--3 in every round: each growth result below holds for every sequence of pass-rate profiles that
meets them, and each stall result constructs one such sequence (Appendix~\ref{app:B_model}). $N(x)$ denotes the neighbours of $x$ and $B(x,k)$ the
ball of radius $k$ around $x$.

\begin{definition}[Admissible prover sequence]\label{def:main-admissible}
Given training distributions $(D_t)_{t\ge0}$ on $V$, a sequence of pass-rate profiles $(s_t)_{t\ge0}$ is admissible at level
$\epsilon\in[0,1)$ if, for every $t$, (i) $|s_t(x)-s_t(y)|\le p/2$ for every edge $\{x,y\}\in E$, and (ii) $D_t(K_t)\ge1-\epsilon$.
\end{definition}

Condition (i) is Chen and Li's Assumption~3, and (ii) bounds the current training error. Nothing relates $s_t$ to
$s_{t-1}$, so the prover may forget any problem outside the support of $D_t$.

From a sample at $x$ in round $t$ a memoryless curriculum proposes $y\sim C_t(x,\cdot)$, which the filter keeps if
$s_t(y)\ge p/2$. From a cold start $D_0$, the training distributions follow
\begin{equation}\label{eq:main-update}
D_{t+1}(y)=\frac{\sum_xD_t(x)\,C_t(x,y)\,\mathbf 1[s_t(y)\ge p/2]}{\sum_{x,y'}D_t(x)\,C_t(x,y')\,\mathbf 1[s_t(y')\ge p/2]},
\end{equation}
the law of the proposal chain conditioned on all its proposals being kept (Lemma~\ref{lem:st-survival}); a round is
defined if its denominator is positive. With replay weight $w\in[0,1]$, the frontier $F_t$ follows
\eqref{eq:main-update} from $F_0=D_0$ and the prover trains on $D_t=(1-w)F_t+wR_t$, with $R_0=F_0$ and
$R_t=\frac1t\sum_{s<t}F_s$; $w=0$ is the dynamics of Chen and Li, and a weight $w>0$ keeps the problems of earlier rounds in
training. The main cold start is $\delta_o$.

\begin{definition}[Forgetting tolerances]\label{def:main-tolerance}
For a curriculum $\mathcal C$ on $G$ and a cold start $D_0$, with admissible prover sequences taken for the dynamics of
$\mathcal C$ from $D_0$, the divergence tolerance $\epsilon^*_{\mathrm{div}}(\mathcal C;G,D_0)$ is the supremum of the
$\epsilon\in[0,1)$ at which every admissible prover sequence has $|K_t|\to\infty$, and the stall threshold
$\epsilon^*_{\mathrm{bdd}}(\mathcal C;G,D_0)$ is the infimum of the $\epsilon\in[0,1)$ at which some admissible prover sequence
has $\sup_t|K_t|<\infty$, with $\sup\emptyset=0$ and $\inf\emptyset=1$. We write $(\mathcal C;G,o)$ for
$(\mathcal C;G,\delta_o)$, and the definitions apply with any given replay weight.
\end{definition}

Admissibility at level $\epsilon$ persists at every higher level, so $\epsilon^*_{\mathrm{div}}\le\epsilon^*_{\mathrm{bdd}}$, and
we write $\epsilon^*$ when the two are equal. Tolerances are without replay unless a replay weight is stated.

\begin{definition}[Memoryless local curricula]\label{def:main-P0}
A curriculum $\mathcal C\in\mathcal P_0(k)$ is a rule $v\mapsto\mathcal C_v$ that assigns a probability distribution on the
vertices of $v$ to every labelled rooted view $v=(B(x,k),x,s|_{B(x,k)})$ of radius $k$ such that (M) (memoryless) a sample is a
single vertex, and the proposal from a sample at $x$ in round $t$ is drawn from $\mathcal C_v$ for
the anonymous view $v=\mathcal V^k_t(x):=(B(x,k),x,s_t|_{B(x,k)})$ of $x$; (L) (local) proposals from $x$ lie in $B(x,k)$; (T) (time-homogeneous) the
rule does not depend on $t$; and (E) (view-equivariant) if $\varphi:v\to v'$ is an isomorphism of labelled rooted views, a graph
isomorphism of the balls that maps root to root and preserves the labels, then $\mathcal C_{v'}(\varphi(A))=\mathcal C_v(A)$
for every set $A$ of vertices of $v$.
\end{definition}

The four conditions say that a proposal is generated from the current sample alone, near it, by the same rule in every round,
and from a law that depends only on the local graph structure and the observed pass rates, without problem identities,
distances to $o$ or coordinates. Every rule $\mathcal C_v(y)\propto q_v(y)\,f(s(y))$ with an equivariant base
proposal $q_v$, a weight $f$ of the pass rate and an equivariant fallback is in $\mathcal P_0(k)$; this form contains
abstractions of the local scoring rules of STP, R-Zero and Absolute Zero applied to local proposals, and the local form of
Chen and Li's diffusion conjecturer with the simple-random-walk baseline and diffusion time $\tau$ is in $\mathcal P_0(\tau+2)$
(Appendix~\ref{app:B_model}). The lazy random walk
$\LRW$, which proposes $x$ itself with probability $1/2$ and a uniform neighbour otherwise, is in every $\mathcal P_0(k)$.

\begin{definition}[One step of memory]\label{def:main-memory}
A curriculum with one step of memory stores each sample as a pair $(x_-,x)$ of a parent $x_-\in N(x)$ and a current vertex
$x$, with $x_-=\bot$ for the start sample, proposes $x'\in N(x)$ from a law $\mathcal M((x_-,x),\cdot)$, applies the filter of
\eqref{eq:main-update} to $x'$ and stores a kept proposal as $(x,x')$. On a tree of minimum degree at least $2$, the non-backtracking
curriculum $\NB$ proposes $x'$ uniformly from $N(x)\setminus\{x_-\}$, and from $(\bot,o)$ a uniform neighbour of $o$.
\end{definition}

Per proposal $\NB$ makes the same oracle calls as $\LRW$ and stores one vertex more
(Lemma~\ref{lem:mem-budget}); the only information it adds to the anonymous view of radius $1$ is which neighbour of $x$, if any,
is $x_-$. The equal-budget subclass $\mathcal P_0^{=}(k)$
consists of the $\mathcal C\in\mathcal P_0(k)$ that, with the same filter, replay, training oracle and proof sampling, use at
least as many neighbour-oracle calls and pass-rate queries per proposal as $\NB$; it contains $\LRW$.

\paragraph{Spectral quantities.} A kernel $P$ is nearest-neighbour if each $P(x,\cdot)$ is supported on $N(x)\cup\{x\}$, and
then $e_{\min}(P)=\inf_x\min_{y\in N(x)}P(x,y)$. If $\pi$ is a reversible measure of $P$, the spectral gap is
$\lambda_0=1-\|P\|_{\ell^2(\pi)}$, the conductance is $\Phi(P)=\inf_A a(\partial A)/\pi(A)$ over finite nonempty $A$, with
$a(\partial A)=\sum_{x\in A,y\notin A}\pi(x)P(x,y)$, and $\|D\|^2_{\ell^2(1/\pi)}=\sum_xD(x)^2/\pi(x)$; the bound
$\lambda_0/e_{\min}(P)$ is read as $+\infty$ when $e_{\min}(P)=0$.

\paragraph{Instances.} The main instance is $T_3^{(L)}$, the $3$-regular tree with every edge replaced by a path with $L$
edges and $o$ a vertex of degree $3$; it grows exponentially, and its simple random walk has conductance $\Phi=1/(4L-1)$
(Lemma~\ref{lem:st-phi}), so it satisfies the isoperimetric assumption of Chen and Li (their Assumption~5) with
isoperimetric constant $4L-1$. A path of length $L$ between two branch vertices is an arm. The level $|x|$ of a vertex is its
distance from $o$, and $S_r=\{x:|x|=r\}$ is the sphere of radius $r$.

\setcounter{theorem}{-1}
\begin{theorem}[\citealp{chenli2026}, Theorem~4.2]\label{thm:chenli}\StmtChenLi\end{theorem}

Theorem~\ref{thm:chenli} is the form of their proof in Appendix~A.2 before the Cheeger step $\lambda_0\ge\Phi^2/2$
\citep[Theorem~10.3 and its proof]{woess2000}, written out in their Theorem~A.14.

\paragraph{Windows.} For finite nonempty $U\subseteq V$ let $\partial_{\mathrm{in}}U=\{x\in U:N(x)\not\subseteq U\}$ and
$\mathrm{int}\,U=U\setminus\partial_{\mathrm{in}}U$. The restriction $P_U$ of a kernel $P$ to $U\times U$, the walk killed on
leaving $U$, has Perron root $\lambda_U$, and we call a probability vector $\nu$ with $\nu P_U=\lambda_U\nu$ a quasi-stationary distribution
(QSD) of $P_U$, which for irreducible $P_U$ with $\lambda_U>0$ is the unique QSD of the killed chain
\citep{collet2013,meleard2012}. Let $\epsilon_D(P)$ be the infimum of $\nu(\partial_{\mathrm{in}}U)$
over all such $U$ and QSDs $\nu$ of $P_U$. The prover holds the window $U$ if in every round $s_t$ equals $p$ on
$\mathrm{int}\,U$, $p/2$ on $\partial_{\mathrm{in}}U$ and $0$ off $U$; then $K_t=\mathrm{int}\,U$, and from a cold start in $U$
holding $U$ is admissible exactly when $D_t(\partial_{\mathrm{in}}U)\le\epsilon$ for all $t$ (Appendix~\ref{app:C_window}).

\section{Main results}\label{sec:results}
\begingroup\setlist[enumerate]{nosep}

\subsection{Memoryless curricula}\label{sec:res-rw}\label{sec:res-uniform}

\begin{theorem}[Random-walk curricula]\label{thm:rw}\StmtRandomWalk\end{theorem}

Part (a) says that a prover that may forget can pin the training distribution, with any replay weight, to a QSD of the walk
killed outside a window, and that the smallest inner-boundary mass of such a QSD lies between $\lambda_0$ and
$\lambda_0/e_{\min}(P)$, which with Theorem~\ref{thm:chenli} places the smallest stall threshold over finitely supported cold starts between the same two
quantities when $\sup_x\pi(x)<\infty$.
On $T_3^{(L)}$ both sides of (b) are of order $\Phi^2$, so the bound $\Phi^2/2$ of Chen and Li has the
correct order also from the fixed start $o$, and replay with any weight leaves the upper bound in place. When the spectral gap
vanishes, as for $\LRW$ on $\mathbb Z^D$, the knowledge that the loop guarantees is still finite: without replay the infimum of
$\liminf_t|K_t|$ over admissible prover sequences from $\delta_0$ is $\Theta_D(\epsilon^{-D/2})$ (Corollary~\ref{cor:plateau-zd}),
and with any replay weight a prover that holds a cube keeps the knowledge below $(\pi\sqrt{D/\epsilon})^D$ from $\delta_0$
(Corollary~\ref{cor:st-ZD}).

\begin{theorem}[Uniform fixed-start bound for $\mathcal P_0(k)$]\label{thm:uniform}\StmtUniform\end{theorem}

Theorem~\ref{thm:uniform} is the counterpart of Theorem~\ref{thm:chenli} for the whole class. For each fixed $k$ and every
curriculum in $\mathcal P_0(k)$ it gives an admissible prover sequence that keeps the knowledge finite once the training
error reaches $C_k/L^2$, whatever use the rule makes of the local graph and of the observed pass rates, at branch vertices and at
the edge of the knowledge included, so it covers every rule $\mathcal C_v(y)\propto q_v(y)\,f(s(y))$ for every weight $f$. For
simple rules the bounds are explicit at every size of the tree. Call $\mathcal C$ lazy if it proposes the root of every view with
probability at least $1/2$, as $\LRW$ and every rule $\frac12\delta_x+\frac12\mathcal C'_v$ with $\mathcal C'\in\mathcal P_0(1)$
do. For every $L\ge400$ every lazy rule in $\mathcal P_0(1)$ has stall threshold at most $12000/L^2$ from $\delta_o$
(Corollary~\ref{cor:un-lazy}), $\LRW$ has the two-sided bounds of Theorem~\ref{thm:rw}(b) for every $L\ge1$, and by
Corollary~\ref{cor:sep}(c) the tolerance of $\NB$ exceeds $3L$ times the stall threshold of $\LRW$ for every $L\ge1$.
Theorem~\ref{thm:uniform} extends the order $L^{-2}$ to every rule of $\mathcal P_0(k)$ for each fixed $k$, and since $\LRW$ attains
this order, the condition of Chen and Li is tight for the whole class up to a factor that depends only on $k$.

The separation between $\NB$ and $\LRW$ and the order $L^{-2}$ in $\mathcal P_0(1)$ also hold against conservative prover
sequences, which off the training support may lower any pass rate but raise one only as far as condition~(i) of
Definition~\ref{def:main-admissible} forces from the support (Definition~\ref{def:conservative}). The stalls of
Theorems~\ref{thm:rw} and~\ref{thm:nb} are conservative, so these theorems and Corollary~\ref{cor:sep}(c) hold as stated
(Corollary~\ref{cor:h-transfer}). A rule in $\mathcal P_0(1)$ is blind at known roots if, at a vertex of the
knowledge, replacing every pass rate in its view by $p$ does not change its proposal law, as for $\LRW$. Every such rule
satisfies the bound of Theorem~\ref{thm:uniform} with $k=1$ against conservative provers from $\delta_o$
(Proposition~\ref{prop:h-blind}). For $L\ge400$, every rule in $\mathcal P_0(1)$ is stalled by a conservative prover at training
error at most $1.21\cdot10^4/L^2$, from $\delta_o$ in the degenerate cases of Corollary~\ref{cor:h-summary}(a) and otherwise from a
quasi-stationary start of its cycle (Theorem~\ref{thm:h-P01}).

\subsection{One stored parent}\label{sec:res-nb}

\begin{theorem}[Exact tolerance of $\NB$]\label{thm:nb}\StmtNB\end{theorem}

\begin{corollary}[Separation at equal per-proposal cost]\label{cor:sep}\StmtSeparation\end{corollary}

With Theorem~\ref{thm:uniform}, Theorem~\ref{thm:nb} and Corollary~\ref{cor:sep} show that for each fixed $k$ one step of
memory raises the attainable tolerance from order $L^{-2}$ to order $L^{-1}$ as $L\to\infty$, against the whole class and, through
the equal-budget subclass $\mathcal P_0^{=}(k)$, at no greater per-proposal oracle cost. The gain comes from direction. At a vertex of
an arm whose view of radius $k$ is a path on which every pass rate equals $p$, every curriculum in $\mathcal P_0(k)$ takes a
symmetric step of range at most $k$ (Lemma~\ref{lem:un-corridor}), so it cannot tell outward from inward there. Under $\NB$ the training distribution of round $t$ lives on the sphere $S_t$, so a prover with training error $\epsilon$
removes at most an $\epsilon$ fraction of the mass per round while the spheres double every $L$ levels. The retained knowledge
therefore grows exponentially when $2(1-\epsilon)^L>1$, and the tolerance solves $2(1-\epsilon)^L=1$. On $T_3^{(L)}$ from $o$, $\NB$
and the outward curriculum of Proposition~\ref{prop:boundary}(b), which is told which neighbours lie farther from $o$, have the same
training distributions for every prover sequence and hence the same tolerances (Remark~\ref{rem:nb-outward}), so the stored parent
supplies the outward direction that the anonymous view lacks, at the per-proposal oracle cost of $\LRW$.

Direction also makes replay effective. With replay weight $1/2$, $\NB$ has divergence tolerance at least $1/2$ on $T_3^{(L)}$
from $o$, independently of $L$, while $\LRW$ with any replay weight has stall threshold below $3(\ln2)^2/(8L^2)$
(Theorem~\ref{thm:rw}(b) and Corollary~\ref{cor:nb-tree-replay}).

\subsection{What lets a curriculum escape}\label{sec:res-boundary}

Proposition~\ref{prop:boundary} places three further rules relative to $\mathcal P_0(k)$. For $\mathcal C\in\mathcal P_0(k)$, the
corridor variance $\sigma^2$ is the variance of the symmetric step law that $\mathcal C$ follows at every vertex of an arm whose
view of radius $k$ is a path on which every pass rate equals $p$ (Lemma~\ref{lem:un-corridor}), and the cycle operator is the
one-cycle kernel of the alternating construction of Section~\ref{sec:overview-uniform}. Batch selection replaces the filter by a
pass-rate weight $\varpi$ that vanishes below $p/2$ and is normalised over the batch; the shell weight of $s\in[p/2,p)$, a pass rate
that passes the filter without being known, is $\varpi(s)/\varpi(p)$. The outward curriculum proposes only neighbours that a novelty
certificate places farther from $o$. The proof-library curriculum stores every proved problem in a library $B_t$, trains the prover
uniformly on $B_t$ in each round and proposes the whole outer boundary of $B_t$.

\begin{proposition}[The boundary of $\mathcal P_0(k)$]\label{prop:boundary}\StmtBoundary\end{proposition}

Part (a) proves that batch selection with a rule of positive corridor variance has stall threshold $O(k^2/L^2)$ from a
quasi-stationary start, and $O(L^{-2})$ from $\delta_o$ if the rule is also lazy and in $\mathcal P_0(1)$, whenever some pass rate in
$[p/2,p)$ receives no more weight than $p$; a rule of zero corridor variance is stalled from $\delta_o$ at every training error
(Proposition~\ref{prop:un-sel}). This includes the filter of Chen and Li and every weight that is positive at $p$ and nondecreasing on
$[p/2,p]$. Parts (b) and (c) reach order $L^{-1}$ with information outside the anonymous view, and $\NB$ obtains the outward
direction from one stored parent. It matches both growth bounds of (b) on $T_3^{(L)}$ (Theorem~\ref{thm:nb} and
Corollary~\ref{cor:nb-tree-replay}), and for $L\ge2$ the lower bound of Theorem~\ref{thm:nb} places its tolerance above the bound
$1/(2L)$ of (c).
\endgroup

\section{Technical overview}\label{sec:overview}

\subsection{Holding a window}\label{sec:overview-window}

\paragraph{Windows and quasi-stationary distributions.} While the prover holds a window $U$, a QSD $\nu$ of $P_U$ with
$\lambda_U>0$ is a fixed point of \eqref{eq:main-update}, with or without replay, so holding $U$ from $\nu$ is admissible exactly
when $\nu(\partial_{\mathrm{in}}U)\le\epsilon$ (Theorem~\ref{thm:st-qsd}). Summing $\nu P_U=\lambda_U\nu$ over $U$ gives
$e_{\min}(P)\,\nu(\partial_{\mathrm{in}}U)\le1-\lambda_U\le\nu(\partial_{\mathrm{in}}U)$ (Lemma~\ref{lem:st-boundary}), and the
Rayleigh formula $\inf_U(1-\lambda_U)=\lambda_0$ turns this into the two-sided spectral bound of Theorem~\ref{thm:rw}(a)
(Corollary~\ref{cor:st-sandwich}).

\paragraph{From the fixed start.} From $\delta_o$ the window must keep $D_t(\partial_{\mathrm{in}}U)\le\epsilon$ in every round,
while the QSD controls only the limit, so a monotone comparison is needed. For a lazy birth--death chain started at its reflecting end and
killed only from its top level, the killing time is a sum of independent geometric variables with the eigenvalues of the killed
kernel as failure probabilities (\citealp{keilson1979}; \citealp[Theorem~1.2]{fill2009}), so its hazard is nondecreasing and the
top-level mass $D_t(\partial_{\mathrm{in}}U)$ increases to its quasi-stationary value (Lemma~\ref{lem:st-keilson} and
Theorem~\ref{thm:st-fixed}). The lazy random walk on a ball of $T_3^{(L)}$ around $o$ lumps into such a chain, and a radial
Rayleigh bound for $B(o,mL)$ bounds the boundary mass of its QSD by a quantity tending to $\frac32(1-2^{-1/(2L)})^2=\Theta(L^{-2})$
as $m\to\infty$ (Lemma~\ref{lem:st-ball} and Corollary~\ref{cor:st-tree}).

\subsection{The uniform bound over \texorpdfstring{$\mathcal P_0(k)$}{P0(k)}}\label{sec:overview-uniform}

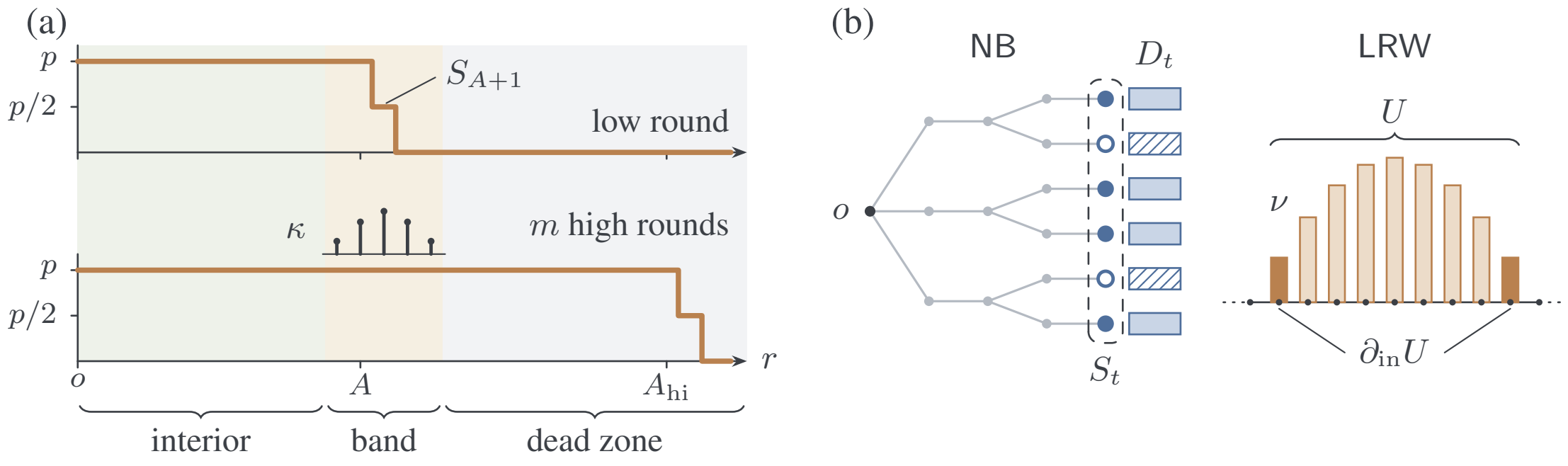
\begin{figure}[t]
\centering
\definecolor{figText}{HTML}{3B3F45}%
\definecolor{figGrey}{HTML}{B4BAC2}%
\definecolor{figLRW}{HTML}{B9814F}%
\definecolor{figLRWfill}{HTML}{ECDCC9}%
\definecolor{figNB}{HTML}{4F6F9C}%
\definecolor{figNBfill}{HTML}{C9D5E6}%
\definecolor{figInt}{HTML}{EEF2EA}%
\definecolor{figBand}{HTML}{F5EFE2}%
\definecolor{figDead}{HTML}{F2F3F5}%
\begin{tikzpicture}[x=1cm,y=1cm,font=\footnotesize,color=figText,line cap=round,
  lev/.style={line width=0.5pt,-{Stealth[length=4pt]}},
  prof/.style={figLRW,line width=1.4pt,line join=round},
  tick/.style={line width=0.5pt},
  brc/.style={decorate,decoration={brace,amplitude=3pt,mirror},line width=0.5pt},
  brcup/.style={decorate,decoration={brace,amplitude=3pt},line width=0.5pt}]
\begin{scope}
  \def\figu{0.22}
  \def\figY{1.95}
  \fill[figInt]  (0,0) rectangle (10.5*\figu,2.95);
  \fill[figBand] (10.5*\figu,0) rectangle (15.5*\figu,2.95);
  \fill[figDead] (15.5*\figu,0) rectangle (6.25,2.95);
  \node[anchor=west,font=\small] at (-0.62,3.15) {(a)};
  \draw[lev] (0,\figY) -- (6.25,\figY);
  \draw[tick] (0,\figY) -- (0,\figY+1.0);
  \foreach \h/\t in {0.85/p,0.425/{p/2}}{
    \draw[tick] (-0.06,\figY+\h) -- (0,\figY+\h);
    \node[anchor=east,font=\scriptsize] at (-0.06,\figY+\h) {$\t$};}
  \foreach \r in {0,12,25}{\draw[tick] (\r*\figu,\figY) -- (\r*\figu,\figY-0.06);}
  \draw[prof] (0,\figY+0.85) -- (12.5*\figu,\figY+0.85) -- (12.5*\figu,\figY+0.425) -- (13.5*\figu,\figY+0.425)
              -- (13.5*\figu,\figY) -- (6.1,\figY);
  \draw[line width=0.4pt] (13.1*\figu,\figY+0.46) -- (3.34,\figY+0.70);
  \node[anchor=west] at (3.30,\figY+0.72) {$S_{A+1}$};
  \node[anchor=east] at (6.22,\figY+0.30) {low round};
  \draw[lev] (0,0) -- (6.25,0);
  \node[anchor=west] at (6.25,0) {$r$};
  \draw[tick] (0,0) -- (0,1.0);
  \foreach \h/\t in {0.85/p,0.425/{p/2}}{
    \draw[tick] (-0.06,\h) -- (0,\h);
    \node[anchor=east,font=\scriptsize] at (-0.06,\h) {$\t$};}
  \foreach \r/\t in {0/o,12/A,25/A_{\mathrm{hi}}}{
    \draw[tick] (\r*\figu,0) -- (\r*\figu,-0.06);
    \node[anchor=north,font=\scriptsize,inner sep=1.5pt] at (\r*\figu,-0.06) {$\t$};}
  \draw[prof] (0,0.85) -- (25.5*\figu,0.85) -- (25.5*\figu,0.425) -- (26.5*\figu,0.425) -- (26.5*\figu,0) -- (6.1,0);
  \node[anchor=east] at (6.22,1.25) {$m$ high rounds};
  \draw[line width=0.4pt] (10.4*\figu,1.0) -- (15.6*\figu,1.0);
  \foreach \d/\h in {-2/0.12,-1/0.30,0/0.40,1/0.30,2/0.12}{
    \draw[line width=0.9pt] ({(13+\d)*\figu},1.0) -- ({(13+\d)*\figu},1.0+\h);
    \fill ({(13+\d)*\figu},1.0+\h) circle (0.032);}
  \node[anchor=east] at (10.3*\figu,1.22) {$\kappa$};
  \draw[brc] (0.02,-0.42) -- (10.5*\figu-0.03,-0.42) node[midway,below=2pt] {interior};
  \draw[brc] (10.5*\figu+0.03,-0.42) -- (15.5*\figu-0.03,-0.42) node[midway,below=2pt] {band};
  \draw[brc] (15.5*\figu+0.03,-0.42) -- (6.22,-0.42) node[midway,below=2pt] {dead zone};
\end{scope}
\begin{scope}[xshift=7.2cm]
  \node[anchor=west,font=\small] at (-0.30,3.15) {(b)};
  \node at (1.35,2.95) {$\NB$};
  \coordinate (o) at (0.2,1.4);
  \foreach \c in {2.24,1.4,0.56}{
    \draw[figGrey,line width=0.6pt] (o) -- (0.75,\c) -- (1.3,\c);
    \fill[figGrey] (0.75,\c) circle (0.045);
    \fill[figGrey] (1.3,\c) circle (0.045);}
  \foreach \c/\a/\b in {2.24/2.45/2.03,1.4/1.61/1.19,0.56/0.77/0.35}{
    \draw[figGrey,line width=0.6pt] (1.3,\c) -- (1.85,\a) -- (2.4,\a);
    \draw[figGrey,line width=0.6pt] (1.3,\c) -- (1.85,\b) -- (2.4,\b);
    \fill[figGrey] (1.85,\a) circle (0.045);
    \fill[figGrey] (1.85,\b) circle (0.045);}
  \fill[figText] (o) circle (0.05);
  \node[anchor=east] at (0.14,1.4) {$o$};
  \draw[dashed,line width=0.5pt,rounded corners=3pt] (2.24,0.17) rectangle (2.56,2.63);
  \node[anchor=north] at (2.4,0.17) {$S_t$};
  \foreach \y in {2.45,1.61,1.19,0.35}{
    \filldraw[figNBfill,draw=figNB,line width=0.5pt] (2.62,\y-0.1) rectangle (3.1,\y+0.1);
    \filldraw[figNB] (2.4,\y) circle (0.07);}
  \foreach \y in {2.03,0.77}{
    \draw[figNB,line width=0.5pt,pattern=north east lines,pattern color=figNB] (2.62,\y-0.1) rectangle (3.1,\y+0.1);
    \filldraw[fill=white,draw=figNB,line width=0.9pt] (2.4,\y) circle (0.07);}
  \node[anchor=south] at (2.86,2.6) {$D_t$};
  \node at (5.1,2.95) {$\LRW$};
  \def\figxo{3.75}
  \def\figdx{0.27}
  \draw[line width=0.5pt] (\figxo,0.55) -- (\figxo+10*\figdx,0.55);
  \draw[line width=0.5pt,dotted] (\figxo-0.25,0.55) -- (\figxo,0.55);
  \draw[line width=0.5pt,dotted] (\figxo+10*\figdx,0.55) -- (\figxo+10*\figdx+0.25,0.55);
  \foreach \j in {2,...,8}{
    \filldraw[figLRWfill,draw=figLRW,line width=0.5pt]
      ({\figxo+\j*\figdx-0.075},0.55) rectangle ({\figxo+\j*\figdx+0.075},{0.55+1.35*sin(18*\j)});}
  \foreach \j in {1,9}{
    \filldraw[figLRW,draw=figLRW,line width=0.5pt]
      ({\figxo+\j*\figdx-0.075},0.55) rectangle ({\figxo+\j*\figdx+0.075},{0.55+1.35*sin(18*\j)});}
  \foreach \j in {0,...,10}{\fill ({\figxo+\j*\figdx},0.55) circle (0.03);}
  \node at ({\figxo+\figdx},1.47) {$\nu$};
  \draw[brcup] ({\figxo+\figdx-0.08},2.02) -- ({\figxo+9*\figdx+0.08},2.02)
    node[midway,above=2pt] {$U$};
  \node (din) at ({\figxo+5*\figdx},0.08) {$\partial_{\mathrm{in}}U$};
  \draw[line width=0.4pt] (din.west) -- ({\figxo+\figdx},0.49);
  \draw[line width=0.4pt] (din.east) -- ({\figxo+9*\figdx},0.49);
\end{scope}
\end{tikzpicture}
\caption{\textbf{The stalling prover alternates holding and forgetting, and $\NB$ trains on the outer sphere.} (a) Pass rate against level $r$ in one cycle. (b) $D_t$ of $\NB$ on $S_t$, with
hollow vertices outside $K_t$, and the QSD $\nu$ of $\LRW$ on $U$.}
\label{fig:overview}
\end{figure}

\paragraph{A fixed window does not suffice.} The rule that proposes uniformly among the vertices of smallest pass rate at least $p/2$ in its view belongs to $\mathcal P_0(k)$ and makes the
inner boundary of every window absorbing, so holding a window is admissible at no level $\epsilon<1$ (Remark~\ref{rem:un-static}).
The prover must instead remove the mass that reaches the edge of the knowledge and return the rest to views in which this edge is
not visible.

\paragraph{The alternating construction.} The prover repeats cycles of one low round followed by $m$ high rounds and alternates
two nested ball profiles (Figure~\ref{fig:overview}(a)). In a low round it knows $B(o,A)$ with $A=\lfloor3L/4\rfloor$ and assigns
pass rate $p/2$ to $S_{A+1}$, so the filter rejects every proposal beyond $S_{A+1}$; the training error of the round is the mass
beyond $S_A$. In a high round it knows the larger ball $B(o,A+1+mk)$, which no mass can leave within the cycle. Every vertex that
carries mass in a high round then sees the constant pass rate $p$ throughout its view, and the rule acts by one fixed kernel of the
local graph (Lemma~\ref{lem:un-blindfold}). The low round removes the mass far beyond $S_A$ and part of the mass in a band of
width $O(k)$ around $S_A$. The kernel of one cycle restricted to kept proposals is the cycle operator.

\paragraph{The corridor law.} At a vertex of an arm whose view of radius $k$ is a path on which every pass rate equals $p$, view
equivariance and the reflection of the path force the curriculum to step by a symmetric law $\kappa$ on $\{-k,\dots,k\}$, the same at
every such vertex, with variance $\sigma^2\le k^2$ (Lemma~\ref{lem:un-corridor}). We call $\kappa$ the corridor law. At these
vertices the curriculum therefore cannot tell outward from inward. If $\sigma^2=0$, a constant ball profile stalls it from $\delta_o$ at every training error
(Theorem~\ref{thm:un-B0}). Otherwise the number $m$ of high rounds is chosen of order $k^2/\sigma^2$, so that one cycle moves the
mass along an arm like a symmetric walk with total variance of order $k^2$ (Lemma~\ref{lem:un-arith}).

\paragraph{The quasi-stationary start.} Take a radial function that is constant near $o$ and equals a cosine on the scale $L$
along the arms. The constant part needs no property of the rule, and on the arms the symmetry of $\kappa$ cancels the first-order
terms, so the function is a subsolution for both kinds of rounds (Lemma~\ref{lem:un-star}). The
Perron root of the cycle operator is therefore at least $1-O(k^2/L^2)$. By the Paley--Zygmund inequality, band mass that survives a
low round reaches the region beyond the band before the next low round with probability bounded below (Lemma~\ref{lem:un-band}).
Together the two facts bound the training error of the low rounds by $O(k^2/L^2)$ from the QSD of the cycle (Theorem~\ref{thm:un-A}).

\paragraph{From the fixed start.} For rules whose level kernels are totally positive of order two, which include every lazy rule in
$\mathcal P_0(1)$, the monotone comparison of Section~\ref{sec:overview-window} applies (Theorem~\ref{thm:un-B}). For a general rule,
a weak Harnack inequality from the start keeps a fixed fraction of the conditioned mass deep inside $B(o,A)$ in every cycle
(Theorem~\ref{thm:fs-WH}), which reduces the training error to the probability that the unconditioned chain of the cycle operator is
in the band or beyond after a number of cycles of order $L^2/k^2$ (Proposition~\ref{prop:fs-RED}). A renewal decomposition at the
visits of this chain to the band reduces this probability to first-passage bounds for the walk with step law $\kappa$
(Proposition~\ref{prop:fs-REN}), which follow from Freedman's inequality \citep{freedman1975}, gambler's-ruin estimates and a
coupling near $o$ (Proposition~\ref{prop:fs-FP}). This gives Theorem~\ref{thm:uniform} (Theorem~\ref{thm:fs-main}). For batch
selection the low-round profile takes the value $s^\ast$ on $S_{A+1}$ and $s^\ast-p/2$ on $S_{A+2}$, and the same argument gives
Proposition~\ref{prop:boundary}(a) (Proposition~\ref{prop:un-sel}).

\paragraph{Conservative provers.} The conservative alternating construction keeps the low rounds of the alternating construction.
Each high round adds only the current training support to the knowledge and gives the adjacent vertices pass rate $p/2$, the
least value that condition~(i) allows, so the known ball grows with the mass (Definition~\ref{def:h-HA}). For $k=1$, band mass that survives a low round still passes beyond the band before the next low
round with probability bounded below (Lemma~\ref{lem:h-band}), which gives Theorem~\ref{thm:h-P01}.

\subsection{One stored parent}\label{sec:overview-nb}

Theorem~\ref{thm:chenli} does not cover the chain of $\NB$ on ordered pairs, so we track the largest point mass. Under $\NB$
every proposal goes to a child of the current vertex and is kept when the current vertex is known, so each vertex of $S_t$ receives mass only from its parent and the
normaliser of \eqref{eq:main-update} is at least $1-\epsilon$ (Lemma~\ref{lem:mem-ST-radial}; Figure~\ref{fig:overview}(b)). The
largest point mass $M_t$ therefore satisfies $M_{t+1}\le M_t/(b_t(1-\epsilon))$ with $b_t$ children per vertex of $S_t$, and the
count $|K_t|\ge(1-\epsilon)/M_t$, an $\ell^\infty$ analogue of the counting bound in Lemma~E.1 of \citet{chenli2026}, gives the
growth bound of Theorem~\ref{thm:nb} (Theorem~\ref{thm:M-tree}). For $\epsilon>1-2^{-1/L}$ the stalling prover keeps the training
distribution uniform on $n_t$ vertices of $S_t$ and forgets at most $\lfloor\epsilon n_t\rfloor$ of them in each round. The tree branches
once every $L$ levels, so over $L$ rounds the frontier is multiplied by about $2(1-\epsilon)^L<1$ and stays bounded
(Theorem~\ref{thm:M-tree-stall}), which gives the exact tolerance (Corollary~\ref{cor:M-tree-threshold}).

\section{Conclusion}\label{sec:conclusion}

We asked whether a self-play loop without external data can keep growing on its own pass-rate signals when the prover may
forget, and how much memory its curriculum needs. On subdivided trees and for each fixed $k$, every memoryless, local,
time-homogeneous and view-equivariant curriculum tolerates training error at most of the order of Chen and Li's spectral condition,
while one stored parent per sample raises the tolerance to exactly $1-2^{-1/L}$, and both orders are attained. On these theorem
graphs the information that separates the two regimes is the direction away from the start, and one stored parent provides it at
the per-proposal oracle cost of the lazy random walk.

Three mathematical questions follow. The first is whether any curriculum with bounded per-proposal oracle cost tolerates more than
order $L^{-1}$ on $T_3^{(L)}$ without replay; $\NB$ reaches this order with the exact value $1-2^{-1/L}$. The second is whether Theorem~\ref{thm:uniform} extends to
curricula that satisfy (M), (L) and (E) and may change their rule between rounds, for instance through a round-by-round bound on
the training error in place of the fixed Perron root of the cycle operator. The third is whether
$\sup_{\mathcal C\in\mathcal P_0(k)}\epsilon^*_{\mathrm{bdd}}(\mathcal C;G,o)$ is of the order of the spectral gap $\lambda_0$ of the
simple random walk on every graph of bounded degree, as it is on $T_3^{(L)}$ (Theorems~\ref{thm:rw} and~\ref{thm:uniform} and Corollary~\ref{cor:st-subdivided}).

\bibliography{references}

\begin{thebibliography}{43}
\providecommand{\natexlab}[1]{#1}
\providecommand{\url}[1]{\texttt{#1}}
\expandafter\ifx\csname urlstyle\endcsname\relax
  \providecommand{\doi}[1]{doi: #1}\else
  \providecommand{\doi}{doi: \begingroup \urlstyle{rm}\Url}\fi

\bibitem[Alon et~al.(2007)Alon, Benjamini, Lubetzky, and
  Sodin]{alon2007nonbacktracking}
Noga Alon, Itai Benjamini, Eyal Lubetzky, and Sasha Sodin.
\newblock Non-backtracking random walks mix faster.
\newblock \emph{Communications in Contemporary Mathematics}, 9\penalty0
  (4):\penalty0 585--603, 2007.
\newblock \doi{10.1142/S0219199707002551}.

\bibitem[Avestimehr et~al.(2026)Avestimehr, Duffy, and
  M{\'e}dard]{avestimehr2026nova}
Salman Avestimehr, Ken Duffy, and Muriel M{\'e}dard.
\newblock {NOVA}: Fundamental limits of knowledge discovery through {AI}.
\newblock arXiv preprint arXiv:2605.15219v3, 2026.

\bibitem[Bailey et~al.(2026)Bailey, Wen, Dong, Hashimoto, and
  Ma]{bailey2026sgs}
Luke Bailey, Kaiyue Wen, Kefan Dong, Tatsunori Hashimoto, and Tengyu Ma.
\newblock Scaling self-play with self-guidance.
\newblock In \emph{Third Conference on Language Modeling}, 2026.
\newblock URL \url{https://openreview.net/forum?id=ehmYi95IV7}.

\bibitem[Budach(1978)]{budach1978}
Lothar Budach.
\newblock Automata and labyrinths.
\newblock \emph{Mathematische Nachrichten}, 86\penalty0 (1):\penalty0 195--282,
  1978.
\newblock \doi{10.1002/mana.19780860120}.

\bibitem[Chen and Li(2026)]{chenli2026}
Thomas Chen and Zhiyuan Li.
\newblock A theoretical framework for self-play theorem proving algorithms.
\newblock arXiv preprint arXiv:2606.01861, 2026.

\bibitem[Collet et~al.(2013)Collet, Mart{\'\i}nez, and
  San~Mart{\'\i}n]{collet2013}
Pierre Collet, Servet Mart{\'\i}nez, and Jaime San~Mart{\'\i}n.
\newblock \emph{Quasi-Stationary Distributions: {M}arkov Chains, Diffusions and
  Dynamical Systems}.
\newblock Probability and Its Applications. Springer, Berlin, Heidelberg, 2013.
\newblock \doi{10.1007/978-3-642-33131-2}.

\bibitem[Coulhon(1996)]{coulhon1996}
Thierry Coulhon.
\newblock Ultracontractivity and {N}ash type inequalities.
\newblock \emph{Journal of Functional Analysis}, 141\penalty0 (2):\penalty0
  510--539, 1996.
\newblock \doi{10.1006/jfan.1996.0140}.

\bibitem[Coulhon and Grigor'yan(1997)]{coulhon1997}
Thierry Coulhon and Alexander Grigor'yan.
\newblock On-diagonal lower bounds for heat kernels and {M}arkov chains.
\newblock \emph{Duke Mathematical Journal}, 89\penalty0 (1):\penalty0 133--199,
  1997.
\newblock \doi{10.1215/S0012-7094-97-08908-0}.

\bibitem[Darroch and Seneta(1965)]{darroch1965}
J.~N. Darroch and E.~Seneta.
\newblock On quasi-stationary distributions in absorbing discrete-time finite
  {M}arkov chains.
\newblock \emph{Journal of Applied Probability}, 2\penalty0 (1):\penalty0
  88--100, 1965.
\newblock \doi{10.2307/3211876}.

\bibitem[Diaz-Bone et~al.(2025)Diaz-Bone, Bagatella, H{\"u}botter, and
  Krause]{diazbone2025discover}
Leander Diaz-Bone, Marco Bagatella, Jonas H{\"u}botter, and Andreas Krause.
\newblock {DISCOVER}: Automated curricula for sparse-reward reinforcement
  learning.
\newblock In \emph{Advances in Neural Information Processing Systems},
  volume~38, pages 24783--24816. Neural Information Processing Systems
  Foundation, Inc., 2025.
\newblock \doi{10.52202/085713-0738}.

\bibitem[Dodziuk(1984)]{dodziuk1984}
Jozef Dodziuk.
\newblock Difference equations, isoperimetric inequality and transience of
  certain random walks.
\newblock \emph{Transactions of the American Mathematical Society},
  284\penalty0 (2):\penalty0 787--794, 1984.
\newblock \doi{10.1090/S0002-9947-1984-0743744-X}.

\bibitem[Dodziuk and Kendall(1986)]{dodziuk1986kendall}
Jozef Dodziuk and Wilfrid~S. Kendall.
\newblock Combinatorial {L}aplacians and isoperimetric inequality.
\newblock In K.~David Elworthy, editor, \emph{From Local Times to Global
  Geometry, Control and Physics}, volume 150 of \emph{Pitman Research Notes in
  Mathematics Series}, pages 68--74. Longman Scientific \& Technical, Harlow,
  1986.

\bibitem[Dong and Ma(2025)]{dong2025stp}
Kefan Dong and Tengyu Ma.
\newblock {STP}: Self-play {LLM} theorem provers with iterative conjecturing
  and proving.
\newblock In \emph{Proceedings of the 42nd International Conference on Machine
  Learning}, volume 267 of \emph{Proceedings of Machine Learning Research},
  pages 14114--14136. PMLR, 2025.

\bibitem[Fill(2009)]{fill2009}
James~Allen Fill.
\newblock The passage time distribution for a birth-and-death chain: Strong
  stationary duality gives a first stochastic proof.
\newblock \emph{Journal of Theoretical Probability}, 22\penalty0 (3):\penalty0
  543--557, 2009.
\newblock \doi{10.1007/s10959-009-0235-5}.

\bibitem[Fraigniaud et~al.(2005)Fraigniaud, Ilcinkas, Peer, Pelc, and
  Peleg]{fraigniaud2005}
Pierre Fraigniaud, David Ilcinkas, Guy Peer, Andrzej Pelc, and David Peleg.
\newblock Graph exploration by a finite automaton.
\newblock \emph{Theoretical Computer Science}, 345\penalty0 (2--3):\penalty0
  331--344, 2005.
\newblock \doi{10.1016/j.tcs.2005.07.014}.

\bibitem[Freedman(1975)]{freedman1975}
David~A. Freedman.
\newblock On tail probabilities for martingales.
\newblock \emph{The Annals of Probability}, 3\penalty0 (1):\penalty0 100--118,
  1975.
\newblock \doi{10.1214/aop/1176996452}.

\bibitem[Ganesh et~al.(2005)Ganesh, Massouli{\'e}, and Towsley]{ganesh2005}
A.~Ganesh, L.~Massouli{\'e}, and D.~Towsley.
\newblock The effect of network topology on the spread of epidemics.
\newblock In \emph{Proceedings IEEE 24th Annual Joint Conference of the IEEE
  Computer and Communications Societies ({INFOCOM} 2005)}, volume~2, pages
  1455--1466. IEEE, 2005.
\newblock \doi{10.1109/INFCOM.2005.1498374}.

\bibitem[Goel et~al.(2006)Goel, Montenegro, and Tetali]{goel2006}
Sharad Goel, Ravi Montenegro, and Prasad Tetali.
\newblock Mixing time bounds via the spectral profile.
\newblock \emph{Electronic Journal of Probability}, 11:\penalty0 1--26, 2006.
\newblock \doi{10.1214/EJP.v11-300}.

\bibitem[Grigor'yan(1994)]{grigoryan1994}
Alexander Grigor'yan.
\newblock Heat kernel upper bounds on a complete non-compact manifold.
\newblock \emph{Revista Matem{\'a}tica Iberoamericana}, 10\penalty0
  (2):\penalty0 395--452, 1994.
\newblock \doi{10.4171/RMI/157}.

\bibitem[Hoeffding(1963)]{hoeffding1963}
Wassily Hoeffding.
\newblock Probability inequalities for sums of bounded random variables.
\newblock \emph{Journal of the American Statistical Association}, 58\penalty0
  (301):\penalty0 13--30, 1963.
\newblock \doi{10.1080/01621459.1963.10500830}.

\bibitem[Horn and Johnson(2013)]{hornjohnson2013}
Roger~A. Horn and Charles~R. Johnson.
\newblock \emph{Matrix Analysis}.
\newblock Cambridge University Press, Cambridge, second edition, 2013.
\newblock \doi{10.1017/CBO9781139020411}.

\bibitem[Huang et~al.(2026)Huang, Yu, Wang, Zhang, Li, Li, Huang, Mi, and
  Yu]{huang2026rzero}
Chengsong Huang, Wenhao Yu, Xiaoyang Wang, Hongming Zhang, Zongxia Li, Ruosen
  Li, Jiaxin Huang, Haitao Mi, and Dong Yu.
\newblock {R-Zero}: Self-evolving reasoning {LLM} from zero data.
\newblock In \emph{The Fourteenth International Conference on Learning
  Representations}, 2026.
\newblock URL \url{https://openreview.net/forum?id=96apU6YzSO}.

\bibitem[Karlin(1968)]{karlin1968}
Samuel Karlin.
\newblock \emph{Total Positivity}, volume~I.
\newblock Stanford University Press, Stanford, CA, 1968.

\bibitem[Keilson(1979)]{keilson1979}
Julian Keilson.
\newblock \emph{{M}arkov Chain Models---Rarity and Exponentiality}, volume~28
  of \emph{Applied Mathematical Sciences}.
\newblock Springer, New York, 1979.
\newblock \doi{10.1007/978-1-4612-6200-8}.

\bibitem[Kierstead and Slobodkin(1953)]{kierstead1953}
Henry Kierstead and L.~B. Slobodkin.
\newblock The size of water masses containing plankton blooms.
\newblock \emph{Journal of Marine Research}, 12\penalty0 (1):\penalty0
  141--147, 1953.

\bibitem[Kim et~al.(2025)Kim, Wu, Lee, and Suzuki]{kim2025metastable}
Juno Kim, Denny Wu, Jason~D. Lee, and Taiji Suzuki.
\newblock Metastable dynamics of chain-of-thought reasoning: Provable benefits
  of search, {RL} and distillation.
\newblock In \emph{Proceedings of the 42nd International Conference on Machine
  Learning}, volume 267 of \emph{Proceedings of Machine Learning Research},
  pages 30791--30825. PMLR, 2025.

\bibitem[Koenig and Simmons(1996)]{koenig1996}
Sven Koenig and Reid~G. Simmons.
\newblock The effect of representation and knowledge on goal-directed
  exploration with reinforcement-learning algorithms.
\newblock \emph{Machine Learning}, 22\penalty0 (1--3):\penalty0 227--250, 1996.
\newblock \doi{10.1007/BF00114729}.

\bibitem[Kreyszig(1978)]{kreyszig1978}
Erwin Kreyszig.
\newblock \emph{Introductory Functional Analysis with Applications}.
\newblock John Wiley \& Sons, New York, 1978.

\bibitem[Loomis and Whitney(1949)]{loomis1949}
L.~H. Loomis and H.~Whitney.
\newblock An inequality related to the isoperimetric inequality.
\newblock \emph{Bulletin of the American Mathematical Society}, 55\penalty0
  (10):\penalty0 961--962, 1949.
\newblock \doi{10.1090/S0002-9904-1949-09320-5}.

\bibitem[M{\'e}l{\'e}ard and Villemonais(2012)]{meleard2012}
Sylvie M{\'e}l{\'e}ard and Denis Villemonais.
\newblock Quasi-stationary distributions and population processes.
\newblock \emph{Probability Surveys}, 9:\penalty0 340--410, 2012.
\newblock \doi{10.1214/11-PS191}.

\bibitem[Mohar(1988)]{mohar1988}
Bojan Mohar.
\newblock Isoperimetric inequalities, growth, and the spectrum of graphs.
\newblock \emph{Linear Algebra and its Applications}, 103:\penalty0 119--131,
  1988.
\newblock \doi{10.1016/0024-3795(88)90224-8}.

\bibitem[Petrov(1975)]{petrov1975}
Valentin~V. Petrov.
\newblock \emph{Sums of Independent Random Variables}, volume~82 of
  \emph{Ergebnisse der Mathematik und ihrer Grenzgebiete}.
\newblock Springer, Berlin, Heidelberg, 1975.
\newblock \doi{10.1007/978-3-642-65809-9}.
\newblock Translated from the Russian by A. A. Brown.

\bibitem[Pu et~al.(2026)Pu, Sun, Liu, Wu, Barsoum, Liu, and
  Wang]{pu2026compassplay}
Sophia~Xiao Pu, Ximeng Sun, Jiang Liu, Jialian Wu, Emad Barsoum, Zicheng Liu,
  and William~Yang Wang.
\newblock {CompassPlay}: Rewarding the proposer for where it moves the solver.
\newblock arXiv preprint arXiv:2609.32228, 2026.

\bibitem[Rajaraman et~al.(2026{\natexlab{a}})Rajaraman, Huang, Dudik, Schapire,
  Foster, and Krishnamurthy]{rajaraman2026curriculum}
Nived Rajaraman, Audrey Huang, Miro Dudik, Rob Schapire, Dylan Foster, and
  Akshay Krishnamurthy.
\newblock Learning to reason with curriculum {I}: Provable benefits of
  autocurriculum.
\newblock In \emph{Proceedings of Thirty Ninth Conference on Learning Theory},
  volume 336 of \emph{Proceedings of Machine Learning Research}, pages
  5518--5555. PMLR, 2026{\natexlab{a}}.

\bibitem[Rajaraman et~al.(2026{\natexlab{b}})Rajaraman, Huang, Dud{\'i}k,
  Schapire, Foster, and Krishnamurthy]{rajaraman2026curriculumii}
Nived Rajaraman, Audrey Huang, Miroslav Dud{\'i}k, Robert Schapire, Dylan
  Foster, and Akshay Krishnamurthy.
\newblock Learning to reason with curriculum {II}: Compositional
  generalization.
\newblock arXiv preprint arXiv:2606.27721, 2026{\natexlab{b}}.

\bibitem[Singh et~al.(1994)Singh, Jaakkola, and Jordan]{singh1994}
Satinder~P. Singh, Tommi Jaakkola, and Michael~I. Jordan.
\newblock Learning without state-estimation in partially observable {M}arkovian
  decision processes.
\newblock In \emph{Machine Learning Proceedings 1994: Proceedings of the
  Eleventh International Conference}, pages 284--292. Morgan Kaufmann, 1994.
\newblock \doi{10.1016/B978-1-55860-335-6.50042-8}.

\bibitem[Skellam(1951)]{skellam1951}
J.~G. Skellam.
\newblock Random dispersal in theoretical populations.
\newblock \emph{Biometrika}, 38\penalty0 (1/2):\penalty0 196--218, 1951.
\newblock \doi{10.1093/biomet/38.1-2.196}.

\bibitem[Thrun(1992)]{thrun1992}
Sebastian~B. Thrun.
\newblock Efficient exploration in reinforcement learning.
\newblock Technical Report CMU-CS-92-102, School of Computer Science, Carnegie
  Mellon University, Pittsburgh, PA, January 1992.

\bibitem[Whitehead(1991)]{whitehead1991}
Steven~D. Whitehead.
\newblock Complexity and cooperation in {Q}-learning.
\newblock In Lawrence~A. Birnbaum and Gregg~C. Collins, editors, \emph{Machine
  Learning Proceedings 1991: Proceedings of the Eighth International Workshop
  ({ML91})}, pages 363--367. Morgan Kaufmann, 1991.
\newblock \doi{10.1016/B978-1-55860-200-7.50075-1}.

\bibitem[Woess(2000)]{woess2000}
Wolfgang Woess.
\newblock \emph{Random Walks on Infinite Graphs and Groups}, volume 138 of
  \emph{Cambridge Tracts in Mathematics}.
\newblock Cambridge University Press, Cambridge, 2000.
\newblock \doi{10.1017/CBO9780511470967}.

\bibitem[Wu et~al.(2026)Wu, Yu, and Yin]{closedloop2026}
Xuening Wu, Shan Yu, and Shenqin Yin.
\newblock Closed-loop knowledge dynamics: An operational framework for
  saturation and escape.
\newblock arXiv preprint arXiv:2607.14185, 2026.

\bibitem[Yang(2026)]{yang2026ancora}
Chengcao Yang.
\newblock {ANCORA}: Learning to question via manifold-anchored self-play for
  verifiable reasoning.
\newblock arXiv preprint arXiv:2604.27644v2, 2026.

\bibitem[Zhao et~al.(2025)Zhao, Wu, Wu, Xu, Yue, Lin, Wang, Wu, Zheng, and
  Huang]{zhao2025absolute}
Andrew Zhao, Yiran Wu, Tong Wu, Quentin Xu, Yang Yue, Matthieu Lin, Shenzhi
  Wang, Qingyun Wu, Zilong Zheng, and Gao Huang.
\newblock {Absolute Zero}: Reinforced self-play reasoning with zero data.
\newblock In \emph{Advances in Neural Information Processing Systems},
  volume~38, pages 117235--117298. Neural Information Processing Systems
  Foundation, Inc., 2025.
\newblock \doi{10.52202/085713-3534}.

\end{thebibliography}

\appendix
\counterwithin*{theorem}{section}
\renewcommand{\thetheorem}{\thesection.\arabic{theorem}}
\section{Additional related work}\label{app:A_related}

\paragraph{Spectral and isoperimetric thresholds.} \citet{ganesh2005} bound the lifetime of an SIS epidemic on a finite graph from
above through the spectral radius of the adjacency matrix (their Theorem~3.1) and from below through a generalised isoperimetric
constant (their Theorem~4.1). Theorem~\ref{thm:rw}(a) has the same two-sided form for self-play growth, with an adversarial prover
choosing where the training error falls. Its isoperimetric side, $\lambda_0\ge\Phi^2/2$ in Theorem~\ref{thm:chenli}, is the discrete
Cheeger inequality on infinite graphs (\citealp[Theorem~2.3]{dodziuk1984}; \citealp[Theorem~10.3 and its proof, p.~111]{woess2000});
\citet[Notes and Remarks to Section~10, p.~137]{woess2000} credit the equivalence of the strong isoperimetric inequality and
$\rho(P)<1$ to \citet{dodziuk1984} and \citet{dodziuk1986kendall} and cite \citet{mohar1988} for graphs, and
\citet[Theorem~A.14]{chenli2026} write the bound in the notation used here.

\paragraph{Critical patch size.} \citet[Section~3.3]{skellam1951} and \citet{kierstead1953} showed that a population that reproduces and
diffuses in a bounded habitat with a lethal boundary dies out when the habitat is smaller than a critical size, at which the principal
Dirichlet eigenvalue of the diffusion equals the growth rate. The windows of Appendix~\ref{app:C_window} and the plateau bounds of
Appendix~\ref{app:G_plateau} rest on a discrete comparison of the same kind, between the Dirichlet eigenvalue $1-\lambda_U$ of a
window and the training error.

\paragraph{Faber--Krahn functions and spectral profiles.} The plateau bounds of Appendix~\ref{app:G_plateau} control the training
distribution through the Faber--Krahn function $\Lambda$ of the walk, the smallest Dirichlet eigenvalue among finite sets of measure at
most $v$. \citet{grigoryan1994} used Faber--Krahn inequalities in place of Sobolev and Nash inequalities and proved their equivalence with
heat-kernel upper bounds (his Definition~1.1 and Theorem~1.1); the truncation in Lemma~\ref{lem:pl-G} is his Lemma~2.1, which he credits
to Gushchin. \citet{coulhon1996} characterises on-diagonal heat-kernel upper bounds by Nash-type inequalities,
\citet[Section~4]{coulhon1997} work with Dirichlet eigenvalues of finite sets of a graph, and \citet{goel2006} bound mixing times through the
spectral profile with the same techniques. Here the Faber--Krahn function bounds the plateau height of random-walk self-play against a
prover that may forget, whose training distribution is conditioned on survival in a set that the prover chooses in every round.

\paragraph{Quasi-stationary distributions.} A QSD of a killed Markov chain is a law that conditioning on survival leaves unchanged.
\citet{darroch1965} studied QSDs of finite absorbing chains; in the monograph of \citet{collet2013} and the survey of \citet{meleard2012}
(finite case: their Theorem~7) the QSD of a finite irreducible chain is its normalised left Perron eigenvector.
Lemma~\ref{lem:st-survival} identifies the filter of \citet{chenli2026} with survival conditioning of the proposal chain, so a QSD of
the window kernel with positive Perron root is a fixed point of the self-play update; it is the cold start of Theorem~\ref{thm:rw}(a).

\paragraph{Closed-loop saturation.} \citet{closedloop2026} describe a closed self-improvement loop as an iteration on an abstract
knowledge state that saturates under a Lyapunov drift condition (their Theorem~1), with a threshold condition for escape (their
Proposition~3). The present paper works in the theorem-graph model of \citet{chenli2026} and bounds the training error that a local
proposal rule tolerates against a prover that may forget, and the change in this tolerance when one parent is stored per sample.

\section{Model and notation}\label{app:B_model}

This appendix states the model of Section~\ref{sec:model} in the form used by the proofs, once for all later appendices:
the prover and admissibility (Section~\ref{sec:B-prover}), the dynamics and the two tolerances (Section~\ref{sec:B-dynamics}), the
class $\mathcal P_0(k)$ (Section~\ref{sec:B-P0}), curricula with one step of memory and the equal-budget accounting
(Section~\ref{sec:B-memory}), and the spectral notation (Section~\ref{sec:B-spectral}).

\subsection{Theorem graph, prover and admissibility}\label{sec:B-prover}

$G=(V,E)$ is a connected, locally finite graph with countably many vertices (problems), in which an edge joins two problems on which
every prover of the class has pass rates within $p/2$ \citep[Assumption~3]{chenli2026}. We write $\mathrm{dist}$ for the graph distance,
$N(x)$ for the neighbours of $x$, $B(x,k)=\{y:\mathrm{dist}(x,y)\le k\}$, $B(K,1)=K\cup N(K)$ for $K\subseteq V$, and, for
$U\subseteq V$, $\partial_{\mathrm{in}}U=\{x\in U:N(x)\not\subseteq U\}$ and $\mathrm{int}\,U=U\setminus\partial_{\mathrm{in}}U$. A start
vertex $o$ is fixed, $|x|=\mathrm{dist}(o,x)$ is the level of $x$ and $S_r=\{x:|x|=r\}$. On $T_3^{(L)}$ the branch vertices (degree $3$)
are the vertices of the $3$-regular tree $T_3$, every other vertex is a subdivision vertex (degree $2$), a subdivided edge is the path
with $L$ edges that replaces an edge of $T_3$, and $o$ is a branch vertex. On $\mathbb Z^D$, $o=0$.

Fix $p\in(0,1)$ and $\epsilon\in[0,1)$. In round $t=0,1,2,\dots$ the prover is summarised by its pass-rate profile $s_t:V\to[0,1]$,
and its knowledge is $K_t=\{x:s_t(x)\ge p\}$.

\begin{definition}[Admissible prover sequence; Definition~\ref{def:main-admissible}]\label{def:admissible}
Given training distributions $(D_t)_{t\ge0}$, a sequence $(s_t)_{t\ge0}$ is admissible at level $\epsilon$ if for every $t$:
(i) $|s_t(x)-s_t(y)|\le p/2$ for every edge $\{x,y\}\in E$; and (ii) $D_t(K_t)\ge 1-\epsilon$.
\end{definition}

No relation between $s_t$ and $s_{t-1}$ is required, so the prover may forget, and it may use global information. Admissibility is
the worst-case reading of Assumptions~2--3 of \citet{chenli2026}: (ii) is their accuracy requirement on the current training
distribution and (i) their $p/2$-Lipschitz condition (the sample-size part of Assumption~2 is not used). So every prover sequence of a
prover class satisfying Assumption~3 with a training oracle that is accurate in every round is admissible, and the growth results apply
to it; conversely, Proposition~\ref{prop:st-realize} realises the stalls of Appendix~\ref{app:C_window} up to any finite horizon $t$, with
probability at least $1-(t+1)\delta$, by such a class and an oracle satisfying Assumption~2. The cold start $D_0=\delta_o$ fixes the
training distribution at time $0$, not the prover's initial competence.

\subsection{Dynamics and the two tolerances}\label{sec:B-dynamics}

A memoryless curriculum with round-$t$ kernel $C_t$ proposes $x'\sim C_t(x,\cdot)$ from a sample at $x$, and Algorithm~2 of
\citet{chenli2026} keeps $x'$ iff $s_t(x')\ge p/2$. From a cold start $D_0\in\Delta(V)$ the training distributions follow
\begin{equation}\label{eq:update}
D_{t+1}(y)=\frac{\sum_x D_t(x)\,C_t(x,y)\,\mathbf 1[s_t(y)\ge p/2]}{Z_t},\qquad Z_t=\sum_{x,y'} D_t(x)\,C_t(x,y')\,\mathbf 1[s_t(y')\ge p/2],
\end{equation}
and round $t$ is defined if $Z_t>0$; $Z_t$ is the acceptance probability of round $t$. With replay weight $w\in[0,1]$ the frontier $F_t$
follows \eqref{eq:update} from $F_0=D_0$ and the prover trains on $D_t=(1-w)F_t+wR_t$, with $R_0=F_0$ and $R_t=\frac1t\sum_{s<t}F_s$
($t\ge1$); $w=0$ is Chen and Li's dynamics. Every comparison of two curricula uses the same filter, replay, proof sampling and per-proposal
oracle operations on both sides, and counts rounds. Chen and Li's Algorithm~1 in addition assumes that the proofs of the problems in
$\mathrm{supp}\,D_0$ are known (their Assumption~1); only the sample-based statements use this premise.

\paragraph{Defined rounds.} Let $\mu_t$ be the distribution to which \eqref{eq:update} is applied ($D_t$ without replay, $F_t$ with
replay). If $C_t(x,\cdot)$ is supported on $N(x)\cup\{x\}$ (a nearest-neighbour curriculum) and $s_t$ satisfies
Definition~\ref{def:admissible}(i), every proposal from $x\in K_t$ has pass rate at least $s_t(x)-p/2\ge p/2$, so $Z_t\ge\mu_t(K_t)$. Without
replay $\mu_t(K_t)\ge1-\epsilon>0$; with replay weight $w<1$ and $\epsilon<1-w$, $(1-w)F_t(K_t)\ge D_t(K_t)-w\ge1-\epsilon-w>0$. So every round of
such a sequence is defined.

\paragraph{Tolerances.} For a curriculum $\mathcal C$ and a cold start $D_0$ let $\mathcal G$ be the set of $\epsilon\in[0,1)$ at which
every admissible prover sequence has $|K_t|\to\infty$, and $\mathcal B$ the set of $\epsilon\in[0,1)$ at which some admissible prover
sequence has $\sup_t|K_t|<\infty$. Then $\epsilon^*_{\mathrm{div}}=\sup\mathcal G$ and $\epsilon^*_{\mathrm{bdd}}=\inf\mathcal B$
(Definition~\ref{def:main-tolerance}, with $\sup\emptyset=0$, $\inf\emptyset=1$), with any given replay weight. A sequence admissible at
level $\epsilon_1$ is admissible at every $\epsilon_2\in[\epsilon_1,1)$, so $\mathcal G$ is closed downwards, $\mathcal B$ upwards, and
$\mathcal G\cap\mathcal B=\emptyset$; hence $\epsilon^*_{\mathrm{div}}\le\epsilon^*_{\mathrm{bdd}}$, and we write $\epsilon^*$ when they are
equal. Two consequences are used throughout. (Stall.) One admissible sequence at level $\epsilon$ with $\sup_t|K_t|<\infty$ gives
$\epsilon\in\mathcal B$ and so $\epsilon^*_{\mathrm{div}}\le\epsilon^*_{\mathrm{bdd}}\le\epsilon$. (Growth.) If every admissible sequence has
$|K_t|\to\infty$ at every $\epsilon<\epsilon_0$, then $\epsilon_0\le\epsilon^*_{\mathrm{div}}\le\epsilon^*_{\mathrm{bdd}}$.

\subsection{Memoryless local curricula}\label{sec:B-P0}

The radius-$k$ labelled rooted view of $x$ under a profile $s$ is $(B(x,k),x,s|_{B(x,k)})$, and $\mathcal V_t^k(x)$ is the view under
$s_t$. An isomorphism of views $(B(x,k),x,s)\to(B(x',k),x',s')$ is a graph isomorphism $\varphi$ of the induced subgraphs with
$\varphi(x)=x'$ and $s'\circ\varphi=s$.

\begin{definition}[$\mathcal P_0(k)$; Definition~\ref{def:main-P0}]\label{def:P0}
$\mathcal C\in\mathcal P_0(k)$ assigns to every labelled rooted view $v$ of radius $k$ a probability distribution $\mathcal C_v$ on the
vertices of $v$ such that (M) a sample is a single vertex and the proposal from $x$ in round $t$ is drawn from
$\mathcal C_{\mathcal V^k_t(x)}$; (L) proposals lie in $B(x,k)$; (T) the rule does not depend on $t$; (E) if $\varphi:v\to v'$ is an
isomorphism of views, then $\mathcal C_{v'}(\varphi(A))=\mathcal C_v(A)$ for every $A$.
\end{definition}

The induced kernel of $\mathcal C$ under a profile $s$ is $\mathcal C^s(x,\cdot):=\mathcal C_{(B(x,k),x,s|_{B(x,k)})}$; by (M) and (T) the
round-$t$ kernel is $C_t=\mathcal C^{s_t}$, so $C_t=\mathcal C^s$ for every $t$ when $s_t=s$ for every $t$. A success-blind curriculum
\citep[Definition~4.1]{chenli2026} with kernel $P$ has $\mathcal C^s=P$ for every $s$; statements made for a curriculum $\mathcal C$
include it. A curriculum is nearest-neighbour if every $\mathcal C^s$ is. Under (E), information from outside the view (the distance to
$o$, a depth label, a parent pointer, a proof library or a task buffer) does not enter, while the observed pass rates may orient the view.

\paragraph{Self-play systems.} Every rule $\mathcal C_v(y)\propto q_v(y)\,f(s(y))$ with an equivariant base proposal $q_v$, a weight $f$
and an equivariant fallback when the normaliser vanishes is in $\mathcal P_0(k)$. So $\mathcal P_0(k)$ contains abstractions of the
local scoring rules of STP, which trains on conjectures with empirical pass rate in $(0,1/4]$ \citep[Section~3.2]{dong2025stp}, of
R-Zero, which rewards uncertainty \citep[Section~2.2]{huang2026rzero}, and of Absolute Zero, which rewards learnability
\citep[Eq.~(4)]{zhao2025absolute}, with the reward written as a function $f$ of the pass rate and the generation of new tasks replaced by
the local base proposal $q_v$; seeds, buffers, batch-level filters and penalties and the learned proposers are outside the abstraction.
The class also contains the local form of Chen and Li's diffusion conjecturer (their Algorithm~4, eq.~(5.8), with the simple-random-walk
baseline and the unnormalised vertex weights $\deg(z)$ in their matrix $M'(x,\tau)$, which is well defined on locally finite graphs). For
diffusion time $\tau$ its proposal law at $x$ is computed from the degrees of the vertices within distance $\tau+1$ of $x$, which the
induced subgraph on $B(x,\tau+2)$ determines. Its feasible set $\{\mu\in\Delta(N(x)):\mu(y)\ge\eta/\deg(x)\ \forall y\}$, $\eta\in[0,1/2]$,
is nonempty, compact and convex and its objective is a continuous convex quadratic form, so the minimisers form a nonempty compact convex
set with a unique element of least Euclidean norm. View isomorphisms permute coordinates and preserve the feasible set and the objective,
so this choice is equivariant and the rule lies in $\mathcal P_0(\tau+2)$.

\subsection{One step of memory and the equal-budget comparison}\label{sec:B-memory}\label{sec:mem-budget}

\begin{definition}[Dynamics with one step of memory; Definition~\ref{def:main-memory}]\label{def:mem-dynamics}
A curriculum with one step of memory is a kernel $\mathcal M$ that assigns to every pair $(x_-,x)$ with $x_-\in N(x)\cup\{\bot\}$ a
probability distribution on $N(x)$. Its state in round $t$ is a distribution $\Pi_t$ on such pairs, $\Pi_0=\delta_{(\bot,o)}$, and
\begin{equation}\label{eq:mem-update}
\Pi_{t+1}\big((x,x')\big)=\frac{1}{Z_t}\sum_{x_-}\Pi_t\big((x_-,x)\big)\,\mathcal M\big((x_-,x),x'\big)\,\mathbf 1[s_t(x')\ge p/2],
\end{equation}
with $Z_t$ the sum of the numerator over all pairs. The frontier $F_t$ is the second marginal of $\Pi_t$, and $D_t=(1-w)F_t+wR_t$ as in
Section~\ref{sec:B-dynamics}. $\NB$ has $\mathcal M((x_-,x),\cdot)$ uniform on $N(x)\setminus\{x_-\}$ on trees of minimum degree at
least $2$, and uniform on $N(x)$ when $x_-=\bot$.
\end{definition}

Equation~\eqref{eq:mem-update} is Algorithm~2 applied to stored pairs: $(x_-,x)$ proposes $x'\sim\mathcal M((x_-,x),\cdot)$, which is kept
iff $s_t(x')\ge p/2$ and stored as $(x,x')$.

\begin{definition}[Primitive operations and equal budget]\label{def:mem-budget}
A system is a curriculum (memoryless or with one step of memory) with the filter of Algorithm~2 and a replay weight $w$. In round $t$ it
performs (B1) one call of the training oracle on $n$ i.i.d.\ samples of $D_t$ \citep[Assumption~2]{chenli2026}, with the same $n$ for all
systems; (B2) Algorithm~2: draw a stored frontier sample, draw a proposal $x'$, query $s_t(x')$, repeat until $s_t(x')\ge p/2$ (each
repetition is one proposal), then draw proofs of $x'$ from the prover and verify them until one is accepted; (B3) storage of the frontier
samples (a vertex, or a pair $(x_-,x)$) and of the replay buffer. The per-proposal cost is the pair of the maximal numbers, over proposals,
of neighbour-oracle calls and of pass-rate queries. Given $x'$, each proof draw succeeds independently with probability
$s_t(x')\ge p/2$, so the number of proof draws is geometric with conditional mean $1/s_t(x')\le2/p$, with the same conditional law for
every system. Two systems run at equal budget if they share $G$, $o$, $p$, $\epsilon$, $n$, the filter, the proof sampling and $w$, and
the per-proposal cost of the first is coordinatewise at most that of the second.
\end{definition}

\begin{lemma}[Per-proposal cost]\label{lem:mem-budget}
\begin{enumerate}
\item[(a)] $\NB$ on a tree of minimum degree at least $2$ uses at most one neighbour-oracle call and exactly one
pass-rate query per proposal, and stores one vertex identifier more than a memoryless sample; with stable reverse-port information, $x_-$
is an index into $N(x)\cup\{\bot\}$, that is, $2$ bits on $T_3^{(L)}$.
\item[(b)] The simple random walk and $\LRW$, both in $\mathcal P_0(1)$, use the same per-proposal operations and store one vertex.
\item[(c)] Every system makes at least one pass-rate query per proposal. A rule in $\mathcal P_0(k)$ can be evaluated after at most
$|B(x,k)|$ pass-rate queries, one of which serves as the filter query; membership in $\mathcal P_0(k)$ forces no label query ($\LRW$ reads none).
\item[(d)] For a nearest-neighbour curriculum, memoryless or $\NB$, with $w\in[0,1)$ and $\epsilon<1-w$, every admissible prover sequence has
$Z_t\ge1-\epsilon/(1-w)>0$ in every round, so a call of Algorithm~2 makes a geometric number of proposals with mean
$1/Z_t\le(1-w)/(1-w-\epsilon)$, followed by proof draws with mean at most $2/p$.
\end{enumerate}
Consequently $\NB$ and $\LRW$ run at equal budget in both directions, and $\NB$ runs at equal budget with every system whose curriculum
makes a neighbour-oracle call in at least one proposal.
\end{lemma}

\begin{proof}
(a) From $(x_-,x)$, $\NB$ proposes a uniform element of $N(x)\setminus\{x_-\}$ (one call for the list $N(x)$); the filter needs one query, and the new pair $(x,x')$ consists of known entries. On $T_3^{(L)}$,
$|N(x)\cup\{\bot\}|\le4$. (b) The simple random walk proposes a uniform neighbour (one call); $\LRW$ proposes
$x$ (no call) or a uniform neighbour. (c) The filter query is part of (B2); a rule in $\mathcal P_0(k)$ is a function of the labels
$s_t(y)$, $y\in B(x,k)$, and $x'\in B(x,k)$ by (L). (d) $F_0(K_0)=D_0(K_0)\ge1-\epsilon$, and for $t\ge1$,
$(1-w)F_t(V\setminus K_t)\le D_t(V\setminus K_t)\le\epsilon$, so $F_t(K_t)\ge1-\epsilon/(1-w)$; the bound $Z_t\ge F_t(K_t)$ is the paragraph
on defined rounds (for $\NB$, the proposal from $(x_-,x)$ lies in $N(x)$). The passes of Algorithm~2 use fresh randomness and each is
accepted with probability $Z_t$.
\end{proof}

\subsection{Kernels and spectral notation}\label{sec:B-spectral}

A kernel on $V$ is a nonnegative matrix with unit row sums; it is nearest-neighbour if $P(x,\cdot)$ is supported on $N(x)\cup\{x\}$, and
then $e_{\min}(P)=\inf_x\min_{y\in N(x)}P(x,y)$. A measure $\pi>0$ is reversible for $P$ if $\pi(x)P(x,y)=\pi(y)P(y,x)$. For reversible $P$
let $a(x,y)=\pi(x)P(x,y)$, $a(\partial A)=\sum_{x\in A,y\notin A}a(x,y)$, $\Phi(P)=\inf\{a(\partial A)/\pi(A):A\text{ finite, nonempty}\}$,
$\langle f,g\rangle_\pi=\sum_x\pi(x)f(x)g(x)$, $\|f\|_\pi^2=\langle f,f\rangle_\pi$, $\|D\|^2_{\ell^2(1/\pi)}=\sum_xD(x)^2/\pi(x)$ and
$\lambda_0=1-\|P\|_{\ell^2(\pi)}$. The lazy version of $P$ is $P^{\mathrm{lazy}}=\frac12(I+P)$. Chen and Li's conjecturer with edge
weights $\omega>0$, $P_\omega(x,y)=\omega(x,y)/\sum_{z\in N(x)}\omega(x,z)$, is reversible with $\pi(x)=\sum_{z\in N(x)}\omega(x,z)$. For the
simple random walk (SRW) $\pi=\deg$, $a(\partial A)$ is the number of edges with one endpoint in $A$, and $e_{\min}=1/\sup_x\deg(x)$; its
lazy version is $\LRW$.

For a substochastic matrix $Q$ on a finite nonempty set, $\lambda(Q)$ is its Perron root, and a QSD of $Q$ is a probability vector
$\nu\ge0$ with $\nu Q=\lambda(Q)\nu$; by the Perron--Frobenius theorem \citep{hornjohnson2013} one exists, and it is unique and positive if
$Q$ is irreducible. For a kernel $P$ and finite nonempty $U$, $P_U$ is the restriction of $P$ to $U\times U$, $\lambda_U=\lambda(P_U)$, and
$\nu_U$ denotes a QSD of $P_U$ (a statement about $\nu_U$ is a statement about every QSD of $P_U$). Finally
$\epsilon_D(P)=\inf\{\nu(\partial_{\mathrm{in}}U):U\text{ finite, nonempty},\ \nu\text{ a QSD of }P_U\}$.

\section{Holding a window: proofs of \texorpdfstring{Theorems~\ref{thm:chenli} and~\ref{thm:rw}}{Theorems 0 and 1}}\label{app:C_window}\label{sec:stall}

\begin{restated}{Theorem~\ref{thm:chenli}}\StmtChenLi\end{restated}

\begin{restated}{Theorem~\ref{thm:rw}}\StmtRandomWalk\end{restated}

\paragraph{Roadmap.} Every stall in this appendix is a prover that holds a fixed window $U$. Section~\ref{sec:st-qsd} shows that the
filter of Chen and Li conditions the proposal chain on survival (Lemma~\ref{lem:st-survival}), so a QSD of the window kernel is a fixed
point of the dynamics and holding $U$ from it is admissible exactly when its inner-boundary mass is at most $\epsilon$
(Theorem~\ref{thm:st-qsd}); Proposition~\ref{prop:st-realize} realises such stalls by systems that satisfy Chen and Li's assumptions.
Section~\ref{sec:st-sandwich} compares the boundary mass with $\lambda_0$ (Corollary~\ref{cor:st-sandwich}), Section~\ref{sec:st-tree}
evaluates both sides on $T_3^{(L)}$ (Corollary~\ref{cor:st-subdivided}), and Section~\ref{sec:st-fixed} passes from the QSD start to
the fixed start $\delta_o$ through a monotone comparison for lazy birth--death chains (Theorem~\ref{thm:st-fixed}), which gives
Theorem~\ref{thm:rw}(b) (Corollary~\ref{cor:st-tree}).

\paragraph{Proof of Theorems~\ref{thm:chenli} and~\ref{thm:rw}.} A reversible nearest-neighbour curriculum kernel $P$ is a
success-blind curriculum, and the hypotheses of both theorems are the standing assumptions of Section~\ref{sec:st-sandwich}. For
Theorem~\ref{thm:chenli}, the display and $\epsilon^*_{\mathrm{div}}(P;G,D_0)\ge\lambda_0$ are Corollary~\ref{cor:st-sandwich}(i), and
$\lambda_0\ge\Phi^2/2$ is Corollary~\ref{cor:st-sandwich}(iv). On $T_3^{(L)}$, $\delta_o$ is finitely supported, so
Corollary~\ref{cor:st-subdivided}(4) gives $\epsilon^*_{\mathrm{div}}(P;T_3^{(L)},o)\ge\Phi(P)^2/2=1/(2(4L-1)^2)$ for the SRW $P$. For
Theorem~\ref{thm:rw}(a), $\lambda_0\le\epsilon_D(P)\le\lambda_0/e_{\min}(P)$ is Corollary~\ref{cor:st-sandwich}(ii); for
$\epsilon\in(\epsilon_D(P),1)$, Corollary~\ref{cor:st-sandwich}(iii) gives a finite $U$ and a QSD $\nu$ of $P_U$ with
$\nu(\partial_{\mathrm{in}}U)<\epsilon$ and $\lambda_U>0$, and since $Q_U=P_U$ for a success-blind curriculum, Theorem~\ref{thm:st-qsd}(a)
shows that from $D_0=\nu$, with any replay weight, holding $U$ is admissible at level $\epsilon$ with $K_t=\mathrm{int}\,U$; the display is
that of Corollary~\ref{cor:st-sandwich}(iii). Theorem~\ref{thm:rw}(b) is Corollary~\ref{cor:st-tree}(b), with $\LRW=P^{\mathrm{lazy}}$.

\subsection{Windows, survival conditioning and the quasi-stationary start}\label{sec:st-qsd}\label{sec:st-conventions}

For a finite $U\subset V$ the window profile is $s_U=p$ on $\mathrm{int}\,U$, $s_U=p/2$ on $\partial_{\mathrm{in}}U$ and $s_U=0$ off $U$.
The prover holds $U$ if $s_t=s_U$ for every $t$. We write $P^U:=\mathcal C^{s_U}$ and $Q_U:=(P^U(x,y))_{x,y\in U}$; these, and the QSDs and
trajectories computed from them, are taken at the fixed threshold $p$ (for a success-blind curriculum $P^U=P$ does not depend on $p$).

\begin{lemma}[Window profile]\label{lem:st-window}
Let $U\subset V$ be finite. (a) No edge joins $\mathrm{int}\,U$ and $V\setminus U$. (b) $s_U$ satisfies
Definition~\ref{def:admissible}(i). (c) $\{s_U\ge p\}=\mathrm{int}\,U$ and $\{s_U\ge p/2\}=U$.
\end{lemma}

\begin{proof}
(a) holds by the definition of $\mathrm{int}\,U$. By (a) every edge lies inside $U$ (values in $\{p/2,p\}$), inside $V\setminus U$
(values $0$), or joins $\partial_{\mathrm{in}}U$ to $V\setminus U$ (values $p/2,0$), which gives (b). (c) holds because $p>p/2>0$.
\end{proof}

While the prover holds $U$, $K_t=\mathrm{int}\,U$ and the filter keeps exactly the proposals in $U$. If $D_0$ is supported in $U$, so is
every $D_t$, and Definition~\ref{def:admissible}(ii) reads
\begin{equation}\label{eq:st-window-adm}
D_t(\partial_{\mathrm{in}}U)\le\epsilon .
\end{equation}
So holding $U$ is admissible at level $\epsilon$ if and only if \eqref{eq:st-window-adm} holds for every $t$, and then
$|K_t|=|\mathrm{int}\,U|$.

\begin{lemma}[Algorithm~2 is survival conditioning]\label{lem:st-survival}
Let $C_t$ be kernels and $s_t$ profiles, $A_t(x,y)=C_t(x,y)\mathbf 1[s_t(y)\ge p/2]$, and $D_{t+1}=D_tA_t/(D_tA_t\mathbf 1)$ as in
\eqref{eq:update}.
\begin{enumerate}
\item[(a)] If $\alpha=DA_t\mathbf 1>0$, Algorithm~2 with input $(D,C_t,s_t,p/2)$ terminates almost surely and returns a point with law
$DA_t/\alpha$; distinct calls are independent.
\item[(b)] Let $(X_t)$ be the Markov chain with $X_0\sim D_0$ and $X_{t+1}\sim C_t(X_t,\cdot)$, and $\tau=\inf\{t\ge1:s_{t-1}(X_t)<p/2\}$.
If $\mathbb P(\tau>t)>0$, then $D_t=D_0A_0\cdots A_{t-1}/(D_0A_0\cdots A_{t-1}\mathbf 1)=\mathbb P(X_t\in\cdot\mid\tau>t)$.
\item[(c)] If $C_t=C$ and $s_t=s$ for every $t$, and $\nu\in\Delta(V)$ satisfies $\nu A=\lambda\nu$ with $\lambda>0$, then $D_0=\nu$ implies
$D_t=\nu$ for every $t$; with replay of any weight, $F_0=\nu$ implies $F_t=R_t=D_t=\nu$ for every $t$.
\item[(d)] If $\mathcal C\in\mathcal P_0(k)$ and $s_t=s$ for every $t$, then $C_t=\mathcal C^s$ for every $t$, and (c) applies with $C=\mathcal C^s$.
\end{enumerate}
\end{lemma}

\begin{proof}
(a) Each pass draws $x\sim D$ and $x'\sim C_t(x,\cdot)$ with fresh randomness, so it returns $y$ with probability $(DA_t)(y)$ and is
accepted with probability $\alpha$; the number of passes is geometric, and $\mathbb P(\text{return }y)=\sum_{j\ge0}(1-\alpha)^j(DA_t)(y)=(DA_t)(y)/\alpha$.
(b) With $\mu_t=D_0A_0\cdots A_{t-1}$, induction gives $D_t=\mu_t/(\mu_t\mathbf 1)$, since $D_tA_t=\mu_{t+1}/(\mu_t\mathbf 1)$. Summing over
paths, $\mathbb P(X_t=y,\tau>t)=\sum_{x_0,\dots,x_{t-1}}D_0(x_0)\prod_{r<t}A_r(x_r,x_{r+1})|_{x_t=y}=\mu_t(y)$ and $\mathbb P(\tau>t)=\mu_t\mathbf 1$;
the latter is positive exactly when the normalisers of \eqref{eq:update} up to round $t$ are.
(c) $\mu_t=\lambda^t\nu$. With replay the frontier follows the same update, so $F_t=\nu$, $R_t=\nu$ and $D_t=\nu$.
(d) holds by (M) and (T).
\end{proof}

So a QSD of the accepted kernel is invariant under \eqref{eq:update}, the defining property of a QSD of a killed chain \citep{collet2013}.

\begin{theorem}[Stall from a quasi-stationary start]\label{thm:st-qsd}
Let $\mathcal C$ be a curriculum (success-blind with kernel $P$, or in $\mathcal P_0(k)$), $U\subset V$ finite, $\nu$ a QSD of $Q_U$ with
$\lambda=\lambda(Q_U)>0$, and $\epsilon'=\nu(\partial_{\mathrm{in}}U)$.
\begin{enumerate}
\item[(a)] For every $\epsilon\in[\epsilon',1)$ and every replay weight $w\in[0,1]$, holding $U$ is an admissible prover sequence for the
dynamics of $\mathcal C$ from $D_0=\nu$, with $D_t=\nu$ and $K_t=\mathrm{int}\,U$ for every $t$. Hence
$\epsilon^*_{\mathrm{div}}(\mathcal C;G,\nu)\le\epsilon^*_{\mathrm{bdd}}(\mathcal C;G,\nu)\le\epsilon'$ for every replay weight.
\item[(b)] If the proofs of the problems in $\mathrm{supp}\,\nu$ are known, the conclusions of Proposition~\ref{prop:st-realize} hold with
$D_0=\nu$ and $D^U_t=\nu$ for every $\epsilon\in(\epsilon',1)$ and $\delta\in(0,1)$.
\end{enumerate}
\end{theorem}

\begin{proof}
Let $A(x,y)=P^U(x,y)\mathbf 1[y\in U]$, the accepted kernel under $s_U$ (Lemma~\ref{lem:st-window}(c)). For $\nu$ supported in $U$,
$(\nu A)(y)=(\nu Q_U)(y)=\lambda\nu(y)$ for $y\in U$ and $(\nu A)(y)=0$ otherwise, so $\nu A=\lambda\nu$, and Lemma~\ref{lem:st-survival}(c)--(d)
gives $D_t=\nu$ with or without replay. Then $D_t(\partial_{\mathrm{in}}U)=\epsilon'\le\epsilon$, so \eqref{eq:st-window-adm} holds; the bound on
the tolerances is the stall consequence of Section~\ref{sec:B-dynamics} (trivial if $\epsilon'=1$). For (b), the trajectory is $D^U_t=\nu$
with acceptance probability $\nu A\mathbf 1=\lambda>0$ and boundary mass $\epsilon'<\epsilon$, so Proposition~\ref{prop:st-realize} applies.
\end{proof}

\begin{proposition}[Realising Chen and Li's assumptions]\label{prop:st-realize}
Let $\mathcal C$ be success-blind or in $\mathcal P_0(k)$, $U\subset V$ finite, $D_0\in\Delta(U)$ with known proofs, and $(D^U_t)$ the
trajectory of \eqref{eq:update} while the prover holds $U$, i.e.\ $D^U_{t+1}=D^U_tA/(D^U_tA\mathbf 1)$ with $A$ as above. Assume
$D^U_tA\mathbf 1>0$ and $D^U_t(\partial_{\mathrm{in}}U)\le\epsilon'$ for every $t$, where $\epsilon'\in[0,1)$. Then for every
$\epsilon\in(\epsilon',1)$ and $\delta\in(0,1)$ there are (1) a prover class $\mathcal F$ whose members have $p/2$-Lipschitz pass rates
(Assumption~3), and (2) a training oracle that, for every $D\in\Delta(V)$, uses $n=\lceil2\ln(1/\delta)/(\epsilon-\epsilon')^2\rceil$ i.i.d.\
samples of $D$ and returns $f\in\mathcal F$ with $\mathbb P_{x\sim D}[\mathrm{succ}(x,f)\ge p]\ge1-\epsilon$ with probability at least
$1-\delta$ (Assumption~2 at accuracy $\epsilon$ and confidence $\delta$), such that Algorithm~1 from $D_0$, with Algorithm~2 as sampler and
the round-$s$ kernel of $\mathcal C$, satisfies for every $t$, with probability at least $1-(t+1)\delta$: for all $s\le t$ the trained prover
has pass rate $s_U$, the round-$s$ training distribution is $D^U_s$, and $K^p(f_s)=\mathrm{int}\,U$. If $\epsilon\ge2\epsilon'$, the
oracle is the member at $(\epsilon,\delta)$ of a family that satisfies Assumption~2 at every accuracy $\tilde\epsilon\in(0,1)$ and confidence
$\tilde\delta\in(0,1)$ with at most $\lceil8\ln(1/\tilde\delta)/(\tilde\epsilon^2(1-p)^2)\rceil$ samples, that is, with $C(\mathcal F)=8$.
\end{proposition}

\begin{proof}
\emph{Prover class.} Fix for each $x$ a verified proof $\mathrm{pf}(x)$ and a rejected string $\bot$. For a profile $s$, the prover $f_s$
outputs $\mathrm{pf}(x)$ with probability $s(x)$ and $\bot$ otherwise, so $\mathrm{succ}(x,f_s)=s(x)$. Let
$\mathcal F=\{f_s:s\text{ satisfies Definition~\ref{def:admissible}(i)}\}$; it contains $f_{s_U}$ (Lemma~\ref{lem:st-window}(b)) and
$f_{\mathbf 1}$.

\emph{Oracle.} From samples $x_1,\dots,x_n$ of $D$ let $\hat\varphi=\frac1n\sum_i\mathbf 1[x_i\notin\mathrm{int}\,U]$, and output $f_{s_U}$ if
$\hat\varphi\le(\epsilon+\epsilon')/2$ and $f_{\mathbf 1}$ otherwise. Let $\mu=D(V\setminus\mathrm{int}\,U)$, so $\mathbb E\hat\varphi=\mu$ and
$\mathbb P_D[s_U\ge p]=1-\mu$. If $\mu\le\epsilon$ both outputs are accurate. If $\mu>\epsilon$, an inaccurate output needs
$\hat\varphi\le(\epsilon+\epsilon')/2<\mu-(\epsilon-\epsilon')/2$, which by Hoeffding's inequality \citep{hoeffding1963} has probability at
most $\exp(-n(\epsilon-\epsilon')^2/2)\le\delta$. For the family: at accuracy $\tilde\epsilon<2\epsilon'$ return $f_{\mathbf 1}$ without
samples; at $\tilde\epsilon\ge2\epsilon'$ use the test with $(\tilde\epsilon,\tilde\delta)$ and $\tilde n=\lceil2\ln(1/\tilde\delta)/(\tilde\epsilon-\epsilon')^2\rceil
\le\lceil8\ln(1/\tilde\delta)/\tilde\epsilon^2\rceil$ samples, since $\tilde\epsilon-\epsilon'\ge\tilde\epsilon/2$.

\emph{Run.} Let $E_s$ be the event that the provers of rounds $0,\dots,s$ are all $f_{s_U}$, and $E_{-1}$ the sure event. On $E_{s-1}$ the
kernels of rounds $r<s$ are $P^U$ (Lemma~\ref{lem:st-survival}(d)), so by Lemma~\ref{lem:st-survival}(a)--(b) the sampler terminates and the
round-$s$ training distribution is $D^U_s$, with $\mu=D^U_s(\partial_{\mathrm{in}}U)\le\epsilon'$; Hoeffding's inequality gives
$\mathbb P(\hat\varphi>(\epsilon+\epsilon')/2)\le\exp(-2n((\epsilon-\epsilon')/2)^2)\le\delta$. So $\mathbb P(E_{s-1}\setminus E_s)\le\delta$ and
$\mathbb P(E_t^c)\le(t+1)\delta$; on $E_t$, $K^p(f_s)=\mathrm{int}\,U$ for $s\le t$. Every accepted proposal $x'$ has
$\mathrm{succ}(x',f_{s_U})\ge p/2$, so line~6 of Algorithm~2 records a verified proof after at most $2/p$ prover calls in expectation, and
every training distribution of the run has known proofs.
\end{proof}

\paragraph{Replay.} With replay weight $w\in[0,1]$, the oracle receiving samples of $D_t=(1-w)F_t+wR_t$ and stored samples keeping their
proofs, the same prover class and oracle give, with probability at least $1-(t+1)\delta$, for all $s\le t$: prover $f_{s_U}$, frontier
$D^U_s$, training distribution $\bar D^U_s:=(1-w)D^U_s+\frac ws\sum_{r<s}D^U_r$ ($\bar D^U_0=D^U_0$), and $K^p(f_s)=\mathrm{int}\,U$. The proof is
the one above: on $E_{s-1}$ the frontiers of rounds $r\le s$ are $D^U_r$, because the frontier update does not involve replay, and
$\bar D^U_s$ is an average of measures supported in $U$ with boundary mass at most $\epsilon'$.

\subsection{The two-sided spectral bound}\label{sec:st-sandwich}

In Lemma~\ref{lem:st-rayleigh}, $P$ is any kernel reversible with respect to $\pi$. Elsewhere in this subsection $P$ is in addition
nearest-neighbour on the connected graph $G$ with $P(x,y)>0$ on every edge.

\begin{lemma}[Rayleigh characterisations]\label{lem:st-rayleigh}
(a) $P$ is a self-adjoint contraction of $\ell^2(\pi)$, and $\|P\|_{\ell^2(\pi)}=\sup\{\langle f,Pf\rangle_\pi/\|f\|_\pi^2\}$ over finitely
supported $f\ne0$. (b) For finite $U$, the eigenvalues of $P_U$ are real, and $\lambda_U$ is the largest one and equals
$\max\{\langle f,Pf\rangle_\pi/\|f\|_\pi^2:f\ne0,\ \mathrm{supp}f\subseteq U\}$. (c) $\sup_U\lambda_U=\|P\|_{\ell^2(\pi)}$ over finite $U$,
i.e.\ $\inf_U(1-\lambda_U)=\lambda_0$, and $U\subseteq U'$ implies $\lambda_U\le\lambda_{U'}$. (d) For finitely supported $f$,
$\|f\|_\pi^2-\langle f,Pf\rangle_\pi=\frac12\sum_{x,y}a(x,y)(f(x)-f(y))^2$.
\end{lemma}

\begin{proof}
(a) Jensen's inequality and $\sum_x\pi(x)P(x,y)=\pi(y)$ give $\|Pf\|_\pi\le\|f\|_\pi$, and reversibility makes the absolutely convergent
series $\sum_{x,y}a(x,y)f(x)g(y)$ symmetric, so $P$ is self-adjoint. For bounded self-adjoint $P$, $\|P\|=\sup_{\|f\|_\pi=1}|\langle f,Pf\rangle_\pi|$
\citep{kreyszig1978}; since $P\ge0$, $|\langle f,Pf\rangle_\pi|\le\langle|f|,P|f|\rangle_\pi$, and the quadratic form is continuous while
finitely supported functions are dense. This is (A.35)--(A.39) in \citet{chenli2026}.
(b) With $\Pi_U=\mathrm{diag}(\pi|_U)$, $S=\Pi_U^{1/2}P_U\Pi_U^{-1/2}$ is symmetric and similar to $P_U$. By the Perron--Frobenius theorem the
spectral radius $\lambda_U$ is an eigenvalue, hence the largest eigenvalue of $S$, which is $\max_g g^\top Sg/g^\top g$; put $g=\Pi_U^{1/2}f$.
(c) A finitely supported $f$ is supported on the finite set $\mathrm{supp}f$, so (a) and (b) give $\sup_U\lambda_U=\|P\|$; monotonicity holds
because the maximum in (b) is over a larger set.
(d) Expand the square and use $\sum_yP(x,y)=1$ and $\sum_x\pi(x)P(x,y)=\pi(y)$.
\end{proof}

\begin{lemma}[Boundary mass of a QSD]\label{lem:st-boundary}
Let $P$ be a nearest-neighbour kernel (not necessarily reversible), $U$ finite and $\nu$ a QSD of $P_U$. Then
$1-\lambda_U=\sum_{x\in\partial_{\mathrm{in}}U}\nu(x)P(x,V\setminus U)$, hence $e_{\min}(P)\,\nu(\partial_{\mathrm{in}}U)\le1-\lambda_U\le\nu(\partial_{\mathrm{in}}U)$.
If $P(x,V\setminus U)=e$ on $\partial_{\mathrm{in}}U$, then $e\,\nu(\partial_{\mathrm{in}}U)=1-\lambda_U$.
\end{lemma}

\begin{proof}
Summing $\nu P_U=\lambda_U\nu$ over $U$ gives $\lambda_U=1-\sum_{x\in U}\nu(x)P(x,V\setminus U)$. For $x\in\mathrm{int}\,U$, $P(x,V\setminus U)=0$;
for $x\in\partial_{\mathrm{in}}U$ some $y\in N(x)\setminus U$ gives $e_{\min}(P)\le P(x,y)\le P(x,V\setminus U)\le1$.
\end{proof}

\begin{lemma}[Laziness]\label{lem:st-lazy}
$P'=\frac12(I+P)$ is reversible and nearest-neighbour with the same $\pi$; for finite $U$, $P'_U$ and $P_U$ have the same QSDs and
$\lambda(P'_U)=\frac12(1+\lambda_U)$; $\lambda_0(P')=\lambda_0(P)/2$, $\epsilon_D(P')=\epsilon_D(P)$, $e_{\min}(P')=e_{\min}(P)/2$ and
$\Phi(P')=\Phi(P)/2$.
\end{lemma}

\begin{proof}
The eigenvalues of $P'_U$ are $\frac12(1+\theta)$ for the real eigenvalues $\theta$ of $P_U$, so by Lemma~\ref{lem:st-rayleigh}(b)
$\lambda(P'_U)=\frac12(1+\lambda_U)$, and $\nu P'_U=\frac12(1+\lambda_U)\nu$ iff $\nu P_U=\lambda_U\nu$. Lemma~\ref{lem:st-rayleigh}(a) gives
$\|P'\|=\frac12(1+\|P\|)$, and $P'(x,y)=P(x,y)/2$ for $y\ne x$.
\end{proof}

\begin{corollary}[Two-sided spectral bound]\label{cor:st-sandwich}
\begin{enumerate}
\item[(i)] (\citealp[Theorem~4.2]{chenli2026}, in the form of their Appendix~A.2 before the Cheeger step.) Assume $\sup_x\pi(x)<\infty$.
For every $\epsilon\in(0,1)$, every $D_0$ with $\|D_0\|_{\ell^2(1/\pi)}<\infty$ and every admissible prover sequence for $P$ from $D_0$ without
replay,
\[
|K_t|\ \ge\ \frac{(1-\epsilon)^2}{\sup_x\pi(x)\,\|D_0\|^2_{\ell^2(1/\pi)}}\Big(\frac{1-\epsilon}{1-\lambda_0}\Big)^{2t}\qquad(t\ge0).
\]
Hence $\epsilon^*_{\mathrm{bdd}}(P;G,D_0)\ge\epsilon^*_{\mathrm{div}}(P;G,D_0)\ge\lambda_0$ for every such $D_0$, in particular every finitely
supported one.
\item[(ii)] For every finite $U$ and QSD $\nu$ of $P_U$, $\lambda_0\le1-\lambda_U$ and
$e_{\min}(P)\,\nu(\partial_{\mathrm{in}}U)\le1-\lambda_U\le\nu(\partial_{\mathrm{in}}U)$. Hence $\lambda_0\le\epsilon_D(P)\le\lambda_0/e_{\min}(P)$.
\item[(iii)] For every $\epsilon\in(\epsilon_D(P),1)$ there are a finite $U$ and a QSD $\nu$ of $P_U$ with $\nu(\partial_{\mathrm{in}}U)<\epsilon$
and $\lambda_U>0$, so Theorem~\ref{thm:st-qsd} applies. If $\sup_x\pi(x)<\infty$, then, with infima over finitely supported $D_0$,
\[
\lambda_0\ \le\ \inf_{D_0}\epsilon^*_{\mathrm{div}}(P;G,D_0)\ \le\ \inf_{D_0}\epsilon^*_{\mathrm{bdd}}(P;G,D_0)\ \le\ \epsilon_D(P)\ \le\ \lambda_0/e_{\min}(P).
\]
\item[(iv)] $\Phi(P)^2/2\le\lambda_0\le\Phi(P)$; the lower bound is the discrete Cheeger inequality
(\citealp[Theorem~10.3 and its proof, p.~111]{woess2000}; \citealp[Theorem~2.3]{dodziuk1984}), written in the present notation as
\citet[Theorem~A.14]{chenli2026}.
\end{enumerate}
\end{corollary}

\begin{proof}
(i) For $D$ with $\|D\|_{\ell^2(1/\pi)}<\infty$ and $g=D/\pi$, reversibility gives $(DP)/\pi=Pg$, so
$\|DP\|_{\ell^2(1/\pi)}\le(1-\lambda_0)\|D\|_{\ell^2(1/\pi)}$ (their (A.41)--(A.47)). The acceptance probability satisfies
$Z_t\ge D_t(K_t)\ge1-\epsilon$ (Section~\ref{sec:B-dynamics}), so $D_{t+1}\le D_tP/(1-\epsilon)$ pointwise and
$\|D_t\|_{\ell^2(1/\pi)}\le(\frac{1-\lambda_0}{1-\epsilon})^t\|D_0\|_{\ell^2(1/\pi)}$ (their (A.50)--(A.54)). By the Cauchy--Schwarz inequality
(their Lemma~E.1), $1-\epsilon\le D_t(K_t)\le(|K_t|\sup\pi)^{1/2}\|D_t\|_{\ell^2(1/\pi)}$, which gives the display. For $\epsilon<\lambda_0$ the
right-hand side tends to infinity, and the growth consequence of Section~\ref{sec:B-dynamics} gives the tolerance bound.
(ii) is Lemma~\ref{lem:st-rayleigh}(c) and Lemma~\ref{lem:st-boundary}; take the infimum over $(U,\nu)$ and use
$\inf_U(1-\lambda_U)=\lambda_0$.
(iii) By the definition of $\epsilon_D$ there is $(U,\nu)$ with $\nu(\partial_{\mathrm{in}}U)<\epsilon$, and $1-\lambda_U\le\nu(\partial_{\mathrm{in}}U)<1$
by (ii). Theorem~\ref{thm:st-qsd}(a) gives $\epsilon^*_{\mathrm{bdd}}(P;G,\nu)<\epsilon$ for the finitely supported $\nu$; letting
$\epsilon\downarrow\epsilon_D(P)$ gives the third inequality (trivial if $\epsilon_D(P)=1$). The first is (i), the second holds for every $D_0$,
the last is (ii).
(iv) For finite $A$, $\langle\mathbf 1_A,P\mathbf 1_A\rangle_\pi=\pi(A)-a(\partial A)$, so Lemma~\ref{lem:st-rayleigh}(b)--(c) give
$\lambda_0\le1-\lambda_A\le a(\partial A)/\pi(A)$. If $\Phi(P)>0$, then $\pi(A)\le\kappa\,a(\partial A)$ with $\kappa=1/\Phi(P)$, and the cited
theorem (for irreducible reversible chains; $P$ is irreducible because $G$ is connected and $P>0$ on $E$) gives
$\|P\|\le1-1/(2\kappa^2)$.
\end{proof}

\subsection{Subdivided trees}\label{sec:st-tree}

Let $T=T_3^{(L)}$, $L\ge1$, $P$ the SRW ($\pi=\deg$) and $a:=2^{-1/(2L)}$. Every edge joins consecutive levels, every vertex at level
$r+1$ has exactly one neighbour at level $r$, level $r\ge1$ consists of branch vertices if $L\mid r$ and of subdivision vertices otherwise,
and the level sizes are
\begin{equation}\label{eq:st-levels}
n(0)=1,\qquad n(r)=3\cdot2^{\lceil r/L\rceil-1}\ (r\ge1),\qquad \tfrac32\,2^{r/L}\le n(r)\le3\cdot2^{(r-1)/L}\ (r\ge1),
\end{equation}
since the vertices at distance $r\in((j-1)L,jL]$ lie one on each of the $3\cdot2^{j-1}$ subdivided edges of skeleton depth $j$.

\begin{lemma}[Conductance of $T_3^{(L)}$]\label{lem:st-phi}
Let $A\subseteq V$ be finite and nonempty with $T[A]$ connected, with $n_b$ branch and $n_p$ subdivision vertices, and let $\partial_EA$ be
the set of edges with exactly one endpoint in $A$. Then $|\partial_EA|=n_b+2$. If $n_b=0$, then $L\ge2$ and $n_p\le L-1$; if $n_b\ge1$, then
$n_p\le(2n_b+1)(L-1)$. Consequently $\Phi(P)=1/(4L-1)$.
\end{lemma}

\begin{proof}
$T[A]$ is a tree with $|A|-1$ edges, so $|\partial_EA|=\sum_{x\in A}\deg(x)-2(|A|-1)=n_b+2$. If $n_b=0$, $A$ is a path of subdivision vertices,
which lies in the interior of one subdivided edge. If $n_b\ge1$, the geodesic from a subdivision vertex of $A$ to a branch vertex of $A$ lies in
$A$, so every subdivided edge containing a subdivision vertex of $A$ has an endpoint in $A$. The subdivided edges with both endpoints in $A$ are
the edges of the subtree of $T_3$ spanned by the branch vertices of $A$, which has $n_b-1$ edges, so $3n_b-(n_b-1)=2n_b+1$ subdivided edges have an
endpoint in $A$, and $n_p\le(2n_b+1)(L-1)$.
For SRW, $a(\partial A)/\pi(A)=|\partial_EA|/\sum_{x\in A}\deg(x)$. If $n_b=0$ the ratio is $2/(2n_p)\ge1/(L-1)>1/(4L-1)$; if $n_b\ge1$ it is at least
$(n_b+2)/(3n_b+2(2n_b+1)(L-1))=(n_b+2)/((4L-1)n_b+2L-2)\ge1/(4L-1)$. For general $A$ the components of $T[A]$ are pairwise non-adjacent, so
both $|\partial_EA|$ and $\pi(A)$ add over components. Hence $\Phi(P)\ge1/(4L-1)$. Taking $A$ to be a connected set of $n_b$ branch vertices
together with the interiors of the $2n_b+1$ subdivided edges incident to it gives equality in both bounds, and the ratio tends to $1/(4L-1)$ as
$n_b\to\infty$.
\end{proof}

\begin{lemma}[Radial Rayleigh bound]\label{lem:st-ball}
For every $N\ge1$ and $U=B(o,N)$, $1-\lambda_U\le(1-a)^2+2a\big(1-\cos\frac{\pi}{N+2}\big)$.
\end{lemma}

\begin{proof}
Let $\theta=\pi/(N+2)$, $\varphi(r)=\sin(\theta(r+1))$ ($0\le r\le N+1$, so $\varphi(N+1)=0$), $g(r)=a^r\varphi(r)$, and $f(x)=g(|x|)$ for $|x|\le N$,
$f=0$ otherwise. By Lemma~\ref{lem:st-rayleigh}(b),(d), $1-\lambda_U\le\mathcal E(f)/\|f\|_\pi^2$ with $\mathcal E(f)=\sum_{\{x,y\}\in E}(f(x)-f(y))^2$.
There are $n(r+1)\le3\cdot2^{r/L}$ edges between levels $r$ and $r+1$ and $2^{r/L}a^{2r}=1$, so
$\mathcal E(f)=\sum_{r=0}^Nn(r+1)(g(r+1)-g(r))^2\le3\sum_{r=0}^N(a\varphi(r+1)-\varphi(r))^2$. Since $\pi(o)=3$, $\deg\ge2$ and
$n(r)\ge\frac32 2^{r/L}$, $\|f\|_\pi^2\ge3\sum_{r=0}^N2^{r/L}g(r)^2=3S$ with $S=\sum_{r=0}^N\varphi(r)^2$. Now
$\sum_{r=0}^N\varphi(r+1)^2=S-\sin^2\theta$, and multiplying $\sin((j-1)\theta)+\sin((j+1)\theta)=2\cos\theta\sin(j\theta)$ by $\sin(j\theta)$ and
summing over $j=1,\dots,N+1$ gives $\sum_{r=0}^N\varphi(r)\varphi(r+1)=\cos\theta\,S$ (the boundary terms vanish). Hence
$\sum_{r=0}^N(a\varphi(r+1)-\varphi(r))^2\le S(1-2a\cos\theta+a^2)$ and $1-\lambda_U\le(1-a)^2+2a(1-\cos\theta)$.
\end{proof}

\begin{corollary}[Subdivided trees]\label{cor:st-subdivided}
\begin{enumerate}
\item[(1)] $\Phi(P)=1/(4L-1)$.
\item[(2)] For $U=B(o,N)$, $N\ge1$, $\partial_{\mathrm{in}}U=S_N$ and $\nu_U(\partial_{\mathrm{in}}U)=\frac{c+1}c(1-\lambda_U)$ with $c=2$ if $L\mid N$
and $c=1$ otherwise; for $N=mL$ this is at most $\frac32\big[(1-a)^2+2a\big(1-\cos\frac{\pi}{mL+2}\big)\big]$.
\item[(3)] $\epsilon_D(P)\le\frac32(1-a)^2<\frac{3(\ln2)^2}{8L^2}$ and $\lambda_0(P)\le(1-a)^2<\frac{(\ln2)^2}{4L^2}$.
\item[(4)] With infima over finitely supported cold starts,
\begin{align*}
\frac1{2(4L-1)^2}&\le\lambda_0(P)\le\inf_{D_0}\epsilon^*_{\mathrm{div}}(P;T,D_0)\le\inf_{D_0}\epsilon^*_{\mathrm{bdd}}(P;T,D_0)\\
&\le\epsilon_D(P)\le\tfrac32(1-a)^2<6(\ln2)^2\,\Phi(P)^2.
\end{align*}
\end{enumerate}
By Lemma~\ref{lem:st-lazy}, (2) and the bound on $\epsilon_D$ in (3) hold for $P^{\mathrm{lazy}}$ with the same QSDs, and
$\lambda_0(P^{\mathrm{lazy}})=\lambda_0(P)/2$.
\end{corollary}

\begin{proof}
(1) is Lemma~\ref{lem:st-phi}. (2) The vertices of $U$ with a neighbour outside $U$ are those of $S_N$; each has $c$ children and degree $c+1$, so
$P(x,V\setminus U)=c/(c+1)$ there, and Lemma~\ref{lem:st-boundary} and Lemma~\ref{lem:st-ball} apply ($P_U$ is irreducible, so $\nu_U$ is unique).
(3) Take the infimum over $m$ in (2), and use Lemma~\ref{lem:st-rayleigh}(c) with Lemma~\ref{lem:st-ball}; finally $1-a<\ln2/(2L)$. (4) The
first inequality is Corollary~\ref{cor:st-sandwich}(iv) with (1), the next three are Corollary~\ref{cor:st-sandwich}(iii) ($\sup\pi=3$), then
(3), and $\frac{3(\ln2)^2}{8L^2}=\frac{3(\ln2)^2(4L-1)^2}{8L^2}\Phi^2<6(\ln2)^2\Phi^2$.
\end{proof}

\subsection{The fixed start}\label{sec:st-fixed}

Theorem~\ref{thm:st-qsd} starts in a QSD. From a single vertex the window must satisfy \eqref{eq:st-window-adm} in every round, while the QSD
controls only the limit. We prove the monotone comparison $D_t(\partial_{\mathrm{in}}U)\le\nu(\partial_{\mathrm{in}}U)$ for lazy birth--death
chains started at their reflecting end, and transfer it to graphs by lumping.

\paragraph{Killed birth--death chains.} Fix $R\ge1$. A killed birth--death chain on $\{0,\dots,R\}$ has up-probabilities $\beta_i$ ($i<R$),
down-probabilities $\gamma_i$ ($i\ge1$, $\gamma_0=0$), holding probabilities $\eta_i$ and killing probability $e_\partial\ge0$ at level $R$ only,
with $\beta_i+\gamma_i+\eta_i=1$ for $i<R$ and $\gamma_R+\eta_R+e_\partial=1$. Let $Q$ be its substochastic matrix, $X_0=0$, $\tau$ the killing time,
$D_t=\mathbb P(X_t\in\cdot\mid\tau>t)$ and $\lambda=\rho(Q)$. Assume
(BD1) $\beta_i>0$ for $i<R$ and $\gamma_i>0$ for $1\le i\le R$, and (BD2) $\eta_i\ge1/2$ for every $i$.
Then $Q$ is irreducible with a unique QSD $\nu$, and $\mathbb P(\tau>t)\ge\eta_0^t>0$.

\begin{theorem}[Fixed-start comparison]\label{thm:st-fixed}
Under (BD1)--(BD2), started at level $0$, $t\mapsto D_t(R)$ is nondecreasing, converges to $\nu(R)$, and $D_t(R)\le\nu(R)$ for every $t\ge0$.
\end{theorem}

For $e_\partial>0$ the proof uses the representation of the killing time as a sum of independent geometric variables whose failure
probabilities are the eigenvalues of $Q$ \citep{keilson1979}, in the discrete-time form of \citet[Theorems~1.1 and~1.2]{fill2009}, which needs
these eigenvalues to be nonnegative; (BD2) provides this. We include the short argument.

\begin{lemma}[Spectrum]\label{lem:st-spectrum}
Under (BD1)--(BD2) the eigenvalues $\theta_0\le\dots\le\theta_R$ of $Q$ lie in $[0,1]$; $\theta_R=\lambda$ is simple with positive left and
right eigenvectors, $\theta_i<\lambda$ for $i<R$, and $\lambda<1$ if $e_\partial>0$.
\end{lemma}

\begin{proof}
With $m_0=1$, $m_{i+1}=m_i\beta_i/\gamma_{i+1}$ and $\Delta=\mathrm{diag}(\sqrt{m_i})$, $S=\Delta Q\Delta^{-1}$ is symmetric. Write $Q=\frac12I+M$;
by (BD2) $M\ge0$ has row sums at most $\frac12$ and $\Delta M\Delta^{-1}$ is symmetric, so the eigenvalues of $M$ are real with modulus at most
$\frac12$, and $\theta_i\in[0,1]$. By (BD1) and the Perron--Frobenius theorem, $\lambda$ is simple with positive eigenvectors, and as the largest
real eigenvalue it is $\theta_R$. If $e_\partial>0$ and $\lambda=1$, a maximiser $i^*$ of a positive right eigenvector $h$ satisfies
$h(i^*)\le h(i^*)\sum_jQ(i^*,j)$ with equality only if the maximum is attained at every neighbour of $i^*$ and the row sum is $1$; by
irreducibility this reaches level $R$, whose row sum is $1-e_\partial<1$.
\end{proof}

\begin{lemma}[Keilson--Fill representation]\label{lem:st-keilson}
Under (BD1)--(BD2) and $e_\partial>0$, $\tau<\infty$ almost surely and $\mathbb E\,z^\tau=\prod_{i=0}^R\frac{(1-\theta_i)z}{1-\theta_iz}$ for
$|z|\le1$. Hence $\tau$ has the law of $G_0+\dots+G_R$ for independent $G_i$ with $\mathbb P(G_i=j)=(1-\theta_i)\theta_i^{\,j-1}$, $j\ge1$.
\end{lemma}

\begin{proof}
$\mathbb P(\tau>t)=e_0^\top Q^t\mathbf 1\to0$ since $\rho(Q)<1$. Killing at step $t+1$ requires $X_t=R$, so for $|z|<1/\rho(Q)$,
$\mathbb E\,z^\tau=z\,e_\partial[(I-zQ)^{-1}]_{0R}=z\,e_\partial\,\mathrm{cof}_{R0}(I-zQ)/\det(I-zQ)$. Deleting row $R$ and column $0$ of the
tridiagonal $I-zQ$ leaves a lower triangular matrix with diagonal $-z\beta_0,\dots,-z\beta_{R-1}$, so $\mathrm{cof}_{R0}=z^R\prod_{i<R}\beta_i$,
and $\det(I-zQ)=\prod_i(1-z\theta_i)$. So $\mathbb E\,z^\tau=c\,z^{R+1}/\prod_i(1-\theta_iz)$, and $\mathbb E\,1^\tau=1$ gives
$c=\prod_i(1-\theta_i)$. Both sides are analytic on a disc of radius greater than $1$, and the right-hand side is the generating function of
$G_0+\dots+G_R$.
\end{proof}

\begin{lemma}[Log-concavity under convolution]\label{lem:st-lc}
Call a probability vector $f$ on $\mathbb Z$ log-concave if its support is an integer interval and $f(k)^2\ge f(k-1)f(k+1)$ for all $k$. If
$f$ and $g$ are log-concave, so is $f*g$.
\end{lemma}

\begin{proof}
This is the closure of P\'olya frequency sequences of order $2$ under convolution \citep{karlin1968}. For log-concave $h$, the ratio
$h(k+1)/h(k)$ is nonincreasing on the support, so $h(b)h(c)\ge h(a)h(d)$ whenever $a\le b,c\le d$ and $a+d=b+c$; hence every $2\times2$ minor
of $T_h(i,j)=h(i-j)$ with increasing rows and columns is nonnegative. Since $T_fT_g=T_{f*g}$, the Cauchy--Binet formula
$(u\cdot x)(v\cdot y)-(u\cdot y)(v\cdot x)=\sum_{k<l}(u_kv_l-u_lv_k)(x_ky_l-x_ly_k)$ for rows $u,v$ of $T_f$ and columns $x,y$ of $T_g$ shows that
the same holds for $T_{f*g}$; rows $m,m+1$ and columns $0,1$ give $(f*g)(m)^2\ge(f*g)(m-1)(f*g)(m+1)$. The support of $f*g$ is the sum of two
integer intervals.
\end{proof}

\begin{lemma}[Hazard]\label{lem:st-hazard}
If $\tau$ takes values in $\{1,2,\dots\}$, has a log-concave law $f$ and $\mathbb P(\tau>t)>0$ for every $t$, then
$h_t:=\mathbb P(\tau=t+1\mid\tau>t)$ is nondecreasing in $t\ge0$.
\end{lemma}

\begin{proof}
The support is $[k_0,\infty)$, and $h_t=0$ for $t+1<k_0$. For $t+1\ge k_0$, $1/h_t=\sum_{j\ge0}f(t+1+j)/f(t+1)$, and
$f(k+j)/f(k)=\prod_{i<j}f(k+i+1)/f(k+i)$ is nonincreasing in $k$.
\end{proof}

\begin{lemma}[Monotone chains]\label{lem:st-monotone}
Under (BD1)--(BD2) and $e_\partial=0$, started at $0$, $t\mapsto\mathbb P(X_t\ge j)$ is nondecreasing for every $j$.
\end{lemma}

\begin{proof}
By (BD2), $\beta_i\le\frac12$ and $\gamma_{i+1}\le\frac12$. Then $Q(i,[j,R])\le Q(i+1,[j,R])$ for all $i<R$ and $j$: both are $1$ for $j\le i-1$;
for $j=i$ they are $1-\gamma_i\le1$; for $j=i+1$ they are $\beta_i\le1-\gamma_{i+1}$; for $j\ge i+2$ the left side is $0$. So $Q$ is stochastically
monotone, and $\delta_0\preceq\delta_0Q$ gives $\delta_0Q^t\preceq\delta_0Q^{t+1}$ by induction.
\end{proof}

\begin{proof}\textbf{of Theorem~\ref{thm:st-fixed}.}
\emph{Case $e_\partial>0$.} By Lemma~\ref{lem:st-survival}(b), $D_t=e_0^\top Q^t/(e_0^\top Q^t\mathbf 1)$, and since killing occurs only from level
$R$, the hazard of $\tau$ is $h_t=e_\partial D_t(R)$. By Lemmas~\ref{lem:st-spectrum} and~\ref{lem:st-keilson}, $\tau$ is a sum of independent
geometric variables, whose laws are log-concave, so $h_t$ is nondecreasing by Lemmas~\ref{lem:st-lc} and~\ref{lem:st-hazard}. With the spectral
decomposition $S=\sum_i\theta_i\phi_i\phi_i^\top$, $\mathbb P(\tau>t)=\sum_ic_i\theta_i^t$ with $c_i=m_0^{-1/2}\phi_i(0)\sum_j\sqrt{m_j}\phi_i(j)$ and
$c_R>0$ because $\phi_R>0$. As $t\to\infty$ for the fixed chain, $0\le\theta_i<\lambda$ ($i<R$) gives $\mathbb P(\tau>t)=c_R\lambda^t(1+o(1))$ and
$\mathbb P(\tau=t+1)=\sum_ic_i\theta_i^t(1-\theta_i)=c_R\lambda^t(1-\lambda)(1+o(1))$, so $h_t\to1-\lambda$. Summing $\nu Q=\lambda\nu$ gives
$1-\lambda=e_\partial\nu(R)$. So $D_t(R)=h_t/e_\partial$ increases to $\nu(R)$.

\emph{Case $e_\partial=0$.} Then $D_t$ is the law of $X_t$, $D_t(R)=\mathbb P(X_t\ge R)$ is nondecreasing by Lemma~\ref{lem:st-monotone}, and
$D_t\to\nu$ because the chain is finite, irreducible and aperiodic ($\eta_0>0$).
\end{proof}

\begin{lemma}[Radial reduction]\label{lem:st-radial}
Let $\Gamma$ be a group of automorphisms of $G$ fixing $o$, $U\ni o$ finite, and $U=U_0\cup\dots\cup U_R$ a partition with $U_0=\{o\}$ such that
$\Gamma$ maps each $U_j$ onto itself, acts transitively on it, and preserves $s_U$. Let $\mathcal C\in\mathcal P_0(k)$ (or a $\Gamma$-invariant
success-blind kernel) and $P^U=\mathcal C^{s_U}$.
\begin{enumerate}
\item[(a)] $P^U(\gamma x,\gamma A)=P^U(x,A)$ for all $x$, $A$ and $\gamma\in\Gamma$, so $\bar Q(i,j):=P^U(x,U_j)$ does not depend on $x\in U_i$.
\item[(b)] The level process of the proposal chain with kernel $P^U$ from $o$, killed at its first proposal outside $U$, is a killed Markov
chain with matrix $\bar Q$ and the same killing time, and $D_t(U_j)=\mathbb P(Y_t=j\mid\tau>t)$.
\item[(c)] If $\bar Q$ is irreducible, the level projection of every QSD $\nu$ of $Q_U$ is the QSD $\bar\nu$ of $\bar Q$, and $\lambda(Q_U)=\lambda(\bar Q)$.
\end{enumerate}
\end{lemma}

\begin{proof}
(a) An automorphism $\gamma$ maps $B(x,k)$ onto $B(\gamma x,k)$ and, as $s_U\circ\gamma=s_U$, is an isomorphism of the labelled views of $x$ and
$\gamma x$; (E) gives the identity for $A\subseteq B(x,k)$, and (L) extends it to all $A$, since $\gamma A\cap B(\gamma x,k)=\gamma(A\cap B(x,k))$.
For $x,x'\in U_i$ take $\gamma x=x'$. (b) Given the history, the next proposal is accepted into $U_j$ with probability $P^U(X_t,U_j)=\bar Q(Y_t,j)$,
which depends only on $Y_t$ (strong lumpability); the identity is the projection of Lemma~\ref{lem:st-survival}(b). (c) Summing
$\nu Q_U=\lambda(Q_U)\nu$ over $U_j$ and using (a) gives $\bar\nu\bar Q=\lambda(Q_U)\bar\nu$ for $\bar\nu(j)=\nu(U_j)$. Pairing with a positive
right Perron vector of $\bar Q$ gives $\lambda(Q_U)=\lambda(\bar Q)$, and $\bar\nu$ is the unique QSD by the Perron--Frobenius theorem.
\end{proof}

When the partition is by distance from $o$ (on $\mathbb Z$ and on trees) and $\mathcal C$ is nearest-neighbour, $\bar Q$ moves at most one
level per step, the holding probability of level $i\ge1$ is $P^U(x,\{x\})$ for $x\in U_i$ (vertices of the same level are not adjacent), and
proposals leave $U$ only from $\partial_{\mathrm{in}}U$. With replay, the frontier is the trajectory without replay, and
$D_t=(1-w)F_t+\frac wt\sum_{r<t}F_r$ is an average of measures whose boundary mass obeys the same bound; so every comparison below holds for
every replay weight.

\begin{corollary}[Fixed start on $T_3^{(L)}$]\label{cor:st-tree}
Let $T=T_3^{(L)}$, $N\ge1$, $U=B(o,N)$.
\begin{enumerate}
\item[(a)] Let $\mathcal C\in\mathcal P_0(k)$ be nearest-neighbour (or a success-blind nearest-neighbour kernel invariant under the
automorphisms of $T$ fixing $o$), and assume that the level chain $\bar Q$ of $P^U$ satisfies (BD1)--(BD2). If the prover holds $U$ from
$\delta_o$, then $D_t(\partial_{\mathrm{in}}U)\le\nu(\partial_{\mathrm{in}}U)$ for every $t$ and every replay weight, for every QSD $\nu$ of $Q_U$.
\item[(b)] For $\LRW=P^{\mathrm{lazy}}$ and every $\epsilon\in(\frac32(1-a)^2,1)$ there is $m\ge1$ such that holding $B(o,mL)$ is admissible
from $\delta_o$ in every round, with any replay weight, so $\sup_t|K_t|=|B(o,mL-1)|$, and
\[
\frac1{4(4L-1)^2}\le\lambda_0(P^{\mathrm{lazy}})\le\epsilon^*_{\mathrm{div}}(P^{\mathrm{lazy}};T,o)\le\epsilon^*_{\mathrm{bdd}}(P^{\mathrm{lazy}};T,o)
\le\tfrac32(1-a)^2<\frac{3(\ln2)^2}{8L^2}
\]
without replay; the upper bound on $\epsilon^*_{\mathrm{bdd}}$ holds for every replay weight. The interval $(\frac32(1-a)^2,1)$ is nonempty since
$\frac32(1-2^{-1/2})^2<0.13$. In particular, for every $c>0$ and $L\ge3(\ln2)^2/(2c)$ with $c\,\Phi(P)<1$, $\LRW$ from $\delta_o$ can be stalled at
$\epsilon=c\,\Phi(P)$.
\end{enumerate}
\end{corollary}

\begin{proof}
(a) Let $\Gamma$ be the automorphisms of $T$ fixing $o$ and $U_j=S_j$. $\Gamma$ preserves levels, hence $U$, each $U_j$ and $s_U$
($\mathrm{int}\,U=B(o,N-1)$, $\partial_{\mathrm{in}}U=S_N$). $\Gamma$ is transitive on each level, because $T$ rooted at $o$ is spherically
symmetric (every vertex at level $r$ has $b_r$ children, $b_0=3$, $b_r=2$ if $L\mid r>0$, $b_r=1$ otherwise): for $x,y\in S_r$ with geodesics
$(x_i)$ and $(y_i)$ from $o$, map $x_i\mapsto y_i$, the other children of $x_i$ bijectively onto those of $y_i$, and each descendant subtree
isomorphically onto the corresponding one; this bijection preserves the parent relation. Lemma~\ref{lem:st-radial} applies, $\bar Q$ is a
birth--death chain killed only from level $N$, and Theorem~\ref{thm:st-fixed} gives $D_t(\partial_{\mathrm{in}}U)\le\bar\nu(N)=\nu(\partial_{\mathrm{in}}U)$.
(b) The level chain of $\LRW$ has $\beta_0=\frac12$, $\beta_r=\frac{b_r}{2(b_r+1)}$, $\gamma_r=\frac1{2(b_r+1)}$, holding $\frac12$ and killing
$\frac{b_N}{2(b_N+1)}>0$, so (BD1)--(BD2) hold. Its QSD on $U$ is that of SRW (Lemma~\ref{lem:st-lazy}), so by
Corollary~\ref{cor:st-subdivided}(2) $\nu(\partial_{\mathrm{in}}B(o,mL))\le\frac32[(1-a)^2+2a(1-\cos\frac\pi{mL+2})]\le\epsilon$ for large $m$. Part (a)
and \eqref{eq:st-window-adm} give admissibility, and the stall consequence of Section~\ref{sec:B-dynamics} gives $\epsilon^*_{\mathrm{bdd}}\le\epsilon$
for every such $\epsilon$. The lower bounds are Corollary~\ref{cor:st-sandwich}(i) with $D_0=\delta_o$ and $\lambda_0(P^{\mathrm{lazy}})=\lambda_0(P)/2\ge
\Phi(P)^2/4$ (Lemma~\ref{lem:st-lazy}, Corollary~\ref{cor:st-subdivided}(4)); the last inequality is Corollary~\ref{cor:st-subdivided}(3). For the
final claim, $L\ge3(\ln2)^2/(2c)$ gives $8cL^2\ge12(\ln2)^2L\ge3(\ln2)^2(4L-1)$, so $\frac32(1-a)^2<\frac{3(\ln2)^2}{8L^2}\le\frac c{4L-1}=c\,\Phi(P)$.
\end{proof}

\begin{corollary}[Fixed start on $\mathbb Z$]\label{cor:st-Z}
Let $G=\mathbb Z$, $R\ge1$, $U=[-R,R]$, and let the curriculum be $\LRW$ ($P(x,x)=\frac12$, $P(x,x\pm1)=\frac14$). The QSD of $P_U$ is
$\nu(x)\propto\cos\frac{\pi x}{2(R+1)}$, $\nu(\partial_{\mathrm{in}}U)=2(1-\cos\frac{\pi}{2(R+1)})\le\frac{\pi^2}{4(R+1)^2}$, and if the prover holds $U$
from $\delta_0$, then $D_t(\partial_{\mathrm{in}}U)\le\nu(\partial_{\mathrm{in}}U)$ for every $t$ and every replay weight.
\end{corollary}

\begin{proof}
Apply Lemma~\ref{lem:st-radial} with $\Gamma=\{\mathrm{id},x\mapsto-x\}$ and $U_j=\{\pm j\}$. The level chain has $\beta_0=\frac12$,
$\beta_i=\gamma_i=\frac14$ for $i\ge1$, holding $\frac12$ and killing $\frac14$ from level $R$, so (BD1)--(BD2) hold, and
Theorem~\ref{thm:st-fixed} gives the comparison. The function $\phi(x)=\cos\frac{\pi x}{2(R+1)}$ is positive on $U$, vanishes at $\pm(R+1)$ and
satisfies $\phi P_U=\frac12(1+\cos\vartheta)\phi$ with $\vartheta=\frac\pi{2(R+1)}$ ($P_U$ is symmetric), so it is proportional to the QSD, and
Lemma~\ref{lem:st-boundary} with $e=\frac14$ gives $\nu(\partial_{\mathrm{in}}U)=4(1-\lambda_U)=2(1-\cos\vartheta)=4\sin^2(\vartheta/2)\le\vartheta^2$.
\end{proof}

\begin{corollary}[Fixed start on $\mathbb Z^D$]\label{cor:st-ZD}
Let $G=\mathbb Z^D$, $R\ge1$, $U=[-R,R]^D$, so $\mathrm{int}\,U=[-R+1,R-1]^D$, and let the curriculum be $\LRW$ ($P(x,x)=\frac12$,
$P(x,x\pm e_i)=\frac1{4D}$). If the prover holds $U$ from $\delta_0$, then $D_t(\partial_{\mathrm{in}}U)\le\nu_U(\partial_{\mathrm{in}}U)=1-(1-\nu_1)^D\le D\nu_1$
for every $t$, where $\nu_1=2(1-\cos\frac{\pi}{2(R+1)})\le\frac{\pi^2}{4(R+1)^2}$. For every $\epsilon\in(0,1)$ the window with
$R+1=\lceil\frac\pi2\sqrt{D/\epsilon}\rceil$ is admissible from $\delta_0$ in every round, with any replay weight, and
$\sup_t|K_t|=(2R-1)^D<(\pi\sqrt{D/\epsilon})^D$.
\end{corollary}

\begin{proof}
\emph{Clock representation.} $\LRW$ is $X_{t+1}=X_t+\xi_{t+1}e_{c_{t+1}}$ with independent $c_t$ uniform on $\{1,\dots,D\}$ and $\xi_t\in\{-1,0,1\}$
with probabilities $\frac14,\frac12,\frac14$. A proposal leaves $U$ exactly when the moved coordinate steps from $\pm R$ to $\pm(R+1)$. Given the
clock sequence $c$, with $n_j(t)$ the number of $r\le t$ with $c_r=j$, the coordinates are independent copies of the one-dimensional killed lazy
walk $Z$ of Corollary~\ref{cor:st-Z} run for $n_j(t)$ steps, and survival is the intersection of their survival events. With $S(n)$ the survival
probability of $Z$ for $n$ steps, $D^{(1)}_n$ its conditional law from $0$, and $W(c)=\prod_jS(n_j(t))$,
\[
D_t(\mathrm{int}\,U)=\frac{\mathbb E\big[W(c)\prod_j(1-D^{(1)}_{n_j(t)}(\{\pm R\}))\big]}{\mathbb E\,W(c)}\ \ge\ (1-\nu_1)^D,
\]
since $D^{(1)}_n(\{\pm R\})\le\nu_1$ for every $n$ by Corollary~\ref{cor:st-Z}.
\emph{QSD of the cube.} $P_U=\frac1D\sum_iI\otimes\dots\otimes P^{(1)}\otimes\dots\otimes I$ with $P^{(1)}$ the killed one-dimensional matrix of Corollary~\ref{cor:st-Z} in position $i$, so the product of one-dimensional QSDs is a
positive left eigenvector of the irreducible $P_U$, hence $\nu_U$, and $\nu_U(\partial_{\mathrm{in}}U)=1-(1-\nu_1)^D\le D\nu_1$. For the stated
$R$ (and $R\ge1$ since $\frac\pi2\sqrt{D/\epsilon}>1$), $D\nu_1\le D\pi^2/(4(R+1)^2)\le\epsilon$, so \eqref{eq:st-window-adm} holds, with replay as
after Lemma~\ref{lem:st-radial}; finally $2R-1=2(R+1)-3<\pi\sqrt{D/\epsilon}-1$.
\end{proof}

\paragraph{Sample-based versions.} Every stall of Corollaries~\ref{cor:st-tree}--\ref{cor:st-ZD} is a window trajectory from $\delta_o$ with
$D^U_t(\partial_{\mathrm{in}}U)\le\nu(\partial_{\mathrm{in}}U)$ and acceptance probability at least $\frac12$ in every round (by laziness). If the
proof of $o$ is known, Proposition~\ref{prop:st-realize} with $D_0=\delta_o$ and $\epsilon'=\nu(\partial_{\mathrm{in}}U)$, and its replay paragraph,
realise these stalls for every $\epsilon\in(\nu(\partial_{\mathrm{in}}U),1)$ and every replay weight, with probability at least $1-(t+1)\delta$ over
rounds $0,\dots,t$.

\sloppy
\section{The uniform bound over \texorpdfstring{$\mathcal P_0(k)$ and the proof of Proposition~\ref{prop:boundary}(a)}{P0(k) and the proof of Proposition 5(a)}}\label{app:D_uniform}\label{sec:uniform}

Throughout Appendices~\ref{app:D_uniform} and~\ref{app:E_fixedstart}, $G=T_3^{(L)}$, $o$ is a branch vertex, the dynamics is
\eqref{eq:update} without replay, $p\in(0,1)$ is arbitrary, and the model and notation are those of Appendix~\ref{app:B_model}.

\paragraph{Roadmap of the proof of Theorem~\ref{thm:uniform}.} The prover sequence is explicit: the alternating construction
(Definition~\ref{def:un-wash}) alternates one low round, in which the knowledge set is the ball $B(o,A)$, with $m$ high rounds, in which
it is larger by $mk$. In the high rounds every vertex that carries mass has a view with constant label $p$, so every curriculum in
$\mathcal P_0(k)$ moves the mass by one fixed kernel $P^\flat$ (Lemma~\ref{lem:un-blindfold}); along the arms of $T_3^{(L)}$ this kernel
is a symmetric walk whose step law $\kappa$ is the corridor law (Lemma~\ref{lem:un-corridor}). If the corridor variance $\sigma^2$
vanishes, a static ball profile stalls the dynamics (Theorem~\ref{thm:un-B0}). Otherwise $m$ is chosen so that one cycle has variance
$\Theta(k^2)$ (Lemma~\ref{lem:un-arith}); a band of width $O(k)$ at the edge of $B(o,A)$ then sends a fixed fraction $q_0$ of its mass
within one cycle into the dead zone, where the next low round kills it (Lemma~\ref{lem:un-band}), and a cosine subsolution on the scale $L$ bounds the mass killed per
cycle by $O(k^2/L^2)$ (Lemma~\ref{lem:un-star}). These two facts bound the tail mass of every quasi-stationary distribution of the cycle by
$12000k^2/L^2$ (Theorem~\ref{thm:un-A}), and, by a monotonicity argument, the tail mass from $\delta_o$ for radially TP2 curricula
(Theorem~\ref{thm:un-B}). Proposition~\ref{prop:un-reduction} reduces Theorem~\ref{thm:uniform} to a bound on the tail mass of the low
rounds from $\delta_o$, and Appendix~\ref{app:E_fixedstart} proves this bound for every $\mathcal C\in\mathcal P_0(k)$ with $\sigma^2>0$
(Theorem~\ref{thm:fs-main}) in three steps: a weak Harnack inequality from the start (Theorem~\ref{thm:fs-WH}) reduces the conditioned
tail mass to unconditioned hitting probabilities (Proposition~\ref{prop:fs-RED}); a renewal decomposition over the relocations of band
mass reduces these to first-passage bounds for the cycle chain stopped at the band (Proposition~\ref{prop:fs-REN}); and these bounds
follow from estimates for the symmetric walk with law $\kappa$ (Proposition~\ref{prop:fs-FP}). The constants are collected in
Definition~\ref{def:un-params} and Lemma~\ref{lem:un-arith}. Section~\ref{sec:un-sel} transfers the construction to batch selection
and proves Proposition~\ref{prop:boundary}(a).

\subsection{Kernels and the geometry of \texorpdfstring{$T_3^{(L)}$}{T3(L)}}\label{sec:un-conventions}\label{sec:un-structure}

\paragraph{Kernels.} For $\mathcal C\in\mathcal P_0(k)$ and a profile $s:V\to[0,1]$ the induced, accepted and flat kernels are
\[
\mathcal C^s(x,\cdot):=\mathcal C_{(B(x,k),x,s|_{B(x,k)})},\quad
P^{\mathrm{acc}}_s(x,y):=\mathcal C^s(x,y)\,\mathbf 1[s(y)\ge p/2],\quad
P^\flat(x,\cdot):=\mathcal C_{(B(x,k),x,\mathbf p)},
\]
where $\mathbf p$ is the constant label $p$ on $B(x,k)$ and laws on $B(x,k)$ are extended by $0$ to $V$. By (L), $\mathcal C^s(x,\cdot)$
and $P^\flat(x,\cdot)$ are probabilities supported in $B(x,k)$ and $P^{\mathrm{acc}}_s$ is substochastic. By (M) and (T) the proposal
kernel of round $t$ is $\mathcal C^{s_t}$, so \eqref{eq:update} reads $D_{t+1}=D_tP^{\mathrm{acc}}_{s_t}/(D_tP^{\mathrm{acc}}_{s_t}\mathbf 1)$.
Measures are row vectors and functions column vectors. A vertex $x$ has a flat view under $s$ if $s\equiv p$ on $B(x,k)$.

\begin{lemma}[Flat views]\label{lem:un-flat}
If $x$ has a flat view under $s$, then $\mathcal C^s(x,\cdot)=P^{\mathrm{acc}}_s(x,\cdot)=P^\flat(x,\cdot)$; every proposal from $x$ is kept.
\end{lemma}

\begin{proof}
The view of $x$ is $(B(x,k),x,\mathbf p)$, and its proposal law is supported in $B(x,k)$, where $s=p\ge p/2$.
\end{proof}

\begin{lemma}[Unnormalised representation]\label{lem:un-products}
Let $W_0,W_1,\dots$ be nonnegative matrices on $V$ with finitely supported rows, $D_0\in\Delta(V)$ finitely supported,
$D_{t+1}=D_tW_t/(D_tW_t\mathbf 1)$ whenever the denominator is positive, and $\mu_t:=D_0W_0\cdots W_{t-1}$. If $\mu_j\mathbf 1>0$ for
all $j\le t$, then rounds $0,\dots,t-1$ are defined and $D_j=\mu_j/(\mu_j\mathbf 1)$ for $j\le t$. If every $W_j$ is substochastic,
$\mu_t\mathbf 1>0$ implies $\mu_j\mathbf 1>0$ for $j\le t$.
\end{lemma}

\begin{proof}
Induction: if $D_j=\mu_j/(\mu_j\mathbf 1)$ with $j<t$, the denominator of round $j$ is $\mu_{j+1}\mathbf 1/\mu_j\mathbf 1>0$ and
$D_{j+1}=\mu_{j+1}/(\mu_{j+1}\mathbf 1)$. Substochastic factors give $\mu_{j+1}\mathbf 1\le\mu_j\mathbf 1$.
\end{proof}

For $x\in V$ let $r(x):=\mathrm{dist}(o,x)$ (the level) and $S_r:=\{x:r(x)=r\}$; functions of the form $f\circ r$ are radial, and
$\Gamma$ is the group of automorphisms of $T_3^{(L)}$ that fix $o$. Branch vertices have degree $3$, the others degree $2$. A
subdivided edge is the path $z_0,\dots,z_L$ that replaces an edge of $T_3$, oriented so that $r(z_L)\ge r(z_0)$; an arm is a subdivided
edge with $z_0=o$.

\begin{lemma}[Geometry of $T_3^{(L)}$]\label{lem:un-geometry}
Let $L\ge1$.
\begin{enumerate}
\item[(a)] Every edge joins $S_r$ and $S_{r+1}$ for some $r$. Every $x\in S_r$, $r\ge1$, has one neighbour in $S_{r-1}$ (its parent)
and $c(r)$ in $S_{r+1}$ (its children), where $c(r)=2$ if $L\mid r$ and $c(r)=1$ otherwise; $o$ has $c(0):=3$ children. On a
subdivided edge $r(z_i)=r(z_0)+i$ and $L\mid r(z_0)$; the branch vertices are the vertices whose level is a multiple of $L$, and every
$x$ with $1\le r(x)\le L-1$ is the vertex $z_{r(x)}$ of an arm.
\item[(b)] $\Gamma$ acts transitively on every $S_r$.
\item[(c)] Let $k\ge1$; call $x$ a corridor vertex if $\mathrm{dist}(x,b)>k$ for every branch vertex $b$. Then $x$ is a corridor vertex
if and only if $x=z_i$ on a subdivided edge with $k<i<L-k$; in that case $B(x,k)=\{z_{i+d}:|d|\le k\}$ and induces the path
$z_{i-k},\dots,z_{i+k}$. In particular the vertex at level $\ell\in[k+1,L-k-1]$ of an arm is a corridor vertex, and corridor
vertices exist if and only if $L\ge2k+2$.
\end{enumerate}
\end{lemma}

\begin{proof}
(a) In a tree, for every edge $\{x,y\}$ one endpoint lies on the path from $o$ to the other, so the levels differ by $1$; and a vertex
$x\ne o$ with two neighbours of level $r(x)-1$ would have two paths from $o$. So $x$ has one parent and $\deg(x)-1$ children. Paths
between branch vertices subdivide paths of $T_3$, so a branch vertex at $T_3$-distance $h$ from $o$ has level $hL$; the endpoints of an
edge of $T_3$ have $T_3$-distances to $o$ that differ by $1$ (the first claim for $T_3$), so $r(z_L)-r(z_0)=L$ and $L\mid r(z_0)$. Each
of the $L$ edges changes the level by $\pm1$ and the total change is $L$, so $r(z_i)=r(z_0)+i$; hence subdivision vertices have levels
not divisible by $L$, which gives $c(r)$. If $1\le r(x)\le L-1$, then $x=z_i$ with $L\mid r(z_0)<L$, so $z_0=o$.

(b) Let $x,y\in S_r$ with paths $o=x_0,\dots,x_r=x$ and $o=y_0,\dots,y_r=y$. Define $\varphi$ level by level: $\varphi(o)=o$, and once
$\varphi$ is a bijection of $S_j$, map the $c(j)$ children of each $w\in S_j$ bijectively onto the $c(j)$ children of $\varphi(w)$,
mapping $x_{j+1}$ to $y_{j+1}$ when $w=x_j$ (then $\varphi(w)=y_j$ by induction). Children of distinct vertices are distinct and every
vertex of $S_{j+1}$ is a child, so $\varphi$ is a bijection of each level; it maps parent--child pairs onto parent--child pairs, hence by
(a) it is an automorphism, and it fixes $o$ and maps $x$ to $y$.

(c) Branch vertices are not corridor vertices. A subdivision vertex is $z_i$, $1\le i\le L-1$, on exactly one subdivided edge, and
$z_1,\dots,z_{L-1}$ have degree $2$, so every path from $z_i$ to a branch vertex passes through $z_0$ or $z_L$, at distances $i$ and $L-i$.
Hence $z_i$ is a corridor vertex if and only if $\min(i,L-i)>k$, and then paths of length at most $k$ from $z_i$ move along
$z_1,\dots,z_{L-1}$, which gives $B(z_i,k)$ and its induced path. For an arm $r(z_\ell)=\ell$ by (a).
\end{proof}

For a corridor vertex $x=z_i$ and $|d|\le k$ write $x_d:=z_{i+d}$, the vertex of $B(x,k)$ at level $r(x)+d$.

\begin{lemma}[Corridor law]\label{lem:un-corridor}
Let $k\ge1$, $L\ge2k+2$ and $\mathcal C\in\mathcal P_0(k)$. There is a probability law $\kappa$ on $\{-k,\dots,k\}$ with
$\kappa(d)=\kappa(-d)$ such that $P^\flat(x,\{x_d\})=\kappa(d)$ for every corridor vertex $x$ and $|d|\le k$. Its variance
$\sigma^2:=\sum_dd^2\kappa(d)$ satisfies $0\le\sigma^2\le k^2$.
\end{lemma}

\begin{proof}
Fix a corridor vertex $x^\circ$ and put $\kappa(d):=P^\flat(x^\circ,\{x^\circ_d\})$. For a corridor vertex $x$, $x^\circ_d\mapsto x_d$ is
an isomorphism of the induced paths (Lemma~\ref{lem:un-geometry}(c)) with $x^\circ\mapsto x$; both views carry the label $p$, so (E) gives
$P^\flat(x,\{x_d\})=\kappa(d)$. The reflection $x^\circ_d\mapsto x^\circ_{-d}$ is an automorphism of the view at $x^\circ$, so
$\kappa(-d)=\kappa(d)$. By (L), $\kappa$ is a probability on $\{-k,\dots,k\}$.
\end{proof}

The law $\kappa$ and the number $\sigma^2$ are the corridor law and the corridor variance of $\mathcal C$ (for the given $p$). At views
with non-constant labels the curriculum is constrained only by (E) and (L).

\begin{lemma}[Radial lumping]\label{lem:un-lumping}
Let $\mathcal C\in\mathcal P_0(k)$, $s=f\circ r$ a radial profile, $\omega:[0,1]\to[0,\infty)$, and $W_s(x,y):=\mathcal C^s(x,y)\,\omega(s(y))$
(for $\omega=\mathbf 1[\,\cdot\ge p/2]$, $W_s=P^{\mathrm{acc}}_s$).
\begin{enumerate}
\item[(a)] $W_s(\varphi x,\varphi y)=W_s(x,y)$ and $\mathcal C^s(\varphi x,\varphi y)=\mathcal C^s(x,y)$ for $x,y\in V$, $\varphi\in\Gamma$.
\item[(b)] $\mathcal R^{\mathrm{prop}}_s(r,r'):=\mathcal C^s(x,S_{r'})$ and $\mathcal R^\omega_s(r,r'):=W_s(x,S_{r'})$ do not depend on
$x\in S_r$; $\mathcal R^\omega_s(r,r')=\mathcal R^{\mathrm{prop}}_s(r,r')\,\omega(f(r'))$, and $\mathcal R^{\mathrm{prop}}_s(r,r')=0$ unless $|r-r'|\le k$.
\item[(c)] If $\mu\ge0$ is finitely supported and $\Gamma$-invariant ($\mu\circ\varphi=\mu$ for $\varphi\in\Gamma$), so is $\mu W_s$, and
$(\mu W_s)(S_{r'})=\sum_r\mu(S_r)\,\mathcal R^\omega_s(r,r')$.
\item[(d)] A $\Gamma$-invariant measure satisfies $\mu(x)=\mu(S_{r(x)})/|S_{r(x)}|$, so it is determined by its level masses $\bar\mu(r):=\mu(S_r)$.
\end{enumerate}
\end{lemma}

\begin{proof}
(a) $\varphi$ preserves distance and fixes $o$, so it restricts to an isomorphism of the labelled rooted views of $x$ and $\varphi x$ under
$s$ (as $s\circ\varphi=s$); (E) gives the claim for $y\in B(x,k)$, both sides vanish otherwise by (L), and $\omega(s(\varphi y))=\omega(s(y))$.
(b) For $x,x'\in S_r$ take $\varphi\in\Gamma$ with $\varphi x=x'$ (Lemma~\ref{lem:un-geometry}(b)); as $\varphi(S_{r'})=S_{r'}$, (a) gives
$W_s(x',S_{r'})=W_s(x,S_{r'})$. On $S_{r'}$, $\omega(s(y))=\omega(f(r'))$; and $|r(y)-r(x)|\le k$ for $y\in B(x,k)$.
(c) By (a), $(\mu W_s)(\varphi y)=\sum_{x'}\mu(\varphi x')W_s(\varphi x',\varphi y)=(\mu W_s)(y)$; summing over $y\in S_{r'}$ and grouping
$x$ by level gives the formula by (b). (d) By Lemma~\ref{lem:un-geometry}(b), $\mu$ is constant on the finite set $S_r$.
\end{proof}

We call $\mathcal R^{\mathrm{prop}}_s$ the proposal level kernel and $\mathcal R_s:=\mathcal R^\omega_s$ with $\omega=\mathbf 1[\,\cdot\ge p/2]$ the
filtered level kernel of $s$.

\subsection{The alternating construction and its parameters}\label{sec:un-wash}\label{sec:un-params}

\begin{remark}[A fixed window does not suffice for $\mathcal P_0(k)$]\label{rem:un-static}
Let $k\ge1$ and $G$ be infinite, connected and locally finite, and let $\mathcal C^{\min}$ propose, at a labelled rooted view $v$ of
radius $k$, the uniform law on the vertices of $v$ whose label is the smallest label of $v$ that is at least $p/2$ (the point mass at the
root if there is none). View isomorphisms preserve this set and the root, so $\mathcal C^{\min}\in\mathcal P_0(k)$. Let the prover hold a
finite window $U$ (Lemma~\ref{lem:st-window}) with $D_0$ supported in $U$. A vertex $x\in U$ whose ball meets $\partial_{\mathrm{in}}U$
proposes in $\partial_{\mathrm{in}}U$; every other $x\in U$ has $B(x,k)\subseteq\mathrm{int}\,U$ (Lemma~\ref{lem:st-window}(a)) and proposes
uniformly on $B(x,k)\supseteq N(x)$; every proposal is kept. So $D_t=D_0Q_U^{\,t}$ for a stochastic matrix $Q_U$ on $U$ for which
$\partial_{\mathrm{in}}U$ is absorbing. Every component of $U$ contains a vertex of $\partial_{\mathrm{in}}U$ (as $G$ is connected and
infinite). Along a simple path in $U$ towards it, the next vertex is proposed with positive probability while the current ball does not
meet $\partial_{\mathrm{in}}U$, and the next proposal lies in $\partial_{\mathrm{in}}U$ once it does; so
$c_U:=\min_{x\in U}Q_U^{\,|U|}(x,\partial_{\mathrm{in}}U)>0$. Hence
$D_t(K_t)=D_t(\mathrm{int}\,U)\le(1-c_U)^{\lfloor t/|U|\rfloor}\to0$, and holding $U$ is admissible at no level $\epsilon<1$. The construction
below alternates two ball profiles instead.
\end{remark}

\paragraph{Ball profiles.} For an integer $N\ge0$ the ball profile $s^{[N]}$ is $p$ on $B(o,N)$, $p/2$ on $S_{N+1}$ and $0$ elsewhere.
It is $p/2$-Lipschitz (an edge joins consecutive levels, Lemma~\ref{lem:un-geometry}(a)), $\{s^{[N]}\ge p\}=B(o,N)$ and
$\{s^{[N]}\ge p/2\}=B(o,N+1)$; balls are finite.

\begin{definition}[Parameters]\label{def:un-params}
Let $k\ge1$, $L\ge400k$, and let $\mathcal C\in\mathcal P_0(k)$ have corridor variance $\sigma^2>0$. Put
\[
R:=\lfloor L/2\rfloor,\ \ a:=2k,\ \ A:=\lfloor 3L/4\rfloor,\ \ M:=\lfloor L/8\rfloor,\ \
m:=\Big\lceil\frac{4(3k+1)^2}{\sigma^2}\Big\rceil,\ \ A_{\mathrm{hi}}:=A+1+mk,\ \ V:=(m+1)\sigma^2,
\]
\[
\theta:=\frac{\pi}{2(R-a)},\qquad\tilde\mu:=1-\frac{\theta^2\sigma^2}{2},\qquad q_0:=\frac{9}{256},\qquad \bar C_k:=12000\,k^2 .
\]
If $L\ge10^5k$, Appendix~\ref{app:E_fixedstart} uses in addition
\[
q:=\frac Lk,\ \ \delta:=\Big\lfloor\frac L{16}\Big\rfloor,\ \ R_\sharp:=A-\delta,\ \ \theta_\sharp:=\frac{\pi}{2(R_\sharp-a)},\ \
\tilde\mu_\sharp:=1-\frac{\theta_\sharp^2\sigma^2}{2},\ \ \tilde\mu_{\sharp,c}:=\tilde\mu_\sharp^{\,m+1},
\]
\[
M':=\Big\lfloor\frac L{32}\Big\rfloor,\ \ J:=\Big\lceil\frac{L^2}{V}\Big\rceil,\ \
\omega:=\frac{\pi}{2(3\delta+1)},\ \ x_0:=12k,\ \ \eta_F:=e^{-q/24}.
\]
\end{definition}

\begin{lemma}[Arithmetic of the parameters]\label{lem:un-arith}
Under Definition~\ref{def:un-params}:
\begin{enumerate}
\item[(a)] $3k<R$, $R+k\le A$ and $R<L-k$;
\item[(b)] $A+M\le L-k-1$ and $A-2k+2-M\ge k+1$;
\item[(c)] $R-a\ge0.45L$ and $\theta k\le\pi/360$;
\item[(d)] $M\ge L/9$;
\item[(e)] $4(3k+1)^2\le m\sigma^2\le65k^2$ and $4(3k+1)^2<V\le66k^2$; hence $\sqrt V\ge6k+2\ge8$ and $k/\sqrt V<1/6$;
\item[(f)] $\tilde\mu\in(1/2,1]$ and $1-\tilde\mu^{m+1}\le(m+1)(1-\tilde\mu)\le403\,k^2/L^2$;
\item[(g)] $65k^2/M^2\le q_0$, and $(1+1/q_0)\cdot403\,k^2\le \bar C_k$;
\item[(h)] $m\ge37$, $L\ge3k+1\ge2k+2$, and $\bar C_k/L^2\le3/40$.
\end{enumerate}
If moreover $L\ge10^5k$:
\begin{enumerate}
\item[(i)] $L/16-1\le\delta\le L/16$, $L/32-1\le M'\le L/32$, $\delta\ge6249k$, $M'\ge3124k$, and $3L/4-3/4\le A\le3L/4$;
\item[(j)] $3k<R_\sharp\le A-k$, $0.6874725L\le R_\sharp-a\le0.6875L$, $2.2847\le\theta_\sharp L\le2.2850$, $\theta_\sharp^2L^2\le5.221$,
$\theta_\sharp k\le2.3\cdot10^{-5}$;
\item[(k)] $8.3771\le\omega L\le8.3785$; $\omega(u+\delta)\in(0,\pi]$ for $u\in[-M',2\delta+M']$ and $\omega(u+\delta)\in(0,\pi/2]$
for $u\in[-k,2\delta+1]$; $\Psi_{\min}:=\sin(\omega(\delta-k))\ge0.4998$; $h_{\mathsf C}:=\sin(\theta_\sharp\delta)\ge0.1422$;
\item[(l)] $L^2/V\le J\le L^2/V+1$, $J\ge1.5\cdot10^8$, and $J\log(1/\tilde\mu_{\sharp,c})\le2.6106$.
\end{enumerate}
\end{lemma}

\begin{proof}
We use $\lfloor y\rfloor\ge y-1$, so $R\ge(L-1)/2$, $A\ge(3L-3)/4$, $M\ge(L-7)/8$, and $L\ge400k$, $k\ge1$.
(a) $R\ge(L-1)/2>3k$; $R+k\le L/2+k\le(3L-3)/4\le A$ as $L\ge4k+3$; $R\le L/2<L-k$.
(b) $A+M\le7L/8\le L-k-1$ as $L/8\ge k+1$; $A-2k+2-M\ge(3L-3)/4-2k+2-L/8=5L/8+5/4-2k\ge k+1$ as $5L/8\ge3k$.
(c) $R-a\ge(L-1)/2-2k\ge0.45L$ as $0.05L\ge20k\ge2k+1/2$; then $\theta k\le\pi k/(0.9L)\le\pi/360$.
(d) $(L-7)/8\ge L/9$ as $L\ge63$.
(e) $m\ge4(3k+1)^2/\sigma^2$ gives the lower bounds ($V=m\sigma^2+\sigma^2>m\sigma^2$). From $m<4(3k+1)^2/\sigma^2+1$,
$m\sigma^2<4(3k+1)^2+\sigma^2$ and $V<4(3k+1)^2+2\sigma^2$; use $(3k+1)^2\le16k^2$ and $\sigma^2\le k^2$ (Lemma~\ref{lem:un-corridor}).
Finally $\sqrt V>2(3k+1)$ and $k/(6k+2)<1/6$.
(f) By (c), $\theta^2\sigma^2/2\le(\pi/360)^2/2<1/2$. For $u\in[0,1]$, $1-u^n\le n(1-u)$. With (c) and (e),
$(m+1)(1-\tilde\mu)=V\pi^2/(8(R-a)^2)\le66\pi^2k^2/(1.62L^2)\le403k^2/L^2$, as $66\pi^2/1.62=402.09\ldots$.
(g) By (d), $65k^2/M^2\le5265k^2/L^2\le9/256$ as $L^2\ge160000k^2$; and $(1+256/9)\cdot403=11866.1\ldots\le12000$.
(h) $m\ge4(3+1/k)^2>36$; $\bar C_k/L^2\le12000/160000$.
(i) $L/16\ge6250k$, $L/32\ge3125k$, $A=\lfloor3L/4\rfloor$.
(j) $R_\sharp=A-\delta\le A-k$ and $R_\sharp\ge3L/4-3/4-L/16>3k$. Next $R_\sharp-a\ge11L/16-3/4-2k\ge(11/16-2.75/q)L$ and
$R_\sharp-a\le3L/4-(L/16-1)-2k\le11L/16$. Hence $\theta_\sharp L\in[\pi/1.375,\pi/1.374945]\subseteq[2.2847,2.2849]$,
$(\pi/1.374945)^2\le5.2208$, and $\theta_\sharp k\le2.285/q$.
(k) $3\delta+1\in[3L/16-2,3L/16+1]$ by (i), which gives the bounds on $\omega L$. For $u\in[-M',2\delta+M']$, $0<\delta-M'\le u+\delta\le3\delta+M'
\le2(3\delta+1)$; for $u\in[-k,2\delta+1]$, $u+\delta\in(0,3\delta+1]$. Next $\omega(\delta-k)\ge8.3771(1/16-2/q)\ge0.5234$, which is at
most $\pi/2$, and $\sin0.5234\ge0.4998$; $\theta_\sharp\delta\ge2.2847(1/16-1/q)\ge0.14277$ and $\sin0.14277\ge0.1422$.
(l) $J\ge L^2/V\ge q^2/66$. With $y:=\theta_\sharp^2\sigma^2/2\le10^{-9}$ (by (j)), $\log\frac1{1-y}\le y(1+y)$, so
$J\log(1/\tilde\mu_{\sharp,c})=J(m+1)\log\frac1{1-y}\le(L^2/V+1)\frac{\theta_\sharp^2V}2(1+10^{-9})\le\frac{5.221}2(1+66q^{-2})(1+10^{-9})\le2.6106$.
\end{proof}

\begin{definition}[Alternating construction]\label{def:un-wash}
Let $k,L,\mathcal C$ be as in Definition~\ref{def:un-params}. The low and high profiles are $s^{\mathrm{lo}}:=s^{[A]}$ and
$s^{\mathrm{hi}}:=s^{[A_{\mathrm{hi}}]}$, and the alternating construction is the prover sequence $s_t:=s^{\mathrm{lo}}$ if $(m+1)\mid t$ and
$s_t:=s^{\mathrm{hi}}$ otherwise. Round $n(m+1)$ is the $n$-th low round, and rounds $n(m+1),\dots,n(m+1)+m$ form the $n$-th cycle. Put
\[
P_{\mathrm{lo}}:=P^{\mathrm{acc}}_{s^{\mathrm{lo}}},\quad P_{\mathrm{hi}}:=P^{\mathrm{acc}}_{s^{\mathrm{hi}}},\quad Q:=P_{\mathrm{lo}}P_{\mathrm{hi}}^m,
\quad V_Q:=B(o,A_{\mathrm{hi}}),
\]
and the interior, band, dead zone and tail
\[
\mathsf I:=\{r\le A-k\},\quad\mathsf B:=\{A-k<r\le A+k+1\},\quad\mathsf Z:=\{r\ge A+k+2\},\quad\mathsf T:=\{r>A\}\subseteq\mathsf B\cup\mathsf Z.
\]
For a cold start $D_0$ put $\mu_n:=D_0Q^n$, $\Lambda_n:=\mu_n\mathbf 1$ and, when $\Lambda_n>0$, $D^{\mathrm{lo}}_n:=\mu_n/\Lambda_n$. For
finitely supported $\mu\ge0$ put $g(\mu):=\sum_{x\in\mathsf B}\mu(x)(P_{\mathrm{lo}}\mathbf 1)(x)$ and $b(\mu):=\mu(\mathsf B)-g(\mu)$, the
band mass kept and rejected in a low round.
\end{definition}

In the low round the prover forgets the levels $A+1,\dots,A_{\mathrm{hi}}$, which it knew in the preceding high rounds; the knowledge
set is $B(o,A)$ in low rounds and $B(o,A_{\mathrm{hi}})$ in high rounds, so $\sup_t|K_t|=|B(o,A_{\mathrm{hi}})|<\infty$.

\begin{lemma}[High-round support]\label{lem:un-blindfold}
\begin{enumerate}
\item[(a)] For every $x$, $P_{\mathrm{lo}}(x,\cdot)$ is supported in $B(o,A+1)$.
\item[(b)] Let $\mu\ge0$ be finitely supported in $B(o,A+1)$ and $0\le j\le m$. Then $\mu P_{\mathrm{hi}}^{\,j}=\mu(P^\flat)^j$, this measure
is supported in $B(o,A+1+jk)$, and $\mu P_{\mathrm{hi}}^{\,j}\mathbf 1=\mu\mathbf 1$. For $j\le m-1$ its support lies in
$B(o,A_{\mathrm{hi}}-k)$, where every vertex has a flat view under $s^{\mathrm{hi}}$.
\item[(c)] For every $x$, $Q(x,\cdot)$ is supported in $V_Q$, and $\mu Q\mathbf 1=\mu P_{\mathrm{lo}}\mathbf 1$ for finitely supported $\mu\ge0$.
\end{enumerate}
\end{lemma}

\begin{proof}
(a) A proposal $y$ is kept in a low round if and only if $r(y)\le A+1$. (b) Induction on $j$: if $\mu(P^\flat)^j$ ($j<m$) is supported in
$B(o,A+1+jk)\subseteq B(o,A_{\mathrm{hi}}-k)$, every vertex $z$ of its support has $B(z,k)\subseteq B(o,A_{\mathrm{hi}})$, where $s^{\mathrm{hi}}=p$, so
$P_{\mathrm{hi}}(z,\cdot)=P^\flat(z,\cdot)$ (Lemma~\ref{lem:un-flat}), a probability supported in $B(o,A+1+(j+1)k)$. (c) Apply (b) with
$j=m$ to $\delta_xP_{\mathrm{lo}}$ and $\mu P_{\mathrm{lo}}$, using (a) and $A+1+mk=A_{\mathrm{hi}}$.
\end{proof}

So in the high rounds every curriculum in $\mathcal P_0(k)$ moves all of its mass by its flat kernel; this is where memorylessness and (E) enter.

\begin{lemma}[Basic properties]\label{lem:un-wash-basic}
Let $D_0$ be finitely supported and $N\ge1$ with $\Lambda_N>0$. Then $\Lambda_0\ge\dots\ge\Lambda_N>0$, every round $t<N(m+1)$ is
defined, and for $n<N$, $D_{n(m+1)}=D^{\mathrm{lo}}_n$ and $D_{n(m+1)+j}=\mu_nP_{\mathrm{lo}}P_{\mathrm{hi}}^{\,j-1}/\Lambda_{n+1}$ for $1\le j\le m$,
a measure supported in $B(o,A_{\mathrm{hi}}-k)$.
\end{lemma}

\begin{proof}
By (M) and (T) the matrices of Lemma~\ref{lem:un-products} in each cycle are $P_{\mathrm{lo}},P_{\mathrm{hi}},\dots,P_{\mathrm{hi}}$, and the product
at round $n(m+1)+j$ is $\mu_n$ for $j=0$ and $\mu_nP_{\mathrm{lo}}P_{\mathrm{hi}}^{\,j-1}$ for $1\le j\le m+1$. By
Lemma~\ref{lem:un-blindfold}(a),(b), $\mu_nP_{\mathrm{lo}}P_{\mathrm{hi}}^{\,j}\mathbf 1=\mu_nP_{\mathrm{lo}}\mathbf 1=\Lambda_{n+1}$ for $0\le j\le m$, with
support in $B(o,A_{\mathrm{hi}}-k)$ for $j\le m-1$, and $\Lambda_{n+1}\le\Lambda_n$ as $P_{\mathrm{lo}}$ is substochastic. Lemma~\ref{lem:un-products}
gives the claim.
\end{proof}

\begin{lemma}[Low-round accounting]\label{lem:un-accounting}
\begin{enumerate}
\item[(i)] For $x\in\mathsf I$, $P_{\mathrm{lo}}(x,\cdot)=P^\flat(x,\cdot)$, a probability supported in $B(o,A)$.
\item[(ii)] For $x\in\mathsf Z$, $P_{\mathrm{lo}}(x,\cdot)=0$.
\item[(iii)] For $x\in\mathsf B$, $(P_{\mathrm{lo}}\mathbf 1)(x)\in[0,1]$ and $P_{\mathrm{lo}}(x,\cdot)$ is supported on levels in $(A-2k,A+1]$.
\end{enumerate}
Consequently, for every finitely supported $\mu\ge0$,
\begin{equation}\label{eq:un-accounting}
\mu Q\mathbf 1=\mu\mathbf 1-\mu(\mathsf Z)-b(\mu),\qquad b(\mu)\ge0,\qquad \mu(\mathsf T)\le g(\mu)+b(\mu)+\mu(\mathsf Z).
\end{equation}
\end{lemma}

\begin{proof}
(i) $B(x,k)\subseteq B(o,A)$, where $s^{\mathrm{lo}}=p$; apply Lemma~\ref{lem:un-flat}. (ii) $B(x,k)\subseteq\{r\ge A+2\}$, where
$s^{\mathrm{lo}}=0$. (iii) A kept proposal $y$ from $x$ has $A-2k<r(x)-k\le r(y)\le A+1$. By Lemma~\ref{lem:un-blindfold}(c),
$\mu Q\mathbf 1=\mu P_{\mathrm{lo}}\mathbf 1=\mu(\mathsf I)+g(\mu)$, while $\mu\mathbf 1=\mu(\mathsf I)+\mu(\mathsf B)+\mu(\mathsf Z)$; and
$\mathsf T\subseteq\mathsf B\cup\mathsf Z$.
\end{proof}

\subsection{The band lemma and the cosine subsolution}\label{sec:un-band}

\paragraph{The step chain.} Let $(\xi_j,U_j)_{j\ge1}$ be i.i.d., with $\xi_j\sim\kappa$ and $U_j$ uniform on $[0,1]$ independent of $\xi_j$, and fix
for every vertex $x$ a map $F_x:[0,1]\to V$ such that $F_x(U_1)$ has law $P^\flat(x,\cdot)$. For a start vertex $x^{(0)}$ put $X_0:=x^{(0)}$ and
$X_{j+1}:=(X_j)_{\xi_{j+1}}$ if $X_j$ is a corridor vertex, $X_{j+1}:=F_{X_j}(U_{j+1})$ otherwise; let $\mathcal G_j:=\sigma(\xi_i,U_i:i\le j)$.
Given $X_j=x$, $X_{j+1}$ is a function of $x$ and $(\xi_{j+1},U_{j+1})$, which is independent of $\mathcal G_j$, and its law is
$\sum_d\kappa(d)\delta_{x_d}=P^\flat(x,\cdot)$ if $x$ is a corridor vertex (Lemma~\ref{lem:un-corridor}) and $P^\flat(x,\cdot)$ otherwise. So $(X_j)$
is a Markov chain with kernel $P^\flat$ with respect to $(\mathcal G_j)$, and, by the strong Markov property of an i.i.d.\ sequence, for every
stopping time $\varsigma$ of $(\mathcal G_j)$, on $\{\varsigma<\infty\}$ the sequence $(\xi_{\varsigma+i},U_{\varsigma+i})_{i\ge1}$ is independent of
$\mathcal G_\varsigma$ and has the law of $(\xi_i,U_i)_{i\ge1}$.

\begin{lemma}[Band to dead zone]\label{lem:un-band}
\begin{enumerate}
\item[(a)] If $A-2k<r(y)\le A+1$, then $(P^\flat)^m(y,\mathsf Z)\ge q_0$.
\item[(b)] For every finitely supported $\mu\ge0$, $(\mu Q)(\mathsf Z)\ge q_0\,g(\mu)$.
\end{enumerate}
\end{lemma}

\begin{proof}
The idea is that $m$ flat steps from the band are, with probability bounded below, a symmetric walk on one arm that moves by more than
$3k+1$ outwards.

(a) \emph{Position.} As $1\le r(y)\le A+1<L$, $y=z_{r(y)}$ on an arm (Lemma~\ref{lem:un-geometry}(a)), and by
Lemma~\ref{lem:un-arith}(b) every level $\ell\in[A-2k+2-M,A+M]$ lies in $[k+1,L-k-1]$, so $z_\ell$ is a corridor vertex and
$(z_\ell)_d=z_{\ell+d}$.

\emph{Coupling.} Run the step chain from $X_0=y$; then $\Pr(X_m\in\mathsf Z)=(P^\flat)^m(y,\mathsf Z)$.

\emph{Good event.} Let $S_j:=\xi_1+\dots+\xi_j$ and $\mathcal G:=\{\max_{1\le j\le m}|S_j|<M\}$. On $\mathcal G$, induction gives
$X_j=z_{r(y)+S_j}$ for $j\le m$: if it holds for $j<m$, the level $r(y)+S_j$ lies in $[A-2k+2-M,A+M]$, so $X_j$ is a corridor vertex and
$X_{j+1}=z_{r(y)+S_{j+1}}$. Hence on $\mathcal G\cap\{S_m\ge3k+1\}$, $r(X_m)\ge(A-2k+1)+(3k+1)=A+k+2$, that is $X_m\in\mathsf Z$.

\emph{Anti-concentration.} $\mathbb E\xi_1=0$, $\mathbb E\xi_1^2=\sigma^2$ and $\xi_1^4\le k^2\xi_1^2$, so expanding,
$\mathbb ES_m^4=m\mathbb E\xi_1^4+3m(m-1)\sigma^4\le mk^2\sigma^2+3m^2\sigma^4$. The Paley--Zygmund inequality
$\Pr(X>\tau\mathbb EX)\ge(1-\tau)^2(\mathbb EX)^2/\mathbb EX^2$ with $X=S_m^2$, $\tau=1/4$ gives
\[
\Pr\big(S_m^2>\tfrac14m\sigma^2\big)\ \ge\ \frac{9/16}{3+k^2/(m\sigma^2)}\ \ge\ \frac9{64},
\]
as $m\sigma^2\ge4(3k+1)^2\ge k^2$ (Lemma~\ref{lem:un-arith}(e)). On this event $|S_m|>\frac12\sqrt m\sigma\ge3k+1$, and $S_m$ is symmetric, so
$\Pr(S_m\ge3k+1)\ge\frac9{128}$.

\emph{Maximal inequality.} Kolmogorov's inequality and Lemma~\ref{lem:un-arith}(e),(g) give $\Pr(\mathcal G^c)\le m\sigma^2/M^2\le65k^2/M^2\le q_0$.
Hence $(P^\flat)^m(y,\mathsf Z)\ge\frac9{128}-\frac9{256}=q_0$.

(b) Let $\nu':=(\mu\mathbf 1_{\mathsf B})P_{\mathrm{lo}}$, of mass $g(\mu)$ and supported on levels in $(A-2k,A+1]$
(Lemma~\ref{lem:un-accounting}(iii)). As $Q\ge0$, $(\mu Q)(\mathsf Z)\ge(\nu'P_{\mathrm{hi}}^m)(\mathsf Z)=(\nu'(P^\flat)^m)(\mathsf Z)\ge q_0g(\mu)$
by Lemma~\ref{lem:un-blindfold}(b) and (a).
\end{proof}

\begin{lemma}[Cosine subsolution on $T_3^{(L)}$]\label{lem:un-star}
Let $R'$ be an integer with $3k<R'\le A-k$ such that $\theta':=\pi/(2(R'-a))$ satisfies $\theta'k\le1/2$, and put
$\tilde\mu':=1-\theta'^2\sigma^2/2\in[7/8,1]$. Let $H(u):=1$ for $u\le a$, $H(u):=\cos(\theta'(u-a))$ for $a<u<R'$, $H(u):=0$ for
$u\ge R'$, and $h:=H\circ r$. Then $0\le h\le1$, $h(o)=1$, $h$ is nonincreasing in $r$, and for every $x\in V$
\[
(P_{\mathrm{lo}}h)(x)\ge\tilde\mu'\,h(x),\qquad(P_{\mathrm{hi}}h)(x)\ge\tilde\mu'\,h(x);
\]
hence $W_1\cdots W_\ell h\ge\tilde\mu'^{\,\ell}h$ for every product of $\ell$ factors $W_i\in\{P_{\mathrm{lo}},P_{\mathrm{hi}}\}$. The radius $R'=R$
is admissible (then $\theta'=\theta$, $\tilde\mu'=\tilde\mu$), and from $D_0=\delta_o$,
$\delta_oQ^nP_{\mathrm{lo}}P_{\mathrm{hi}}^{\,j}\mathbf 1\ge\tilde\mu^{\,n(m+1)+1+j}$ and $\Lambda_n\ge\tilde\mu^{\,n(m+1)}>0$ for $n\ge0$,
$0\le j\le m$; so every round of the alternating construction from $\delta_o$ is defined.
\end{lemma}

\begin{proof}
$\theta'^2\sigma^2/2\le(\theta'k)^2/2\le1/8$. $\theta'(u-a)\in(0,\pi/2)$ for $a<u<R'$ and $\cos(\theta'(R'-a))=0$, so $H$ is nonincreasing with
values in $[0,1]$, and $H(0)=1$. Let $W\in\{P_{\mathrm{lo}},P_{\mathrm{hi}}\}$, $x\in V$ and $r:=r(x)$. If $r\ge R'$, then $h(x)=0\le(Wh)(x)$.
Let $r<R'$. Then $B(x,k)\subseteq B(o,R'-1+k)\subseteq B(o,A)$, so $x$ has a flat view under both profiles and $(Wh)(x)=(P^\flat h)(x)$
(Lemma~\ref{lem:un-flat}).

(i) $r\le k$. Every $y\in B(x,k)$ has $r(y)\le2k=a$, so $(P^\flat h)(x)=1=h(x)$, whatever the law $P^\flat(x,\cdot)$ is.

(ii) $k<r<R'$. As $R'\le A-k<L-k$, $x$ is the vertex at level $r\in[k+1,L-k-1]$ of an arm, a corridor vertex, and
$(P^\flat h)(x)=\sum_d\kappa(d)H(r+d)$ (Lemmas~\ref{lem:un-geometry} and~\ref{lem:un-corridor}).

(ii-1) $a<r<R'$. Let $G(u):=\cos(\theta'(u-a))$. Then $H\ge G$ on $[r-k,r+k]$: for $u\le a$, $H(u)=1$; for $a<u<R'$, equality; for
$R'\le u\le R'-1+k$, $\theta'(u-a)\in[\pi/2,\pi/2+\theta'k]\subseteq[\pi/2,\pi]$, so $G(u)\le0=H(u)$. As $\kappa$ is symmetric,
$\sum_d\kappa(d)\sin(\theta'd)=0$, and with $\cos v\ge1-v^2/2$ and $G(r)=h(x)\ge0$,
\[
(P^\flat h)(x)\ge\sum_d\kappa(d)\cos(\theta'(r-a)+\theta'd)=G(r)\sum_d\kappa(d)\cos(\theta'd)\ge\tilde\mu'\,h(x).
\]

(ii-2) $k<r\le a$. For $u\le r+k\le3k<R'$, $H(u)\ge1-\frac{\theta'^2}{2}((u-a)_+)^2$ (equality for $u\le a$, $\cos v\ge1-v^2/2$ otherwise),
and $(r+d-a)_+\le d_+$. As $\sum_d\kappa(d)d_+^2=\sigma^2/2$ by symmetry,
$(P^\flat h)(x)\ge1-\theta'^2\sigma^2/4\ge\tilde\mu'=\tilde\mu'h(x)$.

\emph{Products.} $W\ge0$ has finitely supported rows, so $f_1\ge f_2$ implies $Wf_1\ge Wf_2$; with $\tilde\mu'\ge0$, induction on $\ell$ gives
$W_1\cdots W_\ell h\ge\tilde\mu'^{\,\ell}h$. For $R'=R$: $3k<R\le A-k$ and $\theta k\le\pi/360$ (Lemma~\ref{lem:un-arith}(a),(c)).
$Q^nP_{\mathrm{lo}}P_{\mathrm{hi}}^{\,j}$ has $n(m+1)+1+j$ factors, and as $\mathbf 1\ge h\ge0$,
$\delta_oQ^nP_{\mathrm{lo}}P_{\mathrm{hi}}^{\,j}\mathbf 1\ge\tilde\mu^{\,n(m+1)+1+j}h(o)$; the same holds for $Q^n$ alone. Since every
$\Lambda_n>0$, Lemma~\ref{lem:un-wash-basic} shows that every round is defined.
\end{proof}

Only the corridor law enters the inequality; the laws at views that contain $o$ and in the band are arbitrary, and no reversibility is used.

\subsection{Quasi-stationary start}\label{sec:un-A}

Let $Q_{V_Q}:=(Q(x,y))_{x,y\in V_Q}$, a nonnegative square matrix with row sums at most $1$.

\begin{theorem}[Quasi-stationary start]\label{thm:un-A}
Let $k\ge1$, $L\ge400k$, $p\in(0,1)$, let $\mathcal C\in\mathcal P_0(k)$ on $T_3^{(L)}$ have $\sigma^2>0$, and consider the alternating
construction.
\begin{enumerate}
\item[(a)] The Perron root satisfies $\tilde\mu^{m+1}\le\lambda(Q_{V_Q})\le1$, and $Q_{V_Q}$ has a QSD.
\item[(b)] Let $\nu$ be a probability vector on $V_Q$ with $\nu Q_{V_Q}=\lambda\nu$ for some $\lambda\ge\tilde\mu^{m+1}$ (for instance
any QSD of $Q_{V_Q}$). Then
\[
\nu(\mathsf T)\ \le\ \Big(1+\frac1{q_0}\Big)(1-\lambda)\ \le\ \Big(1+\frac1{q_0}\Big)\big(1-\tilde\mu^{m+1}\big)\ \le\ \frac{\bar C_k}{L^2}=\frac{12000\,k^2}{L^2}.
\]
\item[(c)] For $\nu$ as in (b) and every $\epsilon\in[\bar C_k/L^2,1)$, the alternating construction started from $D_0=\nu$ is admissible at
level $\epsilon$; every round is defined, $D_{n(m+1)}=\nu$ for every $n$, and $\sup_t|K_t|=|B(o,A_{\mathrm{hi}})|<\infty$. Hence
$\epsilon^*_{\mathrm{div}}(\mathcal C;T_3^{(L)},\nu)\le\epsilon^*_{\mathrm{bdd}}(\mathcal C;T_3^{(L)},\nu)\le12000k^2/L^2$.
\end{enumerate}
\end{theorem}

\begin{proof}
(a) By the Perron--Frobenius theorem for nonnegative matrices (no irreducibility is needed; \citealp{hornjohnson2013}, applied to
$Q_{V_Q}^\top$), $\rho(Q_{V_Q})$ is an eigenvalue of $Q_{V_Q}^\top$ with an eigenvector $\nu\ge0$, $\nu\ne0$; normalised, $\nu$ is a QSD.
Row sums at most $1$ give $\rho(Q_{V_Q})\le1$. Since $Q(x,\cdot)$ is supported in $V_Q$ (Lemma~\ref{lem:un-blindfold}(c)),
$Q_{V_Q}^n(h|_{V_Q})=(Q^nh)|_{V_Q}$; by Lemma~\ref{lem:un-star} its value at $o$ is at least $\tilde\mu^{n(m+1)}$, and $\|h\|_\infty=1$, so
$\|Q_{V_Q}^n\|_{\infty\to\infty}\ge\tilde\mu^{n(m+1)}$; Gelfand's formula
\citep{hornjohnson2013} gives $\rho(Q_{V_Q})\ge\tilde\mu^{m+1}$.

(b) Extend $\nu$ by $0$; then $\nu Q=\lambda\nu$ on $V$. By \eqref{eq:un-accounting}, $1-\lambda=\nu(\mathsf Z)+b(\nu)$ (killing identity). By
Lemma~\ref{lem:un-band}(b), $\lambda\nu(\mathsf Z)=(\nu Q)(\mathsf Z)\ge q_0g(\nu)$, so $g(\nu)\le\nu(\mathsf Z)/q_0$. By \eqref{eq:un-accounting},
$\nu(\mathsf T)\le g(\nu)+b(\nu)+\nu(\mathsf Z)\le(1+1/q_0)(\nu(\mathsf Z)+b(\nu))=(1+1/q_0)(1-\lambda)$. The other two inequalities are
$\lambda\ge\tilde\mu^{m+1}$ and Lemma~\ref{lem:un-arith}(f),(g).

(c) $\mu_n=\lambda^n\nu$ and $\Lambda_n=\lambda^n>0$, so by Lemma~\ref{lem:un-wash-basic} every round is defined, $D_{n(m+1)}=\nu$, and in high
rounds $D_t$ is supported in $B(o,A_{\mathrm{hi}}-k)\subseteq K_t$. In low rounds $D_t(K_t^c)=\nu(\mathsf T)\le\bar C_k/L^2\le\epsilon$ by (b). The
profiles are $p/2$-Lipschitz. The level $\bar C_k/L^2<1$ (Lemma~\ref{lem:un-arith}(h)) lies in the set whose infimum is $\epsilon^*_{\mathrm{bdd}}$,
and $\epsilon^*_{\mathrm{div}}\le\epsilon^*_{\mathrm{bdd}}$ always (Appendix~\ref{app:B_model}).
\end{proof}

Time homogeneity (T) enters through Lemma~\ref{lem:un-wash-basic}: every cycle applies the same matrix $Q$, so $\nu$ is reproduced.

\subsection{Fixed start: reduction, degenerate corridor and total positivity}\label{sec:un-B}\label{sec:un-fsbl}

\begin{proposition}[Per-curriculum reduction]\label{prop:un-reduction}
Let $k\ge1$, $L\ge400k$, $p\in(0,1)$, let $\mathcal C\in\mathcal P_0(k)$ on $T_3^{(L)}$ have $\sigma^2>0$, and let
$\bar\eta:=\sup_{n\ge0}D^{\mathrm{lo}}_n(\mathsf T)$ for the alternating construction from $\delta_o$. For every $\epsilon\in[\bar\eta,1)$ this
sequence is admissible at level $\epsilon$, every round is defined, and $\sup_t|K_t|=|B(o,A_{\mathrm{hi}})|$. Hence
$\epsilon^*_{\mathrm{bdd}}(\mathcal C;T_3^{(L)},o)\le\bar\eta$.
\end{proposition}

\begin{proof}
Every round is defined (Lemma~\ref{lem:un-star}). In a low round $D_t(K_t^c)=D^{\mathrm{lo}}_n(\mathsf T)\le\bar\eta\le\epsilon$; in a high round $D_t$
is supported in $B(o,A_{\mathrm{hi}}-k)\subseteq K_t$ (Lemma~\ref{lem:un-wash-basic}); the profiles are $p/2$-Lipschitz. If $\bar\eta<1$ the level
$\bar\eta$ lies in the set whose infimum is $\epsilon^*_{\mathrm{bdd}}$; if $\bar\eta=1$ the bound holds by definition.
\end{proof}

\begin{theorem}[Degenerate corridor]\label{thm:un-B0}
Let $k\ge1$, $L\ge3k+1$, $p\in(0,1)$, and let $\mathcal C\in\mathcal P_0(k)$ on $T_3^{(L)}$ have $\sigma^2=0$. For every
$\epsilon\in[0,1)$ the constant sequence $s_t:=s^{[3k]}$ is admissible from $D_0=\delta_o$; every round is defined, every proposal is kept,
$D_t$ is supported in $B(o,2k)$ and $K_t=B(o,3k)$ for every $t$. Hence $\epsilon^*_{\mathrm{bdd}}(\mathcal C;T_3^{(L)},o)=0$.
\end{theorem}

\begin{proof}
$L\ge2k+2$, so $\kappa$ is defined, and $\sigma^2=0$ forces $\kappa=\delta_0$. By induction on $t$, $D_t$ is supported in $B(o,2k)$ and
every proposal of round $t$ is kept: for $x\in B(o,2k)$, $B(x,k)\subseteq B(o,3k)$, so $x$ has a flat view and its proposals are kept
(Lemma~\ref{lem:un-flat}); if $r(x)\le k$ they lie in $B(o,2k)$; if $k<r(x)\le2k\le L-k-1$, $x$ is a corridor vertex of an arm and
$P^\flat(x,\cdot)=\delta_x$. So the normaliser is $1$. The ball profile $s^{[3k]}$ is $p/2$-Lipschitz with $K_t=B(o,3k)$, and
$D_t(K_t)=1\ge1-\epsilon$ for every $\epsilon\in[0,1)$, including $\epsilon=0$.
\end{proof}

\paragraph{Total positivity.} A nonnegative matrix $\mathcal M$ on $\{0,\dots,N\}^2$ is TP2 if
$\mathcal M(i,j)\mathcal M(i',j')\ge\mathcal M(i,j')\mathcal M(i',j)$ for all $i<i'$, $j<j'$. For row vectors $\alpha,\beta\ge0$ write
$\alpha\le_{\mathrm{lr}}\beta$ if $\alpha(j)\beta(j')\ge\alpha(j')\beta(j)$ for all $j<j'$.

\begin{lemma}[TP2 calculus]\label{lem:un-tp2}
Let $\alpha,\beta\ge0$ be row vectors and $\mathcal M,\mathcal N\ge0$ square matrices on $\{0,\dots,N\}$.
\begin{enumerate}
\item[(T1)] If $\alpha\le_{\mathrm{lr}}\beta$ and $\mathcal M$ is TP2, then $\alpha\mathcal M\le_{\mathrm{lr}}\beta\mathcal M$.
\item[(T2)] Products of TP2 matrices are TP2; $\Delta\mathcal M\Delta'$ is TP2 for TP2 $\mathcal M$ and nonnegative diagonal $\Delta,\Delta'$;
submatrices of TP2 matrices (rows and columns in their original order) are TP2.
\item[(T3)] If $\mathcal M(i,j)=0$ whenever $|i-j|\ge2$, then $\mathcal M$ is TP2 if and only if
$\mathcal M(i,i)\mathcal M(i+1,i+1)\ge\mathcal M(i,i+1)\mathcal M(i+1,i)$ for $0\le i<N$.
\item[(T4)] $\delta_0\le_{\mathrm{lr}}\beta$ for every $\beta\ge0$. If $\alpha\le_{\mathrm{lr}}\beta$ and $\alpha\mathbf 1,\beta\mathbf 1>0$,
then $\alpha(\{j\ge c\})/\alpha\mathbf 1\le\beta(\{j\ge c\})/\beta\mathbf 1$ for every $c$.
\end{enumerate}
\end{lemma}

\begin{proof}
(T1) For $j<j'$, $(\alpha\mathcal M)(j)(\beta\mathcal M)(j')-(\alpha\mathcal M)(j')(\beta\mathcal M)(j)=\sum_{i,i'}\alpha(i)\beta(i')[\mathcal M(i,j)\mathcal M(i',j')-\mathcal M(i,j')\mathcal M(i',j)]$;
terms with $i=i'$ vanish, and for $i<i'$ the terms $(i,i')$ and $(i',i)$ add up to
$[\alpha(i)\beta(i')-\alpha(i')\beta(i)][\mathcal M(i,j)\mathcal M(i',j')-\mathcal M(i,j')\mathcal M(i',j)]\ge0$.
(T2) $\mathcal M$ is TP2 if and only if $e_i\mathcal M\le_{\mathrm{lr}}e_{i'}\mathcal M$ for $i<i'$; apply (T1) with $\mathcal N$. The minors of
$\Delta\mathcal M\Delta'$ are nonnegative multiples of those of $\mathcal M$, and minors of a submatrix are minors of $\mathcal M$.
(T3) Necessity is the case $i'=i+1$, $j=i$, $j'=i+1$. A minor on rows $i<i'$, columns $j<j'$ can be negative only if
$\mathcal M(i,j')\mathcal M(i',j)>0$, which needs $j'\le i+1$ and $j\ge i'-1\ge i$, hence $j=i$, $j'=i'=i+1$.
(T4) $\delta_0(j')=0$ for $j'>0$. Summing $\alpha(j)\beta(j')\ge\alpha(j')\beta(j)$ over $j<c\le j'$ gives
$\alpha(\{j<c\})\beta(\{j\ge c\})\ge\alpha(\{j\ge c\})\beta(\{j<c\})$; add $\alpha(\{j\ge c\})\beta(\{j\ge c\})$ to both sides.
\end{proof}

\begin{definition}[Radially TP2 curricula]\label{def:un-radTP2}
Let $k,L,\mathcal C$ be as in Definition~\ref{def:un-params}. $\mathcal C$ is radially TP2 for the alternating construction if the restrictions
$\hat{\mathcal R}_{\mathrm{lo}}$ and $\hat{\mathcal R}_{\mathrm{hi}}$ to $\{0,\dots,A_{\mathrm{hi}}\}^2$ of the filtered level kernels
$\mathcal R_{s^{\mathrm{lo}}}$ and $\mathcal R_{s^{\mathrm{hi}}}$ (Lemma~\ref{lem:un-lumping}) are TP2.
\end{definition}

The level kernels exist for every curriculum in $\mathcal P_0(k)$, and their rows at levels whose views are not flat (near $o$, near
branch levels and in the band) may be arbitrary; the definition is a condition on these rows. Since
$\mathcal R_s=\mathcal R^{\mathrm{prop}}_s\mathrm{diag}(\mathbf 1[f(r')\ge p/2])$, it holds whenever the restricted proposal level kernels are TP2
(Lemma~\ref{lem:un-tp2}(T2)).

\begin{theorem}[Radially TP2 curricula from the fixed start]\label{thm:un-B}
Let $k\ge1$, $L\ge400k$, $p\in(0,1)$, and let $\mathcal C\in\mathcal P_0(k)$ on $T_3^{(L)}$ have $\sigma^2>0$ and be radially TP2 for the
alternating construction. Run it from $D_0=\delta_o$.
\begin{enumerate}
\item[(a)] Every $D^{\mathrm{lo}}_n$ is $\Gamma$-invariant, its level law $\bar D_n$ satisfies $\bar D_n\le_{\mathrm{lr}}\bar D_{n+1}$, and
$D^{\mathrm{lo}}_n(\{r\ge c\})$ is nondecreasing in $n$ for every $c$; in particular $D^{\mathrm{lo}}_n(\mathsf T)$ is nondecreasing.
\item[(b)] $\bar D_n$ converges to a probability vector $\bar D_\infty$ on $\{0,\dots,A_{\mathrm{hi}}\}$; the $\Gamma$-invariant lift
$\nu(x):=\bar D_\infty(r(x))/|S_{r(x)}|$ satisfies $\nu Q_{V_Q}=\lambda'\nu$ with $\lambda'\ge\tilde\mu^{m+1}$, and
\[
\sup_nD^{\mathrm{lo}}_n(\mathsf T)=\lim_nD^{\mathrm{lo}}_n(\mathsf T)=\nu(\mathsf T)\le\frac{\bar C_k}{L^2}=\frac{12000\,k^2}{L^2}.
\]
\item[(c)] For every $\epsilon\in[\bar C_k/L^2,1)$, the alternating construction from $\delta_o$ is admissible at level $\epsilon$, every round is
defined, and $\sup_t|K_t|=|B(o,A_{\mathrm{hi}})|<\infty$. Hence $\epsilon^*_{\mathrm{bdd}}(\mathcal C;T_3^{(L)},o)\le12000k^2/L^2$.
\end{enumerate}
\end{theorem}

\begin{proof}
\emph{Levels.} By Lemma~\ref{lem:un-lumping}(c), applied round by round from the $\Gamma$-invariant $\delta_o$, every $\mu_n$ is $\Gamma$-invariant.
All measures in a cycle are supported in $V_Q$ (Lemma~\ref{lem:un-blindfold}), so only rows and columns $\le A_{\mathrm{hi}}$ of the level
kernels are used, and $\bar\mu_n=e_0\bar Q^{\,n}$ with $\bar Q:=\hat{\mathcal R}_{\mathrm{lo}}\hat{\mathcal R}_{\mathrm{hi}}^m$; likewise $\bar\mu\bar Q$ is
the level-mass vector of $\mu Q$ for every $\Gamma$-invariant $\mu$ supported in $V_Q$. By (T2), $\bar Q$ is TP2.

\emph{Monotonicity.} $\Lambda_n>0$ (Lemma~\ref{lem:un-star}), so $\bar D_n=\bar\mu_n/\Lambda_n$, and $D^{\mathrm{lo}}_n$ is the $\Gamma$-invariant measure
with level law $\bar D_n$ (Lemma~\ref{lem:un-lumping}(d)). By (T4), $\bar\mu_0=e_0\le_{\mathrm{lr}}\bar\mu_1$, and (T1) propagates
$\bar\mu_n\le_{\mathrm{lr}}\bar\mu_{n+1}$. The relation is invariant under positive scaling, so $\bar D_n\le_{\mathrm{lr}}\bar D_{n+1}$, and (T4) gives (a).

\emph{Limit.} The tails $\bar D_n(\{r\ge c\})$ are nondecreasing and bounded by $1$; their limits define a probability vector $\bar D_\infty$ with
$\bar D_n\to\bar D_\infty$. Put $\gamma_j:=\bar D_j\bar Q\mathbf 1=\Lambda_{j+1}/\Lambda_j$ and $\lambda':=\bar D_\infty\bar Q\mathbf 1=\lim_j\gamma_j$. By
Lemma~\ref{lem:un-star}, $\frac1n\sum_{j<n}\log\gamma_j=\frac1n\log\Lambda_n\ge(m+1)\log\tilde\mu$; if $\lambda'=0$ the left side would tend to
$-\infty$, so $\lambda'>0$, the Ces\`aro means converge to $\log\lambda'$, and $\lambda'\ge\tilde\mu^{m+1}$. Passing to the limit in
$\bar D_{n+1}=\bar D_n\bar Q/\gamma_n$ gives $\bar D_\infty\bar Q=\lambda'\bar D_\infty$.

\emph{Lift.} $\nu Q$ is $\Gamma$-invariant with level masses $\bar D_\infty\bar Q=\lambda'\bar D_\infty$, so $\nu Q=\lambda'\nu$
(Lemma~\ref{lem:un-lumping}(d)), and Theorem~\ref{thm:un-A}(b) gives $\nu(\mathsf T)\le\bar C_k/L^2$. By (a),
$D^{\mathrm{lo}}_n(\mathsf T)=\bar D_n(\{r\ge A+1\})$ increases to $\nu(\mathsf T)$. Part (c) is Proposition~\ref{prop:un-reduction} with
$\bar\eta\le\bar C_k/L^2$, and $\bar C_k/L^2<1$.
\end{proof}

\begin{corollary}[Every lazy curriculum in $\mathcal P_0(1)$]\label{cor:un-lazy}
Let $L\ge400$, $p\in(0,1)$, and let $\mathcal C\in\mathcal P_0(1)$ on $T_3^{(L)}$ be lazy: $\mathcal C_v(\{x\})\ge1/2$ for every labelled
rooted view $v$ of radius $1$ with root $x$ (it suffices that this holds at the views that occur under $s^{\mathrm{lo}}$ and
$s^{\mathrm{hi}}$). Then for every $\epsilon\in[12000/L^2,1)$ there is an admissible prover sequence from $D_0=\delta_o$ with every round
defined and $\sup_t|K_t|<\infty$: the alternating construction if $\sigma^2>0$, along which $D^{\mathrm{lo}}_n(\mathsf T)$ is nondecreasing,
and the constant sequence $s^{[3]}$ if $\sigma^2=0$. Hence $\epsilon^*_{\mathrm{bdd}}(\mathcal C;T_3^{(L)},o)\le12000/L^2$.
\end{corollary}

\begin{proof}
If $\sigma^2=0$, apply Theorem~\ref{thm:un-B0}. Let $\sigma^2>0$ and $s$ a ball profile. For $x\in S_r$, $B(x,1)=\{x\}\cup N(x)$ with
$N(x)\subseteq S_{r-1}\cup S_{r+1}$, so $\mathcal R^{\mathrm{prop}}_s(r,r')=0$ for $|r-r'|\ge2$ and $\mathcal R^{\mathrm{prop}}_s(r,r)=\mathcal C^s(x,\{x\})\ge1/2$.
Rows sum to $1$, so off-diagonal entries are at most $1/2$ and every consecutive minor is at least $\frac14-\frac14=0$. By (T3) and (T2)
the restrictions of $\mathcal R^{\mathrm{prop}}_s$ and $\mathcal R_s$ to $\{0,\dots,A_{\mathrm{hi}}\}^2$ are TP2. Theorem~\ref{thm:un-B} with $k=1$ applies.
\end{proof}

Laziness is used at every level, including those whose views are not flat. Lazy versions $\frac12\delta_x+\frac12\mathcal C'_v$,
$\mathcal C'\in\mathcal P_0(1)$, of frontier-seeking rules, of selection by low but positive pass rate, of weights that reward uncertainty or
learnability, and of simple random walk satisfy the hypothesis. Without laziness the condition can fail at a flat view: the uniform law
on $B(x,1)$ lies in $\mathcal P_0(1)$, and at the flat views of $o$ and of a level-$1$ vertex its level kernel has consecutive minor
$\frac14\cdot\frac13-\frac34\cdot\frac13<0$ at level $0$. Theorem~\ref{thm:fs-main} covers such curricula; its proof uses no monotonicity.

\subsection{Batch selection with a bounded shell weight}\label{sec:un-sel}

\begin{definition}[Batch selection dynamics]\label{def:un-sel}
A selection weight is a function $\varpi:[0,1]\to[0,\infty)$ with $\varpi(u)=0$ for $u<p/2$. For $\mathcal C\in\mathcal P_0(k)$ and
profiles $(s_t)$, the selection dynamics of $(\mathcal C,\varpi)$ is
\[
D_{t+1}(y)=\frac{\sum_xD_t(x)\,\mathcal C^{s_t}(x,y)\,\varpi(s_t(y))}{\sum_{x,y'}D_t(x)\,\mathcal C^{s_t}(x,y')\,\varpi(s_t(y'))},
\]
defined while the denominator is positive and finite (it is finite for finitely supported cold starts). Admissibility and the two
tolerances are defined with these $D_t$; $\varpi=\mathbf 1[\,\cdot\ge p/2]$ gives \eqref{eq:update}.
\end{definition}

For a selection weight with $\varpi(p)>0$ and $s^\ast\in[p/2,p)$ let the modified low profile $s^{\mathrm{lo}}_\ast$ be $p$ on $B(o,A)$,
$s^\ast$ on $S_{A+1}$, $s^\ast-p/2$ on $S_{A+2}$ and $0$ elsewhere, and call $\beta:=\varpi(s^\ast)/\varpi(p)$ the shell weight (in a low
round, the selection weight of $S_{A+1}$ relative to that of $B(o,A)$). The modified alternating construction uses $s^{\mathrm{lo}}_\ast$ in
low rounds and $s^{\mathrm{hi}}$ in high rounds. Put $\widetilde W_s(x,y):=\mathcal C^s(x,y)\,\varpi(s(y))/\varpi(p)$,
$\widetilde P_{\mathrm{lo}}:=\widetilde W_{s^{\mathrm{lo}}_\ast}$, $\widetilde P_{\mathrm{hi}}:=\widetilde W_{s^{\mathrm{hi}}}$ and
$\widetilde Q:=\widetilde P_{\mathrm{lo}}\widetilde P_{\mathrm{hi}}^m$.

\begin{proposition}[Batch selection]\label{prop:un-sel}
Let $k\ge1$, $p\in(0,1)$, $\mathcal C\in\mathcal P_0(k)$ on $T_3^{(L)}$, and let $\varpi$ be a selection weight with $\varpi(p)>0$ and
$\beta=\varpi(s^\ast)/\varpi(p)\le1$ for some $s^\ast\in[p/2,p)$. Consider the selection dynamics of $(\mathcal C,\varpi)$.
\begin{enumerate}
\item[(a)] If $\sigma^2=0$ and $L\ge3k+1$, the conclusions of Theorem~\ref{thm:un-B0} hold.
\item[(b)] If $\sigma^2>0$ and $L\ge400k$, the modified alternating construction satisfies the conclusions of Theorem~\ref{thm:un-A} with
$\widetilde Q$ in place of $Q$, and of Proposition~\ref{prop:un-reduction} from $\delta_o$, with the same constants.
\item[(c)] If in addition the restrictions to $\{0,\dots,A_{\mathrm{hi}}\}^2$ of $\mathcal R^{\mathrm{prop}}_s\,\mathrm{diag}(\varpi(f_s(r'))/\varpi(p))$,
$s\in\{s^{\mathrm{lo}}_\ast,s^{\mathrm{hi}}\}$, $s=f_s\circ r$, are TP2, the conclusions of Theorem~\ref{thm:un-B} hold. This is the case for every
lazy $\mathcal C\in\mathcal P_0(1)$; for such $\mathcal C$ and every $L\ge400$, the selection dynamics from $\delta_o$ admits, for every
$\epsilon\in[12000/L^2,1)$, an admissible prover sequence with every round defined and $\sup_t|K_t|<\infty$.
\end{enumerate}
\end{proposition}

\begin{proof}
The idea is that, with $\beta\le1$, the selection operators have every property of $P_{\mathrm{lo}},P_{\mathrm{hi}}$ that the proofs above use.

\emph{Normalisation.} The selection dynamics is $D_{t+1}=D_tW_t/(D_tW_t\mathbf 1)$ with $W_t=\varpi(p)\widetilde W_{s_t}$; the factor $\varpi(p)>0$
cancels, so Lemma~\ref{lem:un-products} applies to the $\widetilde W_{s_t}$, and Lemma~\ref{lem:un-lumping} applies with $\omega=\varpi/\varpi(p)$.

\emph{Profile.} Across edges $s^{\mathrm{lo}}_\ast$ changes by $0$, $p-s^\ast\le p/2$, $p/2$, $s^\ast-p/2<p/2$ or $0$, so it is $p/2$-Lipschitz, and
$\{s^{\mathrm{lo}}_\ast\ge p\}=B(o,A)$ as $s^\ast<p$; so $\mathsf T$ and the low-round admissibility condition are unchanged.

\emph{Operators.} (1) For $r(z)\le A_{\mathrm{hi}}-k$, all labels in $B(z,k)$ under $s^{\mathrm{hi}}$ equal $p$, so
$\widetilde P_{\mathrm{hi}}(z,\cdot)=P^\flat(z,\cdot)$ (Lemma~\ref{lem:un-flat}). (2) For $x\in\mathsf I$, $\widetilde P_{\mathrm{lo}}(x,\cdot)=P^\flat(x,\cdot)$
in the same way. (3) For $x\in\mathsf Z$, every label in $B(x,k)\subseteq\{r\ge A+2\}$ is at most $s^\ast-p/2<p/2$, so
$\widetilde P_{\mathrm{lo}}(x,\cdot)=0$. (4) $\widetilde P_{\mathrm{lo}}(x,y)$ equals $\mathcal C^{s^{\mathrm{lo}}_\ast}(x,y)$ if $r(y)\le A$,
$\beta\,\mathcal C^{s^{\mathrm{lo}}_\ast}(x,y)$ if $r(y)=A+1$, and $0$ otherwise; as $\beta\le1$, $\widetilde P_{\mathrm{lo}}$ is substochastic and
supported in $B(o,A+1)$, and its row at $x\in\mathsf B$ lives on levels in $(A-2k,A+1]$.

\emph{Transfer.} The proofs of Lemmas~\ref{lem:un-blindfold}--\ref{lem:un-star} use the low-round operator only through (2)--(4) (for
$P_{\mathrm{lo}}$ these are Lemma~\ref{lem:un-blindfold}(a) and Lemma~\ref{lem:un-accounting}), and the high-round operator only through its rows at
levels $\le A_{\mathrm{hi}}-k$, given by (1); in Lemma~\ref{lem:un-star} the rows at levels $<R\le A_{\mathrm{hi}}-k$ are given by (1), (2), and
elsewhere only nonnegativity is used. The proofs of Theorem~\ref{thm:un-A} and Proposition~\ref{prop:un-reduction} use only these lemmas, the
substochasticity of $\widetilde Q$ (by (4), (1) and Lemma~\ref{lem:un-blindfold}(b), $\widetilde Q=\widetilde P_{\mathrm{lo}}(P^\flat)^m$ on every row),
and Lemma~\ref{lem:un-products}. This gives (b). For (a), the proof of Theorem~\ref{thm:un-B0} uses only rows at flat views, where
$\widetilde W_{s^{[3k]}}=P^\flat$. For (c), the proof of Theorem~\ref{thm:un-B} uses the filtered level kernels only through
Lemma~\ref{lem:un-lumping} and their TP2 property, which is now the hypothesis. For lazy $\mathcal C\in\mathcal P_0(1)$ the proof of
Corollary~\ref{cor:un-lazy} shows that the restricted proposal level kernels are TP2, and the column scaling preserves TP2 (T2); the last
sentence of (c) follows from (a) if $\sigma^2=0$ ($L\ge4$) and from Theorem~\ref{thm:un-B}(c) if $\sigma^2>0$.
\end{proof}

Proposition~\ref{prop:un-sel} covers every selection weight with $\varpi(p)>0$ and $\varpi(s^\ast)\le \varpi(p)$ for some $s^\ast\in[p/2,p)$, for
instance every $\varpi$ with $\varpi(p)>0$ that is nondecreasing on $[p/2,p]$. The condition $\beta\le1$ enters through property (4), where it
makes $\widetilde P_{\mathrm{lo}}$ substochastic.

\begin{restated}{Proposition~\ref{prop:boundary}(a)}
Let $p\in(0,1)$. \StmtBoundaryA
\end{restated}

\paragraph{Proof of Proposition~\ref{prop:boundary}(a).} The weight $\varpi$ satisfies the hypotheses of Proposition~\ref{prop:un-sel}. Let
$\mathcal C\in\mathcal P_0(k)$ have $\sigma^2>0$ and $L\ge400k$. By Proposition~\ref{prop:un-sel}(b) the conclusions of Theorem~\ref{thm:un-A} hold
with $\widetilde Q$ in place of $Q$: $\widetilde Q_{V_Q}$ has a QSD $\nu$, and the modified alternating construction from $D_0=\nu$ is admissible at
every level $\epsilon\in[12000k^2/L^2,1)$ with every round defined and $\sup_t|K_t|<\infty$, so $\epsilon^*_{\mathrm{bdd}}\le12000k^2/L^2$ for the
selection dynamics from $\nu$. For lazy $\mathcal C\in\mathcal P_0(1)$ and $L\ge400$, Proposition~\ref{prop:un-sel}(c) gives such a sequence from
$\delta_o$ for every $\epsilon\in[12000/L^2,1)$, and $12000/L^2\le3/40<1$, so $\epsilon^*_{\mathrm{bdd}}\le12000/L^2$. These are the bounds of
Proposition~\ref{prop:boundary}(a), with the absolute constant $12000$, for $L\ge400k$ and $L\ge400$ respectively.
\fussy

\sloppy
\section{The fixed-start tail bound: proof of \texorpdfstring{Theorem~\ref{thm:uniform}}{Theorem 2}}\label{app:E_fixedstart}\label{sec:fs}

\begin{restated}{Theorem~\ref{thm:uniform}}\StmtUniform\end{restated}

Theorem~\ref{thm:uniform} follows from a bound on the tail mass of the alternating construction (Definition~\ref{def:un-wash}) started from
$\delta_o$, which this appendix proves for every $\mathcal C\in\mathcal P_0(k)$ with positive corridor variance (Theorem~\ref{thm:fs-main}); the
plan of the proof is the roadmap at the start of Appendix~\ref{app:D_uniform}.

\begin{theorem}[Fixed-start tail bound]\label{thm:fs-main}
Let $k\ge1$ and $C_k:=6\cdot10^{31}k^5$. Let $L\ge10^5k$, $p\in(0,1)$, and let $\mathcal C\in\mathcal P_0(k)$ on $T_3^{(L)}$ have corridor
variance $\sigma^2>0$. Then the alternating construction started from $D_0=\delta_o$ satisfies
\[
\sup_{n\ge0}D^{\mathrm{lo}}_n(\mathsf T)\ \le\ \frac{C_k}{L^2}.
\]
\end{theorem}

\begin{theorem}[Theorem~\ref{thm:uniform} with explicit prover sequences]\label{thm:fs-U}
Under the hypotheses of Theorem~\ref{thm:uniform}, with $C_k=6\cdot10^{31}k^5$, the conclusion of Theorem~\ref{thm:uniform} holds with the
following admissible prover sequence: the alternating construction of Definition~\ref{def:un-wash} if $\sigma^2>0$, with
$\sup_t|K_t|=|B(o,A_{\mathrm{hi}})|$, and the constant ball profile $s^{[3k]}$ if $\sigma^2=0$, with $K_t=B(o,3k)$. In particular
$\epsilon^*_{\mathrm{div}}(\mathcal C;T_3^{(L)},o)\le\epsilon^*_{\mathrm{bdd}}(\mathcal C;T_3^{(L)},o)\le6\cdot10^{31}k^5/L^2$.
\end{theorem}

\paragraph{Proof of Theorem~\ref{thm:uniform}.} We prove Theorem~\ref{thm:fs-U}, which contains Theorem~\ref{thm:uniform}. Let $L\ge10^5k$,
$p\in(0,1)$, $\mathcal C\in\mathcal P_0(k)$ and $\epsilon\in[C_k/L^2,1)$. If $\sigma^2=0$, Theorem~\ref{thm:un-B0} (which applies as
$L\ge3k+1$) shows that $s^{[3k]}$ is admissible from $\delta_o$ at level $\epsilon$, with every round defined and $K_t=B(o,3k)$. If
$\sigma^2>0$, Theorem~\ref{thm:fs-main} gives $\bar\eta=\sup_nD^{\mathrm{lo}}_n(\mathsf T)\le C_k/L^2\le\epsilon$ in
Proposition~\ref{prop:un-reduction} (which applies as $L\ge400k$), so the alternating construction is admissible from $\delta_o$ at level
$\epsilon$, with every round defined and $\sup_t|K_t|=|B(o,A_{\mathrm{hi}})|<\infty$. For the tolerances: if $C_k/L^2<1$ the level $C_k/L^2$
lies in the set whose infimum is $\epsilon^*_{\mathrm{bdd}}(\mathcal C;T_3^{(L)},o)$, if $C_k/L^2\ge1$ the bound holds by definition, and
$\epsilon^*_{\mathrm{div}}\le\epsilon^*_{\mathrm{bdd}}$ always (Appendix~\ref{app:B_model}). Theorem~\ref{thm:uniform} is
Theorem~\ref{thm:fs-U} without the names of the two prover sequences.

\paragraph{Standing assumptions and plan.} The rest of this appendix proves Theorem~\ref{thm:fs-main}. Throughout, $k\ge1$, $L\ge10^5k$,
$p\in(0,1)$, $\mathcal C\in\mathcal P_0(k)$ has corridor law $\kappa$ and $\sigma^2>0$, all parameters are those of
Definition~\ref{def:un-params} (including the second group) with the bounds of Lemma~\ref{lem:un-arith}, and the alternating construction is
run from $D_0=\delta_o$. We write $P^m:=(P^\flat)^m$, $D_n:=D^{\mathrm{lo}}_n$ and $\Lambda_n=\mu_n\mathbf 1$. Section~\ref{sec:fs-params}
couples the flat kernel with the walk of law $\kappa$, Section~\ref{sec:fs-walk} proves survival and local bounds for this walk (no constant
depends on $\sigma^2$, on $m$ or on lattice properties of $\kappa$), and Sections~\ref{sec:fs-WH}--\ref{sec:fs-proof} carry out the three
steps of the roadmap and assemble the constants.

\subsection{Coordinates, the step chain and two tools}\label{sec:fs-params}

\begin{definition}[Coordinates and sets]\label{def:fs-params}
The vertex set of $G$ is written $V(G)$ in this appendix ($V$ is the cycle variance). For $x\in V(G)$ let $d(x):=A+1-r(x)$. Then
$\mathsf I=\{d\ge k+1\}$, $\mathsf B=\{-k\le d\le k\}$, $\mathsf Z=\{d\le-k-1\}$ and $\mathsf T=\{d\le0\}$. Put $\mathsf E:=\mathsf B\cup\mathsf Z=\{d\le k\}$,
$N_b:=\{0\le d\le2k\}$, and, as subsets of $V_Q$, $O:=\{x\in V_Q:d(x)\le2\delta\}$ and $\mathsf C:=V_Q\setminus O=\{r\le R_\sharp-\delta\}$.
Finally $\hat G(x,y):=\mathbf 1_{\mathsf I}(x)\,Q(x,y)$ and $\varrho(\mu):=(\mu\mathbf 1_{\mathsf B})P_{\mathrm{lo}}$ for finitely supported $\mu\ge0$, so
that $\varrho(\mu)\mathbf 1=g(\mu)$.
\end{definition}

$N_b$ consists of the levels $A-2k+1,\dots,A+1\in[1,L-1]$, each of which carries one vertex on each arm (Lemma~\ref{lem:un-geometry}(a)),
so $|N_b|=3(2k+1)\le9k$; $\varrho(\mu)$ is supported on $N_b$ (Lemma~\ref{lem:un-accounting}(iii)). By Lemma~\ref{lem:un-geometry}(a),(c), a
vertex at level $r\in[k+1,L-k-1]$ is a corridor vertex of an arm, and by Lemma~\ref{lem:un-arith}(i),
$A+1-d\in[A+2+k-3L/4,\,A-1-k+L/4]\subseteq[k+1,L-k-1]$ whenever $-L/4+k+2\le d\le3L/4-k-1$; hence
\begin{equation}\label{eq:fs-corr}
\text{every vertex $x$ with }-L/4+k+2\le d(x)\le3L/4-k-1\text{ is a corridor vertex of an arm.}
\end{equation}

\begin{lemma}[Subsolution of radius $R_\sharp$]\label{lem:fs-star}
Let $h_\sharp$ be the function $h$ of Lemma~\ref{lem:un-star} with $R'=R_\sharp$. Then $0\le h_\sharp\le1$, $h_\sharp(o)=1$, $h_\sharp$ is
nonincreasing in $r$, $Q^nh_\sharp\ge\tilde\mu_{\sharp,c}^{\,n}h_\sharp$ and $\Lambda_n\ge\tilde\mu_{\sharp,c}^{\,n}$ for $n\ge0$, and
$\min_{\mathsf C}h_\sharp=h_{\mathsf C}=\sin(\theta_\sharp\delta)$.
\end{lemma}

\begin{proof}
$R'=R_\sharp$ is admissible in Lemma~\ref{lem:un-star}: $3k<R_\sharp\le A-k$ and $\theta_\sharp k\le2.3\cdot10^{-5}$
(Lemma~\ref{lem:un-arith}(j)); then $\theta'=\theta_\sharp$ and $\tilde\mu'=\tilde\mu_\sharp$. $Q^n$ is a product of $n(m+1)$ factors, and
$\Lambda_n=\delta_oQ^n\mathbf 1\ge(Q^nh_\sharp)(o)$ as $\mathbf 1\ge h_\sharp\ge0$. Since $a<R_\sharp-\delta$, the minimum over
$\mathsf C=\{r\le R_\sharp-\delta\}$ is $\cos(\theta_\sharp(R_\sharp-a)-\theta_\sharp\delta)=\sin(\theta_\sharp\delta)$.
\end{proof}

\begin{lemma}[Freedman's inequality for the step walk]\label{lem:fs-F}
Let $\xi_1,\xi_2,\dots$ be i.i.d.\ with law $\kappa$ and $S_j:=\xi_1+\dots+\xi_j$. For $N\ge1$ and $x>0$,
\[
\Pr\Big(\max_{j\le N}S_j\ge x\Big)\le\exp\Big(-\frac{x^2}{2(N\sigma^2+kx/3)}\Big)\le e^{-x^2/(4N\sigma^2)}+e^{-3x/(4k)},
\]
and the same holds for $-S_j$. In particular $\Pr(\max_{j\le m+1}|S_j|\ge M')\le\eta_F$.
\end{lemma}

\begin{proof}
$(S_j)$ is a martingale with increments at most $k$ and predictable quadratic variation $j\sigma^2\le N\sigma^2$ for $j\le N$.
Freedman's inequality \citep[Theorem~1.6]{freedman1975}, after scaling the increments by $1/k$, gives
$\Pr(\exists j\le N:S_j\ge x)\le\exp(-v'\,\phi(x/(kv')))$ with $\phi(t)=(1+t)\log(1+t)-t$ and $v'=N\sigma^2/k^2$: that theorem is stated for increments
bounded by $1$, and is applied to $S_j/k$, whose predictable quadratic variation is at most $v'$. Since
$\phi(t)\ge t^2/(2(1+t/3))$ for $t\ge0$, the exponent is at least $\frac{(x/k)^2}{2(v'+x/(3k))}=\frac{x^2}{2(N\sigma^2+kx/3)}$, which is the first bound.
The second uses $e^{-x^2/(2(u+v))}\le\max(e^{-x^2/(4u)},e^{-x^2/(4v)})$ with $u=N\sigma^2$, $v=kx/3$. The law of $-S_j$ is the law of
$S_j$. For the last claim, with $N=m+1$ and $x=M'$: $2((m+1)\sigma^2+kM'/3)\le2(66k^2+kM'/3)\le0.709\,kM'$ because
$M'\ge3124k$, so the probability is at most $2e^{-M'/(0.709k)}\le2e^{-(q/32-1)/0.709}\le e^{-q/24}$, since
$\frac1{32\cdot0.709}-\frac1{24}\ge0.0024$ and $0.0024\,q\ge240>\ln2+\frac1{0.709}$.
\end{proof}

\paragraph{The step chain.} Let $(X_j)$ be the step chain of Section~\ref{sec:un-band}, with filtration $(\mathcal G_j)$. Low-round times
are the multiples of $m+1$, and $Y_n:=X_{n(m+1)}$.

\begin{lemma}[Corridor coupling]\label{lem:fs-coupling}
\begin{enumerate}
\item[(a)] $(X_j)$ is a Markov chain with kernel $P^\flat$ with respect to $(\mathcal G_j)$. For every stopping time $\varsigma$ of
$(\mathcal G_j)$, on $\{\varsigma<\infty\}$ the sequence $(\xi_{\varsigma+i},U_{\varsigma+i})_{i\ge1}$ is independent of
$\mathcal G_\varsigma$ and has the law of $(\xi_i,U_i)_{i\ge1}$.
\item[(b)] If $X_i,\dots,X_{j-1}$ are corridor vertices, then $X_i,\dots,X_j$ lie on one subdivided edge and
$d(X_l)=d(X_i)-(\xi_{i+1}+\dots+\xi_l)$ for $i\le l\le j$.
\item[(c)] Let $c\ge1$. If $X_i$ is a vertex of an arm, every vertex $x$ with $|d(x)-d(X_i)|<c$ on that arm is a corridor vertex, and
$\max_{i<l\le j}|\xi_{i+1}+\dots+\xi_l|<c$, then $X_i,\dots,X_j$ are corridor vertices of that arm and (b) applies.
\item[(d)] For $x\in\mathsf I$, $Q(x,\cdot)=(P^\flat)^{m+1}(x,\cdot)$, a probability supported in $V_Q$. For $z\in\mathsf I$ let
$\tau_{\mathsf E}:=\min\{n\ge1:Y_n\notin\mathsf I\}\in\{1,2,\dots,\infty\}$ for the step chain from $X_0=z$. Then $Y_{\tau_{\mathsf E}}\in\mathsf E$ on
$\{\tau_{\mathsf E}<\infty\}$, and for $u\ge1$ and $y\in V(G)$,
\[
(\delta_z\hat G^u)(y)=\Pr_z(\tau_{\mathsf E}\ge u,\ Y_u=y),\qquad f^z_u:=(\delta_z\hat G^u)(\mathsf E)=\Pr_z(\tau_{\mathsf E}=u).
\]
\end{enumerate}
\end{lemma}

\begin{proof}
(a) is shown in Section~\ref{sec:un-band}.
(b) If $X_l=z_s$ is a corridor vertex of the subdivided edge $(z_0,\dots,z_L)$, then $X_{l+1}=z_{s+\xi_{l+1}}$ is on the same edge
(Lemma~\ref{lem:un-geometry}(c)), and $r(z_{s+\xi})=r(z_s)+\xi$ (Lemma~\ref{lem:un-geometry}(a)); induct on $l$.
(c) Induction on $l$: if $X_i,\dots,X_l$ ($l<j$) are corridor vertices of the arm with $d(X_l)=d(X_i)-(\xi_{i+1}+\dots+\xi_l)$, then
$X_{l+1}$ is on the arm with $|d(X_{l+1})-d(X_i)|<c$ by (b) and the hypothesis, hence a corridor vertex if $l+1<j$.
(d) $P_{\mathrm{lo}}(x,\cdot)=P^\flat(x,\cdot)$ is supported in $B(o,A)$ (Lemma~\ref{lem:un-accounting}(i)), and
Lemma~\ref{lem:un-blindfold}(b) with $j=m$ gives $Q(x,\cdot)=P^\flat(x,\cdot)P^m$, of mass $1$ and supported in
$B(o,A+1+mk)=V_Q$. Since $\hat G(x,\cdot)=\mathbf 1_{\mathsf I}(x)(P^\flat)^{m+1}(x,\cdot)$,
$(\delta_z\hat G^u)(y)=\sum\prod_{i<u}(P^\flat)^{m+1}(y_i,y_{i+1})$ over paths $z=y_0,y_1,\dots,y_u=y$ with $y_0,\dots,y_{u-1}\in\mathsf I$,
which by (a) is $\Pr_z(Y_0,\dots,Y_{u-1}\in\mathsf I,Y_u=y)$. All $Y_n$ with $n\le\tau_{\mathsf E}$ lie in $V_Q$, and
$V_Q\setminus\mathsf I=\mathsf E\cap V_Q$.
\end{proof}

The process $(Y_n)_{0\le n\le\tau_{\mathsf E}}$ for a start $z\in\mathsf I$ is the \emph{free walk}. By (d) it is the cycle chain
$Q$ up to its first low round in $\mathsf E$; it contains the arbitrary laws of $\mathcal C$ at the views that contain $o$. We write
$\Pr_z,\mathbb E_z$ for it.

\subsection{Lemmas for the symmetric walk}\label{sec:fs-walk}

In this subsection $\xi_1,\xi_2,\dots$ are i.i.d.\ with law $\kappa$, $\mathcal F_j:=\sigma(\xi_1,\dots,\xi_j)$, $W_j:=W_0+\xi_1+\dots+\xi_j$ with
$W_0\in\mathbb Z$, $Z_n:=W_{n(m+1)}$ (the cycle walk), and $\zeta$ has the law of $Z_1-Z_0$, that is $\kappa^{*(m+1)}$, with
$\mathbb E\zeta=0$ and $\mathbb E\zeta^2=V$. By symmetry of $\kappa$, $\zeta$ and $-\zeta$ have the same law. For $N\ge1$, $n\ge1$ and $x\ge0$ put
\[
\Xi_N(x):=e^{-x^2/(16N\sigma^2)}+e^{-3x/(8k)},\qquad\bar\Xi_n(x):=\Xi_{n(m+1)}(x)=e^{-x^2/(16nV)}+e^{-3x/(8k)}.
\]
Both are nondecreasing in $N$ (resp.\ $n$) and nonincreasing in $x$, and Lemma~\ref{lem:fs-F} reads
$\Pr(\max_{j\le N}(W_j-W_0)\ge x)\le\Xi_N(2x)$. Two moment facts are used repeatedly: $\mathbb E\xi_1^4\le k^2\sigma^2$, hence for a sum
$S$ of $N$ steps $\mathbb ES^4=N\mathbb E\xi_1^4+3N(N-1)\sigma^4\le N\sigma^2k^2+3(N\sigma^2)^2$, in particular
$\mathbb E\zeta^4\le Vk^2+3V^2\le3.03V^2$ (Lemma~\ref{lem:un-arith}(e)); and $\mathbb E|S|\le\sqrt{N\sigma^2}$,
$\mathbb ES^+=\frac12\mathbb E|S|$.

\begin{lemma}[Local bounds]\label{lem:fs-local}
\begin{enumerate}
\item[(a)] For $N\ge1$, $b\ge k$ and every closed interval $U$ of length $b$, $\Pr(W_N-W_0\in U)\le C_Eb/\sqrt{N\sigma^2}$, $C_E:=2.7$. In
particular $\Pr(W_N-W_0=y)\le C_Ek/\sqrt{N\sigma^2}$ and $\Pr(Z_n-Z_0=y)\le C_Ek/\sqrt{nV}$ for every $y$.
\item[(b)] For $N\ge2$, $\ell\ge k$ and $c\ge0$, $\Pr(W_N-W_0\in[c,c+\ell))\le C_{LG}\,\frac{\ell}{\sqrt{N\sigma^2}}\,\Xi_N(c)$ with $C_{LG}:=8.5$;
the same bound holds for the interval $(-c-\ell,-c]$.
\end{enumerate}
\end{lemma}

\begin{proof}
(a) Esseen's concentration inequality \citep{petrov1975} states
$\sup_x\Pr(x\le X\le x+\lambda)\le(96/95)^2\lambda\int_{|u|\le1/\lambda}|\varphi_X(u)|\,du$ for every random variable $X$ and $\lambda>0$.
Take $X=W_N-W_0$, $\lambda=b$, so $\varphi_X=\hat\kappa^N$ with $\hat\kappa(u)=\sum_d\kappa(d)\cos(ud)$. For $|u|\le1/b\le1/k$ and $|d|\le k$,
$|ud|\le1$, and $1-v^2/2\le\cos v\le1-\frac{11}{24}v^2$ for $|v|\le1$; hence $0<\hat\kappa(u)\le1-\frac{11}{24}u^2\sigma^2\le e^{-11u^2\sigma^2/24}$
and $\int|\varphi_X|\le\int_{\mathbb R}e^{-11Nu^2\sigma^2/24}du=\sqrt{24\pi/(11N\sigma^2)}$. Finally $(96/95)^2\sqrt{24\pi/11}\le2.6735$. A
point lies in a closed interval of length $k$, and $Z_n-Z_0$ is $W_{n(m+1)}-W_0$ with $n(m+1)\sigma^2=nV$.
(b) Let $N_1:=\lfloor N/2\rfloor\ge N/3$ and $N_2:=N-N_1\ge N/2$, and write $W_N-W_0=A'+B'$ with $A'$ the sum of the first $N_1$ steps and
$B'$ of the other $N_2$ steps; they are independent. If $A'+B'\ge c$, then $A'\ge c/2$ or $B'\ge c/2$. With $U:=[c,c+\ell)$,
\begin{align*}
\Pr(A'+B'\in U,\ A'\ge c/2)&=\mathbb E\big[\mathbf 1\{A'\ge c/2\}\Pr(B'\in U-A'\mid A')\big]\\
&\le\Pr(A'\ge c/2)\,\frac{C_E\ell}{\sqrt{N_2\sigma^2}}\le\Xi_N(c)\frac{C_E\ell}{\sqrt{N_2\sigma^2}},
\end{align*}
by (a) and Lemma~\ref{lem:fs-F} ($\Pr(A'\ge c/2)\le\Xi_{N_1}(c)\le\Xi_N(c)$). The same holds with the roles of $A',B'$ exchanged and $N_1$
in place of $N_2$. Hence the probability is at most $C_E(\sqrt2+\sqrt3)\ell\,\Xi_N(c)/\sqrt{N\sigma^2}$, and $2.7(\sqrt2+\sqrt3)\le8.495$. The
second interval follows by symmetry.
\end{proof}

\begin{lemma}[Completion of a cycle, gambler's ruin and exit time]\label{lem:fs-ruin}
\begin{enumerate}
\item[(a)] Let $\varsigma$ be a stopping time of $(\mathcal F_j)$ and $\lambda:=(m+1)\lceil\varsigma/(m+1)\rceil$. Then on $\{\varsigma<\infty\}$
we have $\Pr(|W_\lambda-W_\varsigma|\ge x_0\mid\mathcal F_\varsigma)\le65/144$ and
\[
\Pr\Big(|W_\lambda-W_\varsigma|<x_0,\ \max_{\varsigma<j\le\lambda}|W_j-W_\varsigma|<M'\ \Big|\ \mathcal F_\varsigma\Big)\ge\frac{79}{144}-\eta_F\ge0.548 .
\]
\item[(b)] Let $a<W_0<b$ and $\sigma_{ab}:=\min\{j\ge0:W_j\notin(a,b)\}$. Then $\sigma_{ab}<\infty$ a.s.\ and
\[
\frac{W_0-a}{b-a+k}\le\Pr(W_{\sigma_{ab}}\ge b)\le\frac{W_0-a+k}{b-a+k}.
\]
\item[(c)] Let $J_0\subset\mathbb Z$ have $\max J_0-\min J_0\le\Delta$ with $\Delta\ge1$, let $Z_0\in J_0$ and $\tau:=\min\{n\ge1:Z_n\notin J_0\}$. Then
$\mathbb E\tau\le7.12\,(4\Delta^2/V+1)$.
\end{enumerate}
\end{lemma}

\begin{proof}
(a) $\lambda-\varsigma\le m$ is $\mathcal F_\varsigma$-measurable, and given $\mathcal F_\varsigma$ the increments after $\varsigma$ are i.i.d.\ $\kappa$.
Chebyshev's inequality gives $\Pr(|W_\lambda-W_\varsigma|\ge x_0\mid\mathcal F_\varsigma)\le m\sigma^2/x_0^2\le65k^2/(144k^2)$
(Lemma~\ref{lem:un-arith}(e)), and Lemma~\ref{lem:fs-F} bounds the probability of the maximal deviation by $\eta_F$.
(b) Since $\sigma^2>0$ and $\kappa$ is symmetric, $\kappa(d_1)>0$ for some $d_1\ge1$; in each block of $\lceil(b-a)/d_1\rceil$ consecutive steps the walk
leaves $(a,b)$ if all steps equal $d_1$, so $\sigma_{ab}$ has a geometric tail and $\mathbb E\sigma_{ab}<\infty$. The stopped martingale
$W_{\sigma_{ab}\wedge j}$ is bounded by $|a|+|b|+k$, so $W_0=\mathbb EW_{\sigma_{ab}}$. As $W_{\sigma_{ab}}\in(a-k,a]\cup[b,b+k)$, with
$p_b:=\Pr(W_{\sigma_{ab}}\ge b)$ we get $W_0\le(1-p_b)a+p_b(b+k)$ and $W_0\ge(1-p_b)(a-k)+p_bb$, which are the two bounds.
(c) Put $t_0:=\lceil4\Delta^2/V\rceil$ and $X:=(Z_{t_0}-Z_0)^2$. Then $\mathbb EX=t_0V\ge4\Delta^2$ and $\mathbb EX^2\le t_0Vk^2+3(t_0V)^2$, so by the
Paley--Zygmund inequality $\Pr(X>\mathbb EX/4)\ge\frac{9/16}{3+k^2/(t_0V)}\ge\frac{9/16}{3+1/36}\ge\frac9{64}$. On this event
$|Z_{t_0}-Z_0|>\Delta$, so $Z_{t_0}\notin J_0$ if $Z_0\in J_0$. Applying this at the times $t_0,2t_0,\dots$ (the increments of disjoint blocks are
independent) gives $\Pr(\tau>jt_0)\le(55/64)^j$ and $\mathbb E\tau=\sum_{n\ge0}\Pr(\tau>n)\le t_0\cdot\frac{64}9\le7.12(4\Delta^2/V+1)$.
\end{proof}

\begin{lemma}[Survival at step level]\label{lem:fs-surv}
Let $a<W_0$ and $N\ge1$.
\begin{enumerate}
\item[(a)] $\Pr(\min_{j\le N}W_j>a)\le\Pr(|W_N-W_0|<W_0-a+k)\le2C_E(W_0-a+k)/\sqrt{N\sigma^2}$.
\item[(b)] If $N\ge2$ and $x\ge0$, then
\[
\Pr\Big(\min_{j\le N}W_j>a,\ W_N-W_0\ge x\Big)\le\Pr\big(W_N-W_0\in[x,\,x+\ell_a)\big)\le C_{LG}\frac{\ell_a}{\sqrt{N\sigma^2}}\,\Xi_N(x),
\]
where $\ell_a:=2(W_0-a)+2k$.
\end{enumerate}
\end{lemma}

\begin{proof}
(a) Let $c:=W_0-a>0$, $U_j:=W_0-W_j$ and $\varsigma:=\min\{j:U_j\ge c\}$, so $U_\varsigma\in[c,c+k)$ on $\{\varsigma<\infty\}$ and
$\{\varsigma>N\}=\{\min_{j\le N}W_j>a\}$. Given $\mathcal F_\varsigma$ on $\{\varsigma\le N\}$, $U_N-U_\varsigma$ is a sum of $N-\varsigma$ i.i.d.\ symmetric
steps, so $\Pr(\varsigma\le N,U_N>U_\varsigma)=\Pr(\varsigma\le N,U_N<U_\varsigma)$. If $U_N\ge c+k$ then $\varsigma\le N$ and $U_N>U_\varsigma$. Hence
$\Pr(\varsigma\le N)\ge2\Pr(U_N\ge c+k)$ and, by symmetry of $U_N$, $\Pr(\varsigma>N)\le1-\Pr(U_N\ge c+k)-\Pr(U_N\le-c-k)=\Pr(|U_N|<c+k)$.
Lemma~\ref{lem:fs-local}(a) with an interval of length $2(c+k)\ge k$ gives the last bound.
(b) Let $\varsigma:=\min\{j:W_j\le a\}$; then $W_\varsigma\in(a-k,a]$. By the strong Markov property and symmetry of $W_N-W_\varsigma$ given
$\mathcal F_\varsigma$ on $\{\varsigma\le N\}$,
\[
\Pr(\varsigma\le N,\ W_N\ge W_0+x)=\Pr(\varsigma\le N,\ W_N\le2W_\varsigma-W_0-x)\ge\Pr(W_N\le2a-2k-W_0-x),
\]
because $W_N\le2a-2k-W_0-x$ forces $W_N<a$, hence $\varsigma\le N$, and then $W_N<2W_\varsigma-W_0-x$. By symmetry the right side equals
$\Pr(W_N-W_0\ge x+\ell_a)$. Hence $\Pr(\varsigma>N,W_N-W_0\ge x)\le\Pr(W_N-W_0\ge x)-\Pr(W_N-W_0\ge x+\ell_a)$, which is the interval
probability; apply Lemma~\ref{lem:fs-local}(b) with $c=x$ and $\ell=\ell_a\ge2k$.
\end{proof}

\paragraph{Attempts and excursions.} Let $W_0=h\ge0$, fix $n\ge1$, and let $\tau:=\min\{n'\ge1:Z_{n'}<0\}$. Put $\lambda_0:=0$ and, recursively for
$i\ge1$, $\varsigma_i:=\min\{j\ge\lambda_{i-1}:W_j\le-x_0\}$ and $\lambda_i:=(m+1)\lceil\varsigma_i/(m+1)\rceil$, where $\varsigma_i:=\infty$ if
$\lambda_{i-1}=\infty$ or if $i\ge2$ and $W_{\lambda_{i-1}}<0$. Attempt $i$ \emph{fails} if $\lambda_i<\infty$ and $W_{\lambda_i}\ge0$; write
$\mathrm{fail}_i$ for this event, $R_i:=W_{\lambda_i}-W_{\varsigma_i}$, $s_i:=W_{\lambda_i}$, $s_0:=h$, and
$N_n:=\#\{i\ge1:\varsigma_i<n(m+1)\}$. These objects have the following properties.
\begin{enumerate}
\item[(P1)] $\varsigma_i,\lambda_i$ are stopping times; $\varsigma_1\ge1$ and $\varsigma_{i+1}>\lambda_i$ on $\mathrm{fail}_i$. On $\{\varsigma_i<\infty\}$,
$\Pr(\mathrm{fail}_i\mid\mathcal F_{\varsigma_i})\le65/144\le1/2$, since failure requires $W_{\lambda_i}-W_{\varsigma_i}\ge x_0$ (Lemma~\ref{lem:fs-ruin}(a)).
\item[(P2)] On $\mathrm{fail}_i$: $R_i\ge x_0$ and $0\le s_i\le R_i-x_0<R_i^+$. Given $\mathcal F_{\varsigma_i}$, $R_i$ is a sum of $\lambda_i-\varsigma_i\le m$
i.i.d.\ steps, so $\mathbb E[R_i^+\mid\mathcal F_{\varsigma_i}]\le\frac12\sqrt{m\sigma^2}$, and with $w_0:=1.5834$,
\[
\mathbb E\big[(s_i+13k)\mathbf 1_{\mathrm{fail}_i}\mid\mathcal F_{\varsigma_i}\big]\le\tfrac12\sqrt{m\sigma^2}+\tfrac{13}2k\le w_0\sqrt V
\]
(Lemma~\ref{lem:un-arith}(e): $\sqrt{m\sigma^2}\le\sqrt V$, $13k/(2\sqrt V)<13/12$).
\item[(P3)] On $\{\tau>n,\ N_n=j\}$: attempts $1,\dots,j$ have $\lambda_i\le n(m+1)$ and failed (a success at $\lambda_i\le n(m+1)$, with
$\lambda_i\ge\varsigma_i\ge1$, would give $\tau\le\lambda_i/(m+1)\le n$); $\varsigma_{j+1}>n(m+1)$ (equality would give $Z_n\le-x_0$); and the
excursions $E_0:=[0,\varsigma_1)$, $E_i:=[\lambda_i,\varsigma_{i+1})$ for $1\le i<j$, $E_j:=[\lambda_j,n(m+1)]$ (with $E_0:=[0,n(m+1)]$ if $j=0$),
together with the windows $[\varsigma_i,\lambda_i)$, $i\le j$, partition $\{0,1,\dots,n(m+1)\}$; on every excursion $W>-x_0$.
\item[(P4)] On $\{\tau>n,N_n=j\}$ with $j\le n/2$, some excursion $E_\iota$, $0\le\iota\le j$, contains at least $\lceil T_j\rceil+1$ consecutive
integers, where $T_j:=n(m+1)/(2(j+1))$. Indeed the windows contain at most $jm$ integers, so the excursions contain at least
$n(m+1)+1-jm\ge(j+1)(T_j+1)$ integers, because $n(m+1)/2\ge j(m+1)$.
\end{enumerate}
We call $E_\iota$ \emph{long} if it contains at least $\lceil T_j\rceil+1$ integers. For the last excursion $E_j$ put
$L_j:=n(m+1)-\lambda_j$, a multiple of $m+1$ that is $\mathcal F_{\lambda_j}$-measurable; $E_j$ is long if and only if $L_j\ge\lceil T_j\rceil$.

\begin{lemma}[Survival at cycle level]\label{lem:fs-CF5}
Let $Z_0=h\ge0$ and $\tau:=\min\{n\ge1:Z_n<0\}$. For $n\ge1$, $\Pr(\tau>n)\le C_S(1+h/\sqrt V)\,n^{-1/2}$ with $C_S:=150$.
\end{lemma}

\begin{proof}
Let $c_j:=2C_E\sqrt{2(j+1)}/\sqrt{nV}$. If $E_\iota$ is long and starts at height $s_\iota$, then by Lemma~\ref{lem:fs-surv}(a) with $a=-x_0$
and $N=\lceil T_j\rceil\ge T_j$, conditionally on $\mathcal F_{\lambda_\iota}$ (on $\mathcal F_0$ if $\iota=0$) this has probability at most
$2C_E(s_\iota+x_0+k)/\sqrt{T_j\sigma^2}=c_j(s_\iota+13k)$, as $T_j\sigma^2=nV/(2(j+1))$. Fix $j\le n/2$ and $\iota\le j$, and bound
$\Pr(\mathrm{fail}_1\cap\dots\cap\mathrm{fail}_j\cap\{E_\iota\text{ long}\})$ by conditioning successively, from the last event backwards, on
$\mathcal F_{\varsigma_j},\dots$: each attempt $i>\iota$ contributes a factor $1/2$ by (P1), since all earlier events are
$\mathcal F_{\varsigma_i}$-measurable; $E_\iota$ contributes $c_j(s_\iota+13k)$; for $\iota\ge1$ attempt $\iota$ contributes
$\mathbb E[(s_\iota+13k)\mathbf 1_{\mathrm{fail}_\iota}\mid\mathcal F_{\varsigma_\iota}]\le w_0\sqrt V$ by (P2); and attempts $i<\iota$ contribute $1/2$ each.
This gives $2^{-j}c_j(h+13k)$ for $\iota=0$ and $2^{1-j}c_jw_0\sqrt V$ for $1\le\iota\le j$. By (P3), (P4),
\[
\Pr(\tau>n)\le\sum_{0\le j\le n/2}c_j\big[2^{-j}(h+13k)+j\,2^{1-j}w_0\sqrt V\big]+\Pr\big(\mathrm{fail}_1\cap\dots\cap\mathrm{fail}_{\lfloor n/2\rfloor+1}\big).
\]
The last term is at most $2^{-\lfloor n/2\rfloor-1}\le2^{-n/2}\le0.7072\,n^{-1/2}$. With $S_a:=\sum_{j\ge0}\sqrt{2(j+1)}2^{-j}\le3.811$,
$S_b:=\sum_{j\ge1}j\sqrt{2(j+1)}2^{-j}\le5.504$ and $13k/\sqrt V<13/6$, the sum is at most
$2C_En^{-1/2}[S_a(h/\sqrt V+13/6)+2w_0S_b]$. The total is at most $(20.6\,h/\sqrt V+139.5)\,n^{-1/2}$,
since $2C_ES_a\le20.58$ and $2C_E\big(\frac{13}6S_a+2w_0S_b\big)+0.7072\le139.42$.
\end{proof}

\begin{lemma}[Survival and displacement at cycle level]\label{lem:fs-CF10}
Let $Z_0=h\ge0$ and $\tau:=\min\{n\ge1:Z_n<0\}$. For $n\ge1$ and $x\ge0$,
\[
\Pr(\tau>n,\ Z_n\ge h+x)\le C_P\,(1+h/\sqrt V)\,n^{-1/2}\,\bar\Xi_n(x/2),\qquad C_P:=10^4 .
\]
\end{lemma}

\begin{proof}
We use the objects above and write $\bar\Xi:=\bar\Xi_n(x/2)$. Three facts are needed.

\emph{(i) Window tail.} For $i\ge1$, on $\{\varsigma_i<\infty\}$, Lemma~\ref{lem:fs-F} with at most $m$ steps gives
$\Pr(R_i>x/2\mid\mathcal F_{\varsigma_i})\le e^{-x^2/(16m\sigma^2)}+e^{-3x/(8k)}\le\bar\Xi$, as $m\sigma^2\le V\le4nV$; and, by the
Cauchy--Schwarz inequality, $\|R_i^+\|_2\le\sqrt{m\sigma^2/2}$ and $\sqrt{u+v}\le\sqrt u+\sqrt v$,
\[
\mathbb E\big[(s_i+13k)\mathbf 1_{\mathrm{fail}_i\cap\{R_i>x/2\}}\mid\mathcal F_{\varsigma_i}\big]\le\big(\sqrt{m\sigma^2/2}+13k\big)\big(e^{-x^2/(32m\sigma^2)}+e^{-3x/(16k)}\big)\le\bar w\sqrt V\,\bar\Xi,
\]
with $\bar w:=1/\sqrt2+13/6\le2.874$, since $32m\sigma^2\le64nV$.

\emph{(ii) Final excursion.} Suppose $j\ge1$ and $L_j=0$. Then $\lambda_j=n(m+1)$ and $Z_n=W_{\varsigma_j}+R_j\le R_j-x_0$, so the event
$Z_n\ge h+x$ forces $R_j\ge h+x+x_0>x/2$. Now suppose $j=0$, or $j\ge1$ and $R_j\le x/2$; then $L_j\ge m+1\ge2$. If $j\ge1$, then
$s_j\le R_j-x_0\le x/2$ by (P2), so on the event the walk moves by at least $h+x-s_j\ge x/2$ during $E_j$ while staying above $-x_0$. By
Lemma~\ref{lem:fs-surv}(b) with $a=-x_0$ and $N=L_j$, conditionally on $\mathcal F_{\lambda_j}$ this has probability at most
$C_{LG}(2s_j+26k)(L_j\sigma^2)^{-1/2}\Xi_{L_j}(x/2)$ (for $j=0$: $C_{LG}(2h+26k)(nV)^{-1/2}\Xi_{n(m+1)}(x)$). Write $L_j=\ell(m+1)$ with
$1\le\ell\le n$, so $L_j\sigma^2=\ell V$ and $\Xi_{L_j}=\bar\Xi_\ell$. If $E_j$ is long, $\ell V\ge T_j\sigma^2$ and
$(L_j\sigma^2)^{-1/2}\Xi_{L_j}(x/2)\le\sqrt{2(j+1)}(nV)^{-1/2}\bar\Xi$. In every case
$(\ell V)^{-1/2}\bar\Xi_\ell(x/2)\le e^{1/2}V^{-1/2}\bar\Xi$, by the following bound for $y\ge0$:
$\sup_{1\le\ell\le n}\ell^{-1/2}\bar\Xi_\ell(y)\le e^{1/2}\bar\Xi_n(y)$. [The term $\ell^{-1/2}e^{-3y/(8k)}$ is at most $e^{-3y/(8k)}$. For
$g(\ell):=\ell^{-1/2}e^{-y^2/(16\ell V)}$: if $y^2\ge8nV$, $g$ is nondecreasing on $[1,n]$ and $g(\ell)\le g(n)\le e^{-y^2/(16nV)}$; if
$8V\le y^2<8nV$, $g(\ell)\le\sup_{s>0}g(s)=\sqrt{8V}e^{-1/2}/y\le e^{-1/2}<e^{-y^2/(16nV)}$; if $y^2<8V$, $g(\ell)\le1<e^{1/2}e^{-y^2/(16nV)}$.]

\emph{(iii) Cases.} On $\{\tau>n,Z_n\ge h+x,N_n=j\}$ all $j$ attempts failed. For $j=0$, (ii) gives
$C_{LG}(2h+26k)(nV)^{-1/2}\Xi_{n(m+1)}(x)\le(17h/\sqrt V+36.9)n^{-1/2}\bar\Xi$, using $\Xi_{n(m+1)}(x)\le\bar\Xi$. For $1\le j\le n/2$ let
$c_j:=2C_E\sqrt{2(j+1)}/\sqrt{nV}$ and choose a long excursion $E_\iota$ (P4). We condition backwards in time as in
Lemma~\ref{lem:fs-CF5}; attempts that carry no weight contribute $1/2$ each. The four cases, with their bounds, are:
\begin{itemize}
\item[(L1)] $\iota=j$, $R_j\le x/2$: the final excursion contributes $C_{LG}(2s_j+26k)\sqrt{2(j+1)}(nV)^{-1/2}\bar\Xi$ and attempt $j$ the
weight $\mathbb E[(2s_j+26k)\mathbf 1_{\mathrm{fail}_j}\mid\mathcal F_{\varsigma_j}]\le2w_0\sqrt V$; total
$2^{1-j}\sqrt{2(j+1)}\,(2C_{LG}w_0)\,n^{-1/2}\bar\Xi$.
\item[(L2)] $\iota=j$, $R_j>x/2$: the final excursion is long, so $L_j>0$; bound it by its survival factor $c_j(s_j+13k)$ (Lemma~\ref{lem:fs-surv}(a))
and attempt $j$ by (i); total $2^{1-j}\sqrt{2(j+1)}\,(2C_E\bar w)\,n^{-1/2}\bar\Xi$.
\item[(E1)] $\iota<j$, $R_j\le x/2$: $E_\iota$ contributes $c_j(s_\iota+13k)$, the final excursion $C_{LG}(2s_j+26k)e^{1/2}V^{-1/2}\bar\Xi$ by (ii),
attempt $j$ the weight $2w_0\sqrt V$ and, if $\iota\ge1$, attempt $\iota$ the weight $w_0\sqrt V$. Total for each $\iota\ge1$:
$2^{2-j}\sqrt{2(j+1)}\,(4C_EC_{LG}e^{1/2}w_0^2)\,n^{-1/2}\bar\Xi$; for $\iota=0$:
$2^{1-j}\sqrt{2(j+1)}\,(4C_EC_{LG}e^{1/2}w_0)\frac{h+13k}{\sqrt V}n^{-1/2}\bar\Xi$.
\item[(E2)] $\iota<j$, $R_j>x/2$: bound the final excursion by $1$ (this includes the case $L_j=0$) and attempt $j$ by
$\Pr(R_j>x/2\mid\mathcal F_{\varsigma_j})\le\bar\Xi$. The total for each $\iota\ge1$ is $2^{2-j}\sqrt{2(j+1)}(2C_Ew_0)n^{-1/2}\bar\Xi$, and
for $\iota=0$ it is $2^{1-j}\sqrt{2(j+1)}\cdot2C_E\cdot(h+13k)V^{-1/2}n^{-1/2}\bar\Xi$.
\end{itemize}
By (ii), $L_j>0$ in (L1) and (E1), where Lemma~\ref{lem:fs-surv}(b) is applied. For $j>n/2$ no long excursion is used: the final excursion
and attempt $j$ contribute at most $C_{LG}e^{1/2}\cdot2w_0\,\bar\Xi$ if $R_j\le x/2$ and at most $\bar\Xi$ if $R_j>x/2$, and attempts $1,\dots,j-1$
contribute $2^{1-j}$; since $\sum_{j\ge\lfloor n/2\rfloor+1}2^{1-j}\le2^{3/2-n/2}\le2n^{-1/2}$, these $j$ contribute at most
$2(2C_{LG}e^{1/2}w_0+1)n^{-1/2}\bar\Xi$. Summing, with $A_1:=\sum_{j\ge1}2^{1-j}\sqrt{2(j+1)}\le4.793$ and
$A_2:=\sum_{j\ge2}(j-1)2^{2-j}\sqrt{2(j+1)}\le12.43$ (the number of $\iota\ge1$ is $j-1$), the total is at most
$(7699+1192\,h/\sqrt V)\,n^{-1/2}\bar\Xi$: the coefficient of $h/\sqrt V$ is at most $17+(4C_EC_{LG}e^{1/2}w_0+2C_E)A_1\le1191.6$, and,
using $13k/\sqrt V<13/6$, the constant term is at most
$36.9+(2C_{LG}w_0+2C_E\bar w)A_1+(4C_EC_{LG}e^{1/2}w_0^2+2C_Ew_0)A_2+\frac{13}6(4C_EC_{LG}e^{1/2}w_0+2C_E)A_1+2(2C_{LG}e^{1/2}w_0+1)\le7698.91$.
\end{proof}

\begin{lemma}[Killed cycle kernels]\label{lem:fs-kernel}
For $\mathcal S\subseteq\mathbb Z$ let $K_{\mathcal S}(w,w'):=\Pr(\zeta=w'-w)$ for $w,w'\in\mathcal S$, and let $\tau_{\mathcal S}:=\min\{n\ge1:Z_n\notin\mathcal S\}$.
\begin{enumerate}
\item[(a)] $K_{\mathcal S}^t(w,w')=K_{\mathcal S}^t(w',w)$, $K_{\mathcal S}^t(w,w')=\Pr_w(\tau_{\mathcal S}>t,Z_t=w')$ and
$\sum_{w'\in\mathcal S}K_{\mathcal S}^t(w,w')=\Pr_w(\tau_{\mathcal S}>t)$.
\item[(b)] If $t=t_1+t_2+t_3$ with $t_i\ge1$, then $K_{\mathcal S}^t(w,w')\le\Pr_w(\tau_{\mathcal S}>t_1)\cdot\frac{C_Ek}{\sqrt{t_2V}}\cdot\Pr_{w'}(\tau_{\mathcal S}>t_3)$.
\end{enumerate}
\end{lemma}

\begin{proof}
(a) $K_{\mathcal S}$ is symmetric because $\zeta$ is symmetric, hence so are its powers; the other two identities are the product formula
for the walk killed outside $\mathcal S$. (b) Write $K^t=K^{t_1}K^{t_2}K^{t_3}$, bound $K^{t_2}(x,y)\le\Pr(Z_{t_2}-Z_0=y-x)\le C_Ek/\sqrt{t_2V}$
(Lemma~\ref{lem:fs-local}(a)), and use $\sum_xK^{t_1}(w,x)=\Pr_w(\tau_{\mathcal S}>t_1)$ and $\sum_yK^{t_3}(y,w')=\sum_yK^{t_3}(w',y)$.
\end{proof}

\begin{lemma}[Strip crossing]\label{lem:fs-CF11}
Let $D$ be an integer with $D\ge30\sqrt V$, $\mathcal S:=\{k+1,\dots,D\}$, and for $a',b'\in\mathcal S$ let $h_{a'}:=D-a'$, $h_{b'}:=b'-k-1$ with
$h_{a'}+h_{b'}\le D/2$. For every $t\ge1$,
\[
K_{\mathcal S}^t(a',b')\le C_\star\Big(1+\frac{h_{a'}}{\sqrt V}\Big)\Big(1+\frac{h_{b'}}{\sqrt V}\Big)\frac k{\sqrt V}\,t^{-3/2}\,\bar\Xi_t(D/14),\qquad C_\star:=10^8 .
\]
\end{lemma}

\begin{proof}
$a'-b'=D-k-1-h_{a'}-h_{b'}\ge D/2-k-1\ge0.49D$, because $D\ge30\sqrt V\ge30(6k+2)\ge100(k+1)$.
\emph{Case $t\le2$.} $K^t_{\mathcal S}(a',b')\le\Pr(Z_t-Z_0\le-0.49D)\le\bar\Xi_t(0.98D)\le\bar\Xi_t(D/14)$ by Lemma~\ref{lem:fs-F}, and
$C_\star(k/\sqrt V)t^{-3/2}\ge10^8\cdot2^{-3/2}/\sqrt{66}\ge1$.
\emph{Case $t\ge3$.} Write $t=t_1+t_2+t_3$ with $t_i\in\{\lfloor t/3\rfloor,\lceil t/3\rceil\}$, so $t_i\ge t/5$, and
$K^t_{\mathcal S}(a',b')=\sum_{x,y\in\mathcal S}K^{t_1}(a',x)K^{t_2}(x,y)K^{t_3}(y,b')$. Since $(a'-x)+(x-y)+(y-b')\ge0.49D$, every term has
$a'-x\ge D/7$, $x-y\ge D/7$ or $y-b'\ge D/7$; bound the sum by the sum of the three corresponding partial sums.
(i) $a'-x\ge D/7$: in the coordinate $u_{n'}:=D-Z_{n'}$, started at $h_{a'}\ge0$, the walk from $a'$ that stays in $\mathcal S$ for $t_1$ rounds stays at
$u\ge0$, and $u_{t_1}\ge h_{a'}+D/7$; dropping the constraint at the bottom of $\mathcal S$, Lemma~\ref{lem:fs-CF10} (applied to the walk $-Z$, which
has the same law) bounds $\sum_{x:\,a'-x\ge D/7}K^{t_1}(a',x)$ by $C_P(1+h_{a'}/\sqrt V)t_1^{-1/2}\bar\Xi_{t_1}(D/14)$. For each such $x$,
$\sum_yK^{t_2}(x,y)K^{t_3}(y,b')\le\frac{C_Ek}{\sqrt{t_2V}}\Pr_{b'}(\tau_{\mathcal S}>t_3)\le\frac{C_Ek}{\sqrt{t_2V}}\frac{C_S(1+h_{b'}/\sqrt V)}{\sqrt{t_3}}$ by
Lemma~\ref{lem:fs-kernel} and Lemma~\ref{lem:fs-CF5} in the coordinate $Z-k-1$.
(ii) $x-y\ge D/7$: $K^{t_2}(x,y)\le\Pr(Z_{t_2}-Z_0\in(-(x-y)-k,-(x-y)])\le C_{LG}\frac k{\sqrt{t_2V}}\bar\Xi_{t_2}(D/7)$ by
Lemma~\ref{lem:fs-local}(b) ($t_2(m+1)\ge2$), and the outer sums are survival probabilities bounded by Lemma~\ref{lem:fs-CF5}:
$\sum_xK^{t_1}(a',x)\le C_S(1+h_{a'}/\sqrt V)t_1^{-1/2}$ (coordinate $D-Z$) and $\sum_yK^{t_3}(y,b')\le C_S(1+h_{b'}/\sqrt V)t_3^{-1/2}$.
(iii) $y-b'\ge D/7$: by Lemma~\ref{lem:fs-kernel}(a), $K^{t_3}(y,b')=K^{t_3}(b',y)$, and the argument of (i) in the coordinate $Z-k-1$ started at $h_{b'}$
bounds $\sum_{y:\,y-b'\ge D/7}K^{t_3}(b',y)$ by $C_P(1+h_{b'}/\sqrt V)t_3^{-1/2}\bar\Xi_{t_3}(D/14)$; and
$\sum_xK^{t_1}(a',x)K^{t_2}(x,y)\le\frac{C_Ek}{\sqrt{t_2V}}C_S(1+h_{a'}/\sqrt V)t_1^{-1/2}$.
Since $(t_1t_2t_3)^{-1/2}\le5^{3/2}t^{-3/2}$ and $\bar\Xi_{t_i}(D/14)\le\bar\Xi_t(D/14)$, $\bar\Xi_t(D/7)\le\bar\Xi_t(D/14)$, the three parts sum to at most
$(2C_PC_EC_S+C_S^2C_{LG})5^{3/2}(1+h_{a'}/\sqrt V)(1+h_{b'}/\sqrt V)\frac k{\sqrt V}t^{-3/2}\bar\Xi_t(D/14)$, and
$(2C_PC_EC_S+C_S^2C_{LG})5^{3/2}\le9.3\cdot10^7$.
\end{proof}

\subsection{Strip supersolution and weak Harnack inequality}\label{sec:fs-WH}

The conditioned laws $D_n$ could in principle drift into the band. A sine supersolution on the outer strip $O$ shows that mass which stays
in $O$ decays at rate $\rho<1$ per cycle (Lemma~\ref{lem:fs-SS}), while the subsolution $h_\sharp$ keeps a fixed fraction of the surviving mass in
the core $\mathsf C$; together they keep $D_n(h_\sharp)$ bounded below (Theorem~\ref{thm:fs-WH}), so the survival probability over $J$ cycles is
bounded below and the conditioned bound reduces to an unconditioned one (Proposition~\ref{prop:fs-RED}).

Let $g(u):=\sin(\omega(u+\delta))$ for $u\in\mathbb R$; put $\Psi(x):=g(d(x))$ if $x\in V_Q$ and $-k\le d(x)\le2\delta$, and $\Psi(x):=0$
otherwise. Let $\tilde Q(x,y):=Q(x,y)\mathbf 1[y\notin\mathsf Z]$, $\hat\kappa(v):=\sum_d\kappa(d)\cos(vd)$ and
\[
\rho:=\hat\kappa(\omega)^{m+1}+2\eta_F/g(k+1).
\]
By Lemma~\ref{lem:un-arith}(k), $g$ is positive and concave on $[-M',2\delta+M']$ and nondecreasing on $[-k,2\delta+1]$, so $0\le\Psi\le1$,
$\Psi\ge\Psi_{\min}=g(-k)$ on $O\setminus\mathsf Z=\{x\in V_Q:-k\le d(x)\le2\delta\}$, and $\Psi=0$ on $\mathsf Z\cup \mathsf C$.

\begin{lemma}[Strip supersolution]\label{lem:fs-SS}
For every $x\in O\setminus\mathsf Z$, $(\tilde Q\Psi)(x)\le\rho\,\Psi(x)$. Moreover $1-q_0/2\le\rho<1$ and $\log(1/\rho)\ge0.99\,\omega^2V/2$.
\end{lemma}

\begin{proof}
Since $\Psi=0$ on $\mathsf Z$, $(\tilde Q\Psi)(x)=(Q\Psi)(x)$.
\emph{Interior case, $k+1\le d(x)\le2\delta$.} Then $x\in\mathsf I$ is a corridor vertex of an arm (by \eqref{eq:fs-corr}) and
$Q(x,\cdot)=(P^\flat)^{m+1}(x,\cdot)$ (Lemma~\ref{lem:fs-coupling}(d)). Run the step chain from $X_0=x$, let $S_j:=\xi_1+\dots+\xi_j$ and
$G':=\{\max_{j\le m+1}|S_j|<M'\}$. The vertices of the arm with $|d-d(x)|<M'$ have $d\in(k+1-M',2\delta+M')$ and are corridor vertices
(Lemma~\ref{lem:un-arith}(i) and \eqref{eq:fs-corr}), so on $G'$ Lemma~\ref{lem:fs-coupling}(c) gives $d(X_{m+1})=d(x)-S_{m+1}\in(-M',2\delta+M')$, where $g\ge0$;
hence $\Psi(X_{m+1})\le g(d(x)-S_{m+1})$. Off $G'$, $\Psi(X_{m+1})\le1\le g(d(x)-S_{m+1})+2$. So
$(Q\Psi)(x)\le\mathbb E\,g(d(x)-S_{m+1})+2\Pr(G'^c)$. Now
$\mathbb E\sin(\omega(d(x)+\delta)-\omega S_{m+1})=g(d(x))\,\mathbb E\cos(\omega S_{m+1})-\cos(\omega(d(x)+\delta))\,\mathbb E\sin(\omega S_{m+1})=\hat\kappa(\omega)^{m+1}g(d(x))$,
since $S_{m+1}$ is symmetric and $\mathbb Ee^{i\omega S_{m+1}}=\hat\kappa(\omega)^{m+1}$. With $\Pr(G'^c)\le\eta_F$ (Lemma~\ref{lem:fs-F}) and
$g(d(x))\ge g(k+1)>0$, $(Q\Psi)(x)\le\rho\,g(d(x))=\rho\Psi(x)$.

\emph{Band case, $-k\le d(x)\le k$.} By Lemma~\ref{lem:un-accounting}(iii) and Lemma~\ref{lem:un-blindfold}(b),
$Q(x,\cdot)=\sum_yP_{\mathrm{lo}}(x,y)P^m(y,\cdot)$ with $\sum_yP_{\mathrm{lo}}(x,y)\le1$ and $P_{\mathrm{lo}}(x,\cdot)$ supported on $N_b$; so
$(Q\Psi)(x)\le\max_{y\in N_b}(P^m\Psi)(y)$. Fix $y\in N_b$, a corridor vertex of an arm with $0\le d(y)\le2k$, run the step chain from $y$ and let
$G:=\{\max_{j\le m}|S_j|<M'\}$. On $G$, $d(X_m)=d(y)-S_m\in(-M',2k+M')$ by Lemma~\ref{lem:fs-coupling}(c). If $X_m\in\mathsf Z$ then
$\Psi(X_m)=0$. Otherwise $-k\le d(X_m)\le2k+M'\le2\delta$ and, by concavity of $g$ on $[-M',2\delta+M']$,
$\Psi(X_m)=g(d(X_m))\le g(-k)+g'(-k)(d(X_m)+k)\le g(-k)+\omega(3k+|S_m|)$, as $0\le g'(-k)\le\omega$. Hence, using
Lemma~\ref{lem:un-band}(a) (the level of $y$ lies in $(A-2k,A+1]$) and $\mathbb E|S_m|\le\sqrt{m\sigma^2}\le8.07k$,
\[
(P^m\Psi)(y)\le g(-k)\big(1-P^m(y,\mathsf Z)\big)+\omega(3k+8.07k)+\Pr(G^c)\le(1-q_0)g(-k)+11.07\,k\omega+\eta_F .
\]
Since $\omega(\delta-k)\le\pi/2$ and $\sin v\ge2v/\pi$ on $[0,\pi/2]$, $\frac{q_0}2g(-k)\ge\frac{q_0}\pi\omega(\delta-k)\ge\frac{9\cdot6248}{256\pi}k\omega\ge69.9\,k\omega$,
which exceeds $11.07k\omega+\eta_F$ because $k\omega\ge8.37/q$ and $\eta_F=e^{-q/24}\le q^{-1}$. So
$(Q\Psi)(x)\le(1-q_0/2)g(-k)\le(1-q_0/2)\Psi(x)$, and it remains to show $\rho\ge1-q_0/2$.

\emph{Bounds on $\rho$.} $\cos v\ge1-v^2/2$ gives $\hat\kappa(\omega)\ge1-\omega^2\sigma^2/2>0$, so
$\rho\ge(1-\omega^2\sigma^2/2)^{m+1}\ge1-\omega^2V/2\ge1-2.4\cdot10^{-7}>1-q_0/2$ (Lemma~\ref{lem:un-arith}(e),(k)). For the upper bound let
$Y:=\omega^2V/2$; then $1263\,q^{-2}\le Y\le2.4\cdot10^{-7}$ by $V\ge36k^2$ and Lemma~\ref{lem:un-arith}(k). From $\cos v\le1-v^2/2+v^4/24$
and $\sum_d\kappa(d)d^4\le k^2\sigma^2$, $\hat\kappa(\omega)\le1-\frac{\omega^2\sigma^2}2(1-\frac{\omega^2k^2}{12})$, so
$\hat\kappa(\omega)^{m+1}\le e^{-Y(1-\omega^2k^2/12)}\le e^{-0.99999Y}$. Hence
$e^{-0.99Y}-\hat\kappa(\omega)^{m+1}\ge e^{-Y}(e^{0.00999Y}-1)\ge0.0099\,Y\ge12.5\,q^{-2}$, which exceeds
$2\eta_F/g(k+1)\le4.01e^{-q/24}$ ($g(k+1)\ge g(-k)\ge0.4998$). So $\rho\le e^{-0.99Y}<1$.
\end{proof}

For $x\in V_Q$ and $j\ge1$ let $N_j(x):=\sum Q(x,y_1)Q(y_1,y_2)\cdots Q(y_{j-1},y_j)$ over $y_1,\dots,y_j\in O$ if $x\in O$, and $N_j(x):=0$ if
$x\notin O$; this is the mass of the cycle chain from $x$ that is alive at low round $j$ and has stayed in $O$ at low rounds $0,\dots,j$.

\begin{corollary}[Decay in the outer region]\label{cor:fs-SS}
For every $x\in V_Q$ and $j\ge1$, $N_j(x)\le\rho^{j-1}/\Psi_{\min}$.
\end{corollary}

\begin{proof}
Let $x\in O$. In a nonzero term, $y_1,\dots,y_{j-1}\notin\mathsf Z$, since $Q(y,\cdot)=0$ for $y\in\mathsf Z$ (Lemma~\ref{lem:un-accounting}(ii)); and if
$x\in\mathsf Z$ then $N_j(x)=0$. Summing over $y_j$ first, $N_j(x)\le(\tilde Q_O^{\,j-1}\mathbf 1)(x)$ with
$\tilde Q_O(x',y):=Q(x',y)\mathbf 1[y\in O\setminus\mathsf Z]$. On $O\setminus\mathsf Z$, $\mathbf 1\le\Psi/\Psi_{\min}$, and
$\tilde Q_O\Psi=\tilde Q\Psi\le\rho\Psi$ there (Lemma~\ref{lem:fs-SS}; $\Psi$ vanishes off $O\setminus\mathsf Z$). Since $\tilde Q_O$ only uses
values on $O\setminus\mathsf Z$, induction gives $\tilde Q_O^{\,j-1}\Psi\le\rho^{j-1}\Psi\le\rho^{j-1}$ on $O\setminus\mathsf Z$.
\end{proof}

\begin{theorem}[Weak Harnack inequality from the start]\label{thm:fs-WH}
For every $n\ge0$, $F_n:=D_n(h_\sharp)\ge f_\ast:=\frac12\tilde\mu_{\sharp,c}^{\,2J}h_{\mathsf C}\ge3.83\cdot10^{-4}$, and
\[
\frac{\Lambda_{n+J}}{\Lambda_n}=D_n(Q^J\mathbf 1)\ \ge\ c_{\mathrm{WH}}:=\tilde\mu_{\sharp,c}^{\,J}f_\ast\ \ge\ 2.81\cdot10^{-5}.
\]
\end{theorem}

\begin{proof}
All $D_n$ are probability measures supported in $V_Q$ (Lemma~\ref{lem:un-blindfold}(c), Lemma~\ref{lem:fs-star}). Write $\tilde\mu:=\tilde\mu_{\sharp,c}$.
(1) $F_{n+1}=D_n(Qh_\sharp)/D_n(Q\mathbf 1)\ge\tilde\mu F_n$, since $Qh_\sharp\ge\tilde\mu h_\sharp$ (Lemma~\ref{lem:fs-star}) and $Q\mathbf 1\le\mathbf 1$.

(2) Let $x\in V_Q$. Group the paths $x=y_0,y_1,\dots,y_J$ of $Q^J$ by the first index $i\le J$ with $y_i\in \mathsf C$; let $p_x$ be the total weight
$\sum_{i\le J}\sum Q(y_0,y_1)\cdots Q(y_{i-1},y_i)$ over paths with $y_0,\dots,y_{i-1}\in O$, $y_i\in \mathsf C$. Paths without a visit to $\mathsf C$ have
total weight $N_J(x)$. Since $h_\sharp\ge0$, $Q^{J-i}h_\sharp\ge\tilde\mu^{J-i}h_\sharp\ge\tilde\mu^Jh_{\mathsf C}$ on $\mathsf C$, and $Q^{J-i}\mathbf 1\le1$,
\[
(Q^Jh_\sharp)(x)\ge\tilde\mu^Jh_{\mathsf C}\,p_x,\qquad(Q^J\mathbf 1)(x)\le p_x+N_J(x).
\]
With $c_1:=\tilde\mu^Jh_{\mathsf C}$ and $\eta_J:=\sup_xN_J(x)$ this gives $Q^Jh_\sharp\ge c_1(Q^J\mathbf 1-\eta_J)$ on $V_Q$. As
$D_n(Q^J\mathbf 1)\ge D_n(Q^Jh_\sharp)\ge\tilde\mu^JF_n$ by Lemma~\ref{lem:fs-star},
\[
F_{n+J}=\frac{D_n(Q^Jh_\sharp)}{D_n(Q^J\mathbf 1)}\ge c_1\Big(1-\frac{\eta_J}{D_n(Q^J\mathbf 1)}\Big)\ge c_1\Big(1-\frac{\eta_J}{\tilde\mu^JF_n}\Big).
\]
(3) By Lemma~\ref{lem:fs-SS}, Lemma~\ref{lem:un-arith}(k),(l) and $J\ge L^2/V$,
$(J-1)\log\frac1\rho-2J\log\frac1{\tilde\mu}\ge0.495\,\omega^2(L^2-V)-5.2212\ge29.5>\log\frac4{\Psi_{\min}h_{\mathsf C}}$, the right side being at most
$4.031$. Hence, by Corollary~\ref{cor:fs-SS}, $\eta_J\le\rho^{J-1}/\Psi_{\min}\le\tilde\mu^{2J}h_{\mathsf C}/4=\tilde\mu^Jc_1/4$.
(4) If $F_n\ge c_1/2$, then (2) and (3) give $F_{n+J}\ge c_1(1-\frac12)=c_1/2$. Since $F_0=h_\sharp(o)=1$, induction gives $F_{iJ}\ge c_1/2$ for all
$i\ge0$, and by (1) $F_{iJ+l}\ge\tilde\mu^lc_1/2\ge\tilde\mu^Jc_1/2=f_\ast$ for $0\le l<J$.
(5) $D_n(Q^J\mathbf 1)=\mu_nQ^J\mathbf 1/\Lambda_n=\Lambda_{n+J}/\Lambda_n$, and $D_n(Q^J\mathbf 1)\ge\tilde\mu^JF_n\ge\tilde\mu^Jf_\ast$. By
Lemma~\ref{lem:un-arith}(l), $\tilde\mu^J\ge e^{-2.6106}$, so $f_\ast\ge e^{-5.2212}\cdot0.1422/2\ge3.83\cdot10^{-4}$ and $c_{\mathrm{WH}}\ge2.81\cdot10^{-5}$.
\end{proof}

\begin{proposition}[Reduction to an unconditioned bound]\label{prop:fs-RED}
Suppose that, for a number $C_H$,
\[
\text{(H1)}\ \ \sup_{x\in V_Q}(Q^J\mathbf 1_{\mathsf E})(x)\le C_H\frac{k^2}{L^2},\qquad\text{(H2)}\ \ \max_{1\le n\le J}(Q^n\mathbf 1_{\mathsf E})(o)\le C_H\frac{k^2}{L^2}.
\]
Then $D_n(\mathsf T)\le D_n(\mathsf E)\le C_Hk^2/(c_{\mathrm{WH}}L^2)$ for every $n\ge0$.
\end{proposition}

\begin{proof}
$\mathsf T\subseteq\mathsf E$. For $n=0$, $D_0(\mathsf E)=0$ as $o\in\mathsf I$. For $1\le n\le J$, $D_n(\mathsf E)=(Q^n\mathbf 1_{\mathsf E})(o)/\Lambda_n$ and
$\Lambda_n\ge\tilde\mu_{\sharp,c}^{\,J}\ge e^{-2.6106}\ge c_{\mathrm{WH}}$ (Lemma~\ref{lem:fs-star}). For $n>J$,
$D_n(\mathsf E)=\mu_{n-J}Q^J\mathbf 1_{\mathsf E}/\Lambda_n=D_{n-J}(Q^J\mathbf 1_{\mathsf E})\,\Lambda_{n-J}/\Lambda_n$, and Theorem~\ref{thm:fs-WH} gives
$\Lambda_n/\Lambda_{n-J}\ge c_{\mathrm{WH}}$.
\end{proof}

\subsection{Renewal over relocations}\label{sec:fs-REN}

Mass of the cycle chain that reaches the band is either killed or relocated by the low round to $N_b$ and restarted; a relocation from $N_b$
reaches the dead zone with probability at least $q_0$ (Lemma~\ref{lem:un-band}). Decomposing the cycle chain at these relocations reduces
(H1)--(H2) to first-passage bounds for the free walk.

\begin{lemma}[Decomposition]\label{lem:fs-decomp}
\begin{enumerate}
\item[(a)] For every finitely supported $\mu\ge0$, $\mu Q=\mu\hat G+\varrho(\mu)P^m$, and for $n\ge1$,
$\mu Q^n=\mu\hat G^n+\sum_{u=0}^{n-1}\varrho(\mu\hat G^u)P^mQ^{n-1-u}$.
\item[(b)] For $y\in N_b$ and $t\ge1$ let $\Phi_t(y):=(\delta_yP^mQ^{t-1})(\mathsf E)$, $\phi^y_t:=(\delta_yP^m\hat G^{t-1})(\mathsf E)$ and
$K_b(t)(y,y''):=\varrho(\delta_yP^m\hat G^{t-1})(y'')$ for $y''\in N_b$. Then, as vectors on $N_b$,
\[
\Phi_t=\phi_t+\sum_{t'=1}^{t-1}K_b(t')\,\Phi_{t-t'};
\]
$\sum_{t'\ge1}\sum_{y''}K_b(t')(y,y'')\le1-q_0$; $K_b(t)(y,y'')\le\phi^y_t$; $\sum_{t\ge1}\phi^y_t\le1$; and
$\phi^y_t=\sum_{y'}P^m(y,y')f^{y'}_{t-1}$ with $f^{y'}_0:=\mathbf 1_{\mathsf E}(y')$ and $f^{y'}_s:=0$ for $y'\notin\mathsf I$, $s\ge1$.
\item[(c)] For every probability $\nu$ supported in $V_Q$, $\sum_{s\ge0}(\nu Q^s)(\mathsf B)\le1/q_0$ and $\sum_{s\ge0}(\nu Q^s)(\mathsf Z)\le1$; in particular
$\sum_{t\ge1}\Phi_t(y)\le1+1/q_0$ for $y\in N_b$.
\item[(d)] For $x\in V_Q$, $\sum_{u\ge0}\varrho(\delta_x\hat G^u)\mathbf 1\le1$, and $\varrho(\delta_x\hat G^u)\mathbf 1\le f^x_u$ for $u\ge1$.
\end{enumerate}
\end{lemma}

\begin{proof}
(a) $V(G)=\mathsf I\cup\mathsf B\cup\mathsf Z$. By Lemma~\ref{lem:un-accounting}(ii), $(\mu\mathbf 1_{\mathsf Z})Q=0$; by definition $(\mu\mathbf 1_{\mathsf I})Q=\mu\hat G$; and
$(\mu\mathbf 1_{\mathsf B})Q=\big((\mu\mathbf 1_{\mathsf B})P_{\mathrm{lo}}\big)P_{\mathrm{hi}}^m=\varrho(\mu)P^m$ by Lemma~\ref{lem:un-blindfold}(a),(b). The second
identity follows by induction on $n$: $\mu Q^{n+1}=(\mu\hat G)Q^n+\varrho(\mu)P^mQ^n$, and the induction hypothesis applied to $\mu\hat G$ gives the
terms $u\ge1$ after the shift $u\mapsto u+1$.
(b) For $t=1$ the identity is $\Phi_1=\phi_1$. For $t\ge2$ apply (a) to $\mu:=\delta_yP^m$ with $n=t-1$, evaluate at $\mathsf E$, and put $t':=u+1$:
the term $u$ is $\sum_{y''}\varrho(\mu\hat G^u)(y'')(\delta_{y''}P^mQ^{t-1-t'})(\mathsf E)=\sum_{y''}K_b(t')(y,y'')\Phi_{t-t'}(y'')$. Since
$\varrho(\nu)\mathbf 1=\sum_{x\in\mathsf B}\nu(x)(P_{\mathrm{lo}}\mathbf 1)(x)\le\nu(\mathsf B)$, the row sums satisfy
$\sum_{y''}K_b(t)(y,y'')\le(\mu\hat G^{t-1})(\mathsf B)\le(\mu\hat G^{t-1})(\mathsf E)=\phi^y_t$, which also bounds each entry. By
Lemma~\ref{lem:fs-coupling}(d), for $u\ge1$, $(\mu\hat G^u)(\cdot)=\sum_{z\in\mathsf I}\mu(z)\Pr_z(\tau_{\mathsf E}\ge u,Y_u\in\cdot)$; hence
$\sum_{u\ge0}(\mu\hat G^u)(\mathsf B)\le\mu(\mathsf B)+\mu(\mathsf I)=1-P^m(y,\mathsf Z)\le1-q_0$ by Lemma~\ref{lem:un-band}(a), and
$\sum_{t\ge1}\phi^y_t=\mu(\mathsf E)+\sum_{z\in\mathsf I}\mu(z)\Pr_z(\tau_{\mathsf E}<\infty)\le1$. The formula for $\phi^y_t$ is
$(\mu\hat G^{t-1})(\mathsf E)=\sum_{y'}P^m(y,y')(\delta_{y'}\hat G^{t-1})(\mathsf E)$.
(c) Write $\nu_s:=\nu Q^s$. By \eqref{eq:un-accounting}, $\nu_s(\mathsf Z)+b(\nu_s)=\nu_s\mathbf 1-\nu_{s+1}\mathbf 1$, and by Lemma~\ref{lem:un-band}(b),
$q_0g(\nu_s)\le\nu_{s+1}(\mathsf Z)$; as $\nu_s(\mathsf B)=g(\nu_s)+b(\nu_s)$, $q_0\nu_s(\mathsf B)\le\nu_{s+1}(\mathsf Z)+b(\nu_s)$. Summing over $s$ and
telescoping, $\sum_s\nu_s(\mathsf Z)\le\nu\mathbf 1=1$ and $q_0\sum_s\nu_s(\mathsf B)\le\sum_s(\nu_s(\mathsf Z)+b(\nu_s))\le1$. For $y\in N_b$ take
$\nu=\delta_yP^m$ (Lemma~\ref{lem:un-blindfold}(b)).
(d) For $x\in\mathsf Z$ all terms vanish; for $x\in\mathsf B$ only $u=0$ contributes, at most $1$; for $x\in\mathsf I$, $\varrho(\delta_x)=0$ and for $u\ge1$
$\varrho(\delta_x\hat G^u)\mathbf 1\le(\delta_x\hat G^u)(\mathsf B)=\Pr_x(\tau_{\mathsf E}=u,Y_u\in\mathsf B)\le f^x_u$, with sum at most $1$.
\end{proof}

\begin{proposition}[Renewal bound]\label{prop:fs-REN}
Let $C_1\ge1$ and $C_2>0$, and suppose that
\begin{itemize}
\item[(FP1)] $f^z_u\le C_1(1+d(z)/\sqrt V)\,u^{-3/2}$ for $z\in\mathsf I$ and $1\le u\le2J$;
\item[(FP2)] $f^z_u\le C_2k^2/L^2$ for $z\in\mathsf I$ and $u\ge J/4-1$, $u\ge1$;
\item[(FP3)] $f^o_u\le C_2k^2/L^2$ for all $u\ge1$.
\end{itemize}
Then (H1) and (H2) hold with $C_H:=C_\Phi+C_2(1+9k(1+1/q_0))$, where
$C_\Phi:=q_0^{-1}C_2+9k\cdot6.7C_1\cdot\Sigma_{5/2}\cdot0.0429+9kC_2q_0^{-2}$ and $\Sigma_{5/2}:=\sum_{n\ge1}n^{5/2}(1-q_0)^{n-1}\le4.26\cdot10^5$.
\end{proposition}

\begin{proof}
\emph{Step 0.} For $\tau>0$ let $\phi^\ast_{\ge\tau}:=\sup_{s\ge\tau}\max_{y\in N_b}\phi^y_s$ and $\phi^\ast_{>\tau}:=\sup_{s>\tau}\max_y\phi^y_s$. Let $y\in N_b$
and $s\ge2$. Under $P^m(y,\cdot)$, on $G:=\{\max_{j\le m}|S_j|<M'\}$ the endpoint $y'$ has $d(y')\le2k+|S_m|$ (Lemma~\ref{lem:fs-coupling}(c)), and
$d(y')\le A+1\le L$ always; so $\sum_{y'\in\mathsf I}P^m(y,y')(1+d(y')/\sqrt V)\le1+(2k+\sqrt{m\sigma^2})/\sqrt V+\eta_FL/\sqrt V\le2.334$. If
$s-1\le2J$, (FP1) and $(s-1)^{-3/2}\le2^{3/2}s^{-3/2}$ give $\phi^y_s\le2.334\cdot2^{3/2}C_1s^{-3/2}\le6.7C_1s^{-3/2}$; if $s-1>2J$, (FP2) gives
$\phi^y_s\le C_2k^2/L^2$; and $\phi^y_1\le1\le C_1$. If $s\ge J/4$, then $s\ge2$, $s-1\ge J/4-1$, and (FP2) gives $\phi^y_s\le C_2k^2/L^2$. Hence
$\phi^\ast_{>\tau}\le6.7C_1\tau^{-3/2}+C_2k^2/L^2$ for every $\tau>0$ (for $\tau<1$ the bound exceeds $1$), and $\phi^\ast_{\ge\tau}\le C_2k^2/L^2$ for $\tau\ge J/4$.

\emph{Step 1.} Iterating Lemma~\ref{lem:fs-decomp}(b), a finite recursion,
$\Phi_t=\sum_{n\ge0}\sum K_b(t_1)\cdots K_b(t_n)\phi_s$ over $t_1,\dots,t_n,s\ge1$ with $t_1+\dots+t_n+s=t$. Let $\bar K:=\sum_{t'\ge1}K_b(t')$
(entrywise); its row sums are at most $1-q_0$, so $\bar K^n\mathbf 1\le(1-q_0)^n\mathbf 1$. The terms with $s\ge t/2$ contribute at most
$\sum_n\bar K^n\mathbf 1\,\phi^\ast_{\ge t/2}\le q_0^{-1}\phi^\ast_{\ge t/2}$. In a term with $s<t/2$ we have $n\ge1$ and some $t_i>t/(2n)$; for fixed $n$ and
$i$, replace $K_b(t_i)$ by the larger matrix $\phi^\ast_{>t/(2n)}\mathbf 1\mathbf 1^\top$ (Lemma~\ref{lem:fs-decomp}(b): every entry of $K_b(t_i)$ is at most
$\phi^y_{t_i}$), and sum the other indices without the constraint $\sum t_l+s=t$ (for each choice of them at most one $t_i$ is admissible). Using
$\sum_s\phi^{y''}_s\le1$, $\mathbf 1^\top\bar K^{n-i}\mathbf 1\le|N_b|(1-q_0)^{n-i}$ and $|N_b|\le9k$, the contribution of $(n,i)$ is at most
$9k(1-q_0)^{n-1}\phi^\ast_{>t/(2n)}$. Hence, with $\Phi^\ast_t:=\max_{N_b}\Phi_t$ and Step~0,
\[
\Phi^\ast_t\le q_0^{-1}\phi^\ast_{\ge t/2}+9k\sum_{n\ge1}n(1-q_0)^{n-1}\Big(6.7C_1\big(\tfrac{2n}t\big)^{3/2}+C_2\frac{k^2}{L^2}\Big).
\]
For $t\ge J/2$: $\phi^\ast_{\ge t/2}\le C_2k^2/L^2$; $\sum_nn(1-q_0)^{n-1}=q_0^{-2}$; and $(2/t)^{3/2}\le(4V/L^2)^{3/2}\le4290k^3/L^3\le0.0429k^2/L^2$
(Lemma~\ref{lem:un-arith}(e), $k/L\le10^{-5}$). So $\Phi^\ast_t\le C_\Phi k^2/L^2$ for $t\ge J/2$. The bound on $\Sigma_{5/2}$ is the Cauchy--Schwarz
inequality $\Sigma_{5/2}\le(\sum_nn^2x^{n-1})^{1/2}(\sum_nn^3x^{n-1})^{1/2}=\big(\frac{1+x}{(1-x)^3}\frac{1+4x+x^2}{(1-x)^4}\big)^{1/2}\le\sqrt{12}\,q_0^{-7/2}$
with $x=1-q_0$, and $\sqrt{12}(256/9)^{7/2}\le4.26\cdot10^5$.

\emph{Step 2 (H1).} Let $x\in V_Q$ and $\varrho^x_u:=\varrho(\delta_x\hat G^u)$. By Lemma~\ref{lem:fs-decomp}(a),
\[
(Q^J\mathbf 1_{\mathsf E})(x)=f^x_J+\sum_{u=0}^{J-1}\sum_{y\in N_b}\varrho^x_u(y)\Phi_{J-u}(y),
\]
with $f^x_J:=0$ for $x\notin\mathsf I$. Split the sum at $u=J/2$.
For $u\le J/2$, $\Phi_{J-u}\le C_\Phi k^2/L^2$ by Step~1 and $\sum_u\varrho^x_u\mathbf 1\le1$ (Lemma~\ref{lem:fs-decomp}(d)). For $u>J/2$ we have $u\ge1$,
$\varrho^x_u\mathbf 1\le f^x_u\le C_2k^2/L^2$ by (FP2), and $\sum_{u}\sum_y\Phi_{J-u}(y)\le\sum_{y\in N_b}\sum_t\Phi_t(y)\le9k(1+1/q_0)$
(Lemma~\ref{lem:fs-decomp}(c)). With $f^x_J\le C_2k^2/L^2$ (FP2),
\[
(Q^J\mathbf 1_{\mathsf E})(x)\le\big(C_2+C_\Phi+9kC_2(1+1/q_0)\big)k^2/L^2=C_Hk^2/L^2.
\]

\emph{Step 3 (H2).} For $1\le n\le J$, since $o\in\mathsf I$, $\varrho^o_0=0$ and
$(Q^n\mathbf 1_{\mathsf E})(o)=f^o_n+\sum_{u=1}^{n-1}\sum_y\varrho^o_u(y)\Phi_{n-u}(y)\le\max_uf^o_u\,(1+9k(1+1/q_0))\le C_Hk^2/L^2$ by (FP3).
\end{proof}

\subsection{First-passage bounds for the free walk}\label{sec:fs-FP}

The free walk follows the cycle walk of law $\kappa^{*(m+1)}$ on one arm while it stays in a strip $\{k+1\le d\le D\}$ (Lemma~\ref{lem:fs-corr}).
A last-exit decomposition at the strip splits a first passage into a passage inside the strip, bounded by the killed-kernel estimates of
Section~\ref{sec:fs-walk}, and a crossing of the strip from above, bounded by Green-function and gambler's-ruin estimates; returns from levels
$d>L/4$, where the chain may meet the arbitrary laws near $o$, are exponentially rare (Lemma~\ref{lem:fs-FPL}).

Throughout this subsection $D$ is an integer with $30\sqrt V\le D\le L/4$, $\mathcal S_D:=\{x\in V(G):k+1\le d(x)\le D\}\subseteq\mathsf I$ and
$F_D:=\{x\in\mathsf I:d(x)>D\}$. For the step chain from $y$ let $\nu:=\min\{n\ge1:d(Y_n)\notin[k+1,D]\}$; for $y\in\mathsf I$ the step chain and the
free walk agree up to $\nu\le\tau_{\mathsf E}$. The coupled cycle walk is $Z_n:=d(y)-(\xi_1+\dots+\xi_{n(m+1)})$, with
$\nu^Z:=\min\{n\ge1:Z_n\notin[k+1,D]\}$. By \eqref{eq:fs-corr}, every vertex $y\in\mathsf I$ with $d(y)\le L/4$ is a corridor vertex of an arm,
and the walk below is followed along that arm; all statements depend on the start only through its level. We put
$\varepsilon_c:=e^{-q/7}$, $\varepsilon_B:=2e^{-D/(2k)}$ and $\ell_0:=k-x_0=-11k$.

\begin{lemma}[Corridor errors]\label{lem:fs-corr}
Let $y\in\mathsf I$ with $d(y)\le L/4$. Outside an event of probability at most $\varepsilon_c$, $\nu=\nu^Z$ and $d(Y_n)=Z_n$ for all $n\le\nu^Z$.
Consequently, for every $j\ge1$ and every set $\Gamma\subseteq\mathbb Z^{j+1}$,
$|\Pr_y(\nu=j,(d(Y_n))_{n\le j}\in\Gamma)-\Pr(\nu^Z=j,(Z_n)_{n\le j}\in\Gamma)|\le\varepsilon_c$.
\end{lemma}

\begin{proof}
Let $\mathcal B_n:=\{\max_{1\le l\le m+1}|\xi_{n(m+1)+1}+\dots+\xi_{n(m+1)+l}|\ge L/4-1\}$. If cycle $n$ starts at a vertex of an arm at level
$d\in[k+1,L/4]$, the vertices of that arm with level in $(d-L/4+1,d+L/4-1)$ are corridor vertices (by \eqref{eq:fs-corr}), so on
$\mathcal B_n^c$ Lemma~\ref{lem:fs-coupling}(c) gives that the cycle ends on the same arm at level $d-(\xi_{n(m+1)+1}+\dots+\xi_{(n+1)(m+1)})$. Since
$Z_0=d(y)\le L/4$ and $Z_n\in[k+1,D]$ for $1\le n<\nu^Z$, induction shows that on $\bigcap_{n<\nu^Z}\mathcal B_n^c$ we have $d(Y_n)=Z_n$ for $n\le\nu^Z$,
hence $\nu=\nu^Z$. As $\mathcal B_n$ is independent of $\{\nu^Z>n\}\in\mathcal G_{n(m+1)}$,
$\Pr(\bigcup_{n<\nu^Z}\mathcal B_n)\le\sum_n\Pr(\nu^Z>n)\Pr(\mathcal B_0)=\mathbb E\nu^Z\cdot\Pr(\mathcal B_0)$. If $d(y)\le D$, then $\mathbb E\nu^Z\le7.12(4D^2/V+1)$ (Lemma~\ref{lem:fs-ruin}(c)); if $d(y)>D$, the first cycle may enter $[k+1,D]$, and the
Markov property after it gives $\mathbb E\nu^Z\le1+\sup_{w\in[k+1,D]}\mathbb E_w\nu^Z\le1+7.12(4D^2/V+1)$. In both cases
$\mathbb E\nu^Z\le8.12+7.12\,q^2/144\le0.051q^2$, as $D\le L/4$, $V\ge36k^2$ and $q\ge10^5$. By Lemma~\ref{lem:fs-F} with
$x=L/4-1\ge198k$, $\Pr(\mathcal B_0)\le2(e^{-x^2/(4V)}+e^{-3x/(4k)})\le4e^{-3x/(4k)}\le e^{-q/6}$. Finally $0.051q^2e^{-q/6}\le e^{-q/7}$.
\end{proof}

\begin{lemma}[Gambler's ruin for the free walk]\label{lem:fs-CF8}
Let $z\in\mathcal S_D$. Then $\Pr_z(\nu<\infty,\ d(Y_\nu)>D)\le\big(d(z)-k+3.4\sqrt V\big)/D+2\varepsilon_B+2\eta_F$.
\end{lemma}

\begin{proof}
Let $W'_j:=d(z)-(\xi_1+\dots+\xi_j)$, the coupled step walk in $d$-coordinates (its increments have law $\kappa$). Define attempts for $W'$ as in
Section~\ref{sec:fs-walk} with the deep level $\ell_0$ in place of $-x_0$: $\lambda_0:=0$, $\varsigma_i:=\min\{j\ge\lambda_{i-1}:W'_j\le\ell_0\}$,
$\lambda_i:=(m+1)\lceil\varsigma_i/(m+1)\rceil$, attempt $i$ fails if $W'_{\lambda_i}>k$, and attempt $i+1$ is defined only after a failure of attempt $i$.
Failure requires $W'_{\lambda_i}-W'_{\varsigma_i}>k-\ell_0=x_0$, so $\Pr(\mathrm{fail}_i\mid\mathcal F_{\varsigma_i})\le1/2$ (Lemma~\ref{lem:fs-ruin}(a)).
Let $\mathcal C_1$ be the event that some excursion $[\lambda_{i-1},\varsigma_i)$ ($i\ge1$, after failures of attempts $1,\dots,i-1$) contains a time with
$W'>D$, and $\mathcal C_2$ the event that some window $[\varsigma_i,\lambda_i]$ (after failures of attempts $1,\dots,i-1$) has
$\max_{\varsigma_i<j\le\lambda_i}|W'_j-W'_{\varsigma_i}|\ge M'$ or $W'_{\lambda_i}>D$.
Off $\mathcal C_1\cup\mathcal C_2$, every excursion stays in $(\ell_0,D]$ until it ends at a time with $W'\in(\ell_0-k,\ell_0]$, and every window stays within
$M'$ of its start; all these levels are corridor levels of the arm (by \eqref{eq:fs-corr}), so by Lemma~\ref{lem:fs-coupling}(c) the step chain
follows $W'$ up to the first successful attempt. Before that attempt no low round has $d>D$, and at the low round $\lambda_i$ of the success
$W'_{\lambda_i}\le k$. Every excursion exits the finite interval $(\ell_0,D]$ a.s.\ (Lemma~\ref{lem:fs-ruin}(b)); an exit at the top is in $\mathcal C_1$ and an exit at the bottom ends the excursion. Attempts succeed with conditional probability at least $1/2$, so a.s.\ one
of $\mathcal C_1$, $\mathcal C_2$ or a success occurs; hence $\{\nu<\infty,d(Y_\nu)>D\}\subseteq\mathcal C_1\cup\mathcal C_2$ up to a null set.
For $\mathcal C_1$: the excursion from $W'_0=d(z)$ reaches $W'\ge D+1$ before $W'\le\ell_0$ with probability at most
$(d(z)-\ell_0+k)/(D+1-\ell_0+k)\le(d(z)-k+13k)/D$ (Lemma~\ref{lem:fs-ruin}(b)); after a failure of attempt $i$ not in $\mathcal C_2$, the excursion starts at
$s_i=W'_{\lambda_i}\le\ell_0+R_i$ with $R_i:=W'_{\lambda_i}-W'_{\varsigma_i}$, and reaches the top with conditional probability at most $(R_i^++k)/D$. Since
$\mathbb E[(R_i^++k)\mathbf 1_{\mathrm{fail}_i}\mid\mathcal F_{\varsigma_i}]\le\frac12\sqrt{m\sigma^2}+\frac k2$ and attempts $1,\dots,i-1$ fail with probability at
most $2^{1-i}$, $\Pr(\mathcal C_1)\le(d(z)-k+13k+\sqrt{m\sigma^2}+k)/D\le(d(z)-k+3.34\sqrt V)/D$ by Lemma~\ref{lem:un-arith}(e). For $\mathcal C_2$: each
window has at most $m$ steps, so Lemma~\ref{lem:fs-F} bounds the two events by $\eta_F$ and by
$e^{-D^2/(4m\sigma^2)}+e^{-3D/(4k)}\le\varepsilon_B$ (as $D^2/(4m\sigma^2)\ge7.5D/\sqrt V\ge D/(2k)$ and $W'_{\lambda_i}>D$ needs an increase $>D-\ell_0$);
summing $2^{1-i}$ gives $\Pr(\mathcal C_2)\le2(\eta_F+\varepsilon_B)$.
\end{proof}

\begin{lemma}[Visits to a level window]\label{lem:fs-CF9}
Let $30\sqrt V\le D'\le L/4$ and $W_{D'}:=\{w\in V(G):|d(w)-D'|\le k/2\}$. For every $z\in\mathsf I$,
$\mathbb E_z\#\{n<\tau_{\mathsf E}:Y_n\in W_{D'}\}\le C_GD'/\sqrt V$ with $C_G:=64$.
\end{lemma}

\begin{proof}
It suffices to show that from every $w\in W_{D'}\cap\mathsf I$ the free walk reaches $\mathsf E$ before its next low round in $W_{D'}$ with probability at least
$p_{\mathrm{esc}}:=\sqrt V/(64D')$: the successive low rounds $n<\tau_{\mathsf E}$ with $Y_n\in W_{D'}$ are stopping times, and by the strong Markov property
(Lemma~\ref{lem:fs-coupling}(a)) their number is dominated by a geometric variable with mean $1/p_{\mathrm{esc}}$. The vertex $w$ is a corridor vertex of an
arm with $d(w)\in[D'-k/2,D'+k/2]$. Consider the following three events, each with the stated conditional probability.
(i) $S_{m+1}:=\xi_1+\dots+\xi_{m+1}>\sqrt V/2$ and $\max_{j\le m+1}|S_j|<M'$. By the Paley--Zygmund inequality,
$\Pr(S_{m+1}^2>V/4)\ge\frac{9/16}{3+k^2/V}\ge0.1857$, and by symmetry $\Pr(S_{m+1}>\sqrt V/2)\ge0.0928+\eta_F$; so (i) has probability at least $0.0928$. On (i),
$d(Y_1)=d(w)-S_{m+1}<D'+k/2-\sqrt V/2\le(D'-k/2)-\sqrt V/3$ (as $\sqrt V\ge6k$), so $Y_1\notin W_{D'}$; if $d(Y_1)\le k$ the walk is killed.
(ii) If $d(Y_1)>k$: the coupled step walk from $d(Y_1)$ (increments $-\xi_{m+1+i}$) exits $(\ell_0,b)$, $b:=D'-k/2$, at the bottom (if $d(Y_1)\le\ell_0$
this holds with exit step $m+1$). By
Lemma~\ref{lem:fs-ruin}(b) this has conditional probability at least $(b-d(Y_1))/(b-\ell_0+k)\ge(\sqrt V/3)/(D'+12k)\ge\sqrt V/(3.2D')$, since
$12k\le2\sqrt V\le D'/15$. On this event all steps until the exit are at levels in $(\ell_0-k,b)$, corridor levels of the arm, so the chain follows the
walk; no low round in this stretch lies in $W_{D'}$, and a low round with $d\le k$ kills the walk.
(iii) At the exit step $\varsigma$, Lemma~\ref{lem:fs-ruin}(a) gives conditional probability at least $0.548$ that the next low round $\lambda$ has
$|W'_\lambda-W'_\varsigma|<x_0$ with all intermediate steps within $M'$ of $W'_\varsigma$; then the chain follows the walk and $d(Y)$ at round
$\lambda/(m+1)$ is less than $\ell_0+x_0=k$, while $\lambda$ is the only low round in $[\varsigma,\lambda]$.
On (i)$\cap$(ii)$\cap$(iii) the walk is killed before its next low round in $W_{D'}$, so $p_{\mathrm{esc}}\ge0.0928\cdot0.548\,\sqrt V/(3.2D')\ge\sqrt V/(64D')$.
\end{proof}

\begin{lemma}[Last exit decomposition]\label{lem:fs-FP0}
Let $D\le L/32$. For $z\in\mathsf I$ and $u\ge1$, $f^z_u=A_D(z,u)+B_D(z,u)$, where
\begin{align*}
A_D(z,u)&:=\Pr_z(Y_0,\dots,Y_{u-1}\in\mathcal S_D,\ Y_u\in\mathsf E),\\
B_D(z,u)&:=\sum_{s<u}\sum_{w\in F_D}\Pr_z(\tau_{\mathsf E}>s,Y_s=w)\,e^D_{u-s}(w),
\end{align*}
and $e^D_j(w):=\Pr_w(Y_1,\dots,Y_{j-1}\in\mathcal S_D,\ Y_j\in\mathsf E)$. In particular $A_D(z,u)=0$ for $z\in F_D$.
\end{lemma}

\begin{proof}
On $\{\tau_{\mathsf E}=u\}$ all of $Y_0,\dots,Y_{u-1}$ lie in $\mathsf I=\mathcal S_D\cup F_D$. If none lies in $F_D$ the path is counted in $A_D$; otherwise let $s<u$ be the
last index with $Y_s\in F_D$, and apply the Markov property at the low round $s$.
\end{proof}

\begin{lemma}[Long jumps]\label{lem:fs-FPL}
For every $z\in\mathsf I$, $\mathcal N_z:=\sum_{s\ge0}\Pr_z\big(\tau_{\mathsf E}>s,\ d(Y_s)>L/4,\ d(Y_{s+1})\le L/32\big)\le\varepsilon_L:=e^{-q/13}$.
\end{lemma}

\begin{proof}
Let $H:=\{x\in\mathsf I:d(x)>L/4\}$ and $p_1:=4e^{-0.082q}$.
\emph{Step 1: for every $y\in H$, $\Pr_y(d(Y_1)\le L/32)\le2.001\,p_1\Pr_y(Y_1\notin H)$.} Run the step chain from $y$ and let
$\vartheta:=\min\{j\ge0:d(X_j)\le L/4\}$. Both events require $\vartheta\le m+1$. On this event $d(X_\vartheta)\in(L/4-k,L/4]$, since the previous vertex has
$d>L/4$ and one step changes the level by at most $k$; so $X_\vartheta$ is a corridor vertex of an arm. Let $T_l:=\xi_{\vartheta+1}+\dots+\xi_{\vartheta+l}$,
independent of $\mathcal G_\vartheta$ (Lemma~\ref{lem:fs-coupling}(a)).
(1a) Suppose $d(X_{m+1})\le L/32$. Let $i^\ast$ be the last $j\in[\vartheta,m+1]$ with $d(X_j)>L/4$, or $i^\ast:=\vartheta$ if there is none, and let $j^\ast$ be the first
$j>i^\ast$ with $d(X_j)\le L/32$. Then $d(X_{i^\ast})\in(L/4-k,L/4+k]$ and $d(X_j)\in(L/32,L/4]$ for $i^\ast<j<j^\ast$; these are corridor vertices, so by
Lemma~\ref{lem:fs-coupling}(b) $\xi_{i^\ast+1}+\dots+\xi_{j^\ast}=d(X_{i^\ast})-d(X_{j^\ast})>7L/32-k$. Hence the partial sums $(T_l)$ have range larger than
$7L/32-k$, and $\max_l|T_l|>y_1:=7L/64-k/2$. By Lemma~\ref{lem:fs-F} with at most $m+1$ steps, the conditional probability is at most
$2(e^{-y_1^2/(4V)}+e^{-3y_1/(4k)})\le4e^{-3y_1/(4k)}\le p_1$.
(1b) Let $r':=m+1-\vartheta$. If $T_{r'}\ge0$ and $\max_{l\le r'}|T_l|<L/4-2k$, the chain follows the walk on the arm (all levels within $L/4-2k$ of
$d(X_\vartheta)$ are corridor levels) and $d(X_{m+1})=d(X_\vartheta)-T_{r'}\le L/4$, so $Y_1\notin H$. By symmetry $\Pr(T_{r'}\ge0)\ge1/2$, and
$\Pr(\max_l|T_l|\ge L/4-2k)\le p_1$; so $\Pr(Y_1\notin H\mid\mathcal G_\vartheta)\ge1/2-p_1$.
Taking expectations over $\{\vartheta\le m+1\}$ gives Step~1, since $p_1/(1/2-p_1)\le2.001p_1$.
\emph{Step 2: $\mathbb E_z\#\{s<\tau_{\mathsf E}:Y_s\in H,Y_{s+1}\notin H\}\le9.86\,L/\sqrt V$.} After such an $s$, either $Y_{s+1}\in\mathsf E$ or $Y_{s+1}\in\mathsf I$ with
$k+1\le d(Y_{s+1})\le L/4$. From such a state $y'$ the walk is killed before its next low round in $H$ with probability at least $p_e:=0.1015\sqrt V/L$: as in
the proof of Lemma~\ref{lem:fs-CF9}, with probability at least $0.0928$ the first cycle lowers $d$ by more than $\sqrt V/2$ while staying within $M'$; then
the walk exits $(\ell_0,b)$, $b:=\lfloor L/4\rfloor+1$, at the bottom with conditional probability at least $(\sqrt V/2)/(b+x_0)\ge\sqrt V/(0.5004L)$, without a low round in
$H$ (every integer level below $b$ is $\le L/4$); and by Lemma~\ref{lem:fs-ruin}(a) it is killed at the next low round with conditional probability at least $0.548$; and $0.0928\cdot0.548/0.5004\ge0.1015$. By the strong
Markov property the number of exits from $H$ is dominated by a geometric variable with mean $1/p_e\le9.86L/\sqrt V$.
\emph{Conclusion.} By the Markov property at $s$ and Step~1,
$\mathcal N_z\le\sum_s\mathbb E_z[\mathbf 1\{\tau_{\mathsf E}>s,Y_s\in H\}\,2.001p_1\Pr_{Y_s}(Y_1\notin H)]$, which is $2.001p_1$ times the expectation in Step~2.
With $L/\sqrt V\le q/6$, $\mathcal N_z\le2.001\cdot4\cdot9.86\,(q/6)e^{-0.082q}\le e^{-q/13}$.
\end{proof}

\begin{lemma}[Strip part]\label{lem:fs-FPA}
Let $z\in\mathcal S_D$ and $u\ge4$. Then $A_D(z,u)\le C_Ak\big(1+(d(z)-k)/\sqrt V\big)u^{-3/2}+\varepsilon_c$ with $C_A:=8\cdot10^5$.
\end{lemma}

\begin{proof}
$A_D(z,u)=\Pr_z(\nu=u,d(Y_u)\le k)$. By Lemma~\ref{lem:fs-corr}, $A_D(z,u)\le\Pr(\nu^Z=u,Z_u\le k)+\varepsilon_c$, and with $\mathcal S:=\{k+1,\dots,D\}$,
$\Pr(\nu^Z=u,Z_u\le k)=\sum_{w\in\mathcal S}K_{\mathcal S}^{u-1}(d(z),w)\Pr(\zeta\ge w-k)$. Let $t:=u-1\ge3$, split $t=t_1+t_2+t_3$ with $t_i\ge t/5$
(as in Lemma~\ref{lem:fs-CF11}), and apply Lemma~\ref{lem:fs-kernel}(b) and Lemma~\ref{lem:fs-CF5} in the coordinate $Z-k-1$:
\[
K_{\mathcal S}^{t}(d(z),w)\le C_S^2C_E\,5^{3/2}\,\frac k{\sqrt V}\Big(1+\frac{d(z)-k-1}{\sqrt V}\Big)\Big(1+\frac{w-k-1}{\sqrt V}\Big)t^{-3/2}.
\]
Further $\sum_{w\ge k+1}(1+\frac{w-k-1}{\sqrt V})\Pr(\zeta\ge w-k)=\sum_{i\ge1}(1+\frac{i-1}{\sqrt V})\Pr(\zeta\ge i)=\mathbb E\zeta^++\frac{\mathbb E[\zeta^+(\zeta^+-1)]}{2\sqrt V}\le\frac{\sqrt V}2+\frac{\sqrt V}4$,
and $t^{-3/2}\le(4/3)^{3/2}u^{-3/2}$. Finally $C_S^2C_E5^{3/2}\cdot0.75\cdot(4/3)^{3/2}\le7.85\cdot10^5$.
\end{proof}

\begin{lemma}[Strip crossing from above]\label{lem:fs-FPB}
Let $D\le L/32$. For an integer $\ell'$ with $D<\ell'\le L/4$ let $\bar e(\ell'):=\sup_{j\ge1}\max\{e^D_j(w):d(w)=\ell'\}$. Then
\[
\Sigma_D:=\sum_{\ell'=D+1}^{\lfloor L/4\rfloor}\frac{\ell'}D\,\bar e(\ell')\le C_B\frac{kV^2}{D^3},\qquad C_B:=6\cdot10^{12}.
\]
\end{lemma}

\begin{proof}
Let $\mathcal S:=\{k+1,\dots,D\}$ and, for an integer $\ell'>D$, $e^Z_j(\ell'):=\Pr(\nu^Z=j,Z_j\le k\mid Z_0=\ell')$. For $w$ at level $\ell'\le L/4$,
$e^D_j(w)=\Pr_w(\nu=j,d(Y_j)\le k)\le e^Z_j(\ell')+\varepsilon_c$ (Lemma~\ref{lem:fs-corr}), so $\bar e(\ell')\le e^Z_1(\ell')+e^Z_2(\ell')+\sup_{j\ge3}e^Z_j(\ell')+\varepsilon_c$.
\emph{The error.} $\sum_{\ell'\le L/4}\frac{\ell'}D\varepsilon_c\le\frac{L^2}{31D}\varepsilon_c\le\frac{kV^2}{D^3}\cdot\frac{L^4\varepsilon_c}{31\cdot1024\cdot1296\,k^5}\le\frac{kV^2}{D^3}$, using
$D\le L/32$, $V\ge36k^2$, $k\ge1$ and $q^4e^{-q/7}\le1$.
\emph{Lags $j\le2$.} $e^Z_1(\ell')=\Pr(\zeta\ge\ell'-k)$ and $e^Z_2(\ell')\le\Pr(Z_2-Z_0\le-(\ell'-k))$, so by Lemma~\ref{lem:fs-F}
$e^Z_1+e^Z_2\le2(e^{-(\ell'-k)^2/(8V)}+e^{-3(\ell'-k)/(4k)})$ with $\ell'-k\ge0.994\ell'$; summed with the weights $\ell'/D$ over $\ell'>D\ge30\sqrt V$ this is at
most $10^{-40}kV^2/D^3$. Indeed, put $c:=0.994^2/8\ge0.1235$ and $a:=0.7455/k$; the functions $\ell e^{-c\ell^2/V}$ and $\ell e^{-a\ell}$ decrease on $\ell\ge D$
(as $cD^2/V\ge111$ and $aD\ge134$), so comparison with integrals gives $\sum_{\ell'>D}\frac{\ell'}De^{-c\ell'^2/V}\le\frac V{2cD}e^{-cD^2/V}$ and
$\sum_{\ell'>D}\frac{\ell'}De^{-a\ell'}\le\frac2ae^{-aD}$. With $z:=D/\sqrt V\ge30$ and $v:=D/k\ge180$ the two contributions, divided by $kV^2/D^3$, are at most
$\frac{z^2}{c}e^{-cz^2}\le10^{-44}$ and $\frac{4\cdot 2}{a}\cdot\frac{D^3}{kV^2}e^{-aD}\le10.8\,\frac{v^3}{1296}e^{-0.7455v}\le10^{-50}$ (using $V\ge36k^2$).
\emph{Lags $j\ge3$.} $e^Z_j(\ell')=\sum_{a',b'\in\mathcal S}K(\ell',a')K_{\mathcal S}^{j-2}(a',b')\Pr(\zeta\ge b'-k)$ with $K(\ell',a'):=\Pr(\zeta=\ell'-a')$. Use the notation
$h_{a'}=D-a'$, $h_{b'}=b'-k-1$ of Lemma~\ref{lem:fs-CF11}.
\emph{Pairs with $h_{a'}+h_{b'}\le D/2$.} By Lemma~\ref{lem:fs-CF11} they contribute at most
$C_\star\frac k{\sqrt V}\,\Theta_D\,g(\ell')\sum_{b'}(1+\frac{h_{b'}}{\sqrt V})\Pr(\zeta\ge b'-k)$, where $\Theta_D:=\sup_{t\ge1}t^{-3/2}\bar\Xi_t(D/14)$ and
$g(\ell'):=\sum_{a'}K(\ell',a')(1+h_{a'}/\sqrt V)$. As in Lemma~\ref{lem:fs-FPA} the sum over $b'$ is at most $0.75\sqrt V$. Next,
$\sup_{s>0}s^{-3/2}e^{-y^2/(16sV)}=24^{3/2}e^{-3/2}V^{3/2}y^{-3}\le26.24V^{3/2}y^{-3}$ and, with $v:=D/k$,
$e^{-3D/(112k)}\le112^3e^{-3}v^{-3}\le324.6\,V^{3/2}D^{-3}$ (as $\sup_vv^3e^{-3v/112}=112^3e^{-3}$ and $V^{3/2}\ge216k^3$), so
$\Theta_D\le(26.24\cdot14^3+324.6)V^{3/2}D^{-3}\le7.24\cdot10^4V^{3/2}D^{-3}$. Finally, writing a jump of size $i$ of the first cycle as
$\ell'-a'=(\ell'-D)+h_{a'}$ with $\ell'-D\ge1$, $h_{a'}\ge0$,
\[
\sum_{\ell'>D}\frac{\ell'}Dg(\ell')\le\sum_{i\ge1}\Pr(\zeta=i)\,i\Big(1+\frac iD\Big)\Big(1+\frac i{\sqrt V}\Big)=\mathbb E\Big[\zeta^+\Big(1+\frac{\zeta^+}D\Big)\Big(1+\frac{\zeta^+}{\sqrt V}\Big)\Big]\le1.046\sqrt V,
\]
by $\mathbb E\zeta^+\le\sqrt V/2$, $\mathbb E(\zeta^+)^2=V/2$, $\mathbb E(\zeta^+)^3\le(\mathbb E(\zeta^+)^2\mathbb E(\zeta^+)^4)^{1/2}\le0.871V^{3/2}$ and $D\ge30\sqrt V$.
These pairs contribute at most $C_\star\cdot7.24\cdot10^4\cdot1.046\cdot0.75\,kV^2D^{-3}\le5.69\cdot10^{12}kV^2D^{-3}$ to $\Sigma_D$.
\emph{Pairs with $h_{a'}+h_{b'}>D/2$.} Then $h_{a'}>D/4$ or $h_{b'}>D/4$. In the first case bound $\sum_{b'}K^{j-2}_{\mathcal S}(a',b')\Pr(\zeta\ge b'-k)\le1$; the first
jump exceeds $D/4$, and the weighted sum is at most $\mathbb E[\zeta^+(1+\zeta^+/D)\mathbf 1\{\zeta>D/4\}]$. In the second case bound $K^{j-2}_{\mathcal S}\le1$;
then $\sum_{b':h_{b'}>D/4}\Pr(\zeta\ge b'-k)\le\mathbb E[\zeta^+\mathbf 1\{\zeta>D/4\}]$ and $\sum_{\ell'}\frac{\ell'}D\sum_{a'}K(\ell',a')\le\mathbb E[\zeta^+(1+\zeta^+/D)]\le0.52\sqrt V$.
Together these pairs contribute at most $1.02\sqrt V\,\mathbb E[\zeta^+(1+\zeta^+/D)\mathbf 1\{\zeta>D/4\}]$. With $\Pr(\zeta>y)\le e^{-y^2/(4V)}+e^{-3y/(4k)}$
(Lemma~\ref{lem:fs-F}), $\mathbb E[\zeta^+\mathbf 1\{\zeta>y_0\}]=y_0\Pr(\zeta>y_0)+\int_{y_0}^\infty\Pr(\zeta>y)dy$ and
$\mathbb E[(\zeta^+)^2\mathbf 1\{\zeta>y_0\}]=y_0^2\Pr(\zeta>y_0)+\int_{y_0}^\infty2y\Pr(\zeta>y)dy$ at $y_0=D/4$, and the integrals
$\int_{y_0}^\infty e^{-y^2/(4V)}dy\le\frac{2V}{y_0}e^{-y_0^2/(4V)}$, $\int_{y_0}^\infty2ye^{-y^2/(4V)}dy=4Ve^{-y_0^2/(4V)}$, $\int_{y_0}^\infty e^{-3y/(4k)}dy=\frac{4k}3e^{-3y_0/(4k)}$,
$\int_{y_0}^\infty2ye^{-3y/(4k)}dy=2(\frac{4ky_0}3+\frac{16k^2}9)e^{-3y_0/(4k)}$, the ratio of this contribution to $kV^2D^{-3}$ is at most, with $z:=D/\sqrt V\ge30$ and
$r:=k/\sqrt V\in[1/\sqrt{66},1/6]$,
\[
1.02\frac{z^3}{r}\Big[\Big(\frac{5z}{16}+\frac{12}z\Big)e^{-z^2/64}+\Big(\frac{5z}{16}+2r+\frac{32r^2}{9z}\Big)e^{-3z/(16r)}\Big].
\]
Both terms decrease in $z\ge30$: $(\frac{5z^4}{16}+12z^2)e^{-z^2/64}$ has derivative $e^{-z^2/64}z(\frac54z^2+24-\frac{5z^4}{512}-\frac38z^2)<0$, and the second
term is a polynomial of degree $4$ in $z$ times $e^{-3z/(16r)}$ with $3/(16r)\ge9/8>4/z$. At $z=30$ the first term is at most
$1.02\cdot27000\cdot\sqrt{66}\cdot9.775\,e^{-14.0625}\le1.709$ (worst case $r=1/\sqrt{66}$), and the second is at most
$1.02\cdot27000\cdot6\cdot9.72\,e^{-33.75}\le10^{-8}$ (the factor $r^{-1}e^{-5.625/r}$ increases in $r\le1/6$). So these pairs contribute at most $1.71\,kV^2D^{-3}$, and we use the bound $5kV^2D^{-3}$.
Summing, $\Sigma_D\le(5.69\cdot10^{12}+5+1+1)kV^2D^{-3}\le C_BkV^2D^{-3}$.
\end{proof}

\begin{proposition}[First-passage bounds]\label{prop:fs-FP}
The free walk satisfies (FP1)--(FP3) of Proposition~\ref{prop:fs-REN} with $C_1:=1.42\cdot10^{21}k^2$ and $C_2:=2.25\cdot10^{20}k^2$.
\end{proposition}

\begin{proof}
\emph{Green factor.} Let $D\le L/32$, $z\in\mathsf I$ and $D<\ell'\le L/4$, and put $G_z(\ell'):=\sum_{s\ge0}\Pr_z(\tau_{\mathsf E}>s,d(Y_s)=\ell')$. Since the level
$\ell'$ lies in the window $W_{\ell'}$, Lemma~\ref{lem:fs-CF9} gives $G_z(\ell')\le C_G\ell'/\sqrt V$. If $z\in\mathcal S_D$, a visit to level $\ell'>D$ requires a first
low round $\nu$ with $d(Y_\nu)>D$; by Lemma~\ref{lem:fs-CF8} and the strong Markov property at $\nu$,
$G_z(\ell')\le[(d(z)-k+3.4\sqrt V)/D+2\varepsilon_B+2\eta_F]\,C_G\ell'/\sqrt V\le C_G\big(\frac{d(z)-k}{\sqrt V}+3.41\big)\frac{\ell'}D$, because
$(2\varepsilon_B+2\eta_F)D/\sqrt V\le0.01$ ($D/k\ge180$, $D/\sqrt V\le q$).
\emph{Term $B$.} In $B_D(z,u)$, the vertices $w\in F_D$ with $d(w)\le L/4$ contribute at most $\sum_{\ell'}G_z(\ell')\bar e(\ell')$, since $e^D_{u-s}(w)\le\bar e(d(w))$ and
$\sum_{w:d(w)=\ell'}\sum_s\Pr_z(\tau_{\mathsf E}>s,Y_s=w)=G_z(\ell')$. For $d(w)>L/4$, $e^D_{u-s}(w)\le\Pr_w(d(Y_1)\le D)\le\Pr_w(d(Y_1)\le L/32)$, so by the Markov
property these $w$ contribute at most $\mathcal N_z\le\varepsilon_L$ (Lemma~\ref{lem:fs-FPL}). With Lemma~\ref{lem:fs-FPB},
\begin{align*}
B_D(z,u)&\le C_G\frac D{\sqrt V}\Sigma_D+\varepsilon_L&&(z\in\mathsf I),\\
B_D(z,u)&\le C_G\Big(\frac{d(z)-k}{\sqrt V}+3.41\Big)\Sigma_D+\varepsilon_L&&(z\in\mathcal S_D).
\end{align*}
\emph{(FP2) and (FP3).} Take $D:=\lfloor L/32\rfloor$; then $L/33\le D\le L/32$ and $D\ge30\sqrt V$. For $z\in\mathsf I$ and every $u\ge1$,
$B_D(z,u)\le C_GC_BkV^{3/2}D^{-2}+\varepsilon_L\le64\cdot6\cdot10^{12}\cdot66^{3/2}\cdot1089\,k^4/L^2+\varepsilon_L\le2.2423\cdot10^{20}k^4/L^2+\varepsilon_L$. Let $z\in\mathcal S_D$ and
$u\ge J/4-1$; then $u\ge L^2/(4.01V)\ge4$, and by Lemma~\ref{lem:fs-FPA} with $1+(d(z)-k)/\sqrt V\le1+q/192\le q/191.6$,
$A_D(z,u)\le C_Ak\frac q{191.6}(4.01V/L^2)^{3/2}+\varepsilon_c\le1.8\cdot10^7k^3/L^2+\varepsilon_c$. Since $\varepsilon_L+\varepsilon_c\le q^{-2}$, $f^z_u=A_D+B_D\le2.25\cdot10^{20}k^4/L^2=C_2k^2/L^2$
for $u\ge J/4-1$, which is (FP2). For $z=o$, $d(o)=A+1>D$, so $A_D(o,u)=0$ for all $u\ge1$, which gives (FP3).
\emph{(FP1).} Let $z\in\mathsf I$ and $1\le u\le2J$. If $u\le u_0:=2.3\cdot10^6$, then $f^z_u\le1\le u_0^{3/2}u^{-3/2}\le C_1u^{-3/2}$. If $u>u_0$, take
$D:=\lfloor\sqrt{uV}/50\rfloor$. Then $D\ge(\sqrt{u_0}/50-1/8)\sqrt V\ge30\sqrt V$ (as $\sqrt V\ge8$), $D\ge\sqrt{uV}/51$, and $D\le L/32$ because
$uV\le2JV\le2L^2+2V\le2.001L^2$. Lemma~\ref{lem:fs-FPB} gives $\Sigma_D\le C_B\,51^3kV^2(uV)^{-3/2}\le C_B\,51^3\sqrt{66}\,k^2u^{-3/2}$. If $z\in\mathcal S_D$, Lemmas
\ref{lem:fs-FP0}, \ref{lem:fs-FPA} and the bound on $B_D$ give
$f^z_u\le\big(C_Ak+3.41\,C_GC_B51^3\sqrt{66}\,k^2\big)\big(1+\frac{d(z)}{\sqrt V}\big)u^{-3/2}+\varepsilon_c+\varepsilon_L$; if $d(z)>D$, then $A_D=0$ and
$B_D\le C_G\frac{d(z)}{\sqrt V}\Sigma_D+\varepsilon_L$, which is bounded by the same expression. Finally $\varepsilon_c+\varepsilon_L\le2e^{-q/13}\le76q^{-3}\le(2.001L^2/(36k^2))^{-3/2}\le u^{-3/2}$,
and $8\cdot10^5k+3.41\cdot64\cdot6\cdot10^{12}\cdot51^3\sqrt{66}\,k^2+1\le1.42\cdot10^{21}k^2$.
\end{proof}

\subsection{Proof of \texorpdfstring{Theorem~\ref{thm:fs-main}}{the fixed-start tail bound}}\label{sec:fs-proof}

\begin{proof}
By Proposition~\ref{prop:fs-FP}, (FP1)--(FP3) hold with $C_1=1.42\cdot10^{21}k^2\ge1$ and $C_2=2.25\cdot10^{20}k^2$. Proposition~\ref{prop:fs-REN} gives (H1) and
(H2) with
\[
C_\Phi\le28.45\,C_2+9k\cdot6.7C_1\cdot4.26\cdot10^5\cdot0.0429+810\cdot9kC_2\le1.57\cdot10^{27}k^3,
\]
\[
C_H\le C_\Phi+C_2(1+265.1k)\le1.571\cdot10^{27}k^3,
\]
using $q_0^{-1}\le28.45$, $q_0^{-2}\le810$ and $k^2\le k^3$. Proposition~\ref{prop:fs-RED} and Theorem~\ref{thm:fs-WH} give, for every $n\ge0$,
\[
D^{\mathrm{lo}}_n(\mathsf T)\le\frac{C_H}{c_{\mathrm{WH}}}\frac{k^2}{L^2}\le\frac{1.571\cdot10^{27}}{2.81\cdot10^{-5}}\frac{k^5}{L^2}\le\frac{6\cdot10^{31}k^5}{L^2}=\frac{C_k}{L^2}.
\]
This holds for every $L\ge10^5k$, every $p\in(0,1)$ and every $\mathcal C\in\mathcal P_0(k)$ with $\sigma^2>0$, which proves Theorem~\ref{thm:fs-main}.
\end{proof}

The proof uses the curriculum only through the corridor law $\kappa$ (Lemma~\ref{lem:un-corridor}) and the
structural lemmas of Section~\ref{sec:un-wash}: at views that contain $o$ and in the band, $\mathcal C$ is arbitrary, and no constant depends on
$\sigma^2$, on $m$, on laziness or on the lattice structure of $\kappa$.
\fussy

\section{One step of memory: proofs of \texorpdfstring{Theorem~\ref{thm:nb} and Corollary~\ref{cor:sep}}{Theorem 3 and Corollary 4}}\label{app:F_memory}\label{sec:memory}

\begin{restated}{Theorem~\ref{thm:nb}}\StmtNB\end{restated}

\paragraph{Proof of Theorem~\ref{thm:nb}.} Corollary~\ref{cor:M-tree-threshold} gives the two equalities and the bounds on
$1-2^{-1/L}$. If $\epsilon<1-2^{-1/L}$, Theorem~\ref{thm:M-tree}(c) gives the lower bound on $|K_t|$, and $\epsilon<1-2^{-1/L}$ is equivalent
to $2^{1/L}(1-\epsilon)>1$.

\begin{restated}{Corollary~\ref{cor:sep}}\StmtSeparation\end{restated}

\paragraph{Proof of Corollary~\ref{cor:sep}.} $\epsilon^*(\NB)$ is the common value of Corollary~\ref{cor:M-tree-threshold}, and
$\mathcal P_0^{=}(k)$ is the subclass of Corollary~\ref{cor:M-sep}(d). (a) By Corollary~\ref{cor:M-sep}(a),(d), for $\mathcal P_0(k)$ and for
$\mathcal P_0^{=}(k)$ the ratio is at most $64\ln2\cdot L$ for every $L\ge1$ and at least $L/(9\cdot10^{31}k^5)$ for every $L\ge10^5k$. (b) is
Corollary~\ref{cor:M-sep}(b) with $L_k=9\cdot10^{31}k^5$ and $C_k=6\cdot10^{31}k^5$ as in Theorem~\ref{thm:uniform}; the bound of (ii) grows
exponentially because $2^{1/L}(1-\epsilon)>1$. (c) is Corollary~\ref{cor:M-sep}(c).

\paragraph{Roadmap.} Throughout, $L\ge1$, $G=T_3^{(L)}$, $o$ is a branch vertex, the dynamics is that of
Definition~\ref{def:mem-dynamics}, and $\epsilon_L:=1-2^{-1/L}$. Under $\NB$ the frontier of round $t$ lives on the sphere $S_t$ and every
proposal from a known vertex is kept (Lemma~\ref{lem:mem-ST-radial}), so the largest point mass shrinks by the factor $b_t(1-\epsilon)$ per
round, which gives growth below $\epsilon_L$ (Theorem~\ref{thm:M-tree}). Above $\epsilon_L$ a prover that keeps the frontier uniform on a
bounded number of vertices and forgets an $\epsilon$ fraction of them per round stalls $\NB$ (Theorem~\ref{thm:M-tree-stall}). Together they
give the exact tolerance (Corollary~\ref{cor:M-tree-threshold}), which Corollary~\ref{cor:M-sep} compares with $\mathcal P_0(k)$ using
Theorem~\ref{thm:fs-U} and Corollary~\ref{cor:st-tree}.

\subsection{Geometry and profiles}\label{sec:mem-conventions}\label{sec:mem-tree}

\begin{definition}[Normal-form profile]\label{def:mem-normal-form}
For $K\subseteq V$ let $s^{K}=p$ on $K$, $s^K=p/2$ on $N(K)\setminus K$, and $s^K=0$ elsewhere.
\end{definition}

\begin{lemma}\label{lem:mem-normal-form}
For every $K\subseteq V$, $s^K$ satisfies Definition~\ref{def:admissible}(i), $\{s^K\ge p\}=K$ and $\{s^K\ge p/2\}=B(K,1)$.
\end{lemma}

\begin{proof}
The set identities hold because $0<p/2<p$. For an edge $\{x,y\}$ with $x\notin B(K,1)$, $y\notin K$ (else $x\in N(K)$), so both values lie in
$\{0,p/2\}$; if both endpoints lie in $B(K,1)$ both values lie in $\{p/2,p\}$.
\end{proof}

For $r\ge0$ let $b_r:=3$ if $r=0$, $b_r:=2$ if $r\ge1$ and $L\mid r$, and $b_r:=1$ otherwise.

\begin{lemma}[Spheres of $T_3^{(L)}$]\label{lem:mem-tree-geometry}
(a) Every edge joins $S_r$ and $S_{r+1}$ for some $r$. (b) Every $x\in S_r$, $r\ge1$, has exactly one neighbour in $S_{r-1}$ (its parent) and $b_r$
neighbours in $S_{r+1}$ (its children); $o$ has $b_0=3$ children; distinct vertices of $S_r$ have disjoint sets of children. (c) For $t\ge1$,
$\prod_{r=0}^{t-1}b_r=3\cdot2^{\lfloor (t-1)/L\rfloor}$.
\end{lemma}

\begin{proof}
(a) If $y$ lies on the geodesic from $o$ to $x$, it precedes $x$; otherwise that geodesic followed by $\{x,y\}$ is the geodesic to $y$. (b) Two
neighbours of $x$ in $S_{r-1}$ would give two paths from $o$ to $x$; one exists, so by (a) the other $\deg(x)-1$ neighbours lie in $S_{r+1}$, and
$\deg(x)-1=b_r$ because $S_r$ ($r\ge1$) consists of branch vertices iff $L\mid r$. A common child would have two parents. (c) $b_0=3$, and exactly
$\lfloor(t-1)/L\rfloor$ of $1,\dots,t-1$ are multiples of $L$.
\end{proof}

\begin{lemma}[$\NB$ moves outward]\label{lem:mem-ST-radial}
Run $\NB$ from $\Pi_0=\delta_{(\bot,o)}$ with any profiles, and let $t\ge1$ with $Z_0,\dots,Z_{t-1}>0$. Then $\Pi_t$ is supported on pairs
$(x_-,x)$ with $x\in S_t$ and $x_-$ the parent of $x$. In particular the frontier $F_t$ is supported on $S_t$ and determines $\Pi_t$, the
proposal from a sample at $x\in S_t$ is uniform on the $b_t$ children of $x$, and without replay $D_t=F_t$.
\end{lemma}

\begin{proof}
From $(\bot,o)$ the proposal is uniform on $N(o)=S_1$. If $\Pi_t$ is supported on pairs (parent of $x$, $x$) with $x\in S_t$, the proposal from
$(x_-,x)$ is uniform on $N(x)\setminus\{x_-\}$, the set of children of $x$ (Lemma~\ref{lem:mem-tree-geometry}(b)), and each kept pair is $(x,x')$
with $x$ the parent of $x'\in S_{t+1}$. Since parents are unique, $\Pi_t$ is the image of $F_t$ under $x\mapsto(\text{parent of }x,x)$.
\end{proof}

\subsection{Growth, stall and exact tolerance}

\begin{theorem}[Growth of $\NB$ on $T_3^{(L)}$]\label{thm:M-tree}
Let $\epsilon\in[0,1)$. For every admissible prover sequence for $\NB$ from $\delta_o$ without replay:
\begin{enumerate}
\item[(a)] $Z_t\ge 1-\epsilon$ for every $t\ge0$, so the dynamics is defined for all $t$;
\item[(b)] $\max_x D_t(x)\le (1-\epsilon)^{-t}\big/\prod_{r=0}^{t-1}b_r$ for every $t\ge0$;
\item[(c)] $|K_t|\ge3\cdot2^{\lfloor (t-1)/L\rfloor}(1-\epsilon)^{t+1}\ge\tfrac32(1-\epsilon)\big(2^{1/L}(1-\epsilon)\big)^{t}$ for every $t\ge1$.
\end{enumerate}
In particular, if $\epsilon<\epsilon_L$, then $|K_t|\to\infty$ exponentially fast, and $\epsilon^*_{\mathrm{div}}(\NB;T_3^{(L)},o)\ge\epsilon_L$.
\end{theorem}

\begin{proof}
By induction on $t$ assume $Z_0,\dots,Z_{t-1}>0$, so $D_t$ is supported on $S_t$ (Lemma~\ref{lem:mem-ST-radial}; $S_0=\{o\}$).
(a) A child $y$ of $x\in K_t$ has $s_t(y)\ge s_t(x)-p/2\ge p/2$, so every proposal from $K_t$ is kept and $Z_t\ge D_t(K_t)\ge1-\epsilon$.
(b) Every $y\in S_{t+1}$ has a unique parent $x$, and by \eqref{eq:mem-update} $D_{t+1}(y)=D_t(x)b_t^{-1}\mathbf 1[s_t(y)\ge p/2]/Z_t$, so
$M_{t+1}\le M_t/(b_t(1-\epsilon))$ for $M_t:=\max_xD_t(x)$, with $M_0=1$.
(c) $1-\epsilon\le D_t(K_t)\le|K_t|M_t$, and (b) with Lemma~\ref{lem:mem-tree-geometry}(c) gives the first bound; the second uses
$\lfloor (t-1)/L\rfloor\ge t/L-1$. If $\epsilon<\epsilon_L$, then $2^{1/L}(1-\epsilon)>1$, and the growth consequence of
Section~\ref{sec:B-dynamics} applies.
\end{proof}

The count in (c) is an $\ell^\infty$ analogue of \citet[Lemma~E.1]{chenli2026}; the maximal point mass is the natural potential because the
frontier lives on one sphere.

\begin{theorem}[Stall of $\NB$ above $\epsilon_L$]\label{thm:M-tree-stall}
Let $\epsilon\in(\epsilon_L,1)$, $N:=\max\{3,\lceil 1/(\epsilon-\epsilon_L)\rceil\}$ and $\rho:=1-\epsilon+1/N$. There is an admissible prover
sequence for $\NB$ from $\delta_o$ without replay under which, for every $t\ge0$, $Z_t>0$, $D_t$ is uniform on a set of $n_t$ vertices of $S_t$,
$K_t\subseteq\operatorname{supp}D_t$, and $|K_t|\le n_t\le\max\{2N,N/\rho\}\le N\max\{2,1/(1-\epsilon)\}$.
\end{theorem}

\begin{proof}
\emph{Step 1 (one round).} Let $f(n):=n-\min\{\lfloor\epsilon n\rfloor,(n-N)_+\}$ for integers $n\ge1$. Suppose $D_t$ is uniform on
$X_t\subseteq S_t$, $|X_t|=n_t$. Let $j_t:=n_t-f(n_t)$, choose $K_t\subseteq X_t$ with $|K_t|=f(n_t)$, and put $s_t:=s^{K_t}$. By
Lemma~\ref{lem:mem-normal-form} it satisfies (i) and $D_t(K_t)=1-j_t/n_t\ge1-\epsilon$, so it is admissible. A child $y$ of $x\in X_t$ has
neighbours $x$ and its own children in $S_{t+2}$, so $y\in B(K_t,1)$ iff $x\in K_t$, and the proposal $y$ is kept iff $x\in K_t$. Hence
$Z_t=f(n_t)/n_t>0$ (as $j_t\le\lfloor\epsilon n_t\rfloor<n_t$), and $D_{t+1}$ is uniform on the $n_{t+1}=b_tf(n_t)$ children of the vertices of
$K_t$, which are distinct (Lemma~\ref{lem:mem-tree-geometry}(b)). From $n_0=1$ this defines the sequence, and $|K_t|\le n_t$.

\emph{Step 2 (properties of $f$).} (i) $f(n)\le\max\{N,\rho n\}$: if $n\le N$ then $f(n)=n$; if $n>N$ then
$f(n)=\max\{N,n-\lfloor\epsilon n\rfloor\}$ and $n-\lfloor\epsilon n\rfloor<n(1-\epsilon)+1\le\rho n$. (ii) By induction,
$f^{i}(a)\le\max\{N,\rho^i a\}$, as $\rho\le1$. Moreover $\rho\le1-\epsilon+(\epsilon-\epsilon_L)=2^{-1/L}$, so $2\rho^L\le1$.

\emph{Step 3 (bounded frontier).} $n_1=3f(1)=3$, and rounds $1,\dots,L-1$ have $b=1$ and $f(3)=3$, so $n_t=3$ for $1\le t\le L$. Put
$m_j:=n_{jL}$ ($j\ge1$), so $m_1=3\le N$. Round $jL$ has $b=2$ and rounds $jL+1,\dots,jL+L-1$ have $b=1$, so $n_{jL+1}=2f(m_j)$,
$n_{jL+1+i}=f^{i}(n_{jL+1})$ for $0\le i\le L-1$, and $m_{j+1}=f^{L-1}(n_{jL+1})$. If $m_j\le N/\rho$, then $n_{jL+1}\le2\max\{N,\rho m_j\}\le2N$,
$n_{jL+1+i}\le\max\{N,2\rho^iN\}\le2N$, and $m_{j+1}\le\max\{N,2\rho^L\cdot N/\rho\}\le N/\rho$. So by induction every $n_t\le\max\{2N,N/\rho\}$, and
$\rho\ge1-\epsilon$.
\end{proof}

\begin{corollary}[Exact tolerance of $\NB$]\label{cor:M-tree-threshold}
For every $L\ge1$,
\[
\epsilon^*_{\mathrm{div}}(\NB;T_3^{(L)},o)=\epsilon^*_{\mathrm{bdd}}(\NB;T_3^{(L)},o)=1-2^{-1/L},\qquad\frac{\ln2}{L}-\frac{(\ln2)^2}{2L^2}\le1-2^{-1/L}\le\frac{\ln 2}{L}.
\]
We write $\epsilon^*(\NB)$ for this value.
\end{corollary}

\begin{proof}
Theorem~\ref{thm:M-tree} gives $\epsilon^*_{\mathrm{div}}\ge\epsilon_L$, and Theorem~\ref{thm:M-tree-stall} with the stall consequence of
Section~\ref{sec:B-dynamics} gives $\epsilon^*_{\mathrm{bdd}}\le\epsilon$ for every $\epsilon\in(\epsilon_L,1)$; with
$\epsilon^*_{\mathrm{div}}\le\epsilon^*_{\mathrm{bdd}}$ both equal $\epsilon_L$. With $u=\ln2/L$, $\epsilon_L=1-e^{-u}\in[u-u^2/2,u]$.
\end{proof}

\begin{remark}[Direction read from a certificate]\label{rem:nb-outward}
On $T_3^{(L)}$ with $S_0=\{o\}$, the set $N^+(x)$ of Section~\ref{sec:pl-novelty} is the set of children of $x$. Let $O(x,\cdot)$ be uniform on
$N^+(x)$, the outward curriculum of Proposition~\ref{prop:novelty-replay}(c), run by \eqref{eq:update} from $\delta_o$, and let $\NB$ be run
from $\delta_{(\bot,o)}$, both with the same replay weight $w$. For every prover sequence the two have the same frontiers $F_t$, hence the same
$D_t$ and $K_t$, in every round in which one of them is defined: by induction, $\Pi_t$ is supported on pairs (parent of $x$, $x$) with
$\Pi_t((x_-,x))=F_t(x)$, because $\mathcal M((x_-,x),\cdot)$ is uniform on $N(x)\setminus\{x_-\}=N^+(x)$ (on $N(o)=N^+(o)$ for $x_-=\bot$), so
\eqref{eq:mem-update} gives $\Pi_{t+1}((x,x'))=Z_t^{-1}F_t(x)\,O(x,x')\,\mathbf 1[s_t(x')\ge p/2]$, whose normaliser and second marginal are those
of \eqref{eq:update} with kernel $O$. So the two curricula have the same admissible prover sequences and tolerances; without replay both
tolerances of the outward curriculum equal $1-2^{-1/L}$. The outward curriculum reads the direction from the certificate
$\mathbf 1[\mathrm{dist}(y,S_0)>\mathrm{dist}(x,S_0)]$, while $\NB$ obtains it by storing one parent per sample at the per-proposal cost of
$\LRW$ (Lemma~\ref{lem:mem-budget}).
\end{remark}

\begin{corollary}[$\NB$ with replay on $T_3^{(L)}$]\label{cor:nb-tree-replay}
Let $w\in(0,1)$ and $\epsilon<\min\{w,1-w\}$. Every admissible prover sequence for $\NB$ with replay weight $w$ from $\delta_{(\bot,o)}$
satisfies $|K_t|\ge(1-\epsilon/w)\,t$ for every $t\ge1$; for $w=1/2$, $\epsilon^*_{\mathrm{div}}\ge1/2$. For $\LRW$ with any replay weight,
$\epsilon^*_{\mathrm{bdd}}\le\frac32(1-2^{-1/(2L)})^2<3(\ln2)^2/(8L^2)$.
\end{corollary}

\begin{proof}
With $S_0=\{o\}$ every vertex of $T_3^{(L)}$ has a child, so there are no dead ends and $O$ is an outward kernel. By
Remark~\ref{rem:nb-outward} the bound of Proposition~\ref{prop:novelty-replay}(b) for $O$ transfers to $\NB$; for $w=1/2$ it gives
$|K_t|\to\infty$ at every $\epsilon<1/2$. The statement for $\LRW$ is Corollary~\ref{cor:st-tree}(b).
\end{proof}

\subsection{Separation at equal per-proposal cost}

\begin{corollary}[Separation at equal per-proposal cost]\label{cor:M-sep}
Let $k\ge1$; all tolerances are on $T_3^{(L)}$ from $\delta_o$ without replay, and $\epsilon^*_{\mathrm{bdd}}(\mathcal C)$ stands for
$\epsilon^*_{\mathrm{bdd}}(\mathcal C;T_3^{(L)},o)$.
\begin{enumerate}
\item[(a)] For every $L\ge1$, $\sup_{\mathcal C\in\mathcal P_0(k)}\epsilon^*_{\mathrm{bdd}}(\mathcal C)\ge1/(4(4L-1)^2)$ and
$\epsilon^*(\NB)/\sup_{\mathcal C\in\mathcal P_0(k)}\epsilon^*_{\mathrm{bdd}}(\mathcal C)\le64\ln2\cdot L$. For every $L\ge10^5k$ this ratio is
at least $L/(9\cdot10^{31}k^5)$, so it is $\Theta_k(L)$.
\item[(b)] Let $L_k:=9\cdot10^{31}k^5$. For every $L\ge L_k$ the interval $I_{k,L}:=[6\cdot10^{31}k^5/L^2,\,1-2^{-1/L})$ is nonempty, and for
every $\epsilon\in I_{k,L}$: (i) every $\mathcal C\in\mathcal P_0(k)$ has an admissible prover sequence from $\delta_o$ with every round
defined and $\sup_t|K_t|<\infty$; (ii) every admissible prover sequence for $\NB$ from $\delta_o$ satisfies
$|K_t|\ge\frac32(1-\epsilon)\big(2^{1/L}(1-\epsilon)\big)^t$ for every $t\ge1$, and $2^{1/L}(1-\epsilon)>1$.
\item[(c)] For every $L\ge1$, with $a:=2^{-1/(2L)}$, $\epsilon^*(\NB)/\epsilon^*_{\mathrm{bdd}}(\LRW)\ge\frac{2(1+a)}{3(1-a)}>3L$.
\item[(d)] Let $\mathcal P_0^{=}(k)$ be the set of $\mathcal C\in\mathcal P_0(k)$ such that $\NB$ runs at equal budget with the system of
$\mathcal C$ (Definition~\ref{def:mem-budget}). It contains $\LRW$, with which $\NB$ runs at equal budget in both directions, and every
$\mathcal C$ whose system makes a neighbour-oracle call in at least one proposal. The bounds of (a) hold with $\mathcal P_0^{=}(k)$ in place of
$\mathcal P_0(k)$.
\end{enumerate}
\end{corollary}

\begin{proof}
$\LRW\in\mathcal P_0(k)$ for every $k\ge1$: it proposes $\frac12\delta_x+\frac12\mathrm{Unif}(N(x))$, which lies in $B(x,1)$, does not depend on $t$,
and is preserved by view isomorphisms, which map $N(x)$ onto the neighbours of the image root.
(a) By Corollary~\ref{cor:st-tree}(b), the supremum is at least $\epsilon^*_{\mathrm{bdd}}(\LRW)\ge\epsilon^*_{\mathrm{div}}(\LRW)\ge1/(4(4L-1)^2)$,
and with $\epsilon^*(\NB)\le\ln2/L$ the ratio is at most $4\ln2(4L-1)^2/L\le64\ln2\cdot L$. For $L\ge10^5k$, Theorem~\ref{thm:fs-U} gives
$\epsilon^*_{\mathrm{bdd}}(\mathcal C)\le6\cdot10^{31}k^5/L^2$ for every $\mathcal C\in\mathcal P_0(k)$, and $\epsilon^*(\NB)\ge\frac{\ln2}L(1-\frac{\ln2}{2L})\ge0.693/L$,
so the ratio is at least $0.693L/(6\cdot10^{31}k^5)\ge L/(9\cdot10^{31}k^5)$.
(b) $L_k\ge10^5k$, and $0.693/L>6\cdot10^{31}k^5/L^2$ because $L\ge9\cdot10^{31}k^5>6\cdot10^{31}k^5/0.693$, so $I_{k,L}\ne\emptyset$. (i) is
Theorem~\ref{thm:fs-U}, valid for every $\epsilon\in[6\cdot10^{31}k^5/L^2,1)$; (ii) is Theorem~\ref{thm:M-tree}(c).
(c) By Corollary~\ref{cor:st-tree}(b), $0<\epsilon^*_{\mathrm{bdd}}(\LRW)\le\frac32(1-a)^2$, and $\epsilon^*(\NB)=1-a^2=(1-a)(1+a)$, which gives the
first inequality. With $u:=\ln2/(2L)\in(0,\ln2/2]$ and $a=e^{-u}$, $1-a<u$ and $1+a>2-u$, so
$\frac{2(1+a)}{3(1-a)}>\frac{2(2-u)}{3u}=\frac{8L}{3\ln2}-\frac23>3L$, because $\frac8{3\ln2}-3>0.84$ and $0.84L>\frac23$.
(d) The description is the last statement of Lemma~\ref{lem:mem-budget} with $\LRW\in\mathcal P_0(k)$. Since
$\mathcal P_0^{=}(k)\subseteq\mathcal P_0(k)$, its supremum is at most that over $\mathcal P_0(k)$, which gives the lower bound of (a); the upper
bound of (a) used only $\LRW\in\mathcal P_0^{=}(k)$.
\end{proof}

By Lemma~\ref{lem:mem-budget}(d), at every level $\epsilon<1-w$ one frontier sample costs, on both sides of these comparisons, at most
$(1-w)/(1-w-\epsilon)$ proposals and $2/p$ proof draws in expectation, so counting rounds compares systems at equal budget.

\section{Plateau heights and the proofs of \texorpdfstring{Proposition~\ref{prop:boundary}(b) and~(c)}{Proposition 5(b) and (c)}}\label{app:G_plateau}\label{sec:plateau}

\paragraph{Roadmap.} Sections~\ref{sec:pl-defs}--\ref{sec:pl-zd} bound the plateau height of the lazy random walk through its
Faber--Krahn function. The growth side follows Chen and Li's proof of their Theorem~4.2 but keeps the exact one-round quantity
$\|Pg_t\|^2$ (Lemma~\ref{lem:pl-round}); the truncation lemma of \citet[Lemma~2.1]{grigoryan1994} turns it into a recursion
(Lemma~\ref{lem:pl-G}, Lemma~\ref{lem:pl-recursion}), and the stall side is Theorem~\ref{thm:st-qsd}. Together they give
Theorem~\ref{thm:plateau}, and on $\mathbb Z^D$ the height $\Theta_D(\epsilon^{-D/2})$ (Corollary~\ref{cor:plateau-zd}).
Section~\ref{sec:pl-escape} proves Proposition~\ref{prop:boundary}(b),(c).

\subsection{The Faber--Krahn function}\label{sec:pl-defs}

In Sections~\ref{sec:pl-defs}--\ref{sec:pl-theorem}, $G$ is connected and locally finite with at least one edge and $d:=\sup_x\deg(x)<\infty$;
$P_0$ is the SRW and $P=\frac12(I+P_0)$ is $\LRW$, both reversible for $\pi=\deg$, so $1\le\pi\le d$. Let $\ell_0$ be the finitely supported
functions, $\langle f,h\rangle=\sum_x\pi(x)f(x)h(x)$, $\|f\|^2=\langle f,f\rangle$ and $\|f\|_1=\sum_x\pi(x)|f(x)|$. For a kernel $K$ reversible for
$\pi$ (we use $K\in\{P,P^2\}$) let $\mathcal E_K(f)=\frac12\sum_{x,y}\pi(x)K(x,y)(f(x)-f(y))^2$ for $f\in\ell_0$.

\begin{definition}[Faber--Krahn function]\label{def:pl-fk}
For finite nonempty $U$ and $v,y>0$ (with $\inf\emptyset=+\infty$),
\begin{gather*}
\lambda_1^K(U)=\min\{\mathcal E_K(f)/\|f\|^2:\ f\in\ell_0\setminus\{0\},\ \mathrm{supp}\,f\subseteq U\},\\
\Lambda_K(v)=\inf\{\lambda_1^K(U):\ U\text{ finite, nonempty},\ \pi(U)\le v\},\\ \Lambda_K^{-1}(y)=\sup\{v>0:\Lambda_K(v)\ge y\}.
\end{gather*}
We write $\Lambda:=\Lambda_P$.
\end{definition}

By Lemma~\ref{lem:st-rayleigh}(b),(d), which hold for every kernel reversible for $\pi$, $\lambda_1^K(U)=1-\lambda(K_U)\in[0,1]$; in particular
$\lambda_1^P(U)=1-\lambda_U$, the Dirichlet eigenvalue of the walk killed on leaving $U$. Since $\pi\ge1$, $\Lambda_K=+\infty$ on $(0,1)$, and
$\Lambda_K$ is nonincreasing, so $\{v:\Lambda_K(v)\ge y\}$ is an interval with left end $0$ and
\begin{equation}\label{eq:pl-downclosed}
0<u<\Lambda_K^{-1}(y)\ \Longrightarrow\ \Lambda_K(u)\ge y,\qquad\qquad u>\Lambda_K^{-1}(y)\ \Longrightarrow\ \Lambda_K(u)<y .
\end{equation}

\begin{lemma}[Laziness and the two-step form]\label{lem:pl-comparison}
For every $f\in\ell_0$, $\mathcal E_{P^2}(f)\ge\mathcal E_P(f)$. Hence $\Lambda_{P^2}\ge\Lambda$ and $\Lambda_{P^2}^{-1}\ge\Lambda^{-1}$.
\end{lemma}

\begin{proof}
$P-P^2=\frac14(I-P_0^2)$, and $\langle f,P_0^2f\rangle=\|P_0f\|^2\le\|f\|^2$ by Lemma~\ref{lem:st-rayleigh}(a). So $\langle f,Pf\rangle\ge\langle f,P^2f\rangle$,
and Lemma~\ref{lem:st-rayleigh}(d) for $P$ and $P^2$ gives the inequality; minimise over $f$ and take infima over $U$.
\end{proof}

Laziness is needed here: for $P_0$ and $U=\{x\}$, $\lambda_1^{P_0}(\{x\})=1>1-P_0^2(x,x)=\lambda_1^{P_0^2}(\{x\})$.

\begin{lemma}[Truncation]\label{lem:pl-G}
Let $K$ be a kernel reversible for $\pi$ and $g\in\ell_0$ with $g\ge0$ and $\|g\|_1=1$. Put $a:=\|g\|^2$ and $U_*:=\{x:g(x)>a/4\}$. Then $U_*$ is
finite and nonempty, $\pi(U_*)\le4/a$, and $\mathcal E_K(g)\ge\frac a2\lambda_1^K(U_*)\ge\frac a2\Lambda_K(4/a)$, with $\Lambda_K(4/a)\le1$.
\end{lemma}

\begin{proof}
Let $s=a/4>0$. Markov's inequality gives $s\,\pi(U_*)\le\|g\|_1=1$. Let $h=(g-s)_+$; since $u\mapsto(u-s)_+$ is $1$-Lipschitz and vanishes at $0$,
every term of $\mathcal E_K(h)$ is at most the corresponding term of $\mathcal E_K(g)$, and $\mathrm{supp}\,h=U_*$. Pointwise $g^2\le h^2+2sg$, so
$a\le\|h\|^2+2s=\|h\|^2+a/2$ and $\|h\|^2\ge a/2>0$; in particular $U_*\ne\emptyset$. Hence
$\mathcal E_K(g)\ge\mathcal E_K(h)\ge\lambda_1^K(U_*)\|h\|^2\ge\frac a2\lambda_1^K(U_*)$, and $U_*$ is admissible in the infimum defining $\Lambda_K(4/a)$.
\end{proof}

The truncation $(g-s)_+$ at level $s\asymp\|g\|^2/\|g\|_1$ is that of \citet[Lemma~2.1, after Gushchin]{grigoryan1994}, which passes from a
Faber--Krahn to a Nash-type inequality; the same truncation gives the spectral-profile bound of \citet[Lemma~2.1]{goel2006}.

\subsection{One round and the recursion}\label{sec:pl-round}

\begin{lemma}[One round; Chen and Li's rejection step in exact form]\label{lem:pl-round}
Let the curriculum be $P$, let $D_t$ be finitely supported, and let $s_t$ satisfy Definition~\ref{def:admissible}(i) and $D_t(K_t)\ge1-\eta$ with
$\eta\in[0,1)$. Put $g_t=D_t/\pi$ and $a_t=\|g_t\|^2=\sum_xD_t(x)^2/\pi(x)$. Then $Z_t\ge1-\eta$, $D_{t+1}$ is supported in $B(\mathrm{supp}\,D_t,1)$,
$\|g_t\|_1=1$, and $a_{t+1}\le\|Pg_t\|^2/(1-\eta)^2=(a_t-\mathcal E_{P^2}(g_t))/(1-\eta)^2$.
\end{lemma}

\begin{proof}
$Z_t\ge D_t(K_t)$ by Section~\ref{sec:B-dynamics}, and the support claim holds because $P$ is nearest-neighbour. Reversibility gives
$(D_tP)(y)=\pi(y)(Pg_t)(y)$, so by \eqref{eq:update} $0\le g_{t+1}\le Pg_t/(1-\eta)$; square, multiply by $\pi$ and sum. Finally
$\|Pg_t\|^2=\langle g_t,P^2g_t\rangle=a_t-\mathcal E_{P^2}(g_t)$ by Lemma~\ref{lem:st-rayleigh}(a),(d).
\end{proof}

This is the step (A.49)--(A.54) of \citet{chenli2026} with the exact quantity $\|Pg_t\|^2$ in place of $\|P\|^2_{\ell^2(\pi)}a_t$, which carries no
information on amenable graphs, where $\|P\|_{\ell^2(\pi)}=1$. The count is their Lemma~E.1: by the Cauchy--Schwarz inequality, for finitely
supported $D$ and every $A$, $D(A)^2\le\pi(A\cap\mathrm{supp}\,D)\sum_xD(x)^2/\pi(x)\le d\,|A|\sum_xD(x)^2/\pi(x)$.

\begin{lemma}[The recursion and its invariant region]\label{lem:pl-recursion}
Let $\epsilon\in(0,1)$, let $\Psi:(0,\infty)\to[0,\infty]$ be nonincreasing with $S:=\{u>0:\Psi(u)\ge8\epsilon\}\ne\emptyset$, and $v^*:=\sup S$.
Let $a_t>0$ satisfy
\begin{equation}\label{eq:pl-rec}
a_{t+1}\le a_t\Bigl(1-\tfrac12\Psi(4/a_t)\Bigr)(1-\epsilon)^{-2}\quad\text{and}\quad\Psi(4/a_t)\le2\qquad(t\ge0).
\end{equation}
(a) For every finite $v\in(0,v^*]$, with $M_v:=4/(v(1-\epsilon)^2)$, $a_t\le M_v$ for every $t\ge T_v:=\lceil\ln_+(a_0/M_v)/(2\epsilon)\rceil$.
(b) If $v^*=\infty$, then $a_t\le a_0e^{-2\epsilon t}$ for every $t$.
\end{lemma}

\begin{proof}
Let $r:=(1-4\epsilon)_+/(1-\epsilon)^2$; if $\epsilon\le1/4$, $(1-\epsilon)^2(1-r)=2\epsilon+\epsilon^2\ge2\epsilon(1-\epsilon)^2$, so in all cases
$r\le e^{-2\epsilon}$. $S$ is an interval with left end $0$. If $4/a_t\in S$, then $0\le1-\frac12\Psi(4/a_t)\le(1-4\epsilon)_+$ and $a_{t+1}\le ra_t$. If
$4/a_t\notin S$, then $4/a_t\ge v^*$ and $a_{t+1}\le a_t(1-\epsilon)^{-2}\le4/(v^*(1-\epsilon)^2)\le M_v$. Hence $a_t\le M_v$ implies $a_{t+1}\le M_v$.
If $a_t>M_v$, then $4/a_t<v\le v^*$, so $4/a_t\in S$ and $a_{t+1}\le ra_t$; so $a_t\le e^{-2\epsilon t}a_0$ up to the first time
$\tau$ with $a_\tau\le M_v$, and $\tau\le T_v$ because $e^{-2\epsilon T_v}a_0\le M_v$. This proves (a). If $v^*=\infty$, every step is of the
first kind, which gives (b).
\end{proof}

\subsection{The plateau theorem}\label{sec:pl-theorem}

\begin{theorem}[Plateau height of the lazy random walk]\label{thm:plateau}
Let the curriculum be $P=\LRW$ without replay, $p,\epsilon\in(0,1)$.
\begin{enumerate}[label=(\alph*)]
\item (Growth.) For every finitely supported $D_0$, every admissible prover sequence and every finite $v\in(0,\Lambda^{-1}(8\epsilon)]$,
$|K_t|\ge(1-\epsilon)^4v/(4d)$ for every $t\ge\lceil\ln_+(a_0v(1-\epsilon)^2/4)/(2\epsilon)\rceil$, where $a_0=\sum_xD_0(x)^2/\pi(x)\le1$. If
$\Lambda^{-1}(8\epsilon)=\infty$, then $|K_t|\ge(1-\epsilon)^2e^{2\epsilon t}/(d\,a_0)$ for every $t$. Hence
$\liminf_t|K_t|\ge(1-\epsilon)^4\Lambda^{-1}(8\epsilon)/(4d)$.
\item (Stall.) If $\Lambda(v)<\epsilon/(2d)$, in particular if $v>\Lambda^{-1}(\epsilon/(2d))$, there are a finite nonempty $U$ with $\pi(U)\le v$ and an
admissible prover sequence from $D_0=\nu_U$ with $D_t=\nu_U$ and $K_t=\mathrm{int}\,U$ for every $t$, so $\sup_t|K_t|\le v$.
\item Let $h(\epsilon)$ be the infimum of $\liminf_t|K_t|$ over finitely supported $D_0$ and admissible prover sequences. Then
$\frac{(1-\epsilon)^4}{4d}\Lambda^{-1}(8\epsilon)\le h(\epsilon)\le\Lambda^{-1}(\epsilon/(2d))$.
\end{enumerate}
\end{theorem}

\begin{proof}
(a) By Lemma~\ref{lem:pl-round} with $\eta=\epsilon$, inductively every $D_t$ is finitely supported with $a_t\in(0,\infty)$ and $\|g_t\|_1=1$, and
$a_{t+1}\le(a_t-\mathcal E_{P^2}(g_t))/(1-\epsilon)^2$. Lemma~\ref{lem:pl-G} with $K=P^2$ gives \eqref{eq:pl-rec} for $\Psi=\Lambda_{P^2}$, whose set $S$
contains $(0,1)$ and has $v^*=\Lambda_{P^2}^{-1}(8\epsilon)\ge\Lambda^{-1}(8\epsilon)$ by Lemma~\ref{lem:pl-comparison}; this is the only use of laziness.
For finite $v\le\Lambda^{-1}(8\epsilon)$, Lemma~\ref{lem:pl-recursion}(a) gives $a_t\le4/(v(1-\epsilon)^2)$ for $t\ge T_v$, and the count after
Lemma~\ref{lem:pl-round} with $D_t(K_t)\ge1-\epsilon$ gives $|K_t|\ge(1-\epsilon)^2/(d\,a_t)\ge(1-\epsilon)^4v/(4d)$. Since $\pi\ge1$,
$a_0\le\sum_xD_0(x)=1$. If $\Lambda^{-1}(8\epsilon)=\infty$, use Lemma~\ref{lem:pl-recursion}(b) instead. For the $\liminf$, take
$v=\Lambda^{-1}(8\epsilon)$ when it is finite.
(b) By the definition of $\Lambda$ there is $U$ with $\pi(U)\le v$ and $1-\lambda_U<\epsilon/(2d)$. We may take $U$ connected: $P_U$ is block diagonal
over the components of $U$, so a component $C$ attaining the largest Perron root has $\lambda_C=\lambda_U$ and $\pi(C)\le\pi(U)$. Then $P_U$ is
irreducible with a unique QSD $\nu_U$, Lemma~\ref{lem:st-boundary} with $e_{\min}(P)\ge1/(2d)$ gives $\nu_U(\partial_{\mathrm{in}}U)\le2d(1-\lambda_U)<\epsilon$,
and $\lambda_U>0$. Theorem~\ref{thm:st-qsd}(a) gives the sequence, and $|\mathrm{int}\,U|\le|U|\le\pi(U)\le v$. The second condition implies the
first by \eqref{eq:pl-downclosed}.
(c) The lower bound is (a); for the upper bound apply (b) to every $v>\Lambda^{-1}(\epsilon/(2d))$ (if it is finite).
\end{proof}

\subsection{\texorpdfstring{The lattice $\mathbb Z^D$}{The lattice}}\label{sec:pl-zd}

On $\mathbb Z^D$, $\pi\equiv2D=d$ and $\pi(U)=2D|U|$.

\begin{lemma}[Faber--Krahn function on $\mathbb Z^D$]\label{lem:pl-fk-zd}
For every finite nonempty $U\subseteq\mathbb Z^D$, $1-\lambda_U\ge\frac14|U|^{-2/D}$. Hence $\Lambda(v)\ge\frac14(2D/v)^{2/D}$ and
$\Lambda^{-1}(y)\ge2D(4y)^{-D/2}$ for all $v,y>0$.
\end{lemma}

\begin{proof}
\emph{Form.} $\pi(x)P(x,y)=\frac12$ for $y\in N(x)$, so $\mathcal E_P(f)=\frac12\sum_{\{x,y\}\in E}(f(x)-f(y))^2$ (over unordered edges, including those
leaving $U$), and with $\|f\|^2=2D\sum_xf(x)^2$, $1-\lambda_U=\mu(U)/(4D)$ for
$\mu(U):=\min\{\sum_{\{x,y\}\in E}(f(x)-f(y))^2/\sum_xf(x)^2:f\ne0,\ \mathrm{supp}f\subseteq U\}$.
\emph{Edge isoperimetry.} For finite nonempty $S$, $|\partial_ES|\ge2D|S|^{(D-1)/D}$: with $\pi_i$ deleting the $i$-th coordinate, every line
$\{x:\pi_i(x)=z\}$, $z\in\pi_i(S)$, contributes the two edges leaving the extreme points of its intersection with $S$, so
$|\partial_ES|\ge2\sum_i|\pi_i(S)|$; for $D\ge2$ the discrete Loomis--Whitney inequality $|S|^{D-1}\le\prod_i|\pi_i(S)|$ \citep{loomis1949} and the AM--GM
inequality give $\sum_i|\pi_i(S)|\ge D|S|^{(D-1)/D}$.
\emph{Dirichlet Cheeger bound.} Let $f\ge0$ (replacing $f$ by $|f|$ does not increase the numerator) be supported on $U$ and $S_\tau=\{f^2>\tau\}\subseteq U$.
By the coarea formula and $|S_\tau|^{(D-1)/D}\ge|S_\tau||U|^{-1/D}$,
$\sum_{\{x,y\}}|f(x)^2-f(y)^2|=\int_0^\infty|\partial_ES_\tau|\,d\tau\ge2D|U|^{-1/D}\sum_xf(x)^2$, while the Cauchy--Schwarz inequality and
$(f(x)+f(y))^2\le2f(x)^2+2f(y)^2$ give $\sum_{\{x,y\}}|f(x)^2-f(y)^2|\le(\sum_{\{x,y\}}(f(x)-f(y))^2)^{1/2}(4D\sum_xf(x)^2)^{1/2}$. So
$\mu(U)\ge D|U|^{-2/D}$. Finally $\pi(U)\le v$ gives $|U|\le v/(2D)$, and $v\le2D(4y)^{-D/2}$ gives $\frac14(2D/v)^{2/D}\ge y$.
\end{proof}

\begin{corollary}[Plateau height on $\mathbb Z^D$]\label{cor:plateau-zd}
Let the curriculum be $\LRW$ on $\mathbb Z^D$ without replay, $p\in(0,1)$ and $\epsilon\in(0,1)$.
\begin{enumerate}[label=(\roman*)]
\item For every finitely supported $D_0$ and every admissible prover sequence, $|K_t|\ge\frac14(1-\epsilon)^4(32\epsilon)^{-D/2}$ for every
$t\ge\lceil D\ln_+(1/(32\epsilon))/(4\epsilon)\rceil$.
\item There is an admissible prover sequence from $\delta_0$ with $|K_t|<(\pi^2D/\epsilon)^{D/2}$ for every $t$.
\end{enumerate}
Hence the plateau height from $\delta_0$, the infimum of $\liminf_t|K_t|$ over admissible prover sequences, and $h(\epsilon)$ both lie between
$\frac14(1-\epsilon)^4(32\epsilon)^{-D/2}$ and $(\pi^2D/\epsilon)^{D/2}$, and are $\Theta_D(\epsilon^{-D/2})$ as $\epsilon\to0$.
\end{corollary}

\begin{proof}
(i) Apply Theorem~\ref{thm:plateau}(a) with $v=2D(32\epsilon)^{-D/2}\le\Lambda^{-1}(8\epsilon)$ (Lemma~\ref{lem:pl-fk-zd}) and $d=2D$: then
$(1-\epsilon)^4v/(8D)=\frac14(1-\epsilon)^4(32\epsilon)^{-D/2}$, and $a_0\le1/(2D)$ gives $a_0v(1-\epsilon)^2/4\le\frac14(32\epsilon)^{-D/2}$, so the
waiting time is at most $\lceil D\ln_+(1/(32\epsilon))/(4\epsilon)\rceil$. (ii) is Corollary~\ref{cor:st-ZD}, as $(\pi\sqrt{D/\epsilon})^D=(\pi^2D/\epsilon)^{D/2}$.
\end{proof}

\subsection{The proof-library curriculum and novelty certificates}\label{sec:pl-escape}

\begin{restated}{Proposition~\ref{prop:boundary}(b),(c)}
Let $p\in(0,1)$.
\begin{enumerate}
\item[(b)] \StmtBoundaryB
\item[(c)] \StmtBoundaryC
\end{enumerate}
\end{restated}

\paragraph{Proof of Proposition~\ref{prop:boundary}(b),(c).} Part (a) is proved in Appendix~\ref{app:D_uniform} (Proposition~\ref{prop:un-sel}(b),(c)
with Theorem~\ref{thm:un-A}(c)). The first claim of (b) is Proposition~\ref{prop:novelty-replay}(b), whose hypotheses $w>0$ and $\epsilon<1-w$ hold.
For the second, on $T_3^{(L)}$ with $S_0=\{o\}$, $N^+(x)$ is the set of children of $x$; Proposition~\ref{prop:novelty-replay}(c) with $w=0$ gives
$|K_t|\ge3\cdot2^{\lfloor(t-1)/L\rfloor}(1-\epsilon)^{t+1}\ge\frac32(1-\epsilon)(2^{1/L}(1-\epsilon))^t$, the bound of Theorem~\ref{thm:nb}. Part (c)
is Proposition~\ref{prop:boundary-exposing}(e).

\subsubsection{The proof-library curriculum}\label{sec:pl-boundary}

This curriculum, which lies outside $\mathcal P_0(k)$, stores a library $B_t$ of proved problems; $d=\sup_x\deg(x)<\infty$ and
$d_{\min}=\min_x\deg(x)$. $B_0$ is finite and nonempty; in round $t$ the prover is trained on $D_t=\mathrm{Unif}(B_t)$, giving an admissible $s_t$,
and the curriculum proposes every $y\in N(B_t)\setminus B_t$, applies the filter, samples a proof of every kept proposal and stores it:
$B_{t+1}=B_t\cup\{y\in N(B_t)\setminus B_t:\ s_t(y)\ge p/2\}$. Round $t$ makes at most $d|B_t|$ proposals. Let
$h_{\mathrm{in}}:=\inf_S|\partial_{\mathrm{in}}S|/|S|$ and $\Phi:=\inf_S|\partial_ES|/\mathrm{vol}(S)$ over finite nonempty $S$, with
$\mathrm{vol}(S)=\sum_{x\in S}\deg(x)$.

\begin{proposition}[Proof-library curriculum]\label{prop:boundary-exposing}
Let $p,\epsilon\in(0,1)$.
\begin{enumerate}[label=(\alph*)]
\item (Growth.) If $\epsilon<h_{\mathrm{in}}$, then for every $B_0$ and every admissible prover sequence,
$|B_{t+1}|\ge(1+(h_{\mathrm{in}}-\epsilon)/d)|B_t|$ and $|K_t|\ge(1-\epsilon)|B_t|\ge(1-\epsilon)(1+(h_{\mathrm{in}}-\epsilon)/d)^t|B_0|$.
\item (Stall.) If $|\partial_{\mathrm{in}}S|\le\epsilon|S|$ and $B_0=S$, some admissible prover sequence has $B_t=S$ and $K_t=\mathrm{int}\,S$ for every $t$.
\item (Stall from a point.) If $|\partial_{\mathrm{in}}B(o,R)|\le\epsilon|B(o,R)|$ and $B_0=\{o\}$, some admissible prover sequence has
$B_t=B(o,\min\{t,R\})$ for every $t$ and $K_t=\mathrm{int}\,B(o,R)$ for every $t\ge R$; in particular $\sup_t|K_t|\le|B(o,R)|$.
\item $d_{\min}\Phi/d\le h_{\mathrm{in}}\le d\,\Phi$.
\item On $T_3^{(L)}$, $\frac1{3(2L-1)}\le h_{\mathrm{in}}\le\frac1{2L}$, and for every $\epsilon>\frac1{2L}$ some $R$ satisfies the hypothesis of (c).
Hence, from $B_0=\{o\}$, the knowledge grows exponentially against every admissible prover when $\epsilon<\frac1{3(2L-1)}$, and some admissible
prover keeps it bounded when $\epsilon>\frac1{2L}$.
\end{enumerate}
\end{proposition}

\begin{proof}
(a) Admissibility reads $|B_t\setminus K_t|\le\epsilon|B_t|$, so $|K_t|\ge(1-\epsilon)|B_t|$, and $X:=\partial_{\mathrm{in}}B_t\cap K_t$ has
$|X|\ge(h_{\mathrm{in}}-\epsilon)|B_t|$. Each $x\in X$ has a neighbour $y_x\notin B_t$ with $s_t(y_x)\ge s_t(x)-p/2\ge p/2$, so $y_x\in B_{t+1}\setminus B_t$,
and each $y$ equals $y_x$ for at most $d$ vertices $x$.
(b) Use $s_t=s_S$, which satisfies (i) by Lemma~\ref{lem:st-window}(b). If $B_t=S$, then $D_t(K_t)=1-|\partial_{\mathrm{in}}S|/|S|\ge1-\epsilon$ and every
proposal $y\notin S$ has $s_S(y)=0$, so $B_{t+1}=S$.
(c) For $t<R$ use $s_t=s^{B(o,t)}$ (Definition~\ref{def:mem-normal-form}): if $B_t=B(o,t)$, then $D_t(K_t)=1$ and every proposal lies in
$B(o,t+1)\setminus B(o,t)$ with pass rate $p/2$, so $B_{t+1}=B(o,t+1)$, and $K_t=B(o,t)$. From $t=R$ on, apply (b) with $S=B(o,R)$.
(d) Every edge of $\partial_ES$ has its endpoint in $S$ in $\partial_{\mathrm{in}}S$, and each such vertex is an endpoint of between $1$ and $d$ of them,
so $|\partial_{\mathrm{in}}S|\le|\partial_ES|\le d|\partial_{\mathrm{in}}S|$; also $d_{\min}|S|\le\mathrm{vol}(S)\le d|S|$.
(e) \emph{Lower bound.} For a component $S_j$ of $S$, $\partial_{\mathrm{in}}S\cap S_j=\partial_{\mathrm{in}}S_j$, so it suffices to take $S$ connected, with
$n_b$ branch and $n_p$ subdivision vertices. If $n_b=0$, Lemma~\ref{lem:st-phi} gives $|S|\le L-1$, and $|\partial_{\mathrm{in}}S|\ge1$. If $n_b\ge1$,
Lemma~\ref{lem:st-phi} gives $|\partial_ES|=n_b+2$ and $|S|\le n_b(2L-1)+L-1$, and $|\partial_{\mathrm{in}}S|\ge|\partial_ES|/3$ by the proof of (d), so
$|\partial_{\mathrm{in}}S|/|S|\ge(n_b+2)/(3(n_b(2L-1)+L-1))\ge1/(3(2L-1))$ because $2(2L-1)\ge L-1$.
\emph{Upper bound.} $\partial_{\mathrm{in}}B(o,R)=S_R$, and by \eqref{eq:st-levels}, $|S_{NL}|=3\cdot2^{N-1}$ and $|B(o,NL)|=1+3L(2^N-1)$, whose ratio tends
to $1/(2L)$ as $N\to\infty$. So $h_{\mathrm{in}}\le1/(2L)$, and for $\epsilon>1/(2L)$ some $R=NL$ satisfies the hypothesis of (c). The final claims are
(a) and (c).
\end{proof}

\subsubsection{Novelty certificate and replay}\label{sec:pl-novelty}

Fix a finite nonempty $S_0\subseteq V$ and let $r(x):=\mathrm{dist}(x,S_0)$. The novelty certificate of $y\in N(x)$ is $\mathbf 1[r(y)>r(x)]$, and
$N^+(x):=\{y\in N(x):r(y)=r(x)+1\}$. The graph has no dead ends if $N^+(x)\neq\emptyset$ for every $x$. An outward kernel is a kernel $O$ with
$O(x,N^+(x))=1$ for every $x$; it is memoryless but uses the certificate, which an anonymous local view does not provide in general: on $T_3^{(L)}$ with $S_0=\{o\}$ and $2k<L$, some subdivision vertex has a
reflection-symmetric ball of radius $k$, and under constant pass rates view equivariance forces equal weights on its two neighbours, so no rule of
$\mathcal P_0(k)$ implements an outward kernel for every profile. The frontier follows \eqref{eq:update}
with $C_t=O$ from $F_0=D_0$, a probability on $S_0$, and the prover trains on $D_t=(1-w)F_t+wR_t$ as in Section~\ref{sec:B-dynamics}.

\begin{proposition}[Novelty certificate with replay]\label{prop:novelty-replay}
Let $G$ have no dead ends, $O$ be an outward kernel, $w\in[0,1)$, $\epsilon<1-w$, $\epsilon_F:=\epsilon/(1-w)$, and $(s_t)$ admissible.
\begin{enumerate}[label=(\alph*)]
\item $Z_0\ge1-\epsilon$ and $Z_t\ge1-\epsilon_F$ for $t\ge1$, so the frontier is defined for all $t$, and $\mathrm{supp}\,F_t\subseteq\{r=t\}$.
\item If $w>0$, then $|K_t|\ge(1-\epsilon/w)\,t$ for every $t\ge1$; for $w=\frac12$ and $\epsilon<\frac12$, $|K_t|\ge(1-2\epsilon)t$.
\item On $T_3^{(L)}$ with $S_0=\{o\}$, $D_0=\delta_o$ and $O(x,\cdot)$ uniform on the children of $x$,
$|K_t|\ge3\cdot2^{\lfloor(t-1)/L\rfloor}(1-\epsilon_F)^{t+1}$ for every $t\ge1$.
\end{enumerate}
\end{proposition}

\begin{proof}
For $t\ge1$, $(1-w)F_t(V\setminus K_t)+wR_t(V\setminus K_t)=D_t(V\setminus K_t)\le\epsilon$, so $F_t(K_t)\ge1-\epsilon_F$ and, if $w>0$,
$R_t(K_t)\ge1-\epsilon/w$; also $F_0(K_0)=D_0(K_0)\ge1-\epsilon$.
(a) $N^+(x)\subseteq N(x)$, so proposals from $K_t$ are kept and $Z_t\ge F_t(K_t)>0$; $O(x,\cdot)$ is carried by $\{r=r(x)+1\}$.
(b) By (a), $x$ carries frontier mass only in round $r(x)$, so $R_t(x)\le1/t$ and $|K_t|\ge R_t(K_t)/\max_xR_t(x)\ge(1-\epsilon/w)t$.
(c) Each $y$ with $r(y)=s+1$ has a unique parent and the vertices at distance $s$ have $b_s$ children (Lemma~\ref{lem:mem-tree-geometry}), so
$M_{s+1}\le M_s/(b_sZ_s)$ for $M_s=\max_xF_s(x)$, $M_0=1$. By (a) and Lemma~\ref{lem:mem-tree-geometry}(c),
$M_t\le(3\cdot2^{\lfloor(t-1)/L\rfloor}(1-\epsilon)(1-\epsilon_F)^{t-1})^{-1}$, and $|K_t|\ge F_t(K_t)/M_t$ with $1-\epsilon\ge1-\epsilon_F$.
\end{proof}

\sloppy
\section{Conservative provers}\label{app:H_conservative}

The stall constructions of Appendices~\ref{app:D_uniform} and~\ref{app:E_fixedstart} let the prover raise the pass rates of problems
that the current training distribution does not contain. This appendix restricts the prover so that, off the training support, a pass
rate may only fall or rise as far as the Lipschitz condition forces, and determines which stalls survive. The model and notation are
those of Appendix~\ref{app:B_model}; on $T_3^{(L)}$ we use the levels $r(x)=\mathrm{dist}(o,x)$, the spheres $S_r$, the group $\Gamma$
and the kernels $\mathcal C^s$, $P^{\mathrm{acc}}_s$, $P^\flat$ of Appendix~\ref{app:D_uniform}.

\paragraph{Roadmap.} Section~\ref{sec:h-def} defines conservative provers and shows that Theorems~\ref{thm:rw} and~\ref{thm:nb} and
Corollary~\ref{cor:sep}(c) hold verbatim, because their stalls already satisfy the restriction (Corollary~\ref{cor:h-transfer}).
Section~\ref{sec:h-HA} defines a conservative version of the alternating construction that learns only the training support
(Lemma~\ref{lem:h-HA}); for $k=1$ it coincides with the alternating construction for every rule that ignores the labels at known roots
(Proposition~\ref{prop:h-blind}), and rules that move towards a label in $[p/2,p)$ are stalled at error $0$ (Proposition~\ref{prop:h-pingpong}).
Section~\ref{sec:h-P01} proves that every $\mathcal C\in\mathcal P_0(1)$ has a conservative stall at training error at most $1.21\cdot10^4/L^2$
from a quasi-stationary start of its cycle (Theorem~\ref{thm:h-P01}). The new ingredient is Lemma~\ref{lem:h-band}: the edge of the learned
ball moves outward by one level per round, and band mass still reaches the dead zone with probability at least $9/256$.
Corollary~\ref{cor:h-summary} collects the results.

\subsection{Definition and the results that hold verbatim}\label{sec:h-def}

\begin{definition}[Conservative prover sequence]\label{def:conservative}
Let $(D_t)_{t\ge0}$ be the training distributions of a run (with replay, $D_t=(1-w)F_t+wR_t$; for a curriculum with one step of memory,
$D_t$ is the law of the current vertex). A prover sequence $(s_t)_{t\ge0}$ is conservative admissible at level $\epsilon$ if it satisfies (i) and (ii)
of Definition~\ref{def:admissible} and
\begin{enumerate}
\item[(iii)] for every $t\ge1$ and every $x\notin\mathrm{supp}\,D_t$, $s_t(x)\le\max\{s_{t-1}(x),\ell_t(x)\}$, where
$\ell_t(x):=\sup_{y\in\mathrm{supp}\,D_t}\big(s_t(y)-\tfrac p2\,\mathrm{dist}(x,y)\big)_+$.
\end{enumerate}
The profile $s_0$ is unrestricted. The conservative tolerances $\epsilon^{*,\mathrm{cs}}_{\mathrm{div}}$ and $\epsilon^{*,\mathrm{cs}}_{\mathrm{bdd}}$ are
defined as $\epsilon^*_{\mathrm{div}}$ and $\epsilon^*_{\mathrm{bdd}}$ in Section~\ref{sec:B-dynamics}, with ``admissible'' replaced by
``conservative admissible''.
\end{definition}

Applying (i) along a geodesic gives $s_t(x)\ge s_t(y)-\frac p2\mathrm{dist}(x,y)$ for all $x,y$, so $\ell_t(x)\le s_t(x)$: $\ell_t(x)$ is the
least value that (i) forces at $x$ from the pass rates on the training support. Condition (iii) therefore says that off the training support
a pass rate may only fall, or rise as far as Assumption~3 of \citet{chenli2026} forces. For $K\subseteq V$ let $s^K$ be the profile equal to
$p$ on $K$, to $p/2$ on $N(K)\setminus K$ and to $0$ elsewhere; the ball profile of Appendix~\ref{app:D_uniform} is $s^{[N]}=s^{B(o,N)}$.

\begin{lemma}[Basic properties]\label{lem:h-basic}
\begin{enumerate}
\item[(a)] For every curriculum, graph and cold start, $\epsilon^*_{\mathrm{div}}\le\epsilon^{*,\mathrm{cs}}_{\mathrm{div}}\le\epsilon^{*,\mathrm{cs}}_{\mathrm{bdd}}$
and $\epsilon^*_{\mathrm{bdd}}\le\epsilon^{*,\mathrm{cs}}_{\mathrm{bdd}}$.
\item[(b)] A prover sequence with $s_t=s_{t-1}$ for every $t\ge1$ satisfies (iii).
\item[(c)] For every $K\subseteq V$, $s^K$ is $p/2$-Lipschitz on edges, $\{s^K\ge p\}=K$ and $\{s^K\ge p/2\}=K\cup N(K)$. If $s_t=s^{K_t}$
for every $t$ and $K_t\subseteq K_{t-1}\cup\mathrm{supp}\,D_t$ for every $t\ge1$, then (iii) holds.
\end{enumerate}
\end{lemma}

\begin{proof}
(a) Let $\mathcal G,\mathcal B$ be the sets whose supremum and infimum define $\epsilon^*_{\mathrm{div}},\epsilon^*_{\mathrm{bdd}}$, and
$\mathcal G^{\mathrm{cs}},\mathcal B^{\mathrm{cs}}$ the conservative ones. A conservative admissible sequence is admissible at the same level. Hence, if every
admissible sequence has $|K_t|\to\infty$, so does every conservative admissible one, and $\mathcal G\subseteq\mathcal G^{\mathrm{cs}}$; and a conservative
admissible sequence with $\sup_t|K_t|<\infty$ is admissible, so $\mathcal B^{\mathrm{cs}}\subseteq\mathcal B$. Condition (iii) does not involve
$\epsilon$, so a sequence conservative admissible at level $\epsilon_1$ is conservative admissible at every $\epsilon_2\in[\epsilon_1,1)$; therefore
$\mathcal G^{\mathrm{cs}}$ is closed downwards and $\mathcal B^{\mathrm{cs}}$ upwards. They are disjoint, since a level in $\mathcal B^{\mathrm{cs}}$
has a conservative admissible sequence with $\sup_t|K_t|<\infty$. If $g\in\mathcal G^{\mathrm{cs}}$ and $b\in\mathcal B^{\mathrm{cs}}$ had $g\ge b$,
then $g\in\mathcal B^{\mathrm{cs}}$; so $g<b$, and $\sup\mathcal G^{\mathrm{cs}}\le\inf\mathcal B^{\mathrm{cs}}$ (also when one set is empty, by the
conventions $\sup\emptyset=0$, $\inf\emptyset=1$).
(b) $s_t(x)=s_{t-1}(x)$.
(c) Let $\{x,y\}\in E$. If $x,y\in K\cup N(K)$, both values lie in $\{p/2,p\}$. If $x\notin K\cup N(K)$, then $y\notin K$, so
$s^K(y)\in\{0,p/2\}$ and $s^K(x)=0$. The set identities hold because $0<p/2<p$. For (iii) let $t\ge1$ and $x\notin\mathrm{supp}\,D_t$. If
$x\in K_t$, then $x\in K_{t-1}$ and $s_{t-1}(x)=p=s_t(x)$. If $x\in N(K_t)\setminus K_t$, then $s_t(x)=p/2$; choose $y\in K_t$ adjacent to $x$.
If $y\in K_{t-1}$, then $x\in K_{t-1}\cup N(K_{t-1})$ and $s_{t-1}(x)\ge p/2$; otherwise $y\in\mathrm{supp}\,D_t$ and
$\ell_t(x)\ge s_t(y)-p/2=p/2$. In all other cases $s_t(x)=0$.
\end{proof}

\begin{corollary}[Results that hold verbatim]\label{cor:h-transfer}
With $\epsilon^*_{\mathrm{div}},\epsilon^*_{\mathrm{bdd}}$ replaced by $\epsilon^{*,\mathrm{cs}}_{\mathrm{div}},\epsilon^{*,\mathrm{cs}}_{\mathrm{bdd}}$, and
``admissible'' by ``conservative admissible'', the following hold as stated: Theorem~\ref{thm:rw}, Theorem~\ref{thm:nb} and
Corollary~\ref{cor:sep}(c). In particular $\epsilon^{*,\mathrm{cs}}_{\mathrm{div}}(\NB;T_3^{(L)},o)=\epsilon^{*,\mathrm{cs}}_{\mathrm{bdd}}(\NB;T_3^{(L)},o)=1-2^{-1/L}$
for every $L\ge1$.
\end{corollary}

\begin{proof}
By Lemma~\ref{lem:h-basic}(a), every lower bound on $\epsilon^*_{\mathrm{div}}$ is a lower bound on $\epsilon^{*,\mathrm{cs}}_{\mathrm{div}}$, and a
growth statement for every admissible sequence covers every conservative admissible one. The stalls of Theorem~\ref{thm:rw} hold a window $U$:
in part~(a) the window of Theorem~\ref{thm:st-qsd}(a), in part~(b) the ball $B(o,mL)$ of Corollary~\ref{cor:st-tree}(b), with any replay
weight. These sequences are constant in $t$, hence conservative by Lemma~\ref{lem:h-basic}(b), so the upper bounds of Theorem~\ref{thm:rw} hold
for $\epsilon^{*,\mathrm{cs}}_{\mathrm{bdd}}$. The prover sequence constructed in the proof of Theorem~\ref{thm:M-tree-stall} uses in round $t$
the profile $s^{K_t}$ with $K_t\subseteq\mathrm{supp}\,D_t$, which is conservative by Lemma~\ref{lem:h-basic}(c); so every
$\epsilon\in(1-2^{-1/L},1)$ lies in $\mathcal B^{\mathrm{cs}}$ for $\NB$, and with Theorem~\ref{thm:M-tree} and Lemma~\ref{lem:h-basic}(a) both
conservative tolerances of $\NB$ equal $1-2^{-1/L}$; the growth bound of Theorem~\ref{thm:nb} covers every conservative admissible sequence. For
Corollary~\ref{cor:sep}(c), with $a=2^{-1/(2L)}$, $\epsilon^{*,\mathrm{cs}}(\NB)=1-a^2$ and
$0<\frac1{4(4L-1)^2}\le\epsilon^{*,\mathrm{cs}}_{\mathrm{bdd}}(\LRW)\le\frac32(1-a)^2$ by the transferred Theorems~\ref{thm:nb} and~\ref{thm:rw}(b), and the
arithmetic of the proof of Corollary~\ref{cor:M-sep}(c) applies unchanged.
\end{proof}

The alternating construction of Definition~\ref{def:un-wash} sets $s^{\mathrm{hi}}=p$ on all of $B(o,A_{\mathrm{hi}})$, including vertices
outside the training support whose pass rate was $0$ in the preceding low round, so it does not satisfy (iii). The next section replaces it.

\subsection{The conservative alternating construction}\label{sec:h-HA}

\begin{definition}[Conservative alternating construction]\label{def:h-HA}
Let $k\ge1$, $\mathcal C\in\mathcal P_0(k)$ on $T_3^{(L)}$, integers $A\ge1$ and $m\ge1$, and a finitely supported cold start $D_0$. Call round
$t$ low if $(m+1)\mid t$ and high otherwise. The conservative alternating construction with parameters $(A,m)$ sets $K_t:=B(o,A)$ in low rounds,
$K_t:=K_{t-1}\cup\mathrm{supp}\,D_t$ in high rounds, and $s_t:=s^{K_t}$ in every round.
\end{definition}

Here $D_t$ is computed by \eqref{eq:update} from $D_{t-1}$ and $s_{t-1}$, so $K_t$ is defined in every round that is reached. The low profile
is the low profile $s^{[A]}$ of the alternating construction; in the high rounds the prover learns only the vertices that have carried training
mass during the current cycle, besides $B(o,A)$.

\begin{lemma}[Conservativeness and size]\label{lem:h-HA}
In the setting of Definition~\ref{def:h-HA}, as long as the rounds are defined: (a) $(s_t)$ satisfies (i) and (iii); (b) in a high round
$\mathrm{supp}\,D_t\subseteq K_t$, and in a low round $D_t(V\setminus K_t)=D_t(V\setminus B(o,A))$; (c) in the $j$-th high round of a cycle
($1\le j\le m$), $K_t\subseteq B(o,A+1+(j-1)k)$, so $|K_t|\le|B(o,A+1+(m-1)k)|$ for every $t$.
\end{lemma}

\begin{proof}
(a) Condition (i) is Lemma~\ref{lem:h-basic}(c). For (iii) we check $K_t\subseteq K_{t-1}\cup\mathrm{supp}\,D_t$ for $t\ge1$ and apply
Lemma~\ref{lem:h-basic}(c). In a high round this holds with equality. A low round $t\ge1$ follows a high round, whose knowledge set contains
the knowledge set $B(o,A)$ of the preceding low round, because high rounds only add vertices; so $K_t=B(o,A)\subseteq K_{t-1}$.
(b) holds by definition. (c) In a low round $t_0$, $\{s_{t_0}\ge p/2\}=B(o,A+1)$ by Lemma~\ref{lem:h-basic}(c), so the kept proposals lie in
$B(o,A+1)$ and $\mathrm{supp}\,D_{t_0+1}\subseteq B(o,A+1)$. By (L) every later round moves mass by at most $k$ levels, so
$\mathrm{supp}\,D_{t_0+i}\subseteq B(o,A+1+(i-1)k)$ for $1\le i\le m$, and $K_{t_0+j}=B(o,A)\cup\bigcup_{i\le j}\mathrm{supp}\,D_{t_0+i}$.
\end{proof}

\begin{proposition}[Rules blind at known roots]\label{prop:h-blind}
Let $p\in(0,1)$ and let $\mathcal C\in\mathcal P_0(1)$ on $T_3^{(L)}$ be blind at known roots: $\mathcal C_v=\mathcal C_{v^\flat}$ for every
labelled rooted view $v$ of radius $1$ whose labels are $p/2$-Lipschitz on the edges of the view and whose root has label at least $p$, where
$v^\flat$ is $v$ with every label replaced by $p$.
\begin{enumerate}
\item[(a)] If $\sigma^2>0$ and $L\ge400$, let $A,m$ be the parameters of Definition~\ref{def:un-params} with $k=1$. From $D_0=\delta_o$, the
conservative alternating construction with parameters $(A,m)$ and the alternating construction produce the same $D_t$ in every round, and every
round is defined. The conservative alternating construction is conservative admissible at every level $\epsilon\in[\bar\eta,1)$, where
$\bar\eta:=\sup_nD^{\mathrm{lo}}_n(\mathsf T)$ for the alternating construction from $\delta_o$.
\item[(b)] Consequently $\epsilon^{*,\mathrm{cs}}_{\mathrm{bdd}}(\mathcal C;T_3^{(L)},o)\le6\cdot10^{31}/L^2$ for every $L\ge10^5$, and
$\epsilon^{*,\mathrm{cs}}_{\mathrm{bdd}}(\mathcal C;T_3^{(L)},o)\le12000/L^2$ for every $L\ge400$ if $\mathcal C$ is lazy in the sense of
Corollary~\ref{cor:un-lazy}. If $\sigma^2=0$ and $L\ge4$, then $\epsilon^{*,\mathrm{cs}}_{\mathrm{bdd}}(\mathcal C;T_3^{(L)},o)=0$.
\end{enumerate}
\end{proposition}

\begin{proof}
(a) Induction on $t$: assume both constructions have the same $D_t$. In a low round both use $s^{[A]}$. Let $t$ be the $j$-th high round of
a cycle. For the alternating construction, $D_t$ is supported in $B(o,A_{\mathrm{hi}}-1)$, where every vertex has a flat view under
$s^{\mathrm{hi}}$ (Lemma~\ref{lem:un-blindfold}(b) with $k=1$), so its row at every $x\in\mathrm{supp}\,D_t$ is $P^\flat(x,\cdot)$
(Lemma~\ref{lem:un-flat}). For the conservative construction, $x\in\mathrm{supp}\,D_t\subseteq K_t$ has label $p$, its view is $p/2$-Lipschitz by
Lemma~\ref{lem:h-basic}(c), so blindness gives $\mathcal C^{s_t}(x,\cdot)=P^\flat(x,\cdot)$; this law lives on $B(x,1)\subseteq K_t\cup N(K_t)$,
where $s_t\ge p/2$, so every proposal is kept. Both rows equal $P^\flat(x,\cdot)$ on $\mathrm{supp}\,D_t$, so $D_{t+1}$ and the normaliser
agree. The rounds of the alternating construction from $\delta_o$ are defined by Lemma~\ref{lem:un-star}. By Lemma~\ref{lem:h-HA}, the conservative
construction satisfies (i) and (iii), has error $0$ in high rounds and error $D^{\mathrm{lo}}_n(\mathsf T)\le\bar\eta$ in the $n$-th low round
($K_t=B(o,A)$, $\mathsf T=V\setminus B(o,A)$), and $\sup_t|K_t|<\infty$.
(b) For $L\ge10^5$, Theorem~\ref{thm:fs-main} with $k=1$ gives $\bar\eta\le6\cdot10^{31}/L^2$; if this number is less than $1$ it lies in
$\mathcal B^{\mathrm{cs}}$ by (a), and otherwise the bound holds by definition. For lazy $\mathcal C$ and $L\ge400$, Corollary~\ref{cor:un-lazy}
shows that the alternating construction from $\delta_o$ is admissible at level $12000/L^2<1$; since its low-round knowledge set is $B(o,A)$,
this is $\bar\eta\le12000/L^2$, and (a) applies. If $\sigma^2=0$, the constant profile $s^{[3]}$ of Theorem~\ref{thm:un-B0} is admissible at every
level from $\delta_o$ with $\sup_t|K_t|<\infty$, and it is conservative by Lemma~\ref{lem:h-basic}(b).
\end{proof}

Every rule whose proposal law does not depend on the labels, such as $\LRW$, is blind at known roots, and so is every rule that reads the
labels only at roots with label below $p$.

\begin{proposition}[Rules that chase a label in $\lbrack p/2,p)$]\label{prop:h-pingpong}
Let $k\ge1$, $L\ge1$, $p\in(0,1)$, $c\in[p/2,p)$ and $a\in N(o)$. Let $v_o$ be the view $(B(o,k),o,u)$ with $u(a)=c$ and $u=p$ on
$B(o,k)\setminus\{a\}$, and $v_a$ the view $(B(a,k),a,u')$ with $u'(o)=c$ and $u'=p$ on $B(a,k)\setminus\{o\}$.
\begin{enumerate}
\item[(a)] If $\mathcal C\in\mathcal P_0(k)$ satisfies $\mathcal C_{v_o}(\{a\})=\mathcal C_{v_a}(\{o\})=1$, then for every $\epsilon\in[0,1)$ there is an
conservative admissible prover sequence from $\delta_o$ with every round defined, training error $0$ and $|K_t|=|B(o,k+1)|-1$ for every $t$. Hence
$\epsilon^{*,\mathrm{cs}}_{\mathrm{div}}(\mathcal C;T_3^{(L)},o)=\epsilon^{*,\mathrm{cs}}_{\mathrm{bdd}}(\mathcal C;T_3^{(L)},o)=0$.
\item[(b)] The rule $\mathcal C^{\min}$ of Remark~\ref{rem:un-static} satisfies the hypothesis of (a) for every $c\in[p/2,p)$. It is also
stalled from $\delta_o$ at error $0$ by a constant profile: with $\delta:=\min\{1-p,p/2\}$, the profile equal to $p$ at $o$, to $p+\delta$ on
levels $1,\dots,k$, to $p/2+\delta$ on level $k+1$, to $\delta$ on level $k+2$ and to $0$ beyond.
\end{enumerate}
\end{proposition}

\begin{proof}
(a) Let $W:=B(o,k+1)$; it contains $B(o,k)$, $B(a,k)$ and every neighbour of $o$ or $a$. In even rounds let $s_t$ be $s^W$ with the value at
$a$ replaced by $c$, in odd rounds $s^W$ with the value at $o$ replaced by $c$. The modified vertex has all its neighbours in $W$, where the
value is $p$, and $|p-c|\le p/2$; with Lemma~\ref{lem:h-basic}(c) this gives (i). If $D_t=\delta_o$ with $t$ even, the view of $o$ is $v_o$, so
the proposal is $a$, which has label $c\ge p/2$ and is kept: $D_{t+1}=\delta_a$. Symmetrically, $D_t=\delta_a$ with $t$ odd gives
$D_{t+1}=\delta_o$. So $D_t=\delta_o$ for even and $\delta_a$ for odd $t$, every normaliser is $1$, the supported vertex has label $p$ (error
$0$), and $K_t=W$ minus one vertex. For (iii), consecutive profiles differ only at $o$ and $a$; the vertex among them that is off the support
in round $t$ has $s_t=c<p=s_{t-1}$. Since the sequence works at $\epsilon=0$, $0\in\mathcal B^{\mathrm{cs}}$, and Lemma~\ref{lem:h-basic}(a)
gives both equalities.
(b) At $v_o$ every label is at least $p/2$ and the smallest one is $c$, attained only at $a$; symmetrically at $v_a$. For the constant profile,
edges join consecutive levels (Lemma~\ref{lem:un-geometry}(a)) and the differences across levels are $\delta$, $0$, $p/2$, $p/2$ and
$\delta$, all at most $p/2$. At $o$ every label in $B(o,k)$ is at least $p$ and the smallest, $p$, is attained only at $o$, so
$\mathcal C^{\min}$ proposes $o$: $D_t=\delta_o$ with $s_t(o)=p$ in every round, $K_t\subseteq B(o,k+1)$, and the profile is conservative by Lemma~\ref{lem:h-basic}(b).
\end{proof}

\subsection{Every rule in \texorpdfstring{$\mathcal P_0(1)$}{P0(1)} from a quasi-stationary start}\label{sec:h-P01}

In this section $k=1$, $p\in(0,1)$ and $\mathcal C\in\mathcal P_0(1)$ on $T_3^{(L)}$, with corridor law $\kappa$ and corridor variance
$\sigma^2$ (Lemma~\ref{lem:un-corridor}). Every $x$ with $r(x)\ge1$ has a parent $x^-\in S_{r(x)-1}$, and a vertex of degree $2$ has one
child $x^+$ (Lemma~\ref{lem:un-geometry}(a)). When $L\ge400$ and $\sigma^2>0$ we use the parameters $R=\lfloor L/2\rfloor$, $a=2$,
$A=\lfloor3L/4\rfloor$, $\theta=\pi/(2(R-a))$ and $\tilde\mu=1-\theta^2\sigma^2/2$ of Definition~\ref{def:un-params} with $k=1$, the function
$h:=H\circ r$ with $H(u)=1$ for $u\le a$, $H(u)=\cos(\theta(u-a))$ for $a<u<R$ and $H(u)=0$ for $u\ge R$, the low-round matrix
$P_{\mathrm{lo}}:=P^{\mathrm{acc}}_{s^{[A]}}$, and, as in Definition~\ref{def:un-wash} with $k=1$, $\mathsf I=\{r\le A-1\}$,
$\mathsf B=\{A\le r\le A+2\}$, $\mathsf Z=\{r\ge A+3\}$, $\mathsf T=\{r\ge A+1\}$, $g(\mu)=\sum_{x\in\mathsf B}\mu(x)(P_{\mathrm{lo}}\mathbf 1)(x)$
and $b(\mu)=\mu(\mathsf B)-g(\mu)$. Lemma~\ref{lem:un-star} with the radius $R$ gives $(P_{\mathrm{lo}}h)(x)\ge\tilde\mu h(x)$ for every $x$.

\begin{lemma}[Views of radius one]\label{lem:h-views}
Let $L\ge4$.
\begin{enumerate}
\item[(a)] For every vertex $x$ of degree $2$, $P^\flat(x,\cdot)=\frac{\sigma^2}2\delta_{x^-}+(1-\sigma^2)\delta_x+\frac{\sigma^2}2\delta_{x^+}$.
\item[(b)] There are $c_2,c_2^{\mathrm b}\in[0,1]$ such that for every $N\ge1$ and $x\in S_N$, $\mathcal C^{s^{[N]}}(x,S_{N+1})=c_2$ if $x$
has degree $2$ and $=c_2^{\mathrm b}$ if $x$ is a branch vertex.
\end{enumerate}
\end{lemma}

\begin{proof}
(a) The two neighbours $x^-,x^+$ of $x$ are not adjacent ($T_3^{(L)}$ is a tree), so the view of $x$ is the path $x^-,x,x^+$. A corridor vertex
$x^\circ$ exists because $L\ge4=2k+2$, and its view is the path $x^\circ_{-1},x^\circ,x^\circ_1$ (Lemma~\ref{lem:un-geometry}(c)). With constant
labels, $x^\mp\mapsto x^\circ_{\mp1}$, $x\mapsto x^\circ$ is an isomorphism of labelled rooted views, so (E) and Lemma~\ref{lem:un-corridor} give
$P^\flat(x,\{x^\mp\})=\kappa(\mp1)$ and $P^\flat(x,\{x\})=\kappa(0)$. A symmetric law on $\{-1,0,1\}$ with variance $\sigma^2$ has
$\kappa(\pm1)=\sigma^2/2$ and $\kappa(0)=1-\sigma^2$.
(b) Under $s^{[N]}$ the view of $x\in S_N$ consists of $x$ and its parent, with label $p$, and its children in $S_{N+1}$, with label $p/2$. If
$x$ has degree $2$ the view is a path whose ends carry $p$ and $p/2$; if $x$ is a branch vertex ($N\ge L$) it is a star with three leaves,
one labelled $p$ and two labelled $p/2$. Any two views of the same kind are isomorphic as labelled rooted views, so by (E) the probability
of the set of $p/2$-labelled vertices is the same number for all views of each kind.
\end{proof}

\begin{proposition}[Degenerate rules]\label{prop:h-degenerate}
Let $L\ge4$. (a) If $\sigma^2=0$, the constant profile $s^{[3]}$ is conservative admissible from $\delta_o$ at every level $\epsilon\in[0,1)$, with
every round defined and $K_t=B(o,3)$. (b) If $c_2=0$ or $P^\flat(o,\{o\})=1$, the constant profile $s^{[1]}$ is conservative admissible from
$\delta_o$ at every level $\epsilon\in[0,1)$, with every round defined and $\mathrm{supp}\,D_t\subseteq B(o,1)=K_t$. In both cases
$\epsilon^{*,\mathrm{cs}}_{\mathrm{div}}(\mathcal C;T_3^{(L)},o)=\epsilon^{*,\mathrm{cs}}_{\mathrm{bdd}}(\mathcal C;T_3^{(L)},o)=0$.
\end{proposition}

\begin{proof}
(a) is Theorem~\ref{thm:un-B0} with $k=1$ ($L\ge4=3k+1$) and Lemma~\ref{lem:h-basic}(b). (b) By induction, $D_t$ is supported in $B(o,1)$ and
every proposal is kept. The view of $o$ under $s^{[1]}$ is flat, so its proposal law is $P^\flat(o,\cdot)$, supported in $B(o,1)$; if
$P^\flat(o,\{o\})=1$, then $D_t=\delta_o$ for all $t$. A vertex $x\in S_1$ has degree $2$ ($L\ge2$) and proposes in $\{o,x,x^+\}$, where
$x^+\in S_2$ is proposed with probability $c_2$ (Lemma~\ref{lem:h-views}(b), $N=1$); if $c_2=0$ every proposal lies in $B(o,1)$. All labels on
$B(o,2)$ are at least $p/2$, so the normaliser is $1$, and $D_t(K_t)=1$. The profile is constant, hence conservative, and $0\in\mathcal B^{\mathrm{cs}}$.
\end{proof}

\begin{definition}[Front schedule and cycle operator]\label{def:h-schedule}
Let $L\ge400$ and $\sigma^2>0$. Let $m'$ be the least even integer with $m'\ge64/\sigma^2$, put $F_j:=A+j$ for $j\ge0$ if $c_2^{\mathrm b}>0$ and
$F_j:=\min\{A+j,L\}$ if $c_2^{\mathrm b}=0$, and
\[
P_j:=P^{\mathrm{acc}}_{s^{[F_j]}}\ (1\le j\le m'),\qquad Q_{\mathrm{cs}}:=P_{\mathrm{lo}}P_1\cdots P_{m'},
\]
$V_{\mathrm{cs}}:=B(o,A+m'+1)$ and $q_0:=9/256$.
\end{definition}

Since $A<L$, $F_{j+1}\in\{F_j,F_j+1\}$, and $F_{j+1}=F_j$ only if $F_j=L$ and $c_2^{\mathrm b}=0$. Also $64\le m'\sigma^2<64+2\sigma^2\le66$, and
$L-A\ge L/4\ge100$. The level $F_j$ is the edge of the learned ball in the $j$-th high round; a vertex $x$ is behind the front in round $j$ if
$r(x)<F_j$ and at the front if $r(x)=F_j$.

\begin{lemma}[Front, flat rows and accounting]\label{lem:h-front}
Under Definition~\ref{def:h-schedule}:
\begin{enumerate}
\item[(a)] For $1\le j\le m'$ and $x\in B(o,F_j)$, $P_j(x,\cdot)$ is a probability supported in $B(x,1)\cap B(o,F_{j+1})$, and
$P_j(x,\cdot)=P^\flat(x,\cdot)$ if $r(x)<F_j$. Every $Q_{\mathrm{cs}}(x,\cdot)$ is supported in $V_{\mathrm{cs}}$, and
$\mu Q_{\mathrm{cs}}\mathbf 1=\mu P_{\mathrm{lo}}\mathbf 1$ for finitely supported $\mu\ge0$.
\item[(b)] For finitely supported $\mu\ge0$: $\mu Q_{\mathrm{cs}}\mathbf 1=\mu\mathbf 1-\mu(\mathsf Z)-b(\mu)$, $b(\mu)\ge0$, and
$\mu(\mathsf T)\le g(\mu)+b(\mu)+\mu(\mathsf Z)$.
\item[(c)] $Q_{\mathrm{cs}}^nh\ge\tilde\mu^{\,n(m'+1)}h$ for $n\ge0$; in particular $\delta_oQ_{\mathrm{cs}}^n\mathbf 1\ge\tilde\mu^{\,n(m'+1)}$.
\item[(d)] $Q_{\mathrm{cs}}(\varphi x,\varphi y)=Q_{\mathrm{cs}}(x,y)$ for $\varphi\in\Gamma$. The level matrix $\bar Q(\ell,\ell'):=Q_{\mathrm{cs}}(x,S_{\ell'})$,
$x\in S_\ell$, $0\le\ell,\ell'\le A+m'+1$, is well defined; $(\mu Q_{\mathrm{cs}})(S_{\ell'})=\sum_\ell\mu(S_\ell)\bar Q(\ell,\ell')$ for every
$\mu\ge0$ supported in $V_{\mathrm{cs}}$; and $\mu Q_{\mathrm{cs}}$, $\mu P_{\mathrm{lo}}$, $\mu P_j$ are $\Gamma$-invariant if $\mu$ is.
\end{enumerate}
\end{lemma}

\begin{proof}
(a) The proposals from $x\in B(o,F_j)$ lie in $B(x,1)\subseteq B(o,F_j+1)$, where $s^{[F_j]}\ge p/2$, so all are kept. If $F_{j+1}=F_j+1$
this is the claim. Otherwise $F_j=L$ and $c_2^{\mathrm b}=0$; then $B(x,1)\subseteq B(o,L)$ if $r(x)<L$, and $P_j(x,S_{L+1})=c_2^{\mathrm b}=0$
if $r(x)=L$ (Lemma~\ref{lem:h-views}(b)). If $r(x)<F_j$, the view of $x$ is flat and Lemma~\ref{lem:un-flat} applies. Since $P_{\mathrm{lo}}(x,\cdot)$
is supported in $B(o,A+1)=B(o,F_1)$ (Lemma~\ref{lem:un-blindfold}(a)), induction on $j$ shows that $\delta_xP_{\mathrm{lo}}P_1\cdots P_j$ is
supported in $B(o,F_{j+1})$ with mass $(P_{\mathrm{lo}}\mathbf 1)(x)$; and $F_{m'+1}\le A+m'+1$.
(b) Lemma~\ref{lem:un-accounting}(i)--(iii) concern $P_{\mathrm{lo}}$ only and give $\mu P_{\mathrm{lo}}\mathbf 1=\mu(\mathsf I)+g(\mu)$ and
$b(\mu)\ge0$; with (a) and $\mu\mathbf 1=\mu(\mathsf I)+\mu(\mathsf B)+\mu(\mathsf Z)$ this is the identity, and $\mathsf T\subseteq\mathsf B\cup\mathsf Z$.
(c) Let $W\in\{P_{\mathrm{lo}},P_1,\dots,P_{m'}\}$. If $r(x)\ge R$, then $h(x)=0\le(Wh)(x)$. If $r(x)<R$, then $r(x)\le A-2$
(Lemma~\ref{lem:un-arith}(a)), so $W(x,\cdot)=P^\flat(x,\cdot)=P_{\mathrm{lo}}(x,\cdot)$ by (a) and Lemma~\ref{lem:un-accounting}(i), and
$(Wh)(x)=(P_{\mathrm{lo}}h)(x)\ge\tilde\mu h(x)$. Nonnegative matrices with finitely supported rows preserve pointwise inequalities, so
$Q_{\mathrm{cs}}^nh\ge\tilde\mu^{\,n(m'+1)}h$; at $o$, $\mathbf 1\ge h\ge0$ and $h(o)=1$ give the second claim.
(d) Each factor satisfies $W(\varphi x,\varphi y)=W(x,y)$ (Lemma~\ref{lem:un-lumping}(a)), and so does a product:
$(WW')(\varphi x,\varphi y)=\sum_{z}W(\varphi x,\varphi z)W'(\varphi z,\varphi y)=(WW')(x,y)$. Since $\varphi(S_{\ell'})=S_{\ell'}$ and $\Gamma$ is
transitive on $S_\ell$ (Lemma~\ref{lem:un-geometry}(b)), $Q_{\mathrm{cs}}(x,S_{\ell'})$ does not depend on $x\in S_\ell$. Grouping $x$ by level gives
the formula for $(\mu Q_{\mathrm{cs}})(S_{\ell'})$, and $(\mu W)(\varphi y)=\sum_{x}\mu(\varphi x)W(\varphi x,\varphi y)=(\mu W)(y)$ for $\Gamma$-invariant $\mu$.
\end{proof}

\begin{lemma}[Band to dead zone through the moving front]\label{lem:h-band}
Under Definition~\ref{def:h-schedule}: (a) for every $y$ with $r(y)\in\{A-1,A,A+1\}$, $(P_1\cdots P_{m'})(y,\mathsf Z)\ge q_0$; (b) for every
finitely supported $\mu\ge0$, $(\mu Q_{\mathrm{cs}})(\mathsf Z)\ge q_0\,g(\mu)$.
\end{lemma}

\begin{proof}
\emph{Idea.} A particle behind the front moves by the symmetric walk of Lemma~\ref{lem:h-views}(a). A particle at the front either rides
it, which ends in $\mathsf Z$, or falls behind it and never catches it again before level $L$. In every case the remaining walk is long
enough to carry it to $\mathsf Z$ with a fixed probability, and it cannot come back from level $L-1$ to level $A+2$.

\emph{Coupling.} Let $\xi_1,\dots,\xi_{m'}$ be i.i.d.\ with law $\kappa$ and $U_1,\dots,U_{m'}$ i.i.d.\ uniform on $[0,1]$, independent of the
$\xi_i$, and put $S_j:=\xi_1+\dots+\xi_j$. For each $j$ and $x\in B(o,F_j)$ fix $\Phi_{j,x}:[0,1]\to V$ such that $\Phi_{j,x}(U_1)$ has law
$P_j(x,\cdot)$. Let $Y_0:=y$ and, for $j\ge1$, let $Y_j$ be $Y_{j-1}^-$, $Y_{j-1}$ or $Y_{j-1}^+$ according as $\xi_j=-1,0,1$ if $Y_{j-1}$ has
degree $2$ and $r(Y_{j-1})<F_j$ (a $\kappa$-step), and $Y_j:=\Phi_{j,Y_{j-1}}(U_j)$ otherwise. By Lemma~\ref{lem:h-front}(a) and induction,
$Y_{j-1}\in B(o,F_j)$, and by Lemma~\ref{lem:h-views}(a) and Lemma~\ref{lem:h-front}(a) the conditional law of $Y_j$ given the past is
$P_j(Y_{j-1},\cdot)$ in both cases. So $\Pr(Y_{m'}\in\mathsf Z)=(P_1\cdots P_{m'})(y,\mathsf Z)$. A $\kappa$-step changes the level by
$\xi_j$, and every step changes it by at most $1$.

\emph{The front.} Consider the condition
\[
(\star)_i:\qquad r(Y_{i-1})<F_i,\quad\text{or}\quad r(Y_{i-1})=F_i=L\ \text{and}\ c_2^{\mathrm b}=0 .
\]
If $(\star)_i$ holds, then $r(Y_i)\le F_i$ (in the second case $P_i(Y_{i-1},S_{L+1})=0$); if $F_{i+1}=F_i+1$ this gives $r(Y_i)<F_{i+1}$, and
otherwise $F_i=L$, $c_2^{\mathrm b}=0$ and $F_{i+1}=L\ge r(Y_i)$. So $(\star)_i$ implies $(\star)_{i+1}$. Under $(\star)_i$, if
$1\le r(Y_{i-1})\le L-1$, then $Y_{i-1}$ is behind the front and has degree $2$ (Lemma~\ref{lem:un-geometry}(a)), so step $i$ is a $\kappa$-step.

If $r(y)\in\{A-1,A\}$, then $(\star)_1$ holds ($F_1=A+1$); put $j_*:=0$. If $r(y)=A+1=F_1$, let $J:=\min\{j\le m':r(Y_j)\ne F_j+1\}$
($\min\emptyset:=\infty$). For $j<J$ the particle rides: $r(Y_j)=F_j+1\le F_{j+1}$ by Lemma~\ref{lem:h-front}(a), so $F_{j+1}=F_j+1=r(Y_j)$ and
the particle is at the front in round $j+1$. Hence steps $1,\dots,J$ are not $\kappa$-steps, and $J$ is a function of $(U_i)$ alone. If
$J=\infty$, then $r(Y_{m'})=F_{m'}+1=A+m'+1\ge A+3$. If $J\le m'$, then $r(Y_J)\in\{F_J-1,F_J\}$, so $(\star)_{J+1}$ holds by the argument
above; put $j_*:=J$. In all cases $(\star)_i$ holds for every $i\in(j_*,m']$, and $r(Y_{j_*})\ge A-1$.

\emph{Good event.} Let $\mathcal G:=\{|S_j-S_i|\le L-A-5\text{ for }0\le i\le j\le m'\}$. We claim that on $\mathcal G$ (and $J\le m'$ in the
second case), either $r(Y_{m'})\ge A+3$, or $r(Y_{m'})=r(Y_{j_*})+S_{m'}-S_{j_*}$. If $r(Y_i)\ge L-1$ for some $i\in[j_*,m']$, let $\tau$ be the
largest such $i$. If $\tau=m'$, then $r(Y_{m'})\ge L-1\ge A+3$. Otherwise $r(Y_\tau)=L-1$ and $r(Y_{\tau+1})=L-2$, and by induction on
$i\in[\tau+1,m']$, $r(Y_i)=L-2+S_i-S_{\tau+1}\ge L-2-(L-A-5)=A+3$: if this holds for $i<m'$, then $r(Y_i)\in[A+3,L-2]$, and $(\star)_{i+1}$
makes step $i+1$ a $\kappa$-step. So $r(Y_{m'})\ge A+3$. If instead $r(Y_i)\le L-2$ for all $i\in[j_*,m']$, then by induction
$r(Y_i)=r(Y_{j_*})+S_i-S_{j_*}\ge(A-1)-(L-A-5)=2A-L+4\ge1$, since $A\ge(3L-3)/4$; so $(\star)_{i+1}$ makes every step $i+1>j_*$ a
$\kappa$-step, which proves the claim.

\emph{Probabilities.} Let $E_j:=\{S_{m'}-S_j\ge4\}$ for $j\le3$ and $E_j:=\{S_{m'}-S_j\ge0\}$ for $4\le j\le m'$. On $\{j_*=j\}\cap E_j\cap\mathcal G$
the particle ends in $\mathsf Z$: by the claim, either $r(Y_{m'})\ge A+3$, or $r(Y_{m'})=r(Y_j)+S_{m'}-S_j$, where $r(Y_j)\ge A-1$ for $j\le3$, and
$r(Y_j)\ge F_j-1\ge\min\{A+j,L\}-1\ge A+3$ for $j\ge4$. The event $\{j_*=j\}$ depends only on $(U_i)$ and is independent of $E_j$ and $\mathcal G$, so
\[
\Pr(Y_{m'}\in\mathsf Z)\ \ge\ \Pr(J=\infty)+\sum_{j}\Pr(j_*=j)\big(\Pr(E_j)-\Pr(\mathcal G^c)\big)\ \ge\ \min_j\Pr(E_j)-\Pr(\mathcal G^c).
\]
For $j\le3$, $S_{m'}-S_j$ has the law of $S_n$ with $n=m'-j$ and $n\sigma^2\ge64-3\sigma^2\ge61$. Since $\xi_1\in\{-1,0,1\}$,
$\mathbb E\xi_1^4=\sigma^2$, and expanding, $\mathbb ES_n^4=n\sigma^2+3n(n-1)\sigma^4\le n\sigma^2+3n^2\sigma^4$. The Paley--Zygmund inequality
$\Pr(X>\frac14\mathbb EX)\ge\frac9{16}(\mathbb EX)^2/\mathbb EX^2$ for $X=S_n^2$ gives
$\Pr(S_n^2>\frac14n\sigma^2)\ge\frac{9/16}{3+1/(n\sigma^2)}\ge\frac9{64}$; on this event $|S_n|>\sqrt{61}/2>3$, so $|S_n|\ge4$, and by the
symmetry of $\kappa$, $\Pr(S_n\ge4)\ge\frac9{128}$. For $j\ge4$, $\Pr(S_{m'}-S_j\ge0)\ge\frac12$ by symmetry. If
$\max_{j\le m'}|S_j|<(L-A-4)/2$, then $|S_j-S_i|<L-A-4$, so $\mathcal G$ holds; by Kolmogorov's maximal inequality,
$\Pr(\mathcal G^c)\le4m'\sigma^2/(L-A-4)^2\le4\cdot66/96^2=264/9216$, and $264\cdot256=67584<82944=9\cdot9216$ gives $\Pr(\mathcal G^c)<\frac9{256}$.
Hence $(P_1\cdots P_{m'})(y,\mathsf Z)\ge\frac9{128}-\frac9{256}=q_0$.

(b) Let $\nu':=(\mu\mathbf 1_{\mathsf B})P_{\mathrm{lo}}$. It has mass $g(\mu)$ and is supported on levels $A-1,A,A+1$
(Lemma~\ref{lem:un-accounting}(iii)). Since $\mu Q_{\mathrm{cs}}\ge\nu'P_1\cdots P_{m'}$ entrywise, (a) gives $(\mu Q_{\mathrm{cs}})(\mathsf Z)\ge q_0g(\mu)$.
\end{proof}

\begin{lemma}[Quasi-stationary start of the cycle]\label{lem:h-qsd}
Under Definition~\ref{def:h-schedule}:
\begin{enumerate}
\item[(a)] $\bar Q$ has a left eigenvector $\bar\nu\ge0$, $\bar\nu\mathbf 1=1$, with eigenvalue $\lambda:=\rho(\bar Q)\in[\tilde\mu^{m'+1},1]$, and
$\tilde\mu^{m'+1}>1-q_0$. Its lift $\nu(x):=\bar\nu(r(x))/|S_{r(x)}|$ is a $\Gamma$-invariant probability on $V_{\mathrm{cs}}$ with
$\nu Q_{\mathrm{cs}}=\lambda\nu$.
\item[(b)] If $P^\flat(o,\{o\})<1$, every $\Gamma$-invariant probability $\nu$ on $V_{\mathrm{cs}}$ with $\nu Q_{\mathrm{cs}}=\lambda\nu$ for some
$\lambda>1-q_0$ satisfies $\nu(S_A)>0$.
\item[(c)] Every probability $\nu$ on $V_{\mathrm{cs}}$ with $\nu Q_{\mathrm{cs}}=\lambda\nu$ and $\lambda\ge\tilde\mu^{m'+1}$ satisfies
\[
\nu(\mathsf T)\ \le\ \Big(1+\frac1{q_0}\Big)(1-\lambda)\ \le\ \Big(1+\frac1{q_0}\Big)\big(1-\tilde\mu^{m'+1}\big)\ \le\ \frac{1.21\cdot10^4}{L^2}.
\]
\end{enumerate}
\end{lemma}

\begin{proof}
(a) By the Perron--Frobenius theorem for nonnegative square matrices \citep{hornjohnson2013}, applied to $\bar Q^\top$, $\rho(\bar Q)$ is an
eigenvalue with a left eigenvector $\bar\nu\ge0$, which we normalise. Row sums of $\bar Q$ are $Q_{\mathrm{cs}}(x,V_{\mathrm{cs}})\le1$, so $\rho(\bar Q)\le1$.
By Lemma~\ref{lem:h-front}(d) applied $n$ times, $\bar Q^n(0,\ell)=(\delta_oQ_{\mathrm{cs}}^n)(S_\ell)$, so by Lemma~\ref{lem:h-front}(c)
$\|\bar Q^n\|_{\infty\to\infty}\ge\delta_oQ_{\mathrm{cs}}^n\mathbf 1\ge\tilde\mu^{n(m'+1)}$, and Gelfand's formula gives $\rho(\bar Q)\ge\tilde\mu^{m'+1}$.
By (c) below, $1-\tilde\mu^{m'+1}\le409/L^2<q_0$. The lift $\nu$ is $\Gamma$-invariant; by Lemma~\ref{lem:h-front}(d), $\nu Q_{\mathrm{cs}}$ is
$\Gamma$-invariant with level masses $\bar\nu\bar Q=\lambda\bar\nu$, so $\nu Q_{\mathrm{cs}}=\lambda\nu$ by Lemma~\ref{lem:un-lumping}(d).

(b) Put $\bar\nu(\ell):=\nu(S_\ell)$; by Lemma~\ref{lem:h-front}(d), $\bar\nu\bar Q=\lambda\bar\nu$. \emph{Step 1: $\bar\nu$ charges a level
$\ell\le A-1$.} Rows of $\bar Q$ at levels in $\mathsf Z$ vanish (Lemma~\ref{lem:un-accounting}(ii)). For $x\in\mathsf B$,
Lemma~\ref{lem:h-front}(a) and Lemma~\ref{lem:h-band}(b) with $\mu=\delta_x$ give
$Q_{\mathrm{cs}}(x,V\setminus\mathsf Z)\le(1-q_0)(P_{\mathrm{lo}}\mathbf 1)(x)\le1-q_0$. If $\bar\nu$ vanished on the levels of $\mathsf I$, then
$\lambda\,\nu(\mathsf B)=(\nu Q_{\mathrm{cs}})(\mathsf B)\le(1-q_0)\nu(\mathsf B)$, so $\nu(\mathsf B)=0$; then $\nu$ would live on $\mathsf Z$ and
$\lambda\nu=\nu Q_{\mathrm{cs}}=0$, a contradiction. \emph{Step 2: from such a level, $S_A$ is reached at a low time.} Let $x\in S_\ell$,
$\ell\le A-1$. We give a path $x=v_0,v_1,\dots,v_{n(m'+1)}\in S_A$ in which each step $v_{i-1}\to v_i$ has positive probability under the
matrix of round $i-1$ of the cycle ($P_{\mathrm{lo}}$ if $(m'+1)\mid(i-1)$, otherwise $P_{(i-1)\bmod(m'+1)}$); then $\bar Q^n(\ell,A)=Q_{\mathrm{cs}}^n(x,S_A)>0$.
Every vertex of level at most $A-1$ has a flat view in every round, and so has every vertex of level $A$ in high rounds ($F_j\ge A+1$); by
Lemma~\ref{lem:h-views}(a) such a vertex of degree $2$ moves to its parent and to its child with probability $\sigma^2/2>0$ each, and $o$ moves
to $S_1$ with probability $1-P^\flat(o,\{o\})>0$. So the path can climb from $x$ to a vertex of $S_{A-1}$ and then alternate between levels
$A-1$ and $A$, moving from level $A$ only in high rounds, until it is at level $A$ at a time divisible by $m'+1$. Such a time exists: the path is
at level $A$ at every second time, and since $m'+1$ is odd, consecutive multiples of $m'+1$ have different parities. Hence $\lambda^n\bar\nu(A)\ge
\bar\nu(\ell)\bar Q^n(\ell,A)>0$.

(c) Extend $\nu$ by $0$. By Lemma~\ref{lem:h-front}(b), $\lambda=\nu Q_{\mathrm{cs}}\mathbf 1=1-\nu(\mathsf Z)-b(\nu)$, so
$1-\lambda=\nu(\mathsf Z)+b(\nu)$. By Lemma~\ref{lem:h-band}(b), $q_0g(\nu)\le(\nu Q_{\mathrm{cs}})(\mathsf Z)=\lambda\nu(\mathsf Z)\le\nu(\mathsf Z)$.
Hence $\nu(\mathsf T)\le g(\nu)+b(\nu)+\nu(\mathsf Z)\le(1+1/q_0)\nu(\mathsf Z)+b(\nu)\le(1+1/q_0)(1-\lambda)$. For the numbers,
$1-u^{n}\le n(1-u)$ for $u\in[0,1]$, and $\tilde\mu\in(1/2,1]$ (Lemma~\ref{lem:un-arith}(f)); $m'<64/\sigma^2+2$ gives
$(m'+1)\sigma^2<64+3\sigma^2\le67$; and $R-a\ge(L-1)/2-2\ge0.45L$ as $L\ge50$. So
\[
1-\tilde\mu^{m'+1}\le(m'+1)\frac{\theta^2\sigma^2}2=\frac{(m'+1)\sigma^2\pi^2}{8(R-a)^2}\le\frac{67\pi^2}{1.62\,L^2}\le\frac{409}{L^2},
\]
since $67\pi^2=661.26\ldots\le662.58=409\cdot1.62$. Finally $(1+256/9)\cdot409=108385/9=12042.7\ldots\le1.21\cdot10^4$, and
$409/L^2\le409/160000<9/256$.
\end{proof}

\begin{theorem}[Every rule in $\mathcal P_0(1)$ from the cycle QSD]\label{thm:h-P01}
Let $L\ge400$, $p\in(0,1)$, and let $\mathcal C\in\mathcal P_0(1)$ on $T_3^{(L)}$ have $\sigma^2>0$, $c_2>0$ and $P^\flat(o,\{o\})<1$. Let $\nu$ be
the quasi-stationary distribution of Lemma~\ref{lem:h-qsd}(a). The conservative alternating construction with parameters $(A,m')$ started from
$D_0=\nu$ has every round defined, $D_{n(m'+1)}=\nu$ for every $n$, knowledge $B(o,A)$ in low rounds and $B(o,F_j)$ in the $j$-th high round of
each cycle, and it is conservative admissible at every level $\epsilon\in[\nu(\mathsf T),1)\supseteq[1.21\cdot10^4/L^2,1)$, with
$\sup_t|K_t|\le|B(o,A+m')|<\infty$. Hence
\[
\epsilon^{*,\mathrm{cs}}_{\mathrm{div}}(\mathcal C;T_3^{(L)},\nu)\ \le\ \epsilon^{*,\mathrm{cs}}_{\mathrm{bdd}}(\mathcal C;T_3^{(L)},\nu)\ \le\ \frac{1.21\cdot10^4}{L^2}.
\]
\end{theorem}

\begin{proof}
Let $\lambda$ be the eigenvalue of $\nu$; $\lambda\ge\tilde\mu^{m'+1}>1-q_0$, so $\nu(S_A)>0$ by Lemma~\ref{lem:h-qsd}(b). Suppose that
$D_{t_0}=\nu$ in a low round $t_0$ (true for $t_0=0$). Then $s_{t_0}=s^{[A]}$, $\nu P_{\mathrm{lo}}\mathbf 1=\nu Q_{\mathrm{cs}}\mathbf 1=\lambda>0$
(Lemma~\ref{lem:h-front}(a)), and $D_{t_0+1}=\nu P_{\mathrm{lo}}/\lambda$. Each $x\in S_A$ has degree $2$ and, under $s^{[A]}$, proposes its child
in $S_{A+1}$ with probability $c_2$ (Lemma~\ref{lem:h-views}(b)), which is kept; so $D_{t_0+1}(S_{F_1})\ge c_2\nu(S_A)/\lambda>0$. We show by
induction on $j=1,\dots,m'$ that in round $t=t_0+j$: $D_t$ is $\Gamma$-invariant, supported in $B(o,F_j)$, with $D_t(S_{F_j})>0$ whenever
$F_j>F_{j-1}$, and $K_t=B(o,F_j)$ (once the front has reached $L$, so that $F_j=F_{j-1}=L$, positivity on $S_L$ is not needed: the support stays in
$B(o,L)=K_{t-1}$). Invariance and support follow from Lemma~\ref{lem:h-front}(a),(d). By definition $K_t=K_{t-1}\cup\mathrm{supp}\,D_t$ with
$K_{t-1}=B(o,F_{j-1})$ ($F_0=A$); if $F_j=F_{j-1}+1$, the $\Gamma$-invariant $D_t$ charges every vertex of $S_{F_j}$
(Lemma~\ref{lem:un-lumping}(d)), so in both cases $K_t=B(o,F_j)$. Then $s_t=s^{[F_j]}$, the matrix of round $t$ is $P_j$, the normaliser is $1$
(Lemma~\ref{lem:h-front}(a)), and $D_{t+1}=D_tP_j$. If $F_{j+1}=F_j+1$, every $x\in S_{F_j}$ moves to $S_{F_j+1}$ with probability $c_2$ or
$c_2^{\mathrm b}$ according as $x$ has degree $2$ or is a branch vertex (Lemma~\ref{lem:h-views}(b)); this number is positive, because the
schedule passes a branch level only if $c_2^{\mathrm b}>0$. So $D_{t+1}(S_{F_{j+1}})>0$, which completes the induction. At the end of the cycle
$D_{t_0+m'+1}=\nu Q_{\mathrm{cs}}/\lambda=\nu$. Thus every round is defined and the knowledge sets are as stated.

By Lemma~\ref{lem:h-HA}, the sequence satisfies (i) and (iii), its error is $0$ in high rounds and $\nu(\mathsf T)$ in low rounds, and
$|K_t|\le|B(o,F_{m'})|\le|B(o,A+m')|$. By Lemma~\ref{lem:h-qsd}(c), $\nu(\mathsf T)\le1.21\cdot10^4/L^2\le12100/160000<1$, so this level
lies in $\mathcal B^{\mathrm{cs}}$, and Lemma~\ref{lem:h-basic}(a) gives the display.
\end{proof}

\begin{corollary}[Summary]\label{cor:h-summary}
Let $p\in(0,1)$ and $L\ge400$.
\begin{enumerate}
\item[(a)] For every $\mathcal C\in\mathcal P_0(1)$ on $T_3^{(L)}$ there is a finitely supported cold start $D_0$, namely $\delta_o$ if
$\sigma^2=0$, $c_2=0$ or $P^\flat(o,\{o\})=1$, and the quasi-stationary distribution of Theorem~\ref{thm:h-P01} otherwise, such that for every
$\epsilon\in[1.21\cdot10^4/L^2,1)$ some conservative admissible prover sequence from $D_0$ has every round defined and $\sup_t|K_t|<\infty$. Hence
$\epsilon^{*,\mathrm{cs}}_{\mathrm{bdd}}(\mathcal C;T_3^{(L)},D_0)\le1.21\cdot10^4/L^2$, and $=0$ in the first case.
\item[(b)] If $\mathcal C\in\mathcal P_0(1)$ is blind at known roots (Proposition~\ref{prop:h-blind}), then from $\delta_o$ it satisfies the bound
of Theorem~\ref{thm:uniform} against conservative provers, $\epsilon^{*,\mathrm{cs}}_{\mathrm{bdd}}(\mathcal C;T_3^{(L)},o)\le6\cdot10^{31}/L^2$ for
$L\ge10^5$, and $\epsilon^{*,\mathrm{cs}}_{\mathrm{bdd}}(\mathcal C;T_3^{(L)},o)\le12000/L^2$ if $\mathcal C$ is also lazy.
\item[(c)] $\epsilon^{*,\mathrm{cs}}_{\mathrm{div}}(\NB;T_3^{(L)},o)=\epsilon^{*,\mathrm{cs}}_{\mathrm{bdd}}(\NB;T_3^{(L)},o)=1-2^{-1/L}$. For
$L\ge9\cdot10^{31}$ and every $\epsilon\in[6\cdot10^{31}/L^2,1-2^{-1/L})$, which is nonempty, every $\mathcal C$ as in (b) has a conservative admissible
prover sequence from $\delta_o$ with every round defined and $\sup_t|K_t|<\infty$, while every admissible prover sequence for $\NB$ from $\delta_o$
satisfies $|K_t|\ge\frac32(1-\epsilon)\big(2^{1/L}(1-\epsilon)\big)^t$ for every $t\ge1$, with $2^{1/L}(1-\epsilon)>1$.
\end{enumerate}
\end{corollary}

\begin{proof}
(a) Proposition~\ref{prop:h-degenerate} and Theorem~\ref{thm:h-P01}. (b) Proposition~\ref{prop:h-blind}. (c) The first statement is
Corollary~\ref{cor:h-transfer}. The interval is the interval $I_{1,L}$ of Corollary~\ref{cor:M-sep}(b), nonempty for $L\ge L_1=9\cdot10^{31}$;
the stall is Proposition~\ref{prop:h-blind}(b), and the growth is Theorem~\ref{thm:M-tree}(c).
\end{proof}
\fussy

\end{document}